%% file: main.tex
\documentclass[letterpaper,10pt]{scrartcl}

\usepackage{lineno,hyperref}
\modulolinenumbers[5]

\usepackage[utf8]{inputenc}
\usepackage[english]{babel}
\usepackage{amsmath,amssymb,upref}
\usepackage{amsfonts}
\usepackage{amsthm,array,makecell}
\usepackage{color}
\usepackage{graphicx}
\usepackage{sidecap}
\usepackage{multirow}
\usepackage{booktabs}
\usepackage{fullpage}
\usepackage[title]{appendix}

\usepackage[font=small]{caption}
\usepackage{subcaption}

\usepackage{longtable}
\usepackage{rotating}

\usepackage{algorithm}
\usepackage{algcompatible}

\usepackage{grffile}
\graphicspath{{figures/}}

\usepackage{pgfplots}
\pgfplotsset{width=10cm,compat=1.9}
\usepackage{xcolor}
\definecolor{dark-violet}{RGB}{148,0,211}
\definecolor{sea-green}{RGB}{46, 139,  87}

\DeclareGraphicsExtensions{.pdf,.eps,.png,.jpg,.jpeg}
\newcommand{\R}{\mathbb{R}}

\newcommand{\bfeps}{\boldsymbol \varepsilon}

\newcommand{\bfx}{\boldsymbol x}

\newcommand{\bff}{\boldsymbol f}

\newcommand{\Dcal}{\mathcal{D}}

\newcommand{\Tcal}{\mathcal{T}}
\newcommand{\Ucal}{\mathcal{U}}
\newcommand{\Vcal}{\mathcal{V}}

\newcommand{\bfmu}{\boldsymbol \mu}

\newcommand{\bfq}{\boldsymbol q}

\newcommand{\bfu}{\boldsymbol u}

\newcommand{\bfv}{\boldsymbol v}

\newcommand{\bfz}{\boldsymbol z}

\newcommand{\bfA}{\boldsymbol A}
\newcommand{\bfB}{\boldsymbol B}

\newcommand{\bfD}{\boldsymbol D}

\newcommand{\bfI}{\boldsymbol I}

\newcommand{\bfL}{\boldsymbol L}

\newcommand{\bfO}{\boldsymbol O}
\newcommand{\bfP}{\boldsymbol P}
\newcommand{\bfR}{\boldsymbol R}
\newcommand{\bfU}{\boldsymbol U}
\newcommand{\bfV}{\boldsymbol V}

\newcommand{\bfQ}{\boldsymbol Q}

\newcommand{\nh}{N}
\newcommand{\nr}{n}

\newcommand{\np}{p}

\newcommand{\Qcal}{\mathcal{Q}}

\newcommand{\hbfq}{\hat{\bfq}}

\newcommand{\hbff}{\hat{\bff}}
\newcommand{\hPhi}{\hat{\Phi}}
\newcommand{\hbfz}{\hat{\bfz}}

\newcommand{\bftheta}{\boldsymbol{\theta}}

\newcount\Comments  
\newcommand{\kibitz}[2]{\ifnum\Comments=1\textcolor{#1}{#2}\fi}

\ifpdf
\hypersetup{
  pdftitle={Operator inference with roll outs for learning reduced models from scarce and low-quality data}, 
  pdfauthor={Wayne Isaac Tan Uy and Dirk Hartmann and Benjamin Peherstorfer} 
}
\fi

\newenvironment{keyword}%
   {\begin{trivlist}\item[]{\bfseries\sffamily Keywords:}\ }
   {\end{trivlist}}
   
   \title{Operator inference with roll outs for learning reduced models from scarce and low-quality data}

\author{Wayne Isaac Tan Uy\footnote{Courant Institute of Mathematical Sciences, New York University, 251 Mercer Street, New York, NY 10012, USA} \and Dirk Hartmann\footnote{Siemens Industrial Software GmbH, Munich, Germany} \and Benjamin Peherstorfer\footnotemark[1]}

\begin{document}

\maketitle

\begin{abstract}
Data-driven modeling has become a key building block in computational science and engineering. However, data that are available in science and engineering are typically scarce, often polluted with noise and affected by measurement errors and other perturbations, which makes learning the dynamics of systems challenging.
In this work, we propose to combine data-driven modeling via operator inference with the dynamic training via roll outs of neural ordinary differential equations.
Operator inference with roll outs inherits interpretability, scalability, and structure preservation of traditional operator inference while leveraging the dynamic training via roll outs over multiple time steps to increase stability and robustness for learning from low-quality and noisy data.
Numerical experiments with data describing shallow water waves and surface quasi-geostrophic dynamics demonstrate that operator inference with roll outs provides predictive models from training trajectories even if data are sampled sparsely in time and polluted with noise of up to 10\%.
\end{abstract}


\section{Introduction}
Learning models from data has become a key building block in computational science and engineering. However, data in science and engineering applications often are scarce, perturbed and noisy. To cope with scarce low-quality data, we propose to combine non-intrusive model reduction via operator inference \cite{paper:PeherstorferW2016} with the dynamic training via roll outs of neural ordinary differential equations (Neural ODEs) \cite{NEURIPS2018_69386f6b}. We show that roll outs make operator inference more robust against scarce data and noise. Numerical experiments demonstrate that the proposed operator inference approach with roll outs learns reduced models even when traditional operator inference fails because of too little data and too much noise.

Learning reduced models from data is a widely studied problem. There are dynamic mode decomposition and Koopman-based methods \cite{paper:Schmid2010,FLM:6837872,Tu2014391,NathanBook,Williams2015,doi:10.1137/20M1338289} as well as sparsity-based methods \cite{paper:BruntonPK2016,Schaeffer6634,Schaeffer2018,Rudye1602614}. Related are methods for data-driven closure modeling such as \cite{doi:10.1137/17M1145136,MOU2021113470}. A range of methods for non-intrusive model reduction learn coefficients of low-dimensional approximations from data \cite{Hesthaven2018,DEMO2019873,Guo2018,doi:10.1080/10618562.2014.932352,Guo2019,LARIO2022111475,PhysRevE.100.053306}; see \cite{Demo2018} for a software package. The systems and control community studies data-driven model reduction methods from the frequency-domain perspective, e.g.,
 Loewner methods \cite{ANTOULAS01011986,5356286,Mayo2007634,BeaG12,paper:AntoulasGI2016,paper:GoseaA2018,paper:IonitaA2014,Drmac2022,PSW16TLoewner}, vector fitting \cite{772353,paper:DrmacGB2015}, and eigensystem realization \cite{doi:10.2514/3.20031,KramerERZ}. All of these techniques are static in the sense that the prediction error of the to-be-learned model is penalized based only on the residual from a prediction of a single step into the future, because this typically leads to linear optimization problems.

Operator inference is a non-intrusive model reduction method that learns from time-domain data and is applicable to nonlinear systems \cite{paper:PeherstorferW2016}. It is a building block of other scientific machine learning techniques \cite{QIAN2020132401,https://doi.org/10.48550/arxiv.2110.07653,khodabakhshi2022non,Uy2021,UP20OpInfError,BENNER2020113433,SWK21-HOPINF,doi:10.1080/03036758.2020.1863237,Guo_2022,https://doi.org/10.48550/arxiv.2110.07653} and has been applied to a range of challenging problems in engineering \cite{doi:10.2514/1.J058943}. However, operator inference is prone to perturbations and noise in the data. The work \cite{QianThesis} considers a perturbation analysis for operator inference, which shows that perturbations introduced via the approximation of the time derivatives can have large effects on the model predictions. The work \cite{paper:Peherstorfer2019} shows that even errors in the dynamics due to the projection onto low-dimensional subspaces can lead to large errors in the predicted states. Regularization is one remedy and is considered in \cite{doi:10.2514/1.J058943,doi:10.1080/03036758.2020.1863237,https://doi.org/10.48550/arxiv.2107.02597}; however, determining regularizers that bias against noise and at the same time avoid losing the signal in the data that describe actual dynamics of the underlying systems remains an open problem.
Operator inference from noisy data is also considered in the work \cite{AOpInf}, where an active learning approach is proposed for damping the effect of noise on the prediction error. All these approaches are based on the original formulation of operator inference that relies on the residual of the prediction one time step into the future only.

In contrast, we propose a formulation of operator inference that leverages ideas from differentiable programming and Neural ODEs. We include the accuracy of predictions of the to-be-learned model of multiple time steps into the future in the training objective. Our work is motivated by the preprint \cite{HartmannFailerOpInf} that formulates operator inference in a differentiable setting and analytically derives gradients for the training. The training objective is motivated by Neural ODEs \cite{NEURIPS2018_69386f6b} that parametrize the right-hand side function of an ODE as a deep-neural network and then train the network parameters by integrating the corresponding ODE forward in time.  Neural ODEs have been used in the context of model reduction in, e.g., \cite{NEURIPS2019_42a6845a,doi:10.1098/rspa.2021.0162}, where the network additionally depends on a physical input. There also is work on fitting Neural ODEs to the coefficients of projected data  \cite{https://doi.org/10.48550/arxiv.2202.12373}. Similarly, the work \cite{SAN2019271} trains a network to predict the next state of reduced models; there also is \cite{Fresca2021} that use deep networks to make predictions into the future but no roll outs are used, which avoids differentiating through a time-integration scheme.
Bayesian formulations of analogous concepts have been proposed as well \cite{doi:10.1098/rspa.2020.0290}. There is a line of works \cite{RUDY2019483,https://doi.org/10.48550/arxiv.2205.09479} that aims to identify a noise model of the data, which is different from our approach that aims to regularize against noise via roll outs rather than aiming to explicitly identify a model of the noise.

We leverage the insights gained from the works on Neural ODEs and related methods and combine them with the interpretability and scalability of operator inference. In particular, the parametrization of models via low-degree polynomials of traditional operator inference is beneficial also when using roll outs. Using polynomial models separates linear from nonlinear dynamics, which means that semi-implicit time-integration schemes can be applied. Furthermore, the polynomial structure allows estimating the stability radius of the domain of attraction of the learned models, which provides an analysis of the learned models beyond mere numerical experiments on test sets. Finally, interpolating between polynomial models to quickly derive new models for new physical inputs can be achieved in a structure preserving way by building on widely used techniques from model reduction for interpolating on matrix manifolds \cite{NME:NME2681,panzer_parametric_2010,degroote_interpolation_2010,amsallem_online_2011}.

With roll outs, we show that we gain an increase of robustness of operator inference. Roll outs lead to an implicit stability bias, which can be interpreted as a regularizer that helps when only few data samples are available. Similarly, if data are polluted with noise, roll outs help to prevent overfitting, as our numerical experiments will show. Finally, traditional operator inference typically estimates time derivatives via finite differences, which is prone to inaccuracies in the presence of already little noise and perturbations. Instead, by using roll outs, the time integration scheme is included in the learning process. Thus, the to-be-learned model is discretized in time and only state observations of the underlying systems are required, without time-derivative approximations of the observed states of the underlying high-dimensional system from which data are sampled; see also \cite{doi:10.1098/rspa.2021.0883} for similar observations in the context of dictionary-based learning methods. We present numerical experiments that indicate that operator inference with roll outs can provide predictive models even from coarsely sampled training data trajectories and with noise of up to 10\%.

The manuscript is organized as follows. In Section~\ref{sec:Prelim}, preliminaries are reviewed, such as the traditional formulation of operator inference. Section~\ref{sec:ROpInf} introduces operator inference with roll outs by building on Neural ODEs \cite{NEURIPS2018_69386f6b}. The section also discusses the benefits of polynomial models in terms of interpretability, scalability, and generalization to new physical inputs as well as implementation aspects of operator inference with roll outs. Numerical results are demonstrated in Section~\ref{sec:NumExp} and conclusions are drawn in Section~\ref{sec:Conc}.

\section{Preliminaries}\label{sec:Prelim}
We describe the data collection process in Section~\ref{sec:Prelim:Data} and operator inference \cite{paper:PeherstorferW2016} for learning reduced models from data in Section~\ref{sec:Prelim:StaticOpInf}. The problem formulation is given in Section~\ref{sec:Prelim:ProbForm} to highlight that the operator-inference formulation of  \cite{paper:PeherstorferW2016} is static in the sense that the training loss penalizes the prediction of only a single time step into the future.

\subsection{Collecting data and low-dimensional structure}
\label{sec:Prelim:Data}
Consider a process $(\bfq_t(\bfmu))_{t \geq 0}$ with state $\bfq_t(\bfmu) \in \mathbb{R}^{\nh}$ at time $t$ that is obtained for a control trajectory $(\bfu_t(\bfmu))_{t \geq 0}$ with control $\bfu_t(\bfmu) \in \Ucal \subset \mathbb{R}^p$ at time $t$, initial condition $\bfq_0(\bfmu) \in \Qcal_0 \subseteq \mathbb{R}^{\nh}$, and input $\bfmu \in \Dcal \subset \mathbb{R}^d$. Let $\Dcal_{\text{train}} = \{\bfmu_1, \dots, \bfmu_M\} \subset \Dcal$ be a set of $M$ training inputs. For each training input $\bfmu_i \in \Dcal_{\text{train}}$, there is given an initial condition $\bfq_0(\bfmu_i) \in \Qcal_0$ and a time-discrete control trajectory
\[
\bfU(\bfmu_i) = [\bfu_1(\bfmu_i), \dots, \bfu_K(\bfmu_i)] \in \mathbb{R}^{p \times K}
\]
and the corresponding time-discrete state trajectory
\[
\bfQ(\bfmu_i) = [\bfq_1(\bfmu_i), \dots, \bfq_K(\bfmu_i)] \in \mathbb{R}^{\nh \times K}
\]
at discrete time steps $0 = t_0 < t_1 < \dots < t_K = T$. For ease of exposition, we assume equidistant time steps with time-step size $\delta t$. However, the following methodology extends in a straightforward way to non-equidistant time steps.
Additionally, in the following, we consider processes with a low-dimensional structure so that an accurate reduced model exists. Thus, there is a basis matrix $\bfV = [\bfv_1, \dots, \bfv_{\nr}] \in \mathbb{R}^{\nh \times \nr}$ of the space $\Vcal$ with dimension $\nr \ll \nh$ that is obtained, for example, via the principal component analysis from the snapshots
\[
\bfQ = [\bfq_0(\bfmu_1), \bfQ(\bfmu_1), \bfq_0(\bfmu_2), \bfQ(\bfmu_2), \dots, \bfq_0(\bfmu_M), \bfQ(\bfmu_M)] \in \mathbb{R}^{\nh \times M(K+1)}\,.
\]
In the space $\Vcal$, the states of the process $(\bfq_t(\bfmu))_{t \geq 0}$ can be approximated well; we refer to \cite{book:Antoulas2005,RozzaPateraSurvey,paper:BennerGW2015,P22AMS} for model reduction in general.

\subsection{Static learning of low-dimensional models via operator inference}
\label{sec:Prelim:StaticOpInf}
Operator inference was introduced in \cite{paper:PeherstorferW2016} to learn low-dimensional models from data. It proceeds in three steps:
First, the trajectories are projected onto the reduced space $\Vcal$
\begin{equation}
\bar{\bfQ}(\bfmu_i) =[\bar{\bfq}_1(\bfmu_i), \dots, \bar{\bfq}_K(\bfmu_i)] = \bfV^T\bfQ(\bfmu_i)\,,\qquad i = 1, \dots, M\,.
\label{eq:ProjData}
\end{equation}
Second, the time derivatives of projected states $\bar{\bfq}_j(\bfmu_i)$ at time $t_j$ for $ i = 0, \dots, M$ and $j = 1, \dots, K$ are approximated via, e.g., finite differences, to obtain the approximate time derivatives
\begin{equation}
\bar{\bfQ}'(\bfmu_i) = [\bar{\bfq}_1'(\bfmu_i), \dots, \bar{\bfq}_K'(\bfmu_i)]\,,\qquad i = 1, \dots, M\,.
\label{eq:RegOpInf:TimeDerivatives}
\end{equation}
Third, a low-dimensional model with polynomial structure is fitted to the projected states \eqref{eq:ProjData} and their approximate time derivatives \eqref{eq:RegOpInf:TimeDerivatives} via a least-squares problem.

Operator inference was developed for fitting  polynomial models of degree $L \in \mathbb{N}$,
\[
\frac{\mathrm d}{\mathrm dt} \hat{\bfq}(t; \bfmu) = \sum_{\ell = 1}^{L}\bfA_{\ell}(\bfmu)\hat{\bfq}^{\ell}(t; \bfmu) + \bfB(\bfmu) \bfu(t; \bfmu)
\]
with matrices $\bfA_{\ell}(\bfmu) \in \mathbb{R}^{\nr \times \nr_{\ell}}$ for $\ell = 1, \dots, L$ and control matrix $\bfB(\bfmu) \in \mathbb{R}^{\nr \times \np}$, where $\nr_{\ell} = {\nr + \ell - 1 \choose \ell}$ because $\hat{\bfq}^{\ell}(t; \bfmu)$ contains only combinations that are unique up to commutativity; see \cite{paper:PeherstorferW2016} for details. For all $i = 1, \dots, M$ training inputs, the corresponding model operators $\bfA_1(\bfmu_i), \dots, \bfA_L(\bfmu_i), \bfB(\bfmu_i)$ are fitted via the following linear least-squares problem
\begin{equation}
\min_{\bfA_1, \dots, \bfA_L, \bfB} \frac{1}{K}\sum_{k = 1}^K \left\|\sum_{\ell = 1}^L \bfA_\ell\bar{\bfq}_k^{\ell}(\bfmu_i) + \bfB\bfu_k(\bfmu_i) -\bar{\bfq}_k^{\prime}\right\|^2_2\,,
\label{eq:RegOpInf}
\end{equation}
which gives the inferred operators
\begin{equation}
\bfA_1(\bfmu_i), \dots, \bfA_L(\bfmu_i), \bfB(\bfmu_i)\,.
\label{eq:RegOpInfOps}
\end{equation}
For a new input $\bfmu \in \Dcal \setminus \Dcal_{\text{train}}$, the inferred operators for all $i = 1, \dots, M$ training inputs can be interpolated as shown in \cite{paper:PeherstorferW2016}; see also the work \cite{https://doi.org/10.48550/arxiv.2110.07653} that proposes operator inference for problems with an affine parameter dependence.

The optimization problem \eqref{eq:RegOpInf} is linear in the optimization variables $\bfA_1, \dots, \bfA_L, \bfB$ and thus can be solved efficiently with standard linear algebra routines. This can be seen by reformulating \eqref{eq:RegOpInf} as
\begin{equation}
\bfD(\bfmu_i)\bfO(\bfmu_i) = \bfR(\bfmu_i)\,,
\label{eq:RegOpInfMatForm}
\end{equation}
with matrices $\bfD(\bfmu_i) \in \mathbb{R}^{K \times \bar{\nr}}, \bfO(\bfmu_i) = [\bfA_1^T(\bfmu_i), \dots, \bfA_L^T(\bfmu_i), \bfB^T(\bfmu_i)]^T \in \mathbb{R}^{\bar{\nr} \times \nr}$ and $\bfR(\bfmu_i) \in \mathbb{R}^{K \times \nr}$, where $\bar{\nr} = p + \sum_{\ell = 1}^L \nr_{\ell}$.
The formulation also shows that at least $\bar{\nr}$ rows in $\bfD(\bfmu_i)$ are required, i.e., $K \geq \bar{\nr}$ to obtain a full-rank matrix and thus to avoid an under-determined least-squares problem with non-unique solution; note that if the states are observed with the re-projection sampling scheme, then recovery guarantees can be established under assumptions on the data \cite{paper:Peherstorfer2019}.

If fewer state observations are available, regularization can be helpful, see, e.g., \cite{doi:10.2514/1.J058943,doi:10.1080/03036758.2020.1863237,https://doi.org/10.48550/arxiv.2107.02597}. In the following, we use the minimal-norm solution and thus solve
\begin{equation}
\begin{aligned}
\min_{\bfO(\bfmu_i)}~~~~ & \|\bfO(\bfmu_i)\|_F^2\,,\\
\text{such that~~~~} & \bfD(\bfmu_i)\bfO(\bfmu_i) = \bfR(\bfmu_i)\,,
\end{aligned}
\label{eq:RegOpInfMinNorm}
\end{equation}
for $i = 1, \dots, M$, which can be solved with standard linear algebra routines.

\subsection{Problem formulation}
\label{sec:Prelim:ProbForm}
We refer to a formulation such as \eqref{eq:RegOpInf} as static in the following because the prediction of the learned model is taken into account only for one time step into the future during training. This means, the loss function only compares the residual of the learned model from one time step at a time, without predicting the states during the training process. This has several drawbacks that are inherent to such static formulations:

First, the problem formulation is affected by the sensitivity of least-squares regression to perturbation and noise, which is discussed in the context of operator inference in the work \cite{QianThesis} and later in \cite{AOpInf}. For example, the perturbation can come from the approximation of the time derivatives \eqref{eq:RegOpInf:TimeDerivatives} and from partial state observations \cite{Uy2021}. Another source of perturbations can be a wrong model frame (parametrization) in the sense that, for example, a quadratic model is fitted to data that are observed from a system with cubic dynamics. Noise can be introduced because the observed states are polluted with noise and other numerical perturbations such as early stopping in iterative solvers, as considered in \cite{AOpInf}. These works show that even small perturbations can lead to large errors in the predictions of the learned models as well as to instabilities.

Second, without regularization, problem \eqref{eq:RegOpInf} has multiple solutions if too few data points are available to obtain a full-rank data matrix $\bfD$ in \eqref{eq:RegOpInfMatForm}. While regularization has been proposed for operator inference, it introduces a bias by design and the effect of the bias on the model prediction is still actively studied \cite{doi:10.2514/1.J058943,doi:10.1080/03036758.2020.1863237,https://doi.org/10.48550/arxiv.2107.02597}.
The sensitivity to perturbations in the data and the potentially high requirements of training data can be limitations when learning models with operator inference in practice.

\section{Operator inference with roll outs}\label{sec:ROpInf}
\begin{figure}
\begin{subfigure}[t]{0.48\textwidth}
\resizebox{1.0\columnwidth}{!}{
\begin{tikzpicture}
\begin{axis}[
    height = 7cm,
    width = 10cm,
    axis lines = left,
    xlabel = {time $t$},
    tick style={draw=none},
    yticklabels={,,},
    xticklabels={,,},
    legend pos=south east,
    legend cell align={left}
]
\addplot [
    domain=0.1:1.5,
    samples=100,
    color=black,
    line width=2pt,
]
{ln(x)};
\addplot [
    domain=1.5:2,
    samples=100,
    color=black,
    line width=2pt,
    dotted
]
{ln(x)};
\addplot[
    color=sea-green,
     mark=*,
     line width=2pt,
    ]
    coordinates {
    (0.1, {ln(0.1)})(0.25, {ln(0.35)})
    };
\addplot[
    color=sea-green,
     mark=*,
     line width=2pt,
    ]
    coordinates {
    (0.25, {ln(0.25)})(0.5, {ln(0.6)})
    };
\addplot[
    color=sea-green,
     mark=*,
     line width=2pt,
    ]
    coordinates {
    (0.5, {ln(0.5)})(0.75, {ln(0.84)})
    };
\addplot[
    color=sea-green,
     mark=*,
     line width=2pt,
    ]
    coordinates {
    (0.75, {ln(0.75)})(1.0, {ln(1.1)})
    };
\addplot[
    color=sea-green,
    mark=*,
     line width=2pt,
    ]
    coordinates {
    (1.0, {ln(1.0)})(1.25, {ln(1.35)})
    };

\legend{truth $q(t)$,,traditional OpInf}
\end{axis}
\end{tikzpicture}}
\caption{traditional operator inference}
\end{subfigure}
\begin{subfigure}[t]{0.48\textwidth}
\resizebox{1.0\columnwidth}{!}{\begin{tikzpicture}
\begin{axis}[
    height = 7cm,
    width = 10cm,
    axis lines = left,
    xlabel = {time $t$},
    tick style={draw=none},
    yticklabels={,,},
    xticklabels={,,},
    legend pos=south east,
    legend cell align={left}
]
\addplot [
    domain=0.1:1.5,
    samples=100,
    color=black,
    line width=2pt,
]
{ln(x)};
\addplot [
    domain=1.5:2,
    samples=100,
    color=black,
    line width=2pt,
    dotted
]
{ln(x)};
\addplot[
    color=dark-violet,
    mark=*,
    line width=2pt
    ]
    coordinates {
    (0.1, {ln(0.1)})(0.25, {ln(0.35)})(0.5, {ln(0.75)})
    };
\addplot[
     color=dark-violet,
    mark=*,
    line width=2pt
    ]
    coordinates {
    (0.25, {ln(0.25)})(0.5, {ln(0.60)})(0.75, {ln(1.0)})
    };
\addplot[
    color=dark-violet,
    mark=*,
    line width=2pt
    ]
    coordinates {
    (0.5, {ln(0.5)})(0.75, {ln(0.90)})(1.0, {ln(1.25)})
    };
\addplot[
     color=dark-violet,
    mark=*,
    line width=2pt
    ]
    coordinates {
    (0.75, {ln(0.75)})(1.0, {ln(1.10)})(1.25, {ln(1.35)})
    };

\legend{truth $q(t)$,,OpInf + roll outs}
\end{axis}
\end{tikzpicture}}
\caption{operator inference with roll outs of length two}
\end{subfigure}
\caption{The loss of operator inference with roll outs penalizes model predictions of more than one time step into the future, which increases robustness against instabilities, noise, and perturbations in our numerical experiments. In contrast, traditional operator inference minimizes a loss that takes into account predictions of only one time step into the future and thus cannot penalize errors in long-term predictions of the learned models.}
\label{fig:OpInfVsRollOuts}
\end{figure}
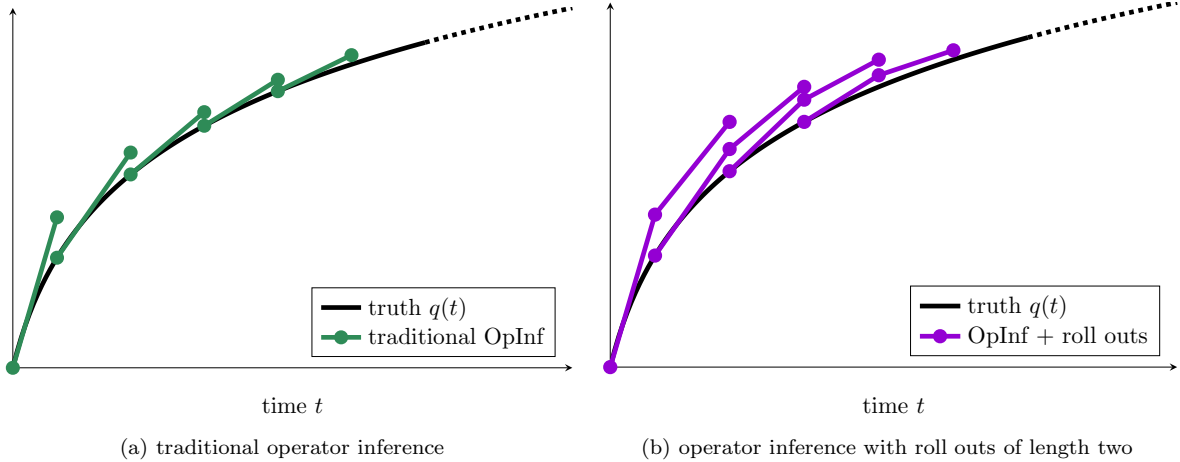

We to roll out the learned model during the training process so that the error of the trajectory predicted over multiple time steps into the future by the learned model is penalized rather than the residual at the current time step alone; cf.~Figure~\ref{fig:OpInfVsRollOuts}. Note that the concept of roll outs are extensively used in Neural ODEs~\cite{NEURIPS2018_69386f6b}. Roll outs provide an inherent stability prior because models that are unstable during the training roll outs get strongly penalized. Similarly, our numerical experiments will demonstrate that the new formulation of operator inference is more robust against perturbations in the data because the whole predicted trajectory is considered in the training loss rather than only a single time step. The drawback is that the training problem becomes more challenging; however, we leverage the recent advances in automatic differentiation to efficiently compute the gradients of the learned model.

We formulate operator inference with rollouts in Section~\ref{sec:DOpfInf:General} and Section~\ref{sec:OpInfRollOutMPC} and discuss polynomial models in the context of differentiable operator inference in Section~\ref{sec:DiffOpInf:ParamPoly}. The stability bias introduced by the roll outs is analyzed in Section~\ref{sec:OIRoll:StabBias}. The computational procedure is discussed in Section~\ref{sec:DOpfInf:CompAsp}.

\subsection{General setup of differentiable non-intrusive model reduction}
\label{sec:DOpfInf:General}
Consider a function $\hbff: \mathbb{R}^{\nr} \times \Ucal \times \Dcal \times \Theta \to \mathbb{R}^{\nr}$
that depends on a parameter $\bftheta \in \Theta \subset \mathbb{R}^{p}$, which we want to learn so that $\hbff$ describes the dynamics of the observed data,
\begin{equation}
\frac{\mathrm d}{\mathrm dt} \hbfq(t; \bfmu) = \hbff(\hbfq(t; \bfmu), \bfu(t), \bfmu; \bftheta)\,.
\label{eq:ModelToFit}
\end{equation}
Notice that this setup follows, e.g., Neural ODEs where the function $\hbff$ is given by a deep neural network \cite{NEURIPS2018_69386f6b}; see also \cite{doi:10.1098/rspa.2021.0162,https://doi.org/10.48550/arxiv.2202.12373}. Based on model \eqref{eq:ModelToFit}, define the discrete flow map
\[
\hbfq_{k}(\bfmu) = \hPhi_{\delta t}(\hbfq_{k-1}(\bfmu), \bfu_{k - 1}(\bfmu), \bfmu; \bftheta)\,,\qquad k = 1, \dots, K
\]
with time-step size $\delta t > 0$, which depends on a time-discretization scheme. Explicit and implicit discretization schemes can be used and the discrete flow map can depend on a history of past time steps rather than just the previous one.
The discrete flow map $\hPhi_{\delta t}$ depends on the input $\bftheta$ through the function $\hbff$.

Consider now a roll-out length $R \in \mathbb{N}$ and the objective function $J: \Theta \times \Dcal \to \mathbb{R}$ defined as
\begin{equation}
J(\bftheta; \bfmu) = \sum_{k = 0}^{K-R}\sum_{r = 1}^R \left\| \bar{\bfq}_{k + r}(\bfmu) - \hbfz_{k,r}(\bfmu)\right\|^2\,,
\label{eq:DiffOpInfObj}
\end{equation}
where
\[
\hbfz_{k,r}(\bfmu) = \hat{\Phi}_{\delta t}(\hbfz_{k,r - 1}(\bfmu), \bfu_{k + r - 1}(\bfmu), \bfmu; \bftheta)\,,\quad \hbfz_{k,0}(\bfmu) = \bar{\bfq}_{k}(\bfmu)
\]
with $\bar{\bfq}_k(\bfmu)$ the projected observed state at time step $k$ with input $\bfmu$; cf.~Eq.~\eqref{eq:ProjData}.
Instead of penalizing the misfit between data and model predictions at all $R$ time steps as in \eqref{eq:DiffOpInfObj}, we can also instead penalize the misfit at $\hat{R}$ time increments $r_1,\dots,r_{\hat{R}}$ such that $r_1 \leq \dots \leq r_{\hat{R}} \leq R$. The corresponding objective is
    \begin{equation}\label{eq:ROpInf:ObjHatR}
J_{\hat{R}}(\bftheta; \bfmu) = \sum_{k = 0}^{K -R}\sum_{\ell = 1}^{\hat{R}} \left\| \bar{\bfq}_{k + r_{\ell}}(\bfmu) - \hbfz_{k,r_{\ell}}(\bfmu)\right\|^2\,.
\end{equation}

The training problem of operator inference with roll outs is the optimization problem
\begin{equation}
\min_{\theta \in \Theta} J(\bftheta;\bfmu)\,,
\label{eq:DiffOpInfOptiProblem}
\end{equation}
which is solved for all training inputs $\bfmu_1, \dots, \bfmu_M$ to obtain the model parameters $\bftheta(\bfmu_1), \dots, \bftheta(\bfmu_M)$.
The objective in \eqref{eq:DiffOpInfOptiProblem} includes the flow map of the to-be-trained model and thus the objective takes into account the predictions of the to-be-trained model, in contrast to the traditional operator inference problem \eqref{eq:RegOpInf}.

For each training input $\bfmu_i \in \Dcal_{\text{train}}$, one obtains a low-dimensional model \eqref{eq:ModelToFit} that one can integrate with new control $\bfu(t) \in \mathbb{R}^p$ and new initial condition $\hbfq_0 \in \mathbb{R}^{\nr}$. To compute a low-dimensional model at a new input $\bfmu \in \Dcal \setminus \Dcal_{\text{train}}$, the parameters $\bftheta(\bfmu_1), \dots, \bftheta(\bfmu_M)$ are interpolated at $\bfmu$; note that this requires the model \eqref{eq:ModelToFit} to be chosen such that interpolating the model parameters indeed leads to models that describe well the dynamics at the new inputs; which we will discuss in the context of polynomial models in detail in Section~\ref{sec:DiffOpInf:ParamPoly}.

\subsection{Model predictive control and operator inference with roll outs}\label{sec:OpInfRollOutMPC}
We formulate operator inference with rollouts as  \eqref{eq:DiffOpInfOptiProblem}, which is a classical optimization problem that follows the concept of unrolling of Neural ODEs~\cite{NEURIPS2018_69386f6b}; however, the formulation with objective \eqref{eq:DiffOpInfObj} is equivalent to solving the following  constrained optimization problem
\begin{align}
    \min_{\theta\in\Theta} & ~~~\sum_{k=0}^{K-R} \sum_{r=1}^{R}
    \left\lVert \bar{\bfq}_{k+r}(\bfmu)-\hat{\bfz}_{k,r}\right\rVert^2 \\
    \text{such that} &
    ~~~\hat{\bfz}_{k,r}=\hat{\Phi}_{\delta t}(\hat{\bfz}_{k,r-1},u_{k+r-1}(\bfmu),\bfmu;\bftheta)),\\
    & ~~~~\hat{\bfz}_{k,0}=\bar{\bfq}_k(\bfmu)\,.
\end{align}
Such constrained optimization problems are well studied in the context of parameter calibration and optimal control \cite{rawlings2017model}, offering alternative solution strategies using Lagrange formulations. In particular, Lagrange formulations allow efficient realizations in software frameworks without automatic differentiation capabilities, with the corresponding adjoint formulations being straightforward to implement\footnote{Solving the adjoint formulation means solving a linear ODE backwards in time to determine the Lagrange multipliers.}. We also want to highlight the connection to so called solver-in-the-loop approaches, which currently gain attention in the context of enriching solvers of PDEs with two- and three-dimensional spatial domains with machine-learning models \cite{um2020solver,kochkov2021machine}, mostly leveraging the JAX library \cite{schoenholz2020jax}.

\subsection{Parametrizations via polynomials in model reduction}
\label{sec:DiffOpInf:ParamPoly}
A wide range of parametrizations can be used for formulating the function $\hbff$ in \eqref{eq:ModelToFit} such as deep networks, as has been done in \cite{NEURIPS2018_69386f6b,doi:10.1098/rspa.2021.0162}. However, in this work, we consider $\hbff$ to be a polynomial in the state and control, to mimic operator inference\footnote{In general, approximating nonlinear behavior with truncated Taylor expansion is wide engineering practice.}. Thus, in the following, the function $\hbff$ is given by
\[
\hbff(\hbfq, \bfu, \bfmu; \bftheta) = \sum_{\ell = 1}^{L}\bfA_{\ell}(\bfmu)\hbfq^{\ell} + \bfB(\bfmu)\bfu\,,
\]
where we have a linear dependence on the control $\bfu$ and thus
\[
\bftheta(\bfmu) = [\bfA_1(\bfmu), \dots, \bfA_{L}(\bfmu), \bfB(\bfmu)]\,.
\]

The use of polynomial functions has several advantages.
First, interpolating an operator $\bfA_i(\bfmu)$ between the parameters $\bftheta(\bfmu_1), \dots, \bftheta(\bfmu_M)$ at a new input $\bfmu \in \Dcal \setminus \Dcal_{\text{train}}$ means to interpolate among the operators $\{\bfA_i(\bfmu_1), \dots, \bfA_i(\bfmu_M)\}$ for each $i = 1, \dots, \ell$, which often leads to an interpolated parameter $\bftheta(\bfmu)$ that indeed leads to a predictive model of the process corresponding to $\bfmu$; this has been successfully leveraged in model reduction \cite{NME:NME2681,panzer_parametric_2010,degroote_interpolation_2010,amsallem_online_2011} and operator inference \cite{paper:PeherstorferW2016} for a long time.

Second, in case of polynomial models, the model parameter $\bftheta(\bfmu) = [\bfA_1(\bfmu), \dots, \bfA_{\ell}(\bfmu), \bfB(\bfmu)]$ is interpretable. For example, the operators provide insights into the stability radius of the model; cf.~\cite{genesio1989stability,411100,K2020_stability_domains_QBROMs,kaptanoglu2021promoting}, which we will also demonstrate in the numerical experiments.

Third, the number of degrees of freedom of a polynomial model, i.e., the dimension of $\bftheta(\bfmu)$ is low, as long as the degree $L$ of the polynomial stays low. As shown theoretically and numerically in \cite{5991229,QIAN2020132401}, low-degree polynomials are sufficient for describing the dynamics of a wide range of applications of interest in science and engineering. Additionally, it is wide engineering practice to approximate nonlinear behavior with low-degree polynomials derived from truncated Taylor expansions.

Fourth, polynomial models explicitly split the dynamics into linear and nonlinear dynamics via the matrices $\bfA_1$ and $\bfA_i, i > 1$. This is helpful when applying semi-implicit time integration schemes that are explicit in the nonlinear dynamics and implicit in the linear ones. A decomposition into linear and nonlinear dynamics typically has to be explicitly enforced in other, more generic nonlinear parametrizations.

\subsection{Stability of polynomial models obtained with operator inference with roll outs}\label{sec:OIRoll:StabBias}
To give insights into the stability bias imposed on the polynomial models by the roll outs, it is informative to look at the objective \eqref{eq:DiffOpInfObj}. For making the following arguments, it is sufficient  consider an autonomous linear model of the form $\hat{\bff}(\hbfq, u, \bfmu; \theta) = \bfA_1\hbfq$, where we assume that $\bfA_1$ is symmetric. We use a forward Euler time discretization scheme with time-step size $\delta t > 0$. The corresponding loss with $\bftheta = [\bfA_1]$ is
\begin{equation}\label{eq:ObjAutoLinModel}
J(\bftheta) = \sum_{k = 0}^{K -R}\sum_{r = 1}^R \left\|\hbfq_{k + r} - \left(\bfI + \delta t \bfA_1\right)^r\hbfq_k\right\|_2^2\,.
\end{equation}
The loss \eqref{eq:ObjAutoLinModel} shows that unstable $\bfA_1$---in the sense that $\bfI + \delta t \bfA_1$ has eigenvalues with positive real parts---are penalized because of their unbounded grow with increasing roll-out length $R$. Similar observations can be made for other parametrizations than autonomous linear polynomial models, with an empirical demonstration provided by our numerical experiments in the next section.

\subsection{Computational aspects}
\label{sec:DOpfInf:CompAsp}
The objective \eqref{eq:DiffOpInfObj} in the training problem is nonlinear in the optimization variable $\bftheta$, which is in stark contrast to the linear objective of the traditional, static operator inference problem \eqref{eq:RegOpInf}. In the following, we use derivative-based optimization methods to numerically solve \eqref{eq:DiffOpInfOptiProblem} and so to train the low-dimensional models. Thus, we have to differentiate through the flow map $\Phi_{\delta t}$ with respect to the parameter $\bftheta(\bfmu)$. Gradients and also higher-order derivative information can be derived explicitly for polynomial models in many cases; see \cite{HartmannFailerOpInf} for a derivation of the gradients in the context of operator inference. However, we instead opt to use the widely available automatic differentiation libraries  such as \cite{47008,jax2018github} for the computations in this work.

The training costs of differentiable operator inference can be higher than in traditional, static operator inference. First, the number of iterations required by derivative-based optimization methods such as gradient descent and Broyden-Fletcher-Goldfarb-Shanno to reduce the gradient norm below a threshold is a priori unknown in many cases. Second, at each objective evaluation, the to-be-learned model is integrated in time, which is computationally more expensive than computing the residual after one time step as in traditional operator inference. However, as we will empirically demonstrate with the numerical experiments in the following section, by investing higher training costs, one can obtain models with higher robustness against noise and perturbations; even from fewer training data, which is well in line with other works such as \cite{rahman2022neural}.

\section{Numerical experiments}\label{sec:NumExp}
The numerical examples in this section demonstrate the following key points:
\begin{itemize}
\item Operator inference with roll outs compensates for scarce training data by penalizing the misfit of model predictions at times far out in the future, resulting oftentimes in more stable and accurate models than traditional operator inference that solves a possibly underdetermined least squares problem in the scarce data regime that leads to unreliable predictions.

\item Increasing the roll-out length can improve the accuracy of the predictions of the learned models and help to mitigate the effect of noise by preventing overfitting when learning models from noisy data.

\item If trajectories are sparsely sampled in time, then roll outs can provide meaningful values between sampled states and so lead to more accurate models, whereas traditional operator inference is required to take coarser approximations of the time derivatives.

\item Computation of the stability radius of low-dimensional polynomially nonlinear models learned via operator inference with and without roll outs offer interpretability beyond mere numerical experiments on test data. The results show that roll outs help to increase upper bounds of stability radii, which indicates that roll outs lead to models that are more robust against disturbances and are applicable to a wider range of initial conditions.
\end{itemize}

Section~\ref{subsec:Implement} discusses the numerical implementation of the proposed approach as well as strategies for training the low-dimensional models with operator inference with roll outs. Numerical experiments are conducted with data sampled from models that describe shallow water waves in Section~\ref{subsec:shallow} and geophysical fluids in Section~\ref{subsec:pyqg}. The hyperparameters of the training procedure mentioned above are tuned using validation data. We chose the hyperparameter configuration that achieved the smallest error on the validation data in all of the learned models to ensure a fair comparison.

\subsection{Numerical implementation} \label{subsec:Implement}
We implemented operator inference with roll outs in JAX\footnote{\href{https://jax.readthedocs.io/en/latest/}{https://jax.readthedocs.io/en/latest/}} \cite{jax2018github} to leverage automatic differentiation when solving the training problem \eqref{eq:DiffOpInfOptiProblem}.
The Adam, adaptive moment estimation, optimizer \cite{ADAMOptimizer} is used to minimize the objective function.
We choose learning rates logarithmically spaced between $10^{-5}$ and $10^{-1}$.
For initial conditions to the non-convex optimization problem \eqref{eq:DiffOpInfOptiProblem}, we either start with zeros or use the model operators from traditional operator inference.
For obtaining models with traditional operator inference, we solve the problem \eqref{eq:RegOpInfMinNorm}, which gives the minimal-norm solution in case of an underdetermined system.

\subsection{Dynamics of shallow water waves}
\label{subsec:shallow}
We consider waves governed by the shallow water equations and demonstrate learning a model with operator inference with roll outs from noisy and scarce data.

\subsubsection{Shallow water equations: Setup}
Consider the two-dimensional spatial domain $\Omega = (-4, 4)^2 \subset \mathbb{R}^2$ and the time domain $\Tcal = (0, T) \subset \mathbb{R}$ with $T = 0.15$. The shallow water equations are given by
\begin{align*}
\partial_t q_h + \nabla \cdot (q_h \nabla q_{\phi}) = & 0\,, & \text{ in } \Omega \times \Tcal\,,\\
\partial_t q_{\phi} + \frac{1}{2} |\nabla q_{\phi}|^2 + q_h = & 0\,, & \text{ in } \Omega \times \Tcal\,.
\end{align*}
The system consists of the scalar potential of the fluid given by $q_{\phi}$ and the height of the free-surface $q_h$, which is normalized by its mean value.

The input $\bfmu = [\mu_1, \mu_2] \in \Dcal = [0.2,0.5] \times [1.1,1.7] $ determines the initial condition
\[
q_{h}^0(\bfx; \bfmu) = 1 + \mu_1 \mathrm e^{-\mu_2\|\bfx\|_2^2}\,,\qquad q_{\phi}^0(\bfx; \bfmu) = 0\,,
\]
where $q_h(0, \bfx; \bfmu) = q_h^0(\bfx; \bfmu)$ and $q_{\phi}(0, \bfx; \bfmu) = q_{\phi}^0(\bfx; \bfmu)$.
We impose periodic boundary conditions. The equations are discretized in space with second-order finite difference schemes on 60 equidistant grid points in each direction so that $N = 7200$.  The time integration scheme used is the implicit midpoint rule for Hamiltonian systems with the time-step size being $\delta t = 0.0004$.

We learn a model of dimension $n=25$ in this example. To obtain the low-dimensional basis $\bfV$, we generate snapshot trajectories using  12 initial conditions with inputs $[\mu_1^{\text{basis}},\mu_2^{\text{basis}}]$ where $\mu_1^{\text{basis}} \in \{0.2, 0.35, 0.5\}$ and $\mu_2^{\text{basis}} \in \{1.1, 1.3, 1.5, 1.7\}$. Training data are obtained by generating trajectories whose initial conditions correspond to three nested sets of training inputs representing increasing levels of information. The training set of inputs is constructed as follows:
\begin{align*}
\bfmu^{\text{train-I}} = & \{(0.225, 1.15),(0.225, 1.65),(0.475, 1.15),(0.475, 1.65)\}\,,\\ \bfmu^{\text{train-II}} = & \bfmu^{\text{train-I}} \cup \{(0.225, 1.4), (0.35, 1.4), (0.475, 1.4)\}\,,\\
\bfmu^{\text{train-III}} = &  \bfmu^{\text{train-II}} \cup \{(0.35, 1.65), (0.35, 1.15)\}\,.
\end{align*}
The hyperparameters are tuned using a validation data set, which contains trajectories whose initial conditions correspond to the validation inputs
\[
\bfmu^{\text{valid}} = \{(0.2875, 1.4),(0.35, 1.525),(0.35, 1.275),(0.4125, 1.4)\}\,.
\]
The test data used to investigate the performance of the learned model is generated using initial conditions with inputs $[\mu_1^{\text{test}},\mu_2^{\text{test}}]$ where $\mu_1^{\text{test}} \in \{0.2875, 0.4125\}$ and $\mu_2^{\text{test}}\in \{1.275, 1.525\}$.

An implicit-explicit scheme is utilized to integrate the learned low-dimensional model over $t\in (0,T)$ with $\delta t = 0.0004$ in which the linear and quadratic terms are evaluated implicitly and explicitly, respectively. The same implicit-explicit scheme with the same time-steps size is also used for the roll outs.
In the experiments below, we chose \eqref{eq:DiffOpInfObj}  as the objective function. In addition to penalizing the misfit between the data and model predictions at all $R$ time points as in \eqref{eq:DiffOpInfObj}, we will also consider the objective \eqref{eq:ROpInf:ObjHatR}. In the latter, we measure the misfit at $\hat{R}=25$ time points that are equidistant between 1 and $R$ in the logarithmic scale.

When training low-dimensional models using operator inference with roll outs, we select the hyperparameter configuration which results in the lowest validation error given by
\begin{align} \label{eq:ValidRelErr}
        \frac{1}{M_{\text{valid}}}\sum_{j=1}^{M_{\text{valid}}} \frac{\sum_{k=0}^K \|\bfq_k (\bfmu^{\text{valid}}_j) - \bfV \hbfq_k(\bfmu^{\text{valid}}_j)\|_F}{\sum_{k=0}^K \|\bfq_k (\bfmu^{\text{valid}}_j) \|_F}\,,
    \end{align}
where $M_{\text{valid}}$ denotes the number of test parameters. The learned low-dimensional operators from this configuration are then used to assemble a model, which is used for predictions at test parameters and test initial conditions.

\subsubsection{Shallow water equations: Scarce data}

\begin{figure}
\begin{subfigure}[t]{0.24\textwidth}
\includegraphics[width=1.0\linewidth]{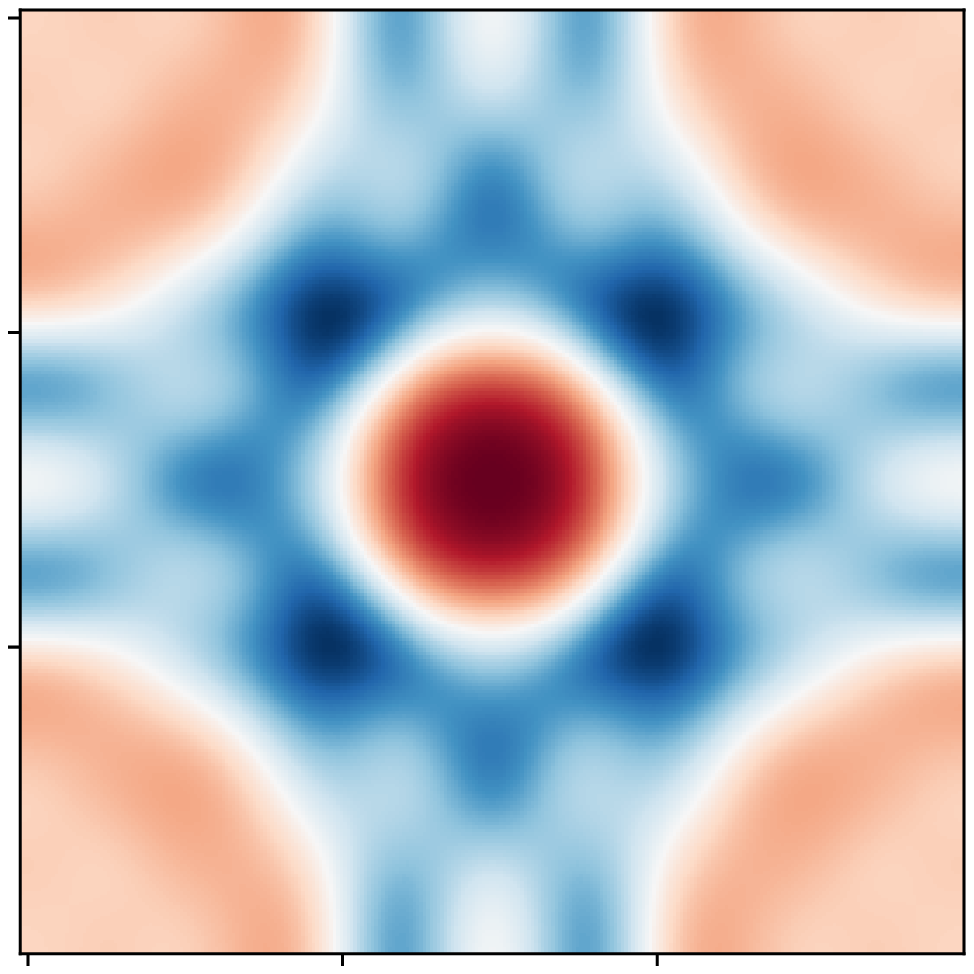} 
\caption{ground truth, $\bfmu^{\text{test}}_1$}
\end{subfigure}
\begin{subfigure}[t]{0.24\textwidth}
\includegraphics[width=1.0\linewidth]{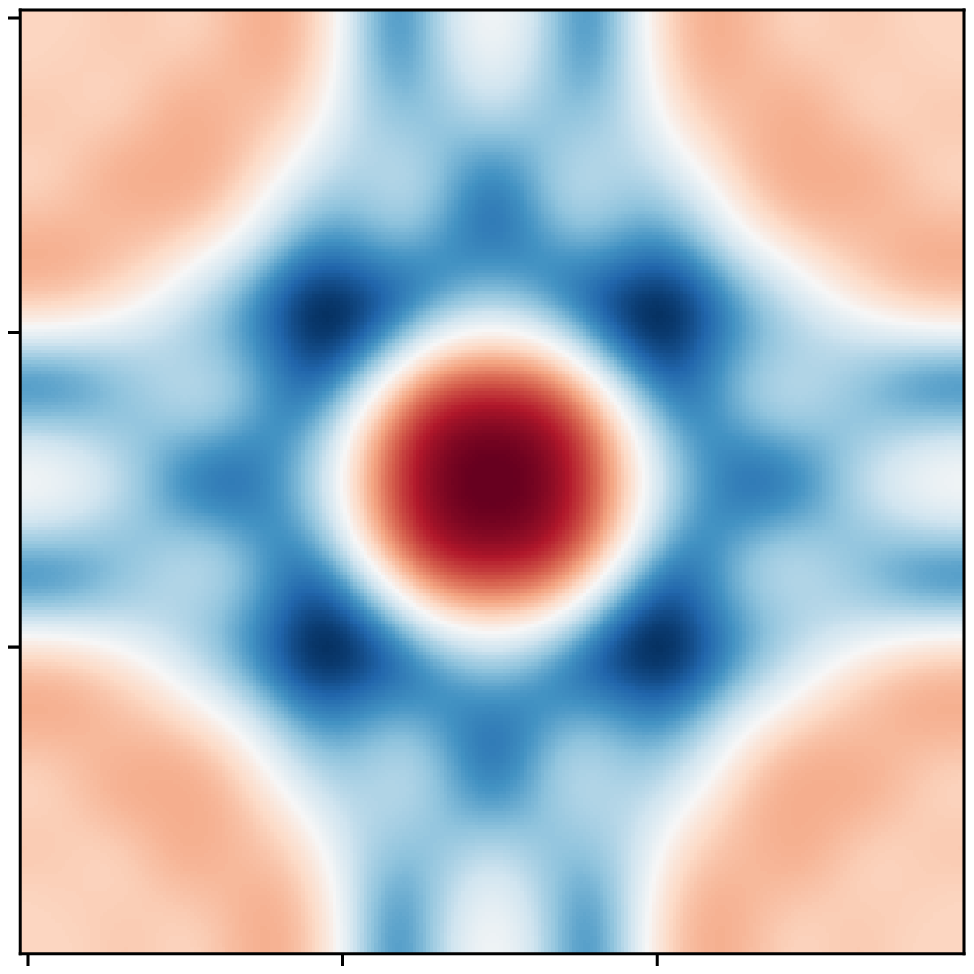} 
\caption{projection, $\bfmu^{\text{test}}_1$}
\end{subfigure}
\begin{subfigure}[t]{0.24\textwidth}
\includegraphics[width=1.0\linewidth]{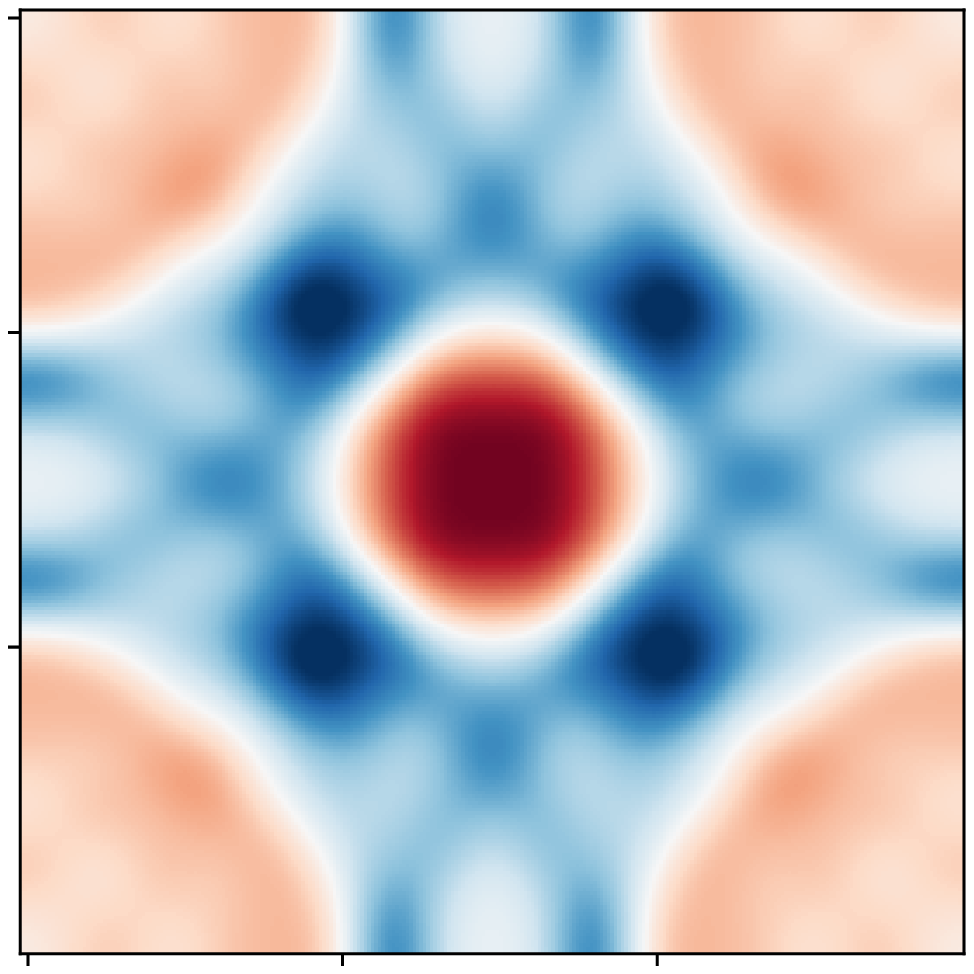} 
\caption{OpInf + roll outs, $\bfmu^{\text{test}}_1$}
\end{subfigure}
\begin{subfigure}[t]{0.24\textwidth}
\includegraphics[width=1.0\linewidth]{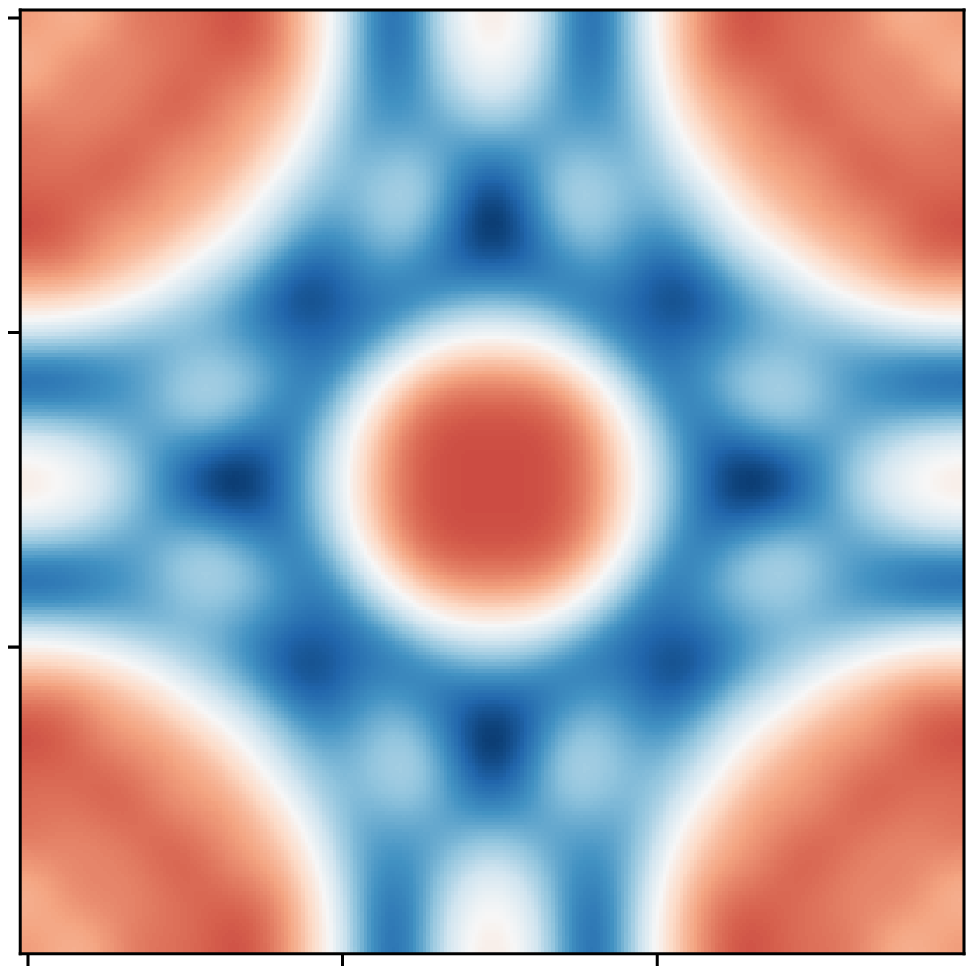} 
\caption{traditional OpInf, $\bfmu^{\text{test}}_1$}
\end{subfigure}
\begin{subfigure}[t]{0.24\textwidth}
\includegraphics[width=1.0\linewidth]{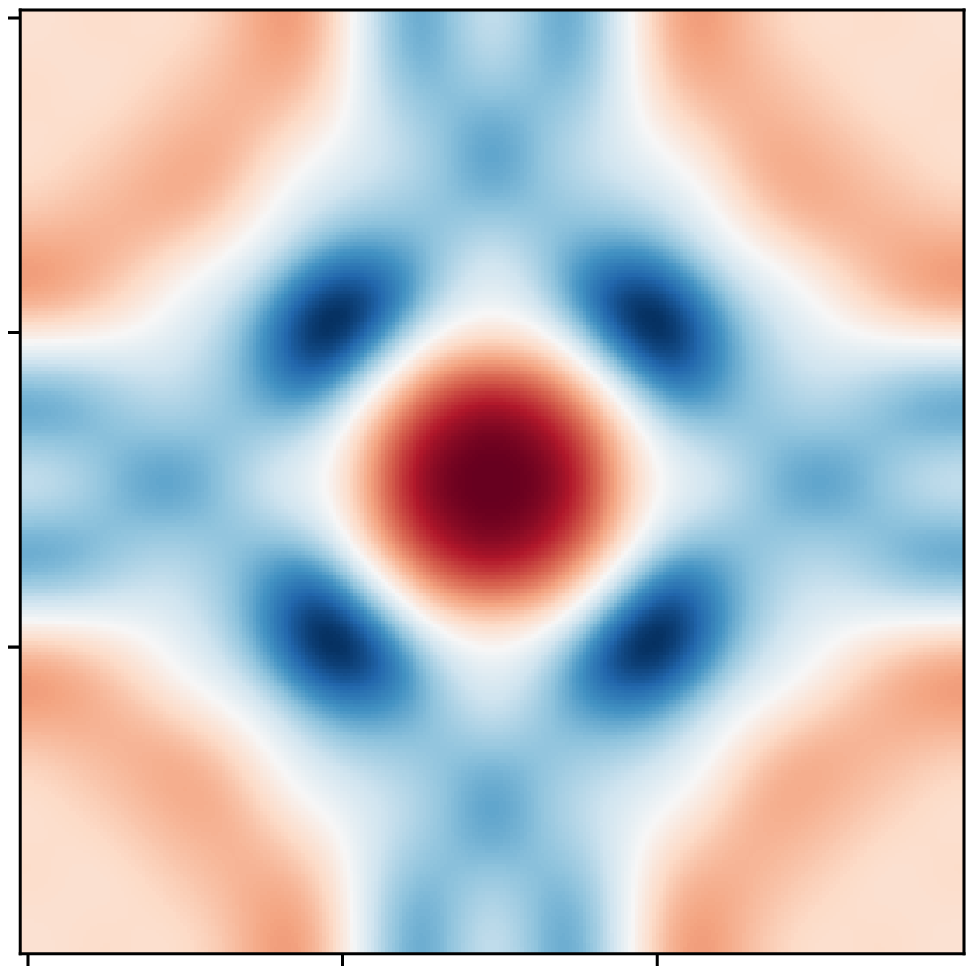} 
\caption{ground truth, $\bfmu^{\text{test}}_2$}
\end{subfigure}
\begin{subfigure}[t]{0.24\textwidth}
\includegraphics[width=1.0\linewidth]{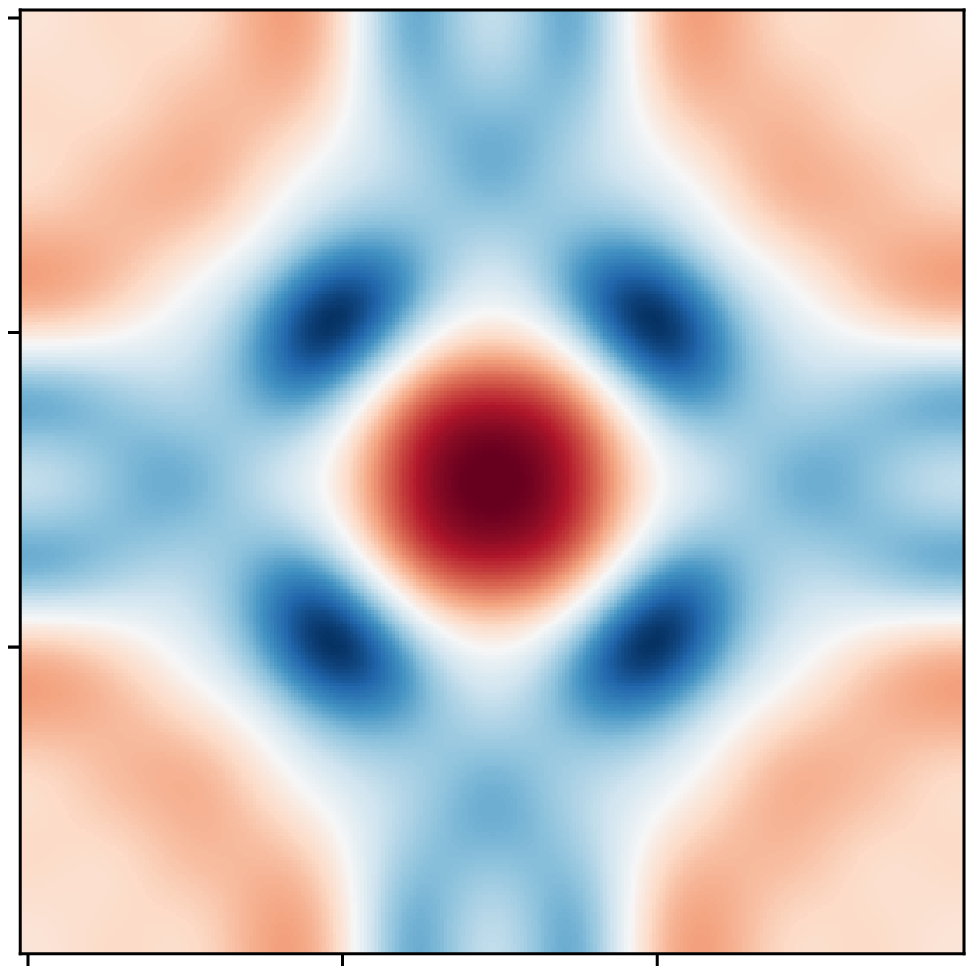} 
\caption{projection, $\bfmu^{\text{test}}_2$}
\end{subfigure}
\begin{subfigure}[t]{0.24\textwidth}
\includegraphics[width=1.0\linewidth]{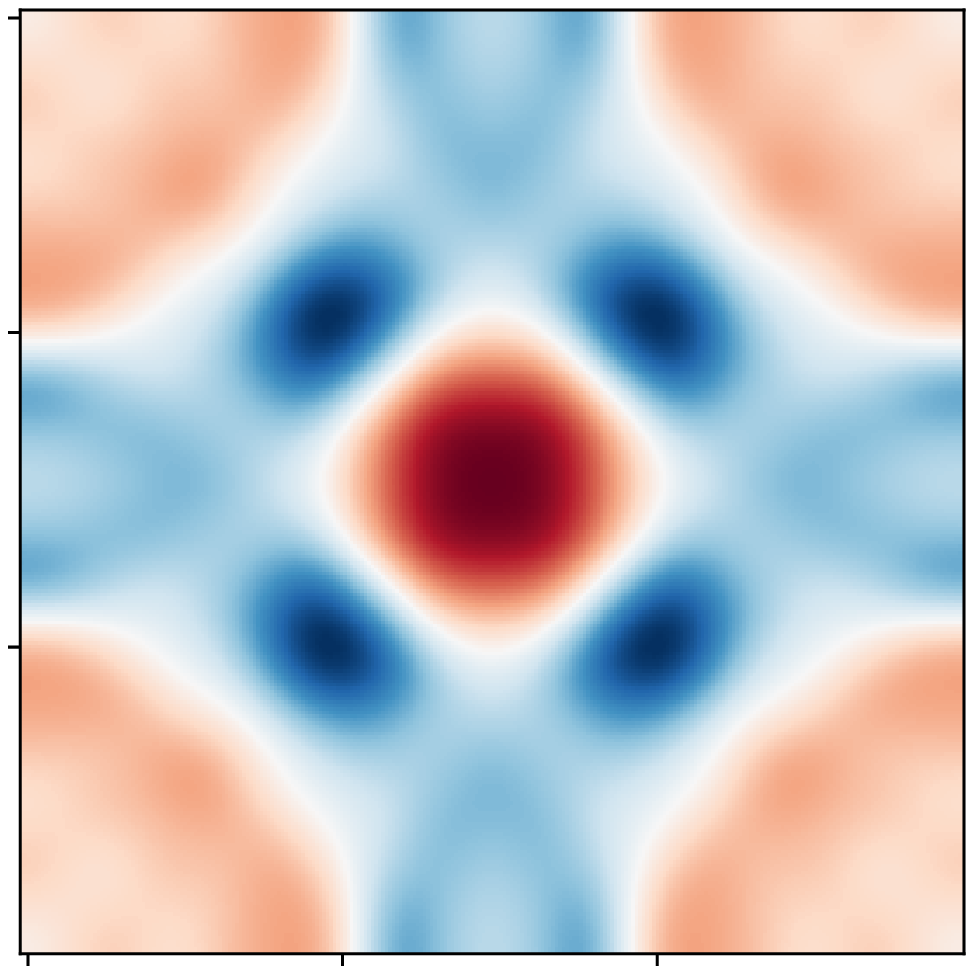} 
\caption{OpInf + roll outs, $\bfmu^{\text{test}}_2$}
\end{subfigure}
\begin{subfigure}[t]{0.24\textwidth}
\includegraphics[width=1.0\linewidth]{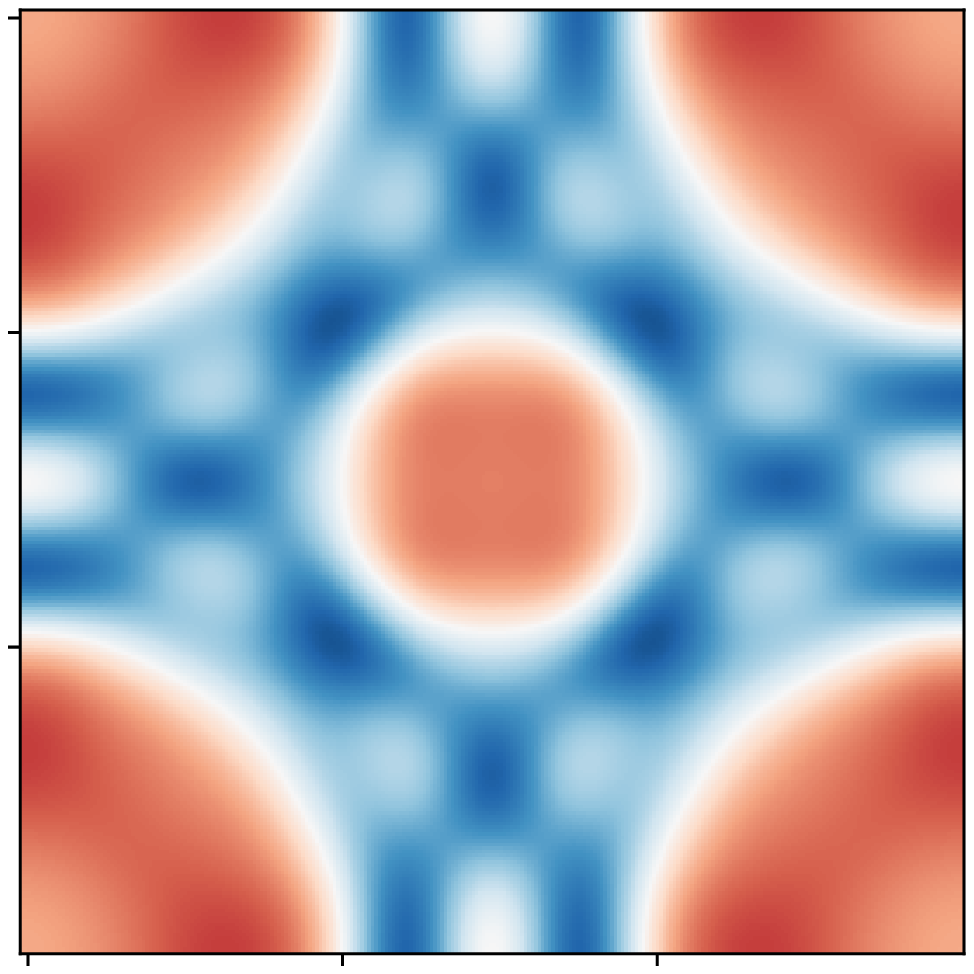} 
\caption{traditional OpInf, $\bfmu^{\text{test}}_2$}
\end{subfigure}
\begin{subfigure}[t]{0.24\textwidth}
\includegraphics[width=1.0\linewidth]{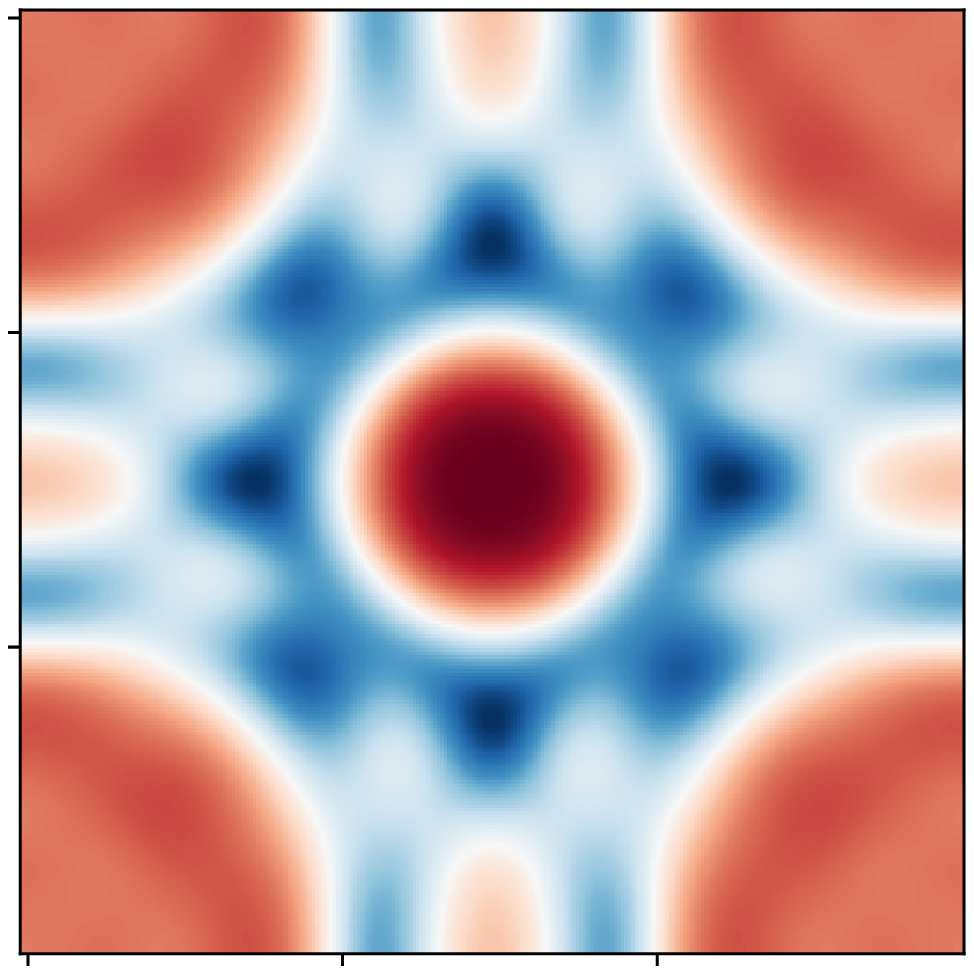} 
\caption{ground truth, $\bfmu^{\text{test}}_3$}
\end{subfigure}
\begin{subfigure}[t]{0.24\textwidth}
\includegraphics[width=1.0\linewidth]{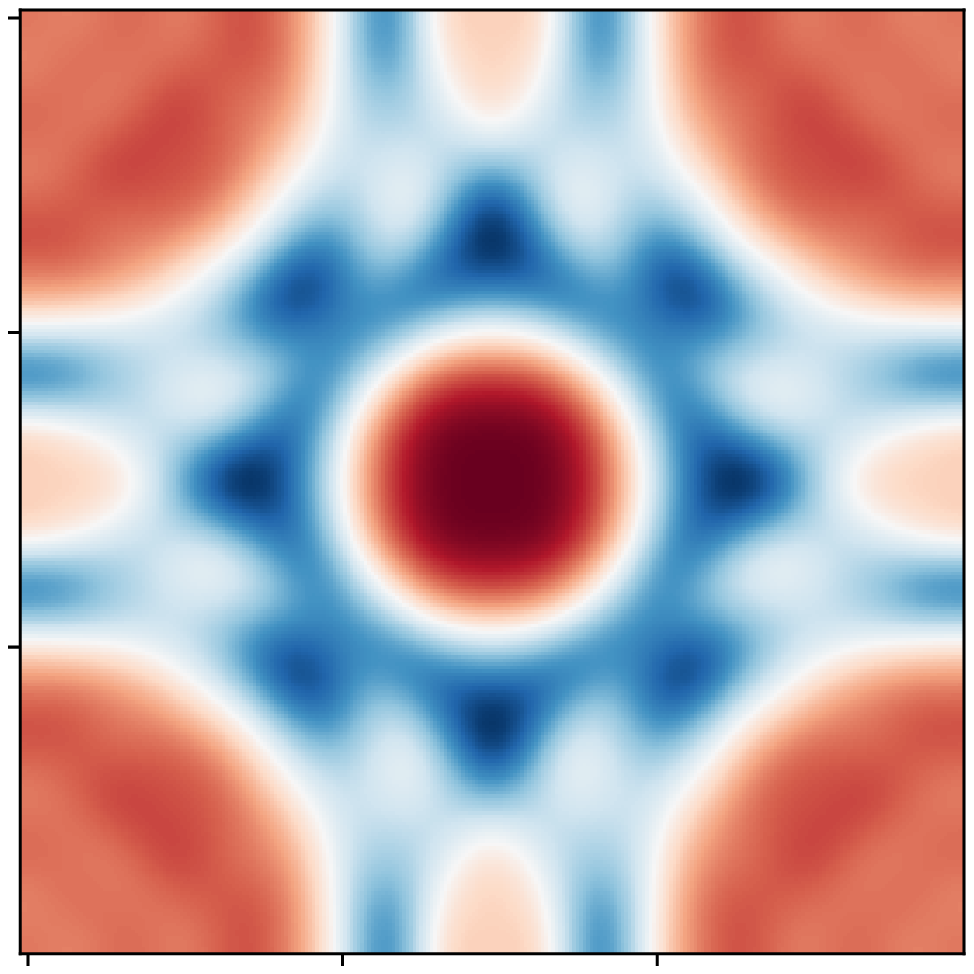} 
\caption{projection, $\bfmu^{\text{test}}_3$}
\end{subfigure}
\begin{subfigure}[t]{0.24\textwidth}
\includegraphics[width=1.0\linewidth]{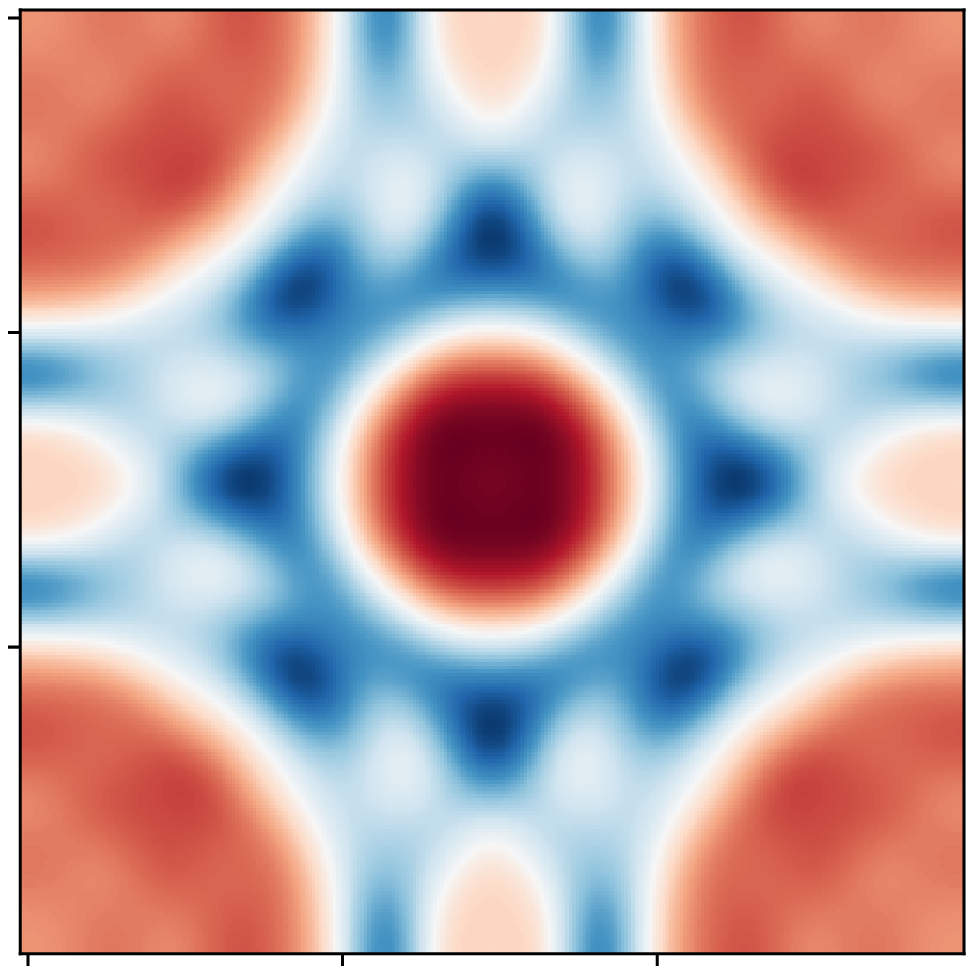} 
\caption{OpInf + roll outs, $\bfmu^{\text{test}}_3$}
\end{subfigure}
\begin{subfigure}[t]{0.24\textwidth}
\includegraphics[width=1.0\linewidth]{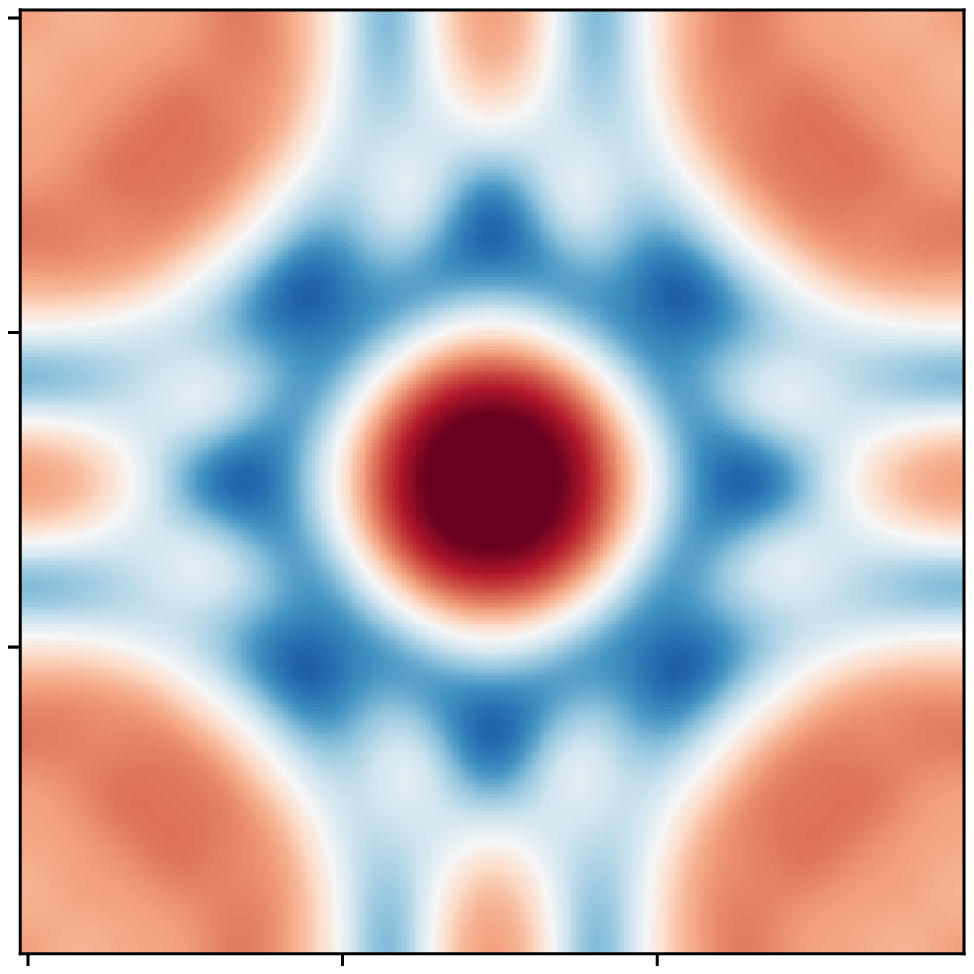} 
\caption{traditional OpInf, $\bfmu^{\text{test}}_3$}
\end{subfigure}
\,
\begin{subfigure}[t]{0.24\textwidth}
\includegraphics[width=1.0\linewidth]{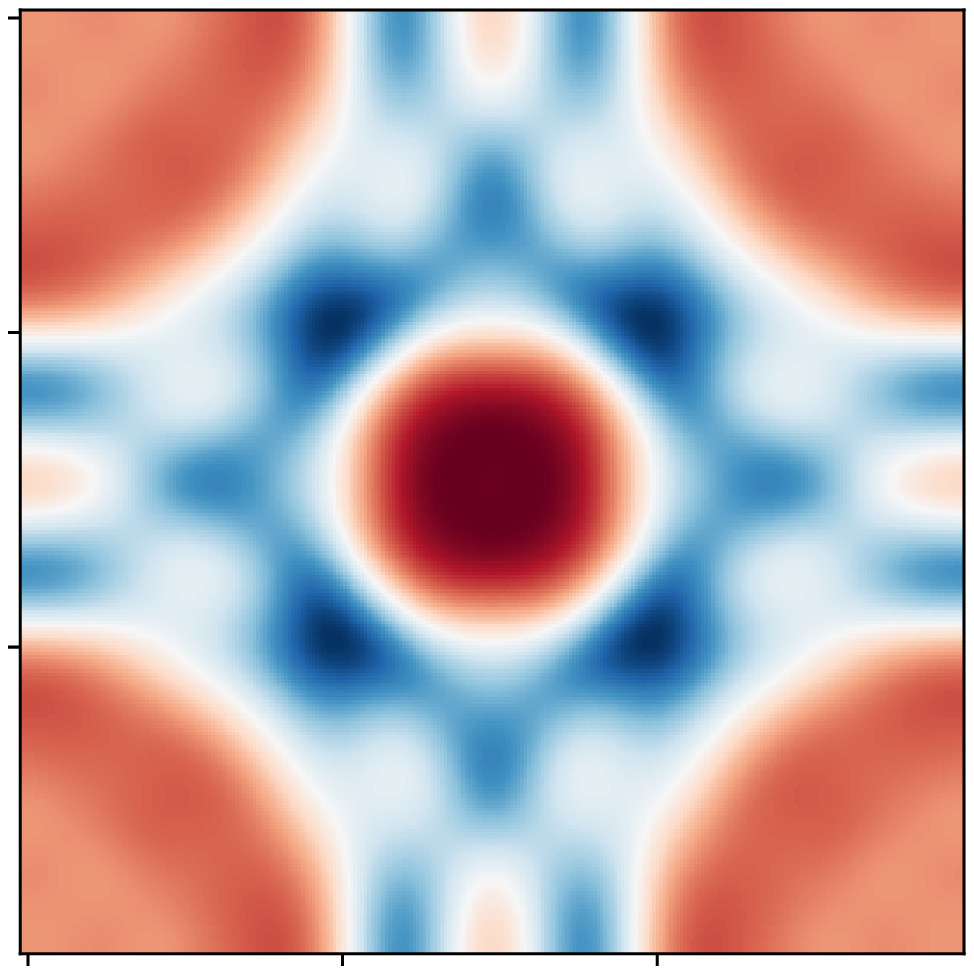} 
\caption{ground truth, $\bfmu^{\text{test}}_4$}
\end{subfigure}
\begin{subfigure}[t]{0.24\textwidth}
\includegraphics[width=1.0\linewidth]{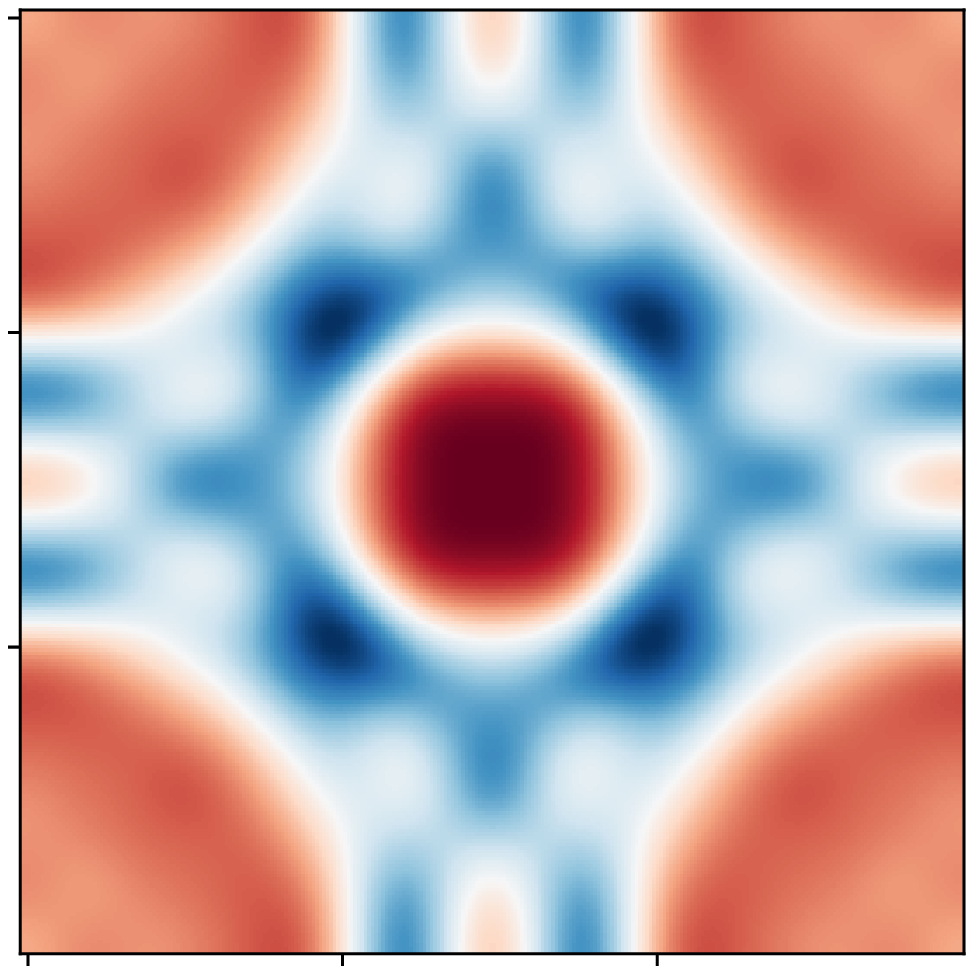} 
\caption{projection, $\bfmu^{\text{test}}_4$}
\end{subfigure} \,
\begin{subfigure}[t]{0.24\textwidth}
\includegraphics[width=1.0\linewidth]{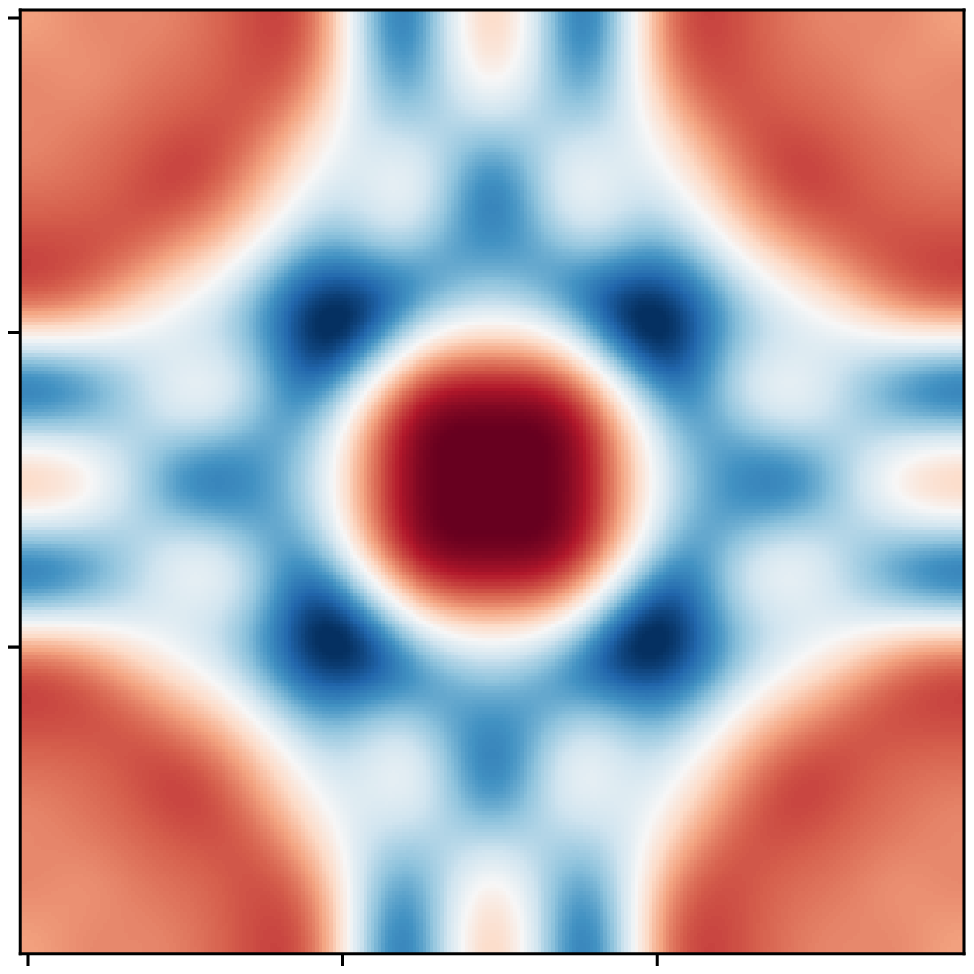} 
\caption{OpInf + roll outs, $\bfmu^{\text{test}}_4$}
\end{subfigure}
\begin{subfigure}[t]{0.24\textwidth}
\includegraphics[width=1.0\linewidth]{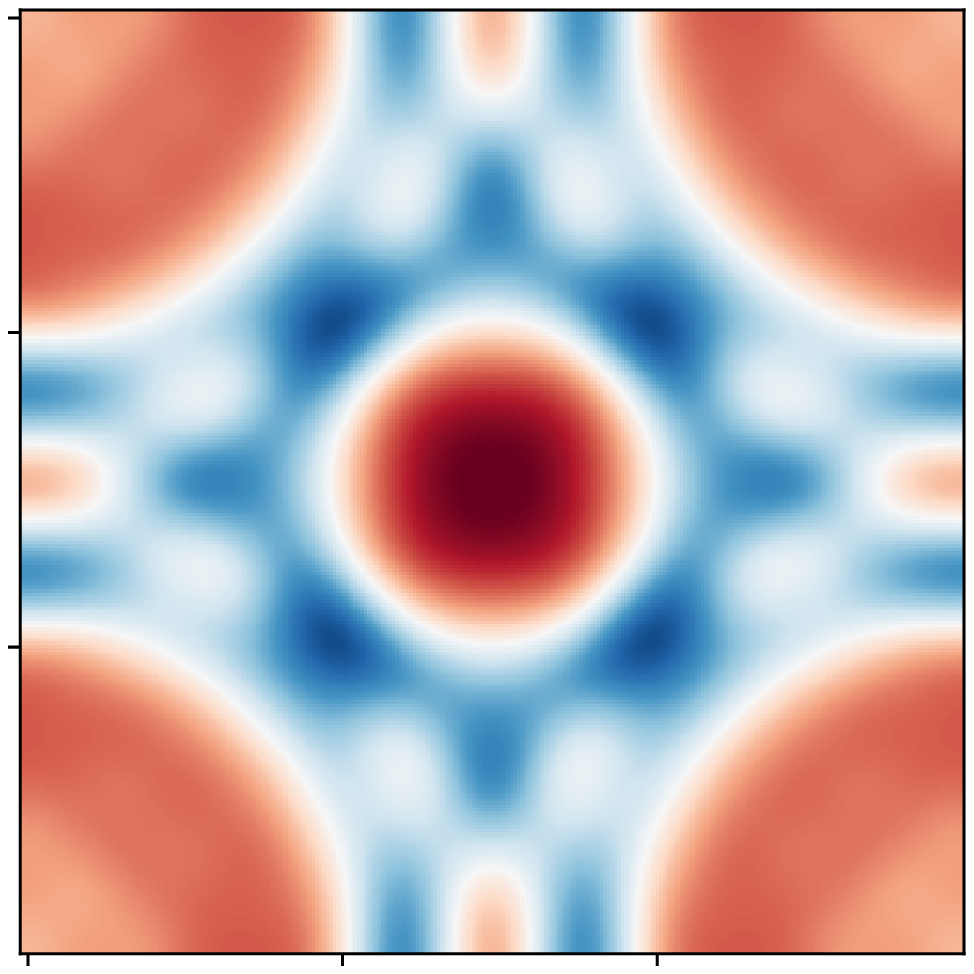} 
\caption{traditional OpInf, $\bfmu^{\text{test}}_4$}
\end{subfigure}
\caption{Shallow water equations (Section~\ref{subsec:shallow}): In this experiment with training data set $\bfmu^{\text{train-I}}$ and roll length $R=100$, operator inference with roll outs leads to a predictive model on the test data set, whereas the model learned with traditional operator inference  provides only inaccurate predictions. The plots corresponding to ``projection'' show the ground-truth data projected onto the reduced space $\Vcal$, which provides the best approximation that any low-dimensional model seeking a linear approximation in $\Vcal$ can achieve. }
\label{fig:shallow_IC}
\end{figure}

We demonstrate the performance of the low-dimensional models learned with traditional operator inference and operator inference with roll outs in the scarce data regime by varying the number of initial conditions utilized to generate state trajectories for training data. First, we only train with trajectories obtained from four initial conditions and we set the roll length to $R=100$. Figure~\ref{fig:shallow_IC} shows ground-truth test data of the free-surface height computed with the truth model, the projection of the ground-truth data onto the low-dimensional subspace $\Vcal$, which gives the best approximation that can be achieved, the predictions of the model with traditional, static operator inference, and the predictions of the model obtained with roll outs at time $t = 0.15$ at the test parameter values. The plots show that the model learned with roll outs captures the complex patterns of the ground truth while static operator inference appears to overfit to certain patterns in this example.

\begin{figure}
\begin{subfigure}[b]{0.48\textwidth}
\begin{center}
{{\Large\resizebox{1\columnwidth}{!}{\input{shallow_errVsnTrainParam}}}}
\end{center}
\caption{roll-out length $R=100$}
\label{fig:shallow_ErrVsnParam}
\end{subfigure}
\begin{subfigure}[b]{0.48\textwidth}
\begin{center}
{{\Large\resizebox{1\columnwidth}{!}{\input{shallow_errVsRollLen}}}}
\end{center}
\caption{four training trajectories}
\label{fig:shallow_rollVsErr}
\end{subfigure}
\caption{Shallow water equations (Section~\ref{subsec:shallow}): The plots show that roll outs during training help operator inference to learn a model that makes more accurate predictions on the test data set than traditional operator inference without roll outs, especially in the scarce data regime with a low number of training trajectories.}
\label{fig:shallow_nParamRollLen}
\end{figure}
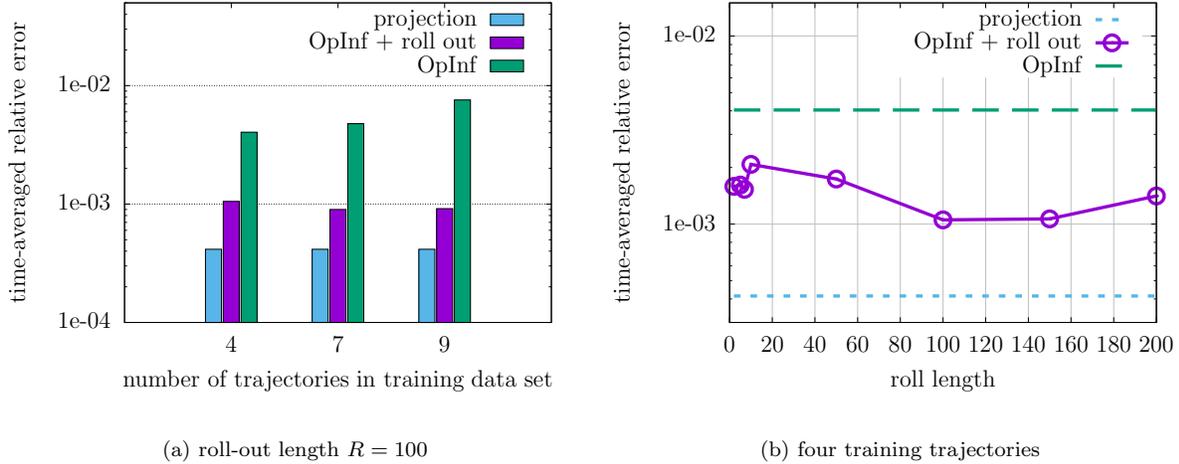

We now investigate how the number of trajectories used for training as well as the roll-out length $R$ influence the performance of operator inference. Figure~\ref{fig:shallow_nParamRollLen} shows the relative error averaged over the test inputs over time of the low-dimensional models as a function of the number of training trajectories $M_{\text{train}}$ and the roll-out length $R$. The plotted error is computed as
\begin{align} \label{eq:TestRelErr}
    \frac{1}{M_{\text{test}}}\sum_{j=1}^{M_{\text{test}}} \frac{\sum_{k=0}^K \|\bfq_k (\bfmu^{\text{test}}_j) - \bfV \hbfq_k(\bfmu^{\text{test}}_j)\|_F}{\sum_{k=0}^K \|\bfq_k (\bfmu^{\text{test}}_j) \|_F}
\end{align}
where $M_{\text{test}}$ denotes the number of test parameters. The plots in Figure~\ref{fig:shallow_nParamRollLen} show that roll outs during training help operator inference to learn models that make more accurate predictions on the test data set than traditional operator inference without roll outs, especially for the scarce data regime with a low number of training trajectories.

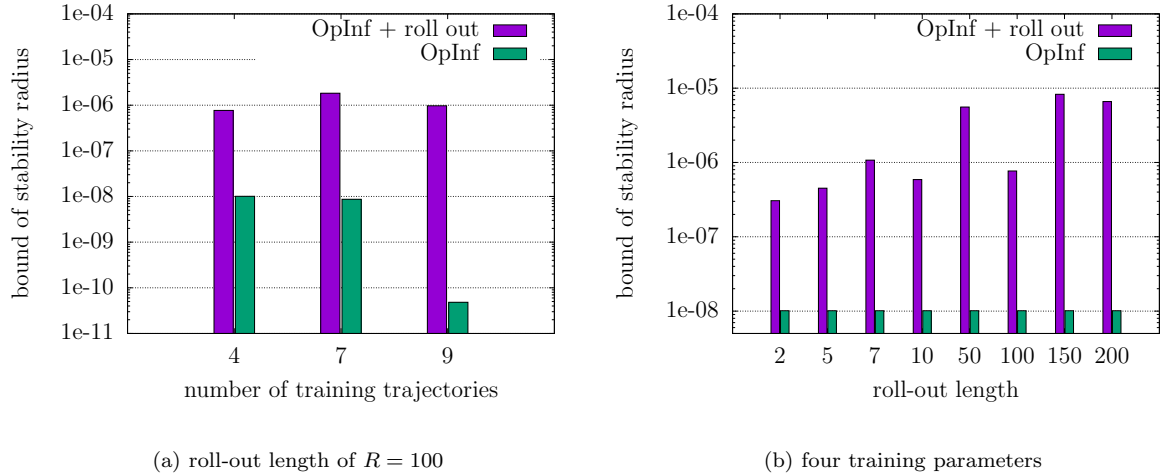
\begin{figure}[t]
\begin{subfigure}[b]{0.48\textwidth}
\begin{center}
{{\Large\resizebox{1\columnwidth}{!}{\input{shallow_radiusnParam}}}}
\end{center}
\caption{roll-out length of $R=100$}
\label{fig:shallow_nParamVsRadius}
\end{subfigure}
\begin{subfigure}[b]{0.48\textwidth}
\begin{center}
{{\Large\resizebox{1\columnwidth}{!}{\input{shallow_radiusRollLen}}}}
\end{center}
\caption{four training parameters}
\label{fig:shallow_RollVsRadius}
\end{subfigure}
\caption{Shallow water equations (Section~\ref{subsec:shallow}): The bounds of the stability radii of models learned with roll outs are orders of magnitude higher than for the models learned with traditional operator inference without roll outs. This suggests that training with roll outs promotes robustness in the learned low-dimensional model especially when training data are scarce.}
\label{fig:SWE:StabilityRadius}
\end{figure}

\begin{figure}[!h]
\begin{subfigure}[b]{0.33\textwidth}
\begin{center}
{{\huge\resizebox{1\columnwidth}{!}{\input{shallow_errVsNoise_nTrain4}}}}
\end{center}
\caption{\#training trajectories $M_{\text{train}}=4$}
\label{fig:shallow_noiseVsErr_nTrain4}
\end{subfigure}
\begin{subfigure}[b]{0.33\textwidth}
\begin{center}
{{\huge\resizebox{1\columnwidth}{!}{\input{shallow_errVsNoise_nTrain7}}}}
\end{center}
\caption{\#training trajectories $M_{\text{train}}=7$}
\label{fig:shallow_noiseVsErr_nTrain7}
\end{subfigure}
\begin{subfigure}[b]{0.33\textwidth}
\begin{center}
{{\huge\resizebox{1\columnwidth}{!}{\input{shallow_errVsNoise_nTrain9}}}}
\end{center}
\caption{\#training trajectories $M_{\text{train}}=9$}
\label{fig:shallow_noiseVsErr_nTrain9}
\end{subfigure}
\caption{Shallow water equations (Section~\ref{subsec:shallow}): Operator inference with roll outs learns models that achieve accurate predictions that are close to the best approximation (projection) in this example even when training data are polluted by 10\% noise. In contrast, traditional operator inference without roll outs fails to make meaningful predictions especially for large noise in the training data.}
\label{fig:shallow_noiseVsErr}
\end{figure}
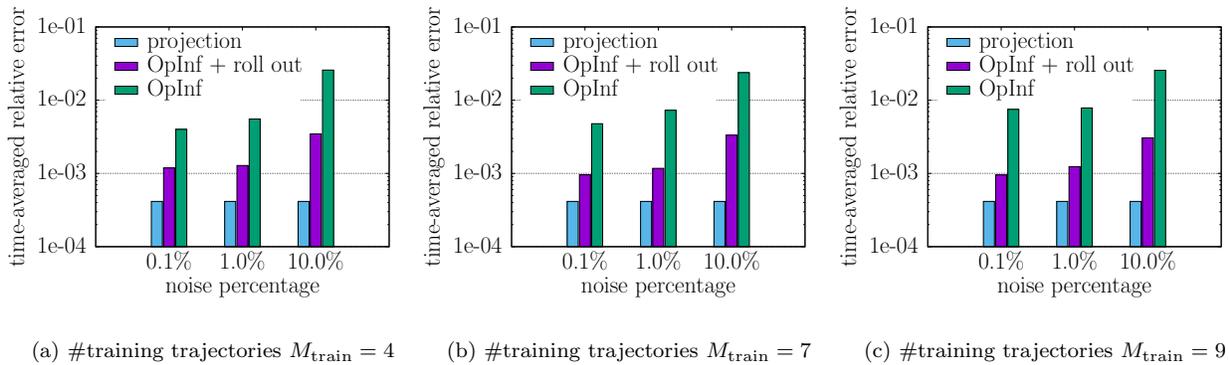

Parametrizing low-dimensional models via quadratic polynomials offers a tractable representation that permits us to study properties of the learned models such as stability. For quadratic models, we compute a bound of the stability radius~\cite{genesio1989stability,411100,kaptanoglu2021promoting} that is derived in \cite{K2020_stability_domains_QBROMs} and given by
\begin{align*}
        \gamma = \frac{\sigma_{\text{min}}(\bfL)}{2 \sqrt{\|\bfP\|_F} \|\bfA_2\|_F}
    \end{align*}
where $\bfP \in \R^{n \times n}$ satisfies the continuous Lyapunov equation $\bfA_1^T \bfP + \bfP \bfA_1 = - \bfL \bfL^T $ for specified $\bfL \in \R^{n \times n}$. We generate 1000 realizations of a random matrix with independent standard normal entries for $\bfL$ and then we compute the average stability radius of the learned quadratic models. The bounds of the stability radii of the learned low-dimensional models are shown in Figure~\ref{fig:SWE:StabilityRadius}. The results indicate that the stability radii of the learned models via operator inference with roll outs are orders of magnitude larger than those learned with traditional operator inference. These results agree with the observation discussed in Section~\ref{sec:ROpInf} that roll outs impose a stability bias. The plots are also consistent with the fact that operator inference with roll outs offers more accurate predictions in this example than models learned with traditional operator inference.

\subsubsection{Shallow water equations: Noisy and scarce data} \label{subsubsec:SWE_scarceNoisy}

\begin{figure}
\begin{subfigure}[t]{0.24\textwidth}
\includegraphics[width=1\linewidth]{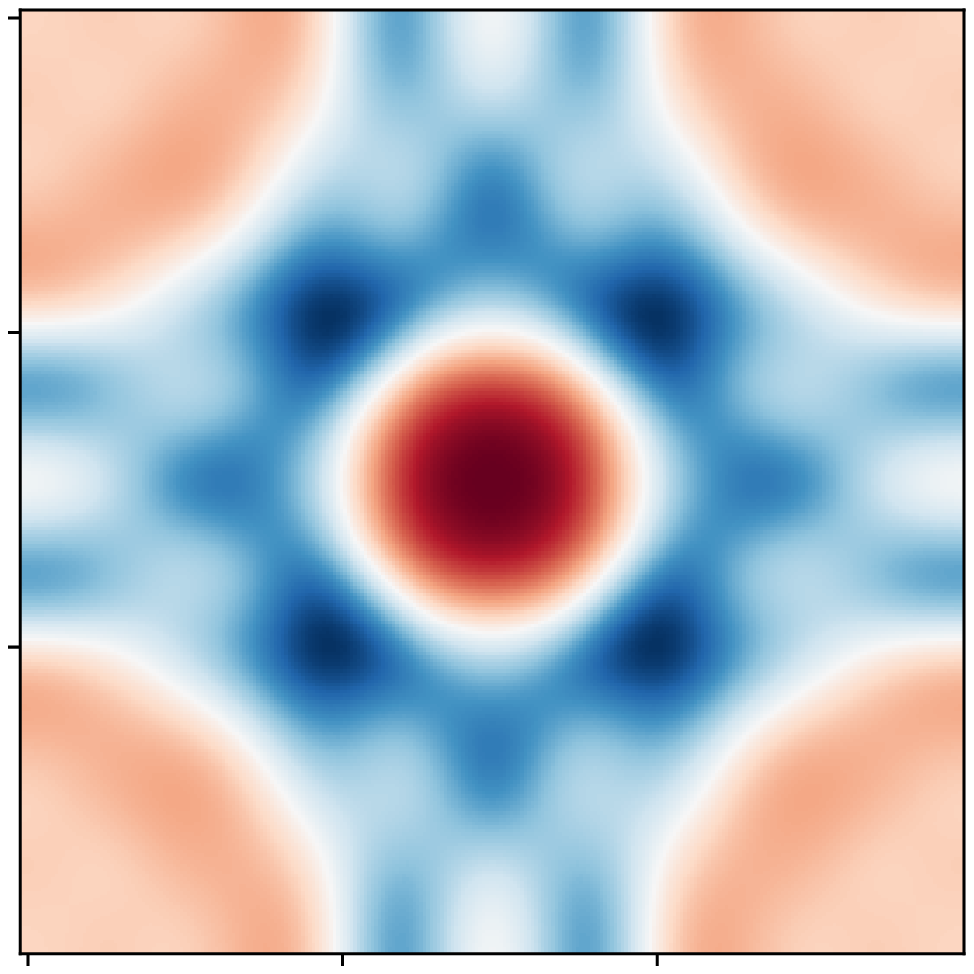} 
\caption{ground truth, $\bfmu^{\text{test}}_1$}
\end{subfigure}
\begin{subfigure}[t]{0.24\textwidth}
\includegraphics[width=1\linewidth]{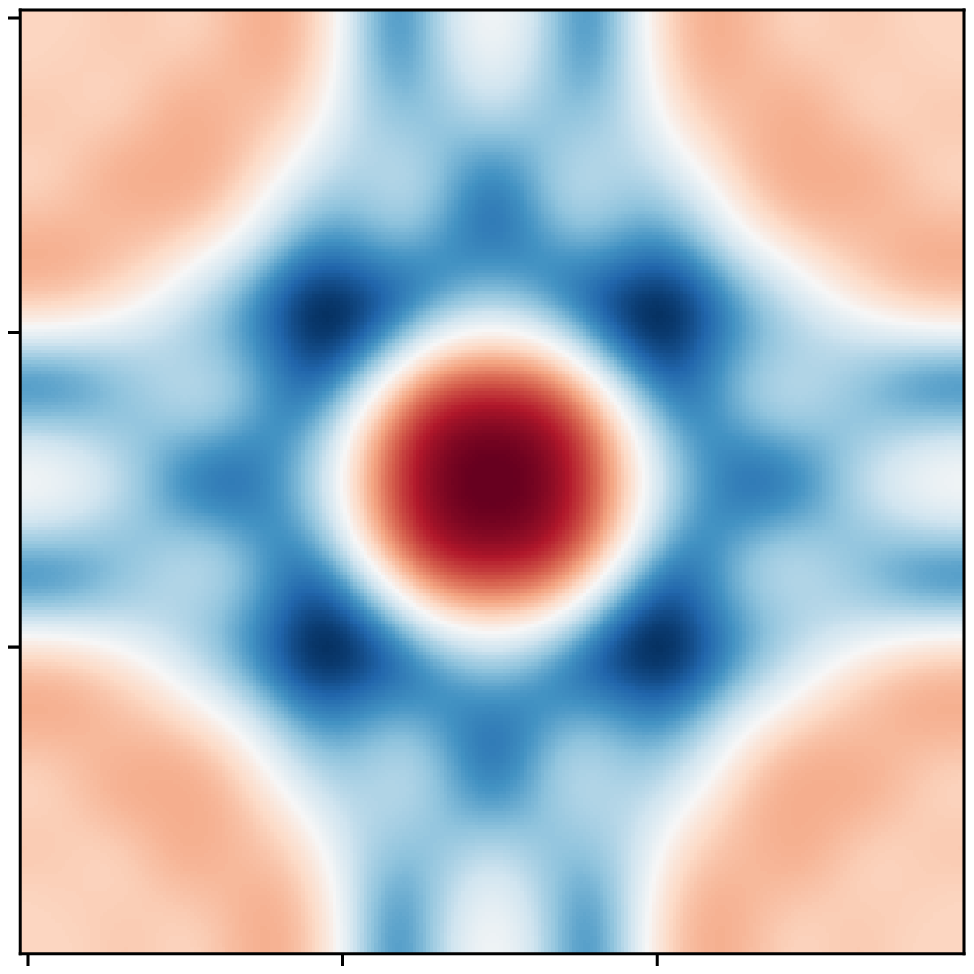} 
\caption{projection, $\bfmu^{\text{test}}_1$}
\end{subfigure}
\begin{subfigure}[t]{0.24\textwidth}
\includegraphics[width=1\linewidth]{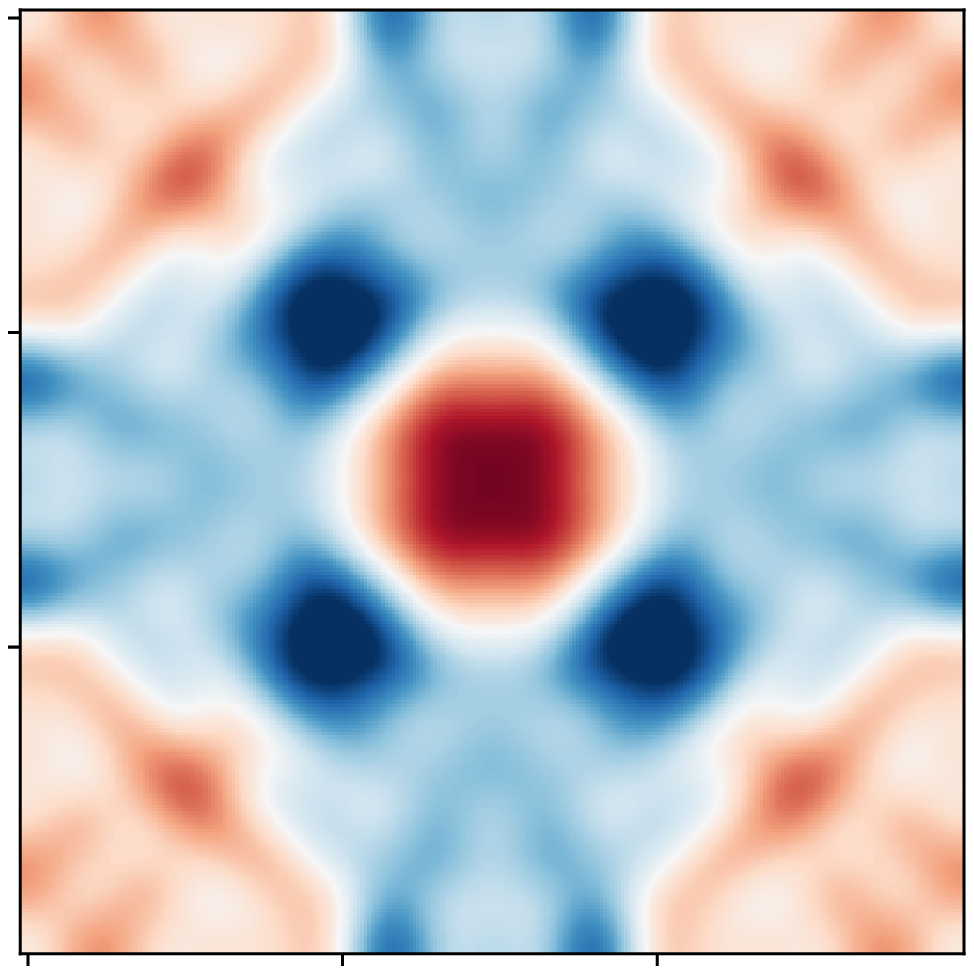} 
\caption{OpInf + roll outs, $\bfmu^{\text{test}}_1$}
\end{subfigure}
\begin{subfigure}[t]{0.24\textwidth}
\includegraphics[width=1\linewidth]{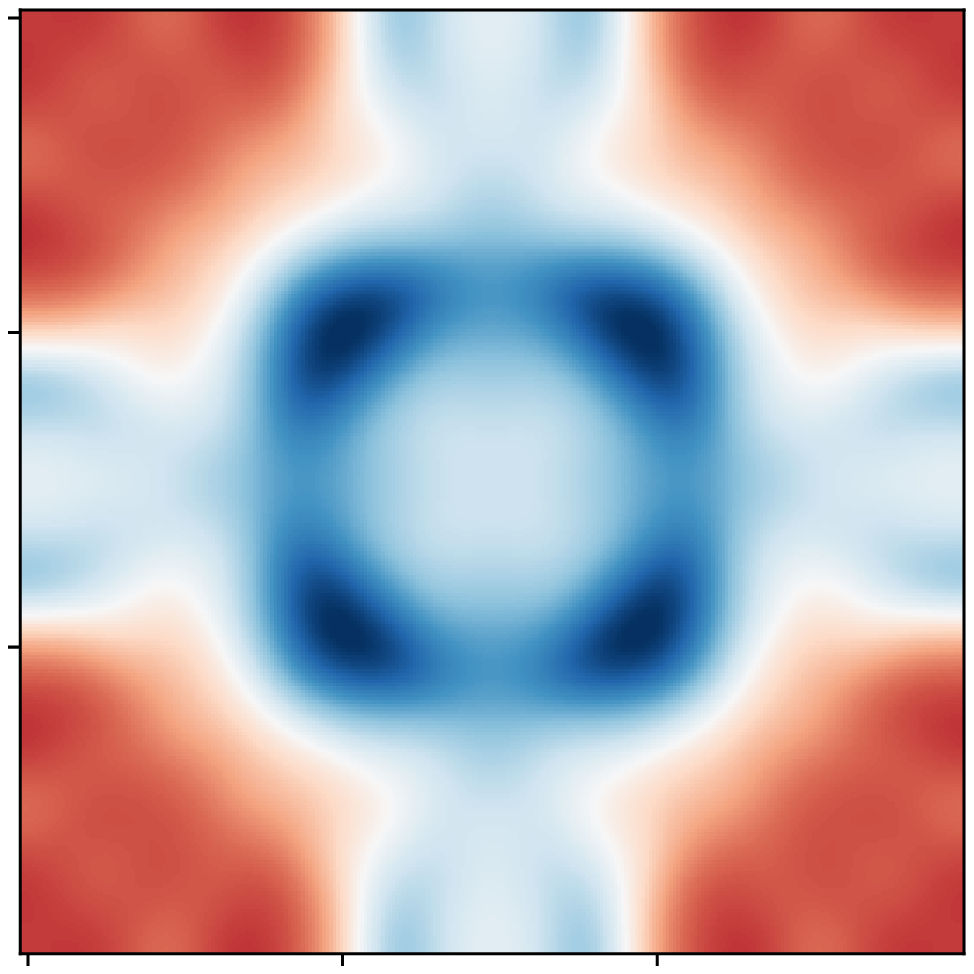} 
\caption{traditional OpInf, $\bfmu^{\text{test}}_1$}
\end{subfigure}

\begin{subfigure}[t]{0.24\textwidth}
\includegraphics[width=1\linewidth]{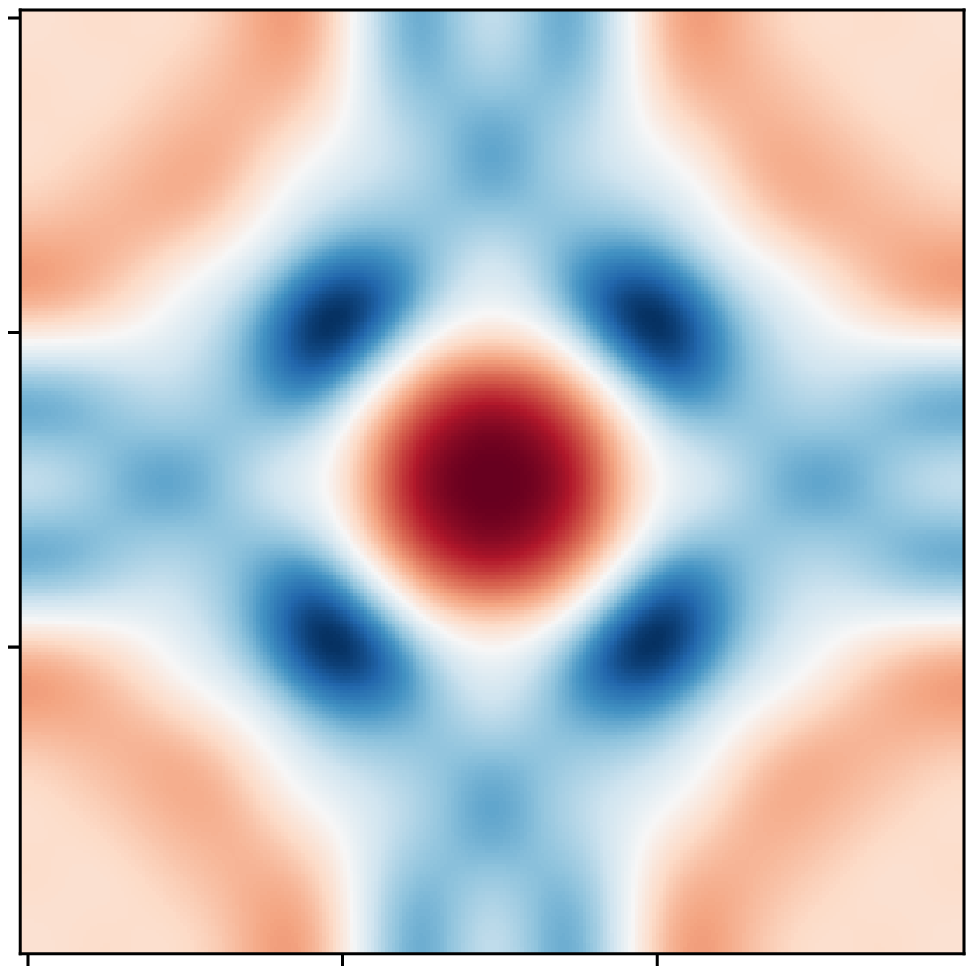} 
\caption{ground truth, $\bfmu^{\text{test}}_2$}
\end{subfigure}
\begin{subfigure}[t]{0.24\textwidth}
\includegraphics[width=1\linewidth]{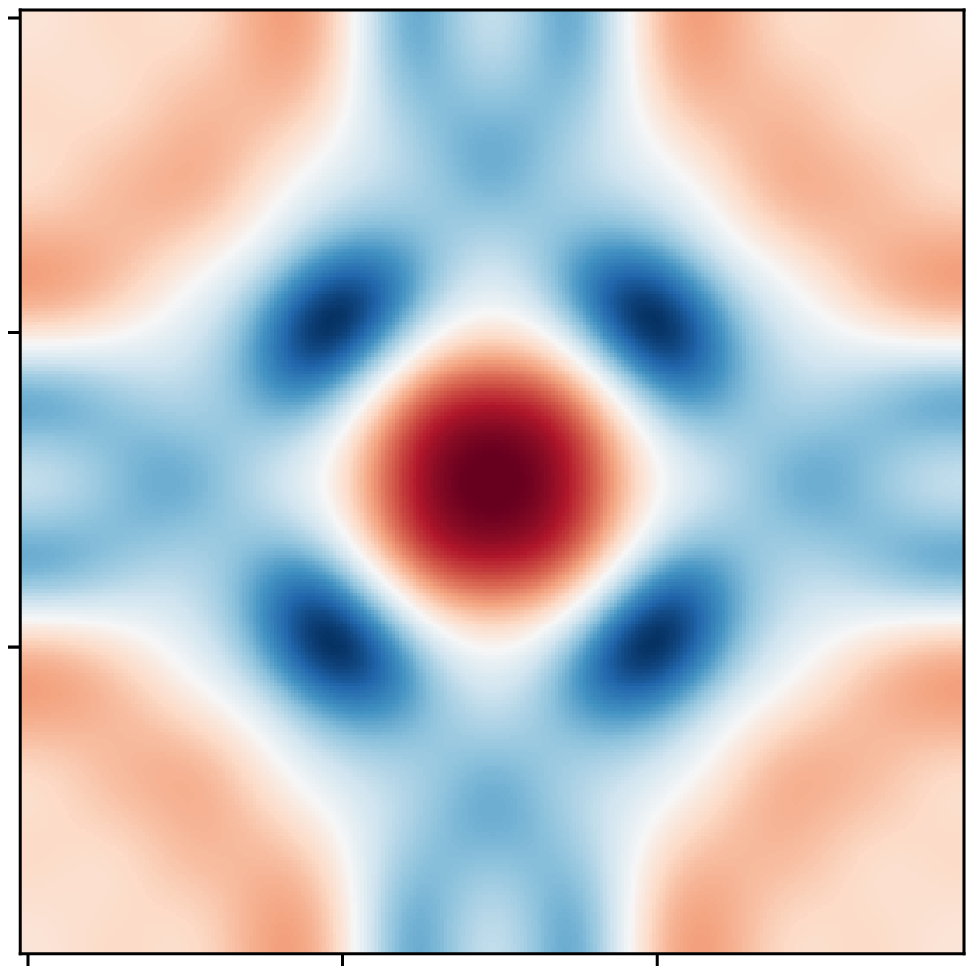} 
\caption{projection, $\bfmu^{\text{test}}_2$}
\end{subfigure}
\begin{subfigure}[t]{0.24\textwidth}
\includegraphics[width=1\linewidth]{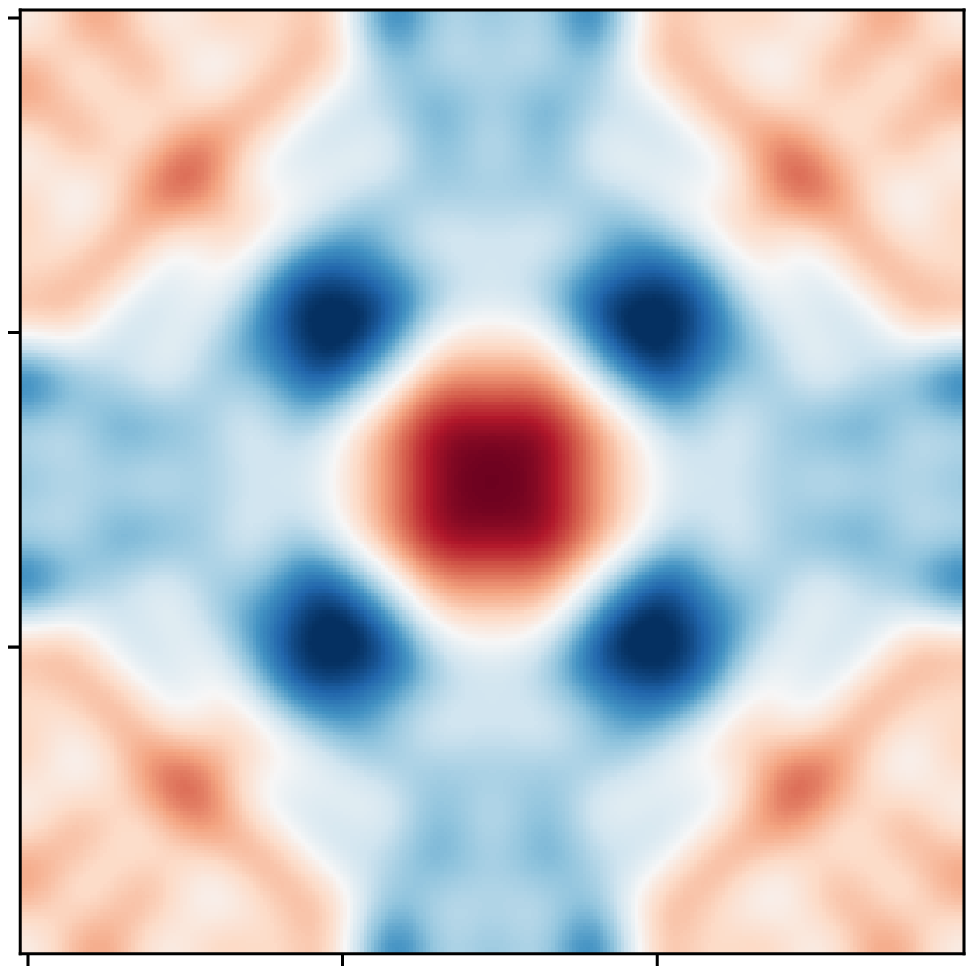} 
\caption{OpInf + roll outs, $\bfmu^{\text{test}}_2$}
\end{subfigure}
\begin{subfigure}[t]{0.24\textwidth}
\includegraphics[width=1\linewidth]{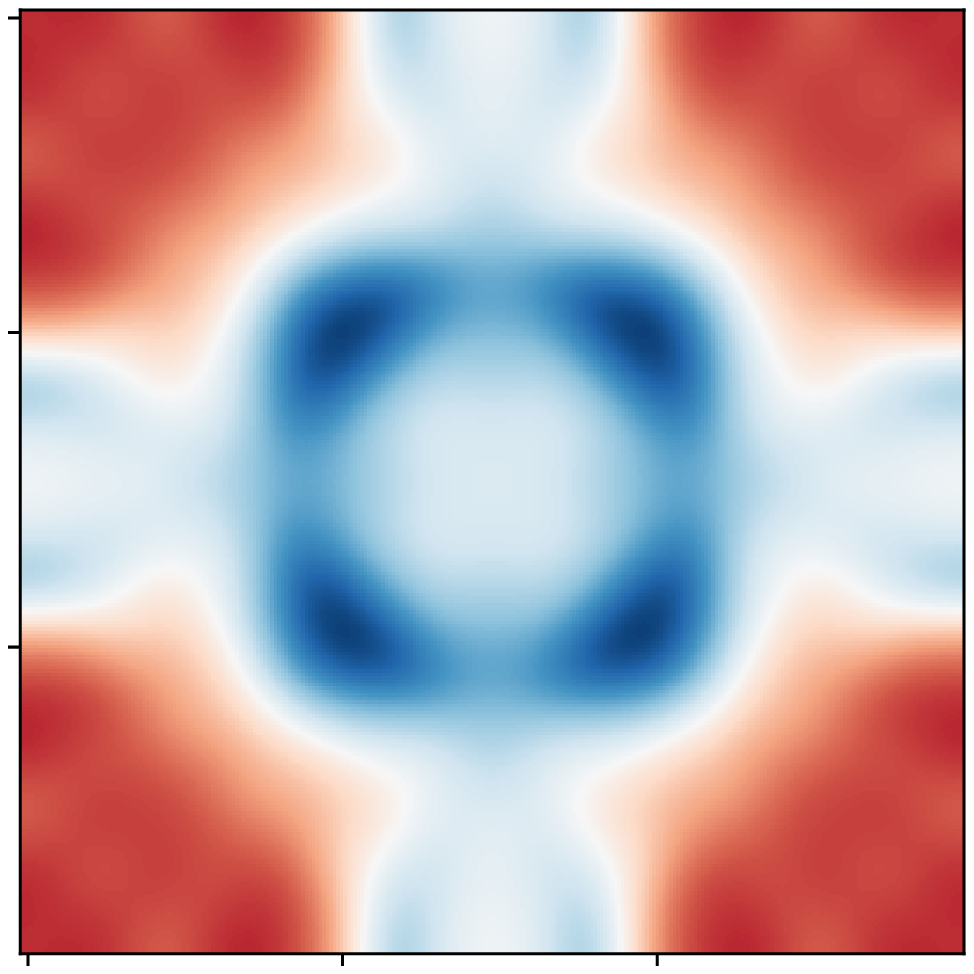} 
\caption{traditional OpInf, $\bfmu^{\text{test}}_2$}
\end{subfigure}

\begin{subfigure}[t]{0.24\textwidth}
\includegraphics[width=1\linewidth]{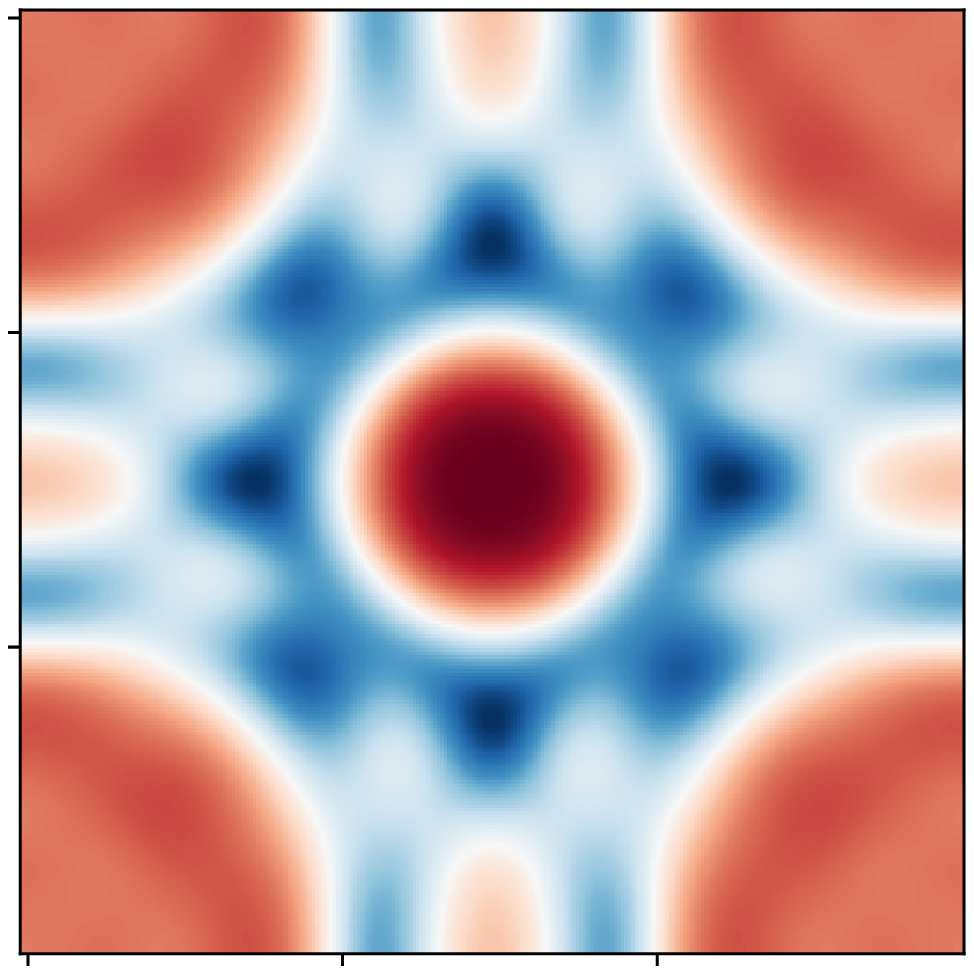} 
\caption{ground truth, $\bfmu^{\text{test}}_3$}
\end{subfigure}
\begin{subfigure}[t]{0.24\textwidth}
\includegraphics[width=1\linewidth]{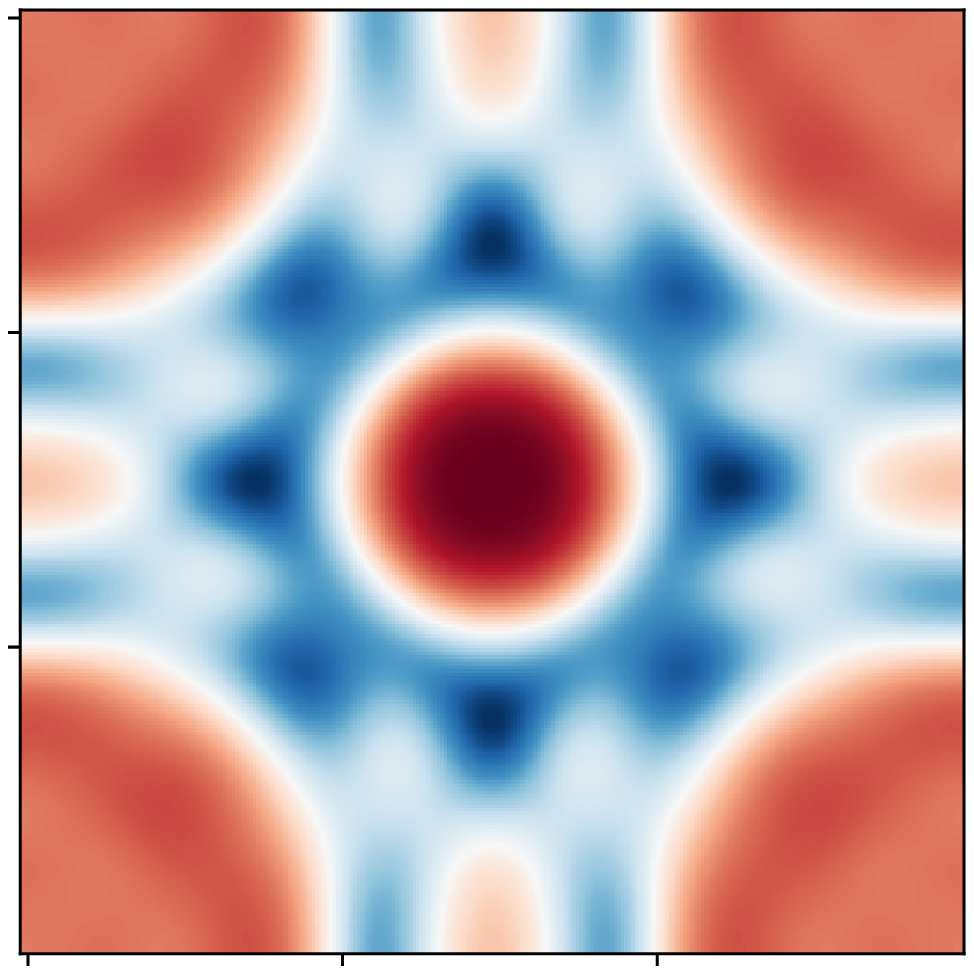} 
\caption{projection, $\bfmu^{\text{test}}_3$}
\end{subfigure}
\begin{subfigure}[t]{0.24\textwidth}
\includegraphics[width=1\linewidth]{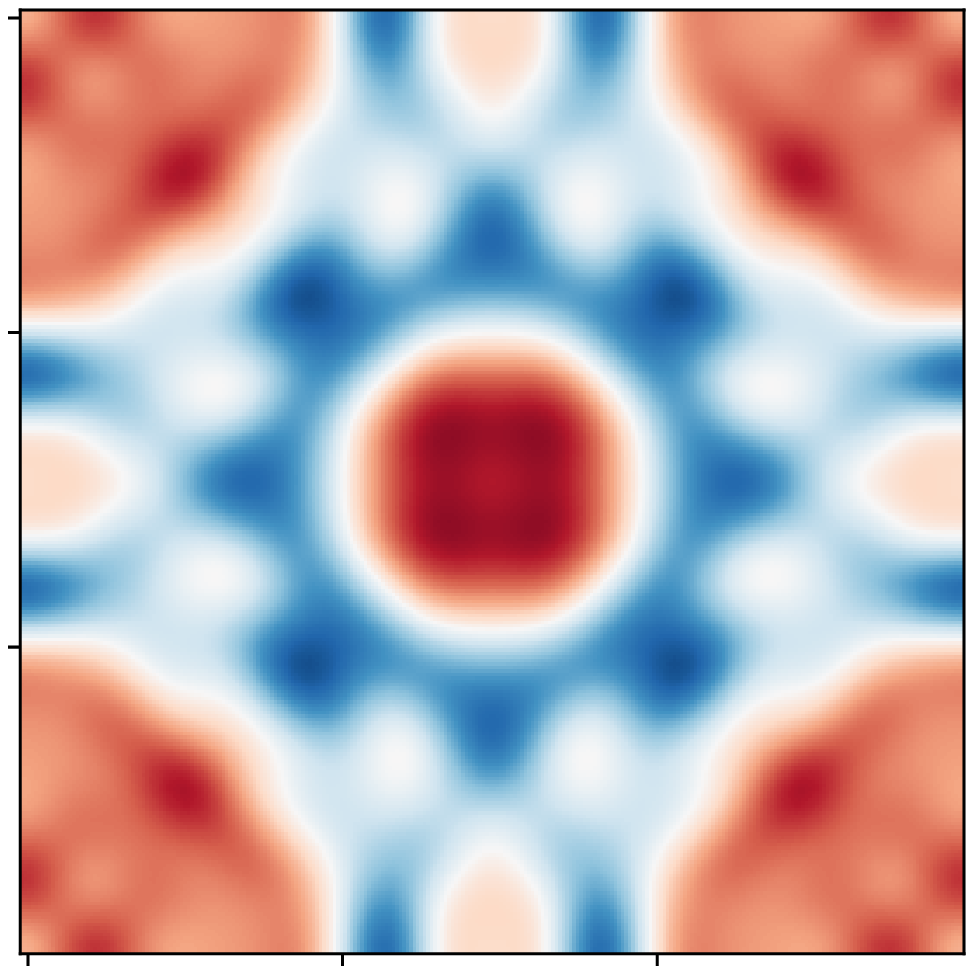} 
\caption{OpInf + roll outs, $\bfmu^{\text{test}}_3$}
\end{subfigure}
\begin{subfigure}[t]{0.24\textwidth}
\includegraphics[width=1\linewidth]{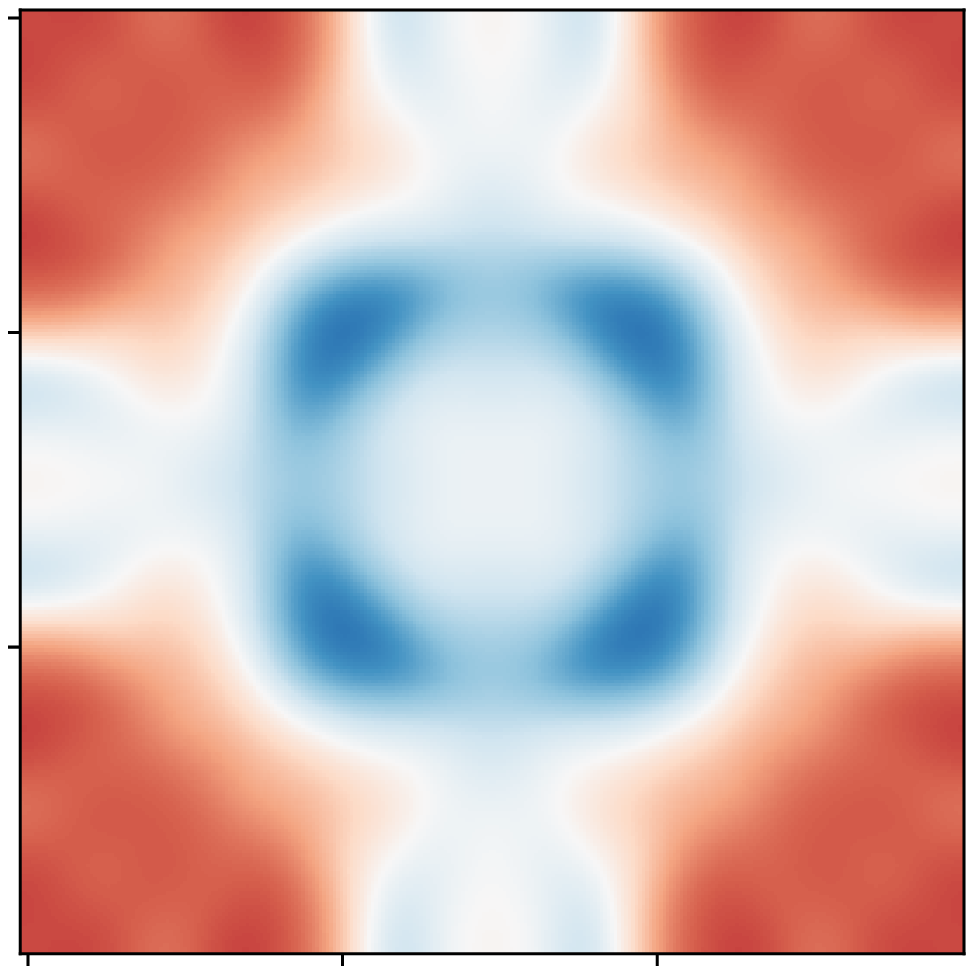} 
\caption{traditional OpInf, $\bfmu^{\text{test}}_3$}
\end{subfigure}

\begin{subfigure}[t]{0.24\textwidth}
\includegraphics[width=1\linewidth]{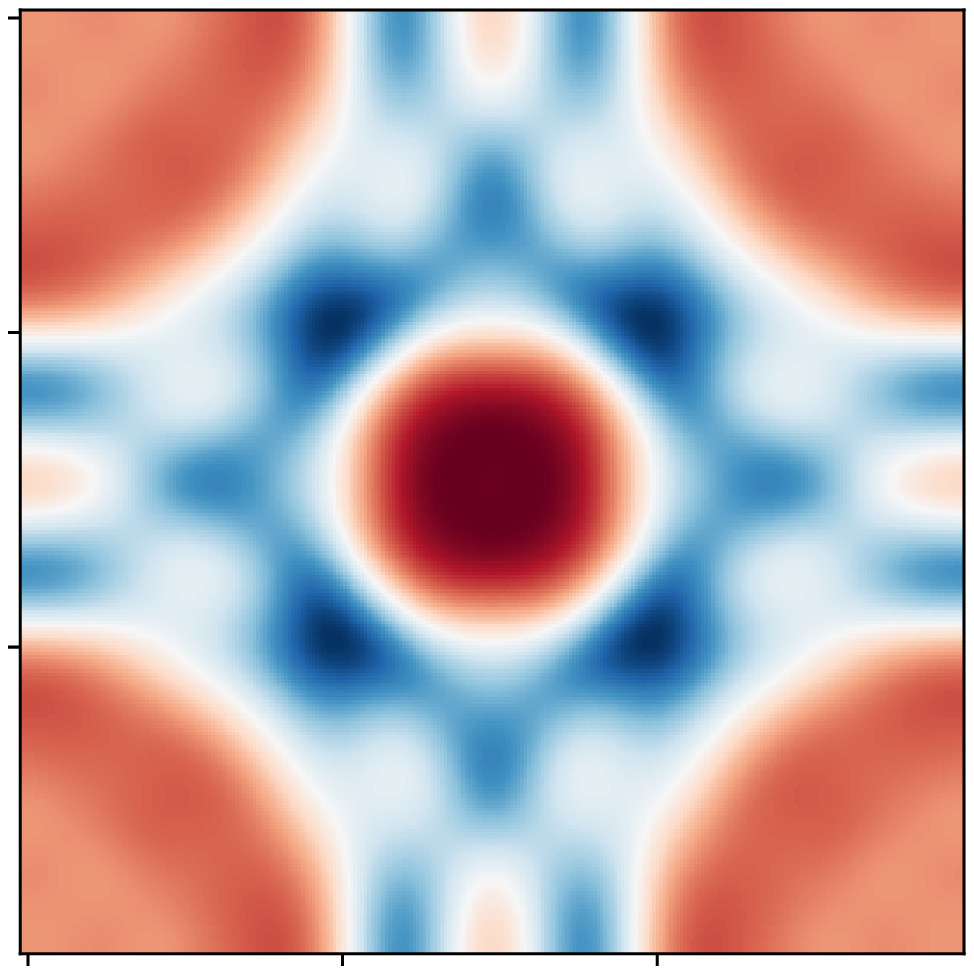} 
\caption{ground truth, $\bfmu^{\text{test}}_4$}
\end{subfigure}
\begin{subfigure}[t]{0.24\textwidth}
\includegraphics[width=1\linewidth]{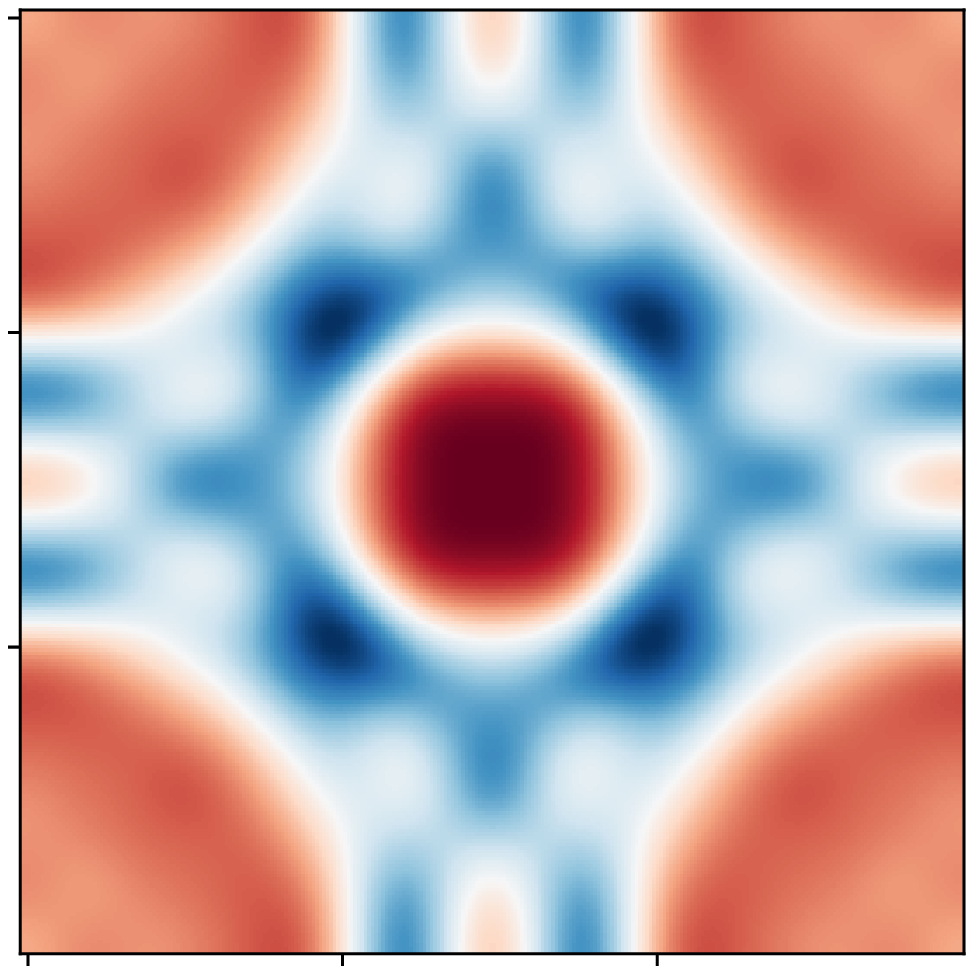} 
\caption{projection, $\bfmu^{\text{test}}_4$}
\end{subfigure}
\begin{subfigure}[t]{0.24\textwidth}
\includegraphics[width=1\linewidth]{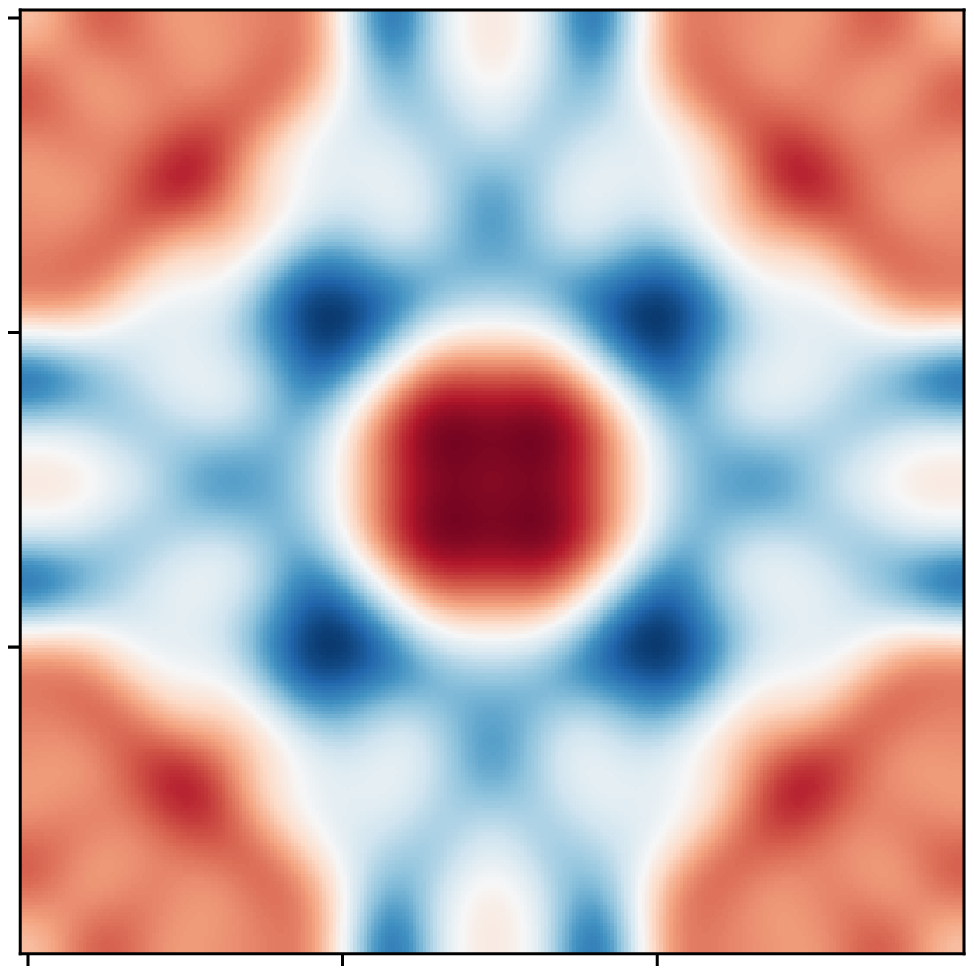} 
\caption{OpInf + roll outs, $\bfmu^{\text{test}}_4$}
\end{subfigure}
\begin{subfigure}[t]{0.24\textwidth}
\includegraphics[width=1\linewidth]{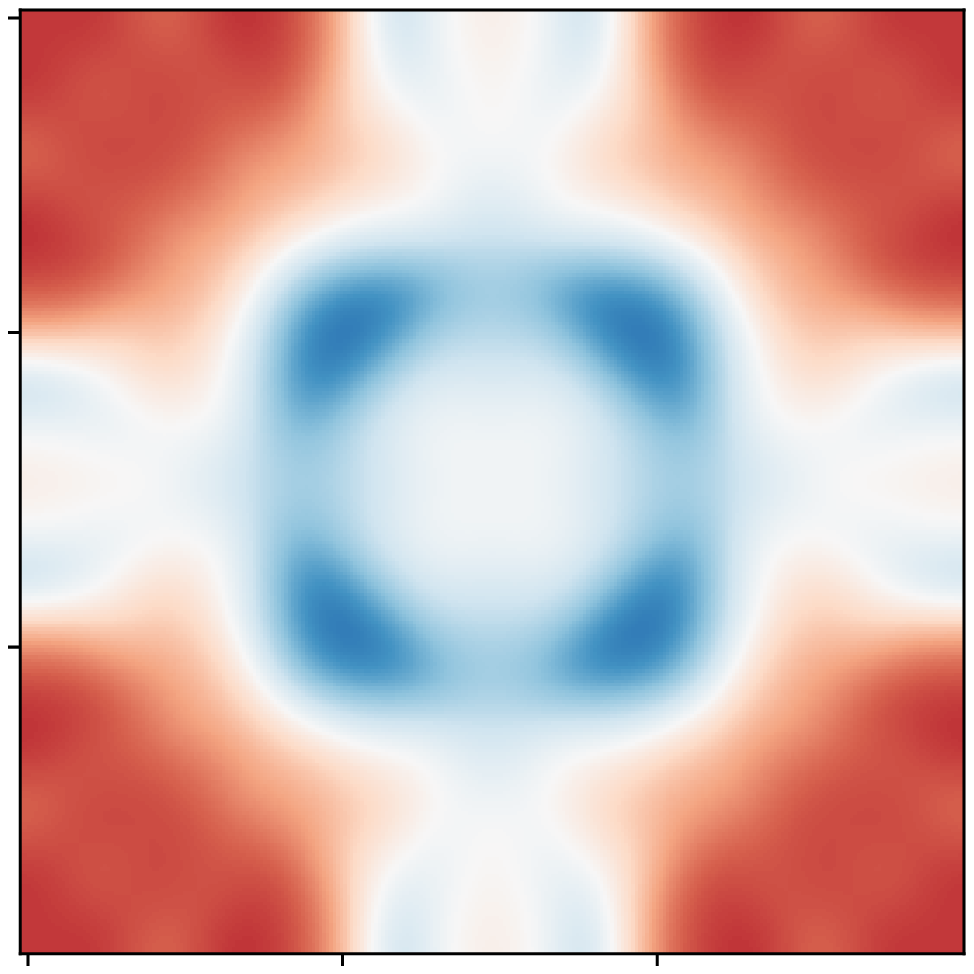} 
\caption{traditional OpInf, $\bfmu^{\text{test}}_4$}
\end{subfigure}

\caption{Shallow water equations (Section~\ref{subsec:shallow}): In this experiment with nine training parameters and roll length $R=150$ and noise $\rho=10\%$, operator inference with roll outs leads to a predictive model for $q_h$ for the test parameters, whereas the model learned traditional operator inference  provides inaccurate predictions. }
\label{fig:shallowNoise_IC0}
\end{figure}

We now consider noisy data that are generated as
\[
\bar{\bfq}_{k}(\bfmu) + \rho \bfeps_k \odot |\bar{\bfq}_{k}(\bfmu)|\,,\qquad k=1,\dots,K-1\,,
\]
where $\bfeps_k \sim N(0,\bfI)$ is standard normal and  $\rho \in \{0.1\%, 1\%, 10\%\}$ controls the standard deviation of the noise. The $\odot$ means component-wise multiplication. The plots in Figure~\ref{fig:shallow_noiseVsErr} show the time-averaged relative error \eqref{eq:TestRelErr} of the model predictions across various noise levels for learning from $M_{\text{train}} \in \{4,7,9\}$ noisy training trajectories with roll-out length $R=150$. Despite having noisy training data, operator inference with roll outs still learns a numerically stable low-dimensional model. The error of the model learned with roll outs increases much slower than the model learned without roll outs, which indicates that the roll outs reduce the effect of noise because they prevent the model from overfitting to the noise in the polluted training data. In contrast, the predictions from several models obtained with traditional operator inference are unstable (and thus replaced by the initial condition in the error computations).

\begin{figure}
\begin{subfigure}[b]{0.48\textwidth}
\begin{center}
{{\Large\resizebox{1\columnwidth}{!}{\input{shallow_errVsRollLen_noise1}}}}
\end{center}
\caption{\#training trajectories $M_{\text{train}}=4$, noise $\rho = 1\%$}
\label{fig:shallow_noise_RollVsErr0.01}
\end{subfigure}
\begin{subfigure}[b]{0.48\textwidth}
\begin{center}
{{\Large\resizebox{1\columnwidth}{!}{\input{shallow_errVsRollLen_noise10}}}}
\end{center}
\caption{\#training trajectories $M_{\text{train}}=9$, noise $\rho = 10\%$}
\label{fig:shallow_noise_RollVsErr0.1}
\end{subfigure}
\caption{Shallow water equations (Section~\ref{subsec:shallow}): Increasing the roll-out length improves the prediction accuracy of the learned model when training on noisy data. }
\label{fig:shallow_NoiseRollErr}
\end{figure}
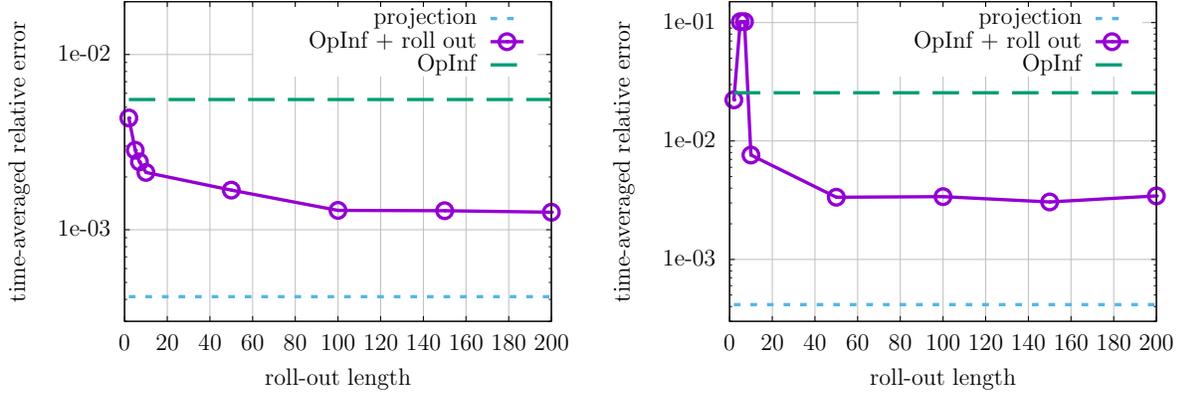

Figure~\ref{fig:shallowNoise_IC0} shows the ground-truth free-surface height computed with the high-dimensional model and the various low-dimensional models  at time $t=0.15$ for different test parameter values with $M_{\text{train}}=9, R=150, \rho=10\%$. Observe that despite the high noise level, the model learned with roll outs is still able to capture the patterns in the ground truth. The model learned via traditional operator inference that does not use roll outs seems to have memorized a pattern for all parameter values in the test set, thereby failing to make meaningful predictions. Figure~\ref{fig:shallow_NoiseRollErr} shows the effect of the roll-out length on the prediction accuracy of the learned model.  In this example, increasing the roll-out length aids in obtaining  more accurate model predictions in the presence of noise. Increasing the roll-out length also increases the number of misfits penalized in the objective function~\eqref{eq:DiffOpInfObj}, or more generally, it lengthens the time window over which we assess the misfit for the objective function~\eqref{eq:ROpInf:ObjHatR}, which makes the learned model more robust against overfitting.

\begin{figure}
\begin{subfigure}[b]{0.33\textwidth}
\begin{center}
{{\huge\resizebox{1\columnwidth}{!}{\input{shallow_errVsNoise_radius_nTrain4}}}}
\end{center}
\caption{\#training trajectories  $M_{\text{train}}=4$}
\label{fig:shallow_noiseVsErr_radius_nTrain4}
\end{subfigure}
\begin{subfigure}[b]{0.33\textwidth}
\begin{center}
{{\huge\resizebox{1\columnwidth}{!}{\input{shallow_errVsNoise_radius_nTrain7}}}}
\end{center}
\caption{\#training trajectories $M_{\text{train}}=7$}
\label{fig:shallow_noiseVsErr_radius_nTrain7}
\end{subfigure}
\begin{subfigure}[b]{0.33\textwidth}
\begin{center}
{{\huge\resizebox{1\columnwidth}{!}{\input{shallow_errVsNoise_radius_nTrain9}}}}
\end{center}
\caption{\#training trajectories $M_{\text{train}}=9$}
\label{fig:shallow_noiseVsErr_radius_nTrain9}
\end{subfigure}
\caption{Shallow water equations (Section~\ref{subsec:shallow}): For small noise intensity, the stability radii of the learned model with roll outs is up to two orders of magnitude higher than the model obtained with traditional operator inference. For larger noise intensity, the stability radii of the two approaches are comparable, owing to the increase in the learned model errors.}
\label{fig:shallow_noiseVsErr_radius}
\end{figure}
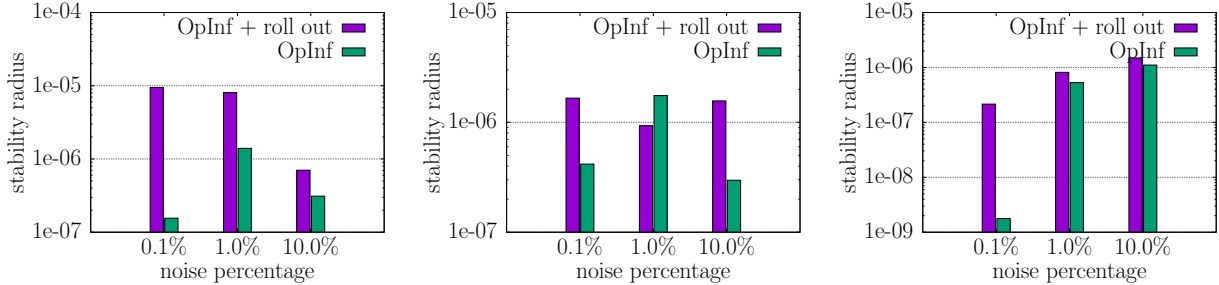

We now compare the bounds of the stability radii of models learned from noisy data, see Figure~\ref{fig:shallow_noiseVsErr_radius}. We vary the number of training trajectories as $M_{\text{train}} \in \{4,7,9\}$ and set the roll-out length to $R=150$. For noise with $\rho=0.1\%$, the bounds of stability radii of the models learned with operator inference with roll outs is up to two orders of magnitude larger than the bounds corresponding to the models learned with traditional operator inference. This is in agreement with the error shown in Figure~\ref{fig:shallow_noiseVsErr}, where for $M_{\text{train}} = 9$ training trajectories and noise with $\rho=0.1\%$, the models learned with roll outs achieve about one order of magnitude higher accuracy. For high noise with $\rho = 1.0\%$ and $\rho = 10\%$, the bounds of the stability radii of the models inferred under both approaches tend to be comparable in magnitude, which mirrors the increase in the error of the models learned with roll outs as $\rho$ is increased to $10\%$ in Figure~\ref{fig:shallow_noiseVsErr}.

\subsection{Surface quasi-geostrophic dynamics} \label{subsec:pyqg}

Surface quasi-geostrophic equations model the dynamics of buoyancy on horizontal boundaries and is a special case of the quasi-geostrophic equations which characterize fluid motion in the limit of strong rotation and stratification. We follow the problem as described in \cite{held_pierrehumbert_garner_swanson_1995} and implemented in the PyQG code\footnote{https://github.com/pyqg/pyqg}.

\subsubsection{Surface quasi-geostrophic dynamics: Setup}
For  $z \in (-\infty,0]$ and  $x,y \in \Omega$ where $\Omega = (-\pi,\pi) \times (-\pi,\pi)$, let $\psi$ be the stream function such that  $b(t, x,y) = \frac{\partial \psi}{\partial z}(t, x, y, 0)$ is the buoyancy. The governing equation of the surface buoyancy for $t \in (0,T) \subset \R$ is given by
\begin{align}\label{eq:PyQG}
    \frac{\partial b}{\partial t} + \left( -\frac{\partial \psi}{\partial y} \bigg \rvert_{z=0}, \frac{\partial \psi}{\partial x} \bigg\rvert_{z=0} \right) \cdot \nabla b = 0
\end{align}
such that
\begin{align*}
    \frac{\partial^2 \psi}{\partial x^2} + \frac{\partial^2 \psi}{\partial y^2} + \frac{\partial^2 \psi}{\partial z^2} & = 0\,,\\
    \lim_{z \rightarrow -\infty} \frac{\partial \psi}{\partial z} & = 0.
\end{align*}
The initial condition is
\begin{align*}
    b(0, x,y) = -\exp \biggl\{ -\frac{9}{4 \pi^2} ((\mu_1 (x-0.3))^2 + (\mu_2 y)^2) \biggr\} -\exp \biggl\{ -\frac{9}{4 \pi^2} ((\mu_1 (x+0.3))^2 + (\mu_2 y)^2) \biggr\}
\end{align*}
with inputs $\bfmu = [\mu_1, \mu_2] \in \Dcal$ where $\Dcal = [2.5,4] \times [3,4.5] \cup [4,5.5] \times [1.5,3]$.

We generate data using the PyQG code which discretizes \eqref{eq:PyQG} using a spectral method and advances the corresponding system of ordinary differential equations in time using a third-order Adams-Bashford scheme. We set $\delta t = 0.005$ and $T = 5$. The square domain is partitioned into a regular grid with 200 equidistant points along each dimension such that at time step $k$, the components of the high-dimensional state $\bfq_k(\bfmu) \in \R^N, N = 40000,$ correspond to the buoyancy at each point in the grid. We use an implicit-explicit scheme for integrating the low-dimensional models over $t \in (0,T)$ with $\delta t = 0.005$. The linear term is handled implicitly while the quadratic term is evaluated explicitly.

We aim to learn a low-dimensional model of dimension $n=60$ for this example. The low-dimensional basis $\bfV$ of the space $\Vcal$ is computed from snapshot trajectories that are obtained from nine different initial conditions corresponding to the inputs \[
\bfmu^{\text{basis}} \in \{(2.5,3), (4,4.5), (4,1.5), (5.5,3), (4,3), (3.25,3.75),\\ (4.75,2.25), (5.5,1.5), (2.5,4.5)\}.
\]
Training data correspond to three nested sets of inputs given by
\begin{align*}
\bfmu^{\text{train-I}} = & \{(2.5,3.75),(5.5, 2.25),(4, 3.75),(4, 2.25)\}\,,\\
\bfmu^{\text{train-II}} = & \bfmu^{\text{train-I}} \cup \{(3,3), (3.5,4.5), (4.5, 3), (5,1.5)\}\,,\\
\bfmu^{\text{train-III}} = & \bfmu^{\text{train-II}} \cup \{(3, 4.5), (3.5,3), (4.5,1.5), (5,3)\}\,.
\end{align*}
Hyperparameters are tuned using trajectories on the validation data set corresponding to the inputs
\begin{multline}
\bfmu^{\text{valid}} \in \{(2.75,3.375), (3.75, 3.375), (2.75,4.125), (3.75, 4.125),\\ (4.25, 1.875), (5.25, 1.875), (4.25, 2.625), (5.25,2.625)\}.
\end{multline}
We then assess the performance of the learned low-dimensional models using test data corresponding to the inputs $\bfmu^{\text{test}} \in \{(3.25,4.125), (3.25, 3.375), (4.75,1.875), (4.75, 2.625)\}.$

\subsubsection{Surface quasi-geostrophic dynamics: Scarce data because of sparsely sampled trajectories in time} In this example, we will also demonstrate another advantage of operator inference with roll outs: the objective function is flexible enough to accommodate the situation where the training data is not observed at all time points. This would pose a challenge for traditional operator inference when it approximations time derivatives via finite differences, for example. In particular, operator inference with roll outs is well-equipped to handle data at irregularly-spaced time points. To keep computations manageable, in this work, we will consider scarce training data that are obtained at equidistant time points; however, the methodology applies to non-equidistant time points as well. The sampling period is $\xi \in \mathbb{N}$, which corresponds to the number of time steps between two consecutive measurements of training states. Thus, with a sampling period $\xi$,   the training data are collected at $t \in \{\xi \Delta t | 0 \le \xi \Delta t \le T, \xi \in \mathbb{Z}, \xi > 0\}$. Setting $\xi = 1$ means that data are not sparsified. For traditional operator inference, we change the objective function as
\begin{equation*}
\min_{\bfA_1, \dots, \bfA_L, \bfB} \frac{1}{\lfloor K/\xi \rfloor}\sum_{k = 1}^{\lfloor K/\xi \rfloor} \left\|\sum_{\ell = 1}^L \bfA_\ell\bar{\bfq}_{k\xi}^{\ell}(\bfmu_i) + \bfB\bfu_{k\xi}(\bfmu_i) -\bar{\bfq}_{k \xi}^{\prime}\right\|^2_2\,,
\end{equation*}
where the derivative $\bar{\bfq}_{k \xi}^{\prime}$ is approximated via finite difference using the scarce data, i.e.
\begin{align*}
    \bar{\bfq}_{k \xi}^{\prime} = \frac{\bar{\bfq}_{(k+1) \xi} - \bar{\bfq}_{k \xi}}{\xi \Delta t}.
\end{align*}
For operator inference with roll outs, the objective function is now
\begin{equation*}
J(\bftheta; \bfmu) = \sum_{k = 0}^{\lfloor (K-R)/\xi \rfloor}\sum_{r = 1}^{\lfloor R/\xi \rfloor} \left\| \bar{\bfq}_{(k + r)\xi}(\bfmu) - \hbfz_{k \xi,r \xi}(\bfmu)\right\|^2\,.
\end{equation*}

\subsubsection{Surface quasi-geostrophic dynamics: Results for learning from scarce data}
\label{sec:NumExp:Surface:Scarce}

\begin{figure}
\begin{subfigure}[b]{0.24\textwidth}
\includegraphics[width=1.0\linewidth]{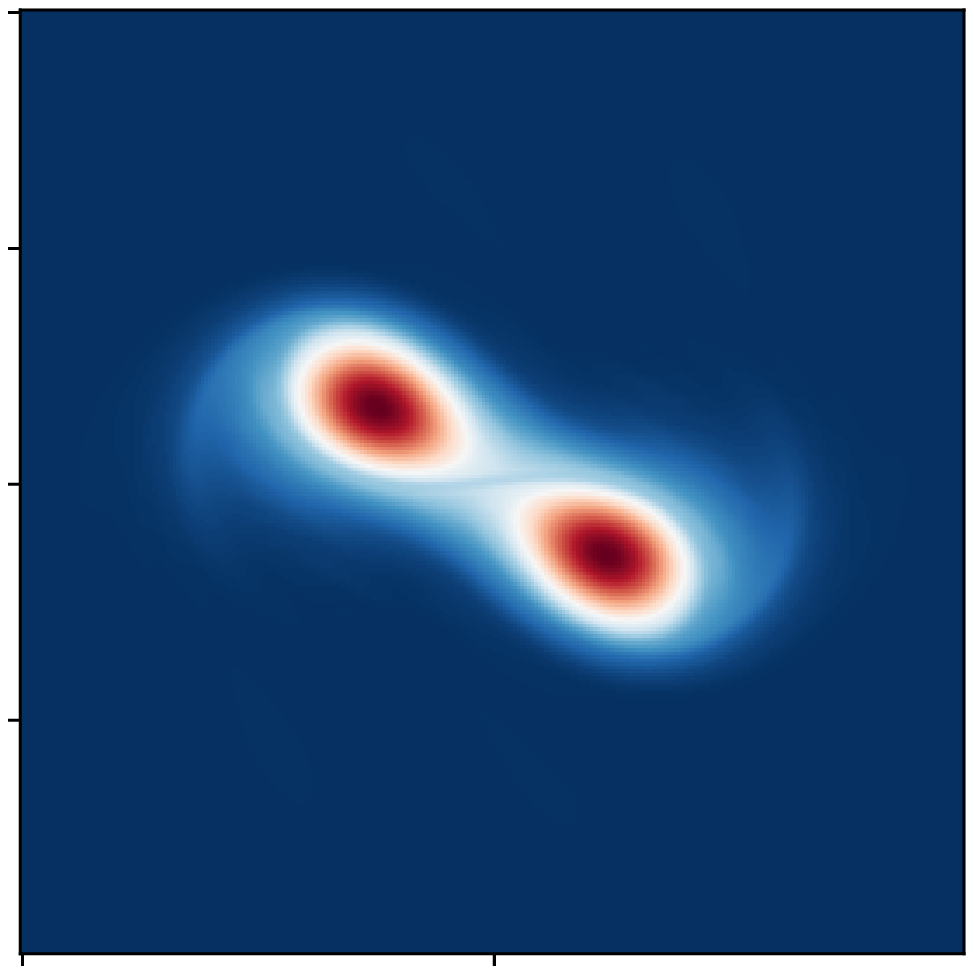} 
 \caption{ground truth, $\bfmu^{\text{test}}_1$}
\end{subfigure}
\begin{subfigure}[b]{0.24\textwidth}
\includegraphics[width=1.0\linewidth]{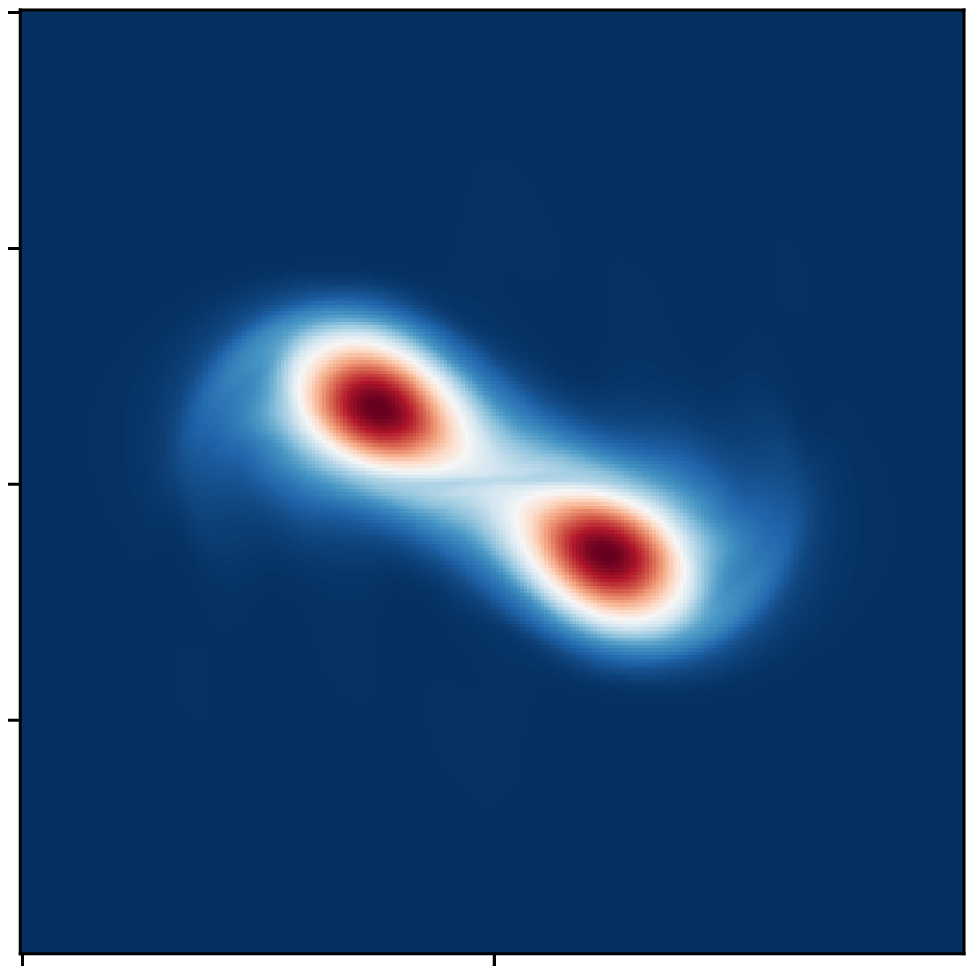} 
 \caption{projection, $\bfmu^{\text{test}}_1$}
\end{subfigure}
\begin{subfigure}[b]{0.24\textwidth}
\includegraphics[width=1.0\linewidth]{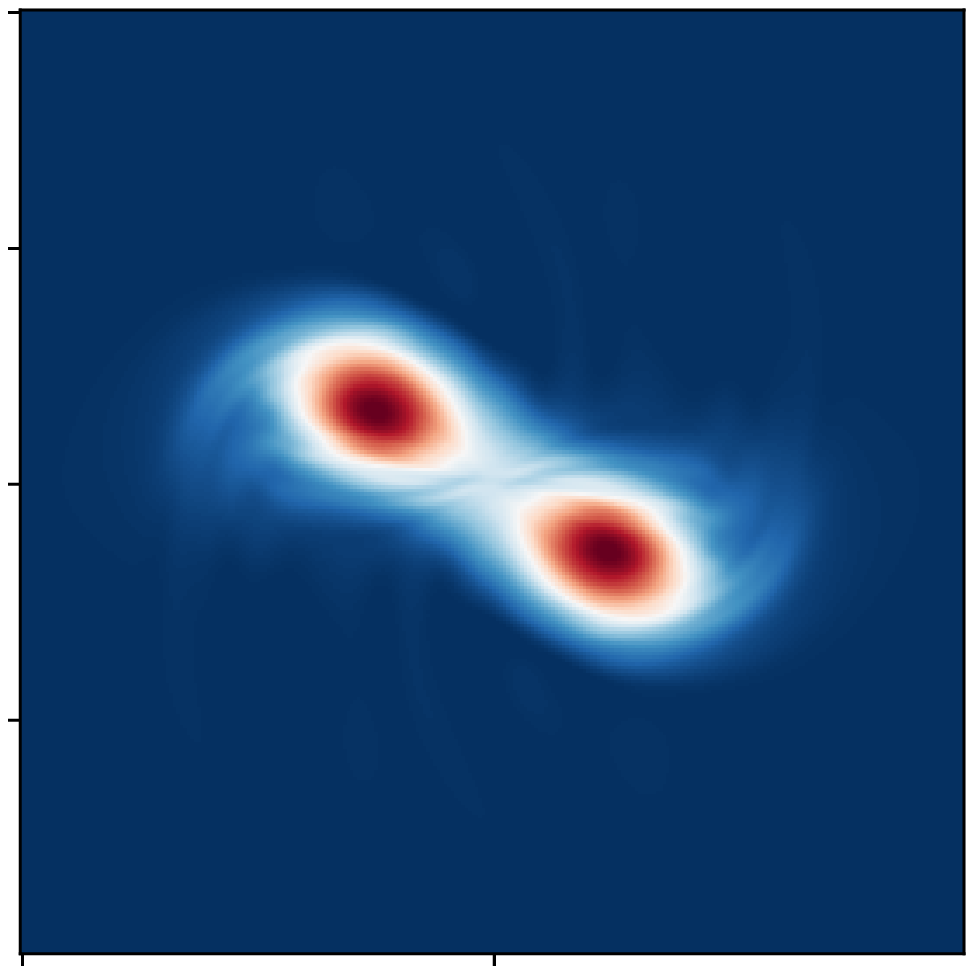} 
\caption{OpInf + roll outs, $\bfmu^{\text{test}}_1$}
\end{subfigure}
\begin{subfigure}[b]{0.24\textwidth}
\includegraphics[width=1.0\linewidth]{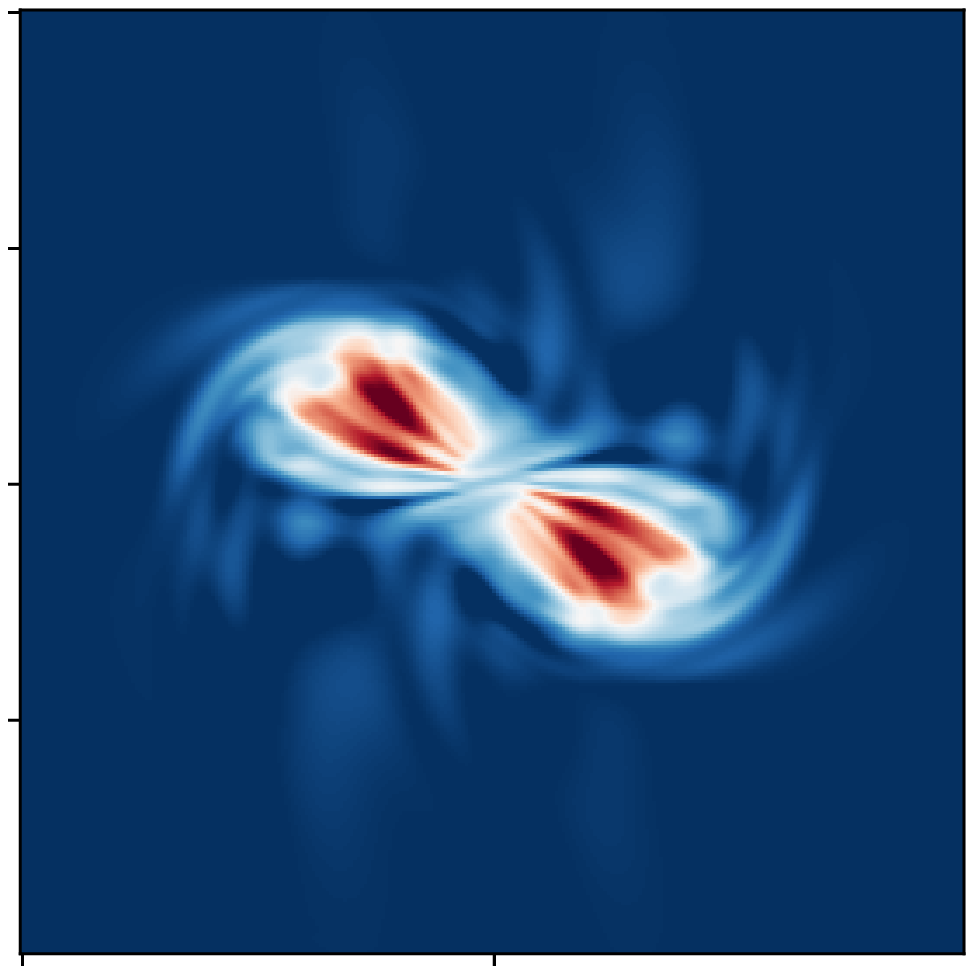} 
\caption{traditional OpInf, $\bfmu^{\text{test}}_1$}\label{fig:sqgstates:TOpInfUnstable}
\end{subfigure}

\begin{subfigure}[b]{0.24\textwidth}
\includegraphics[width=1.0\linewidth]{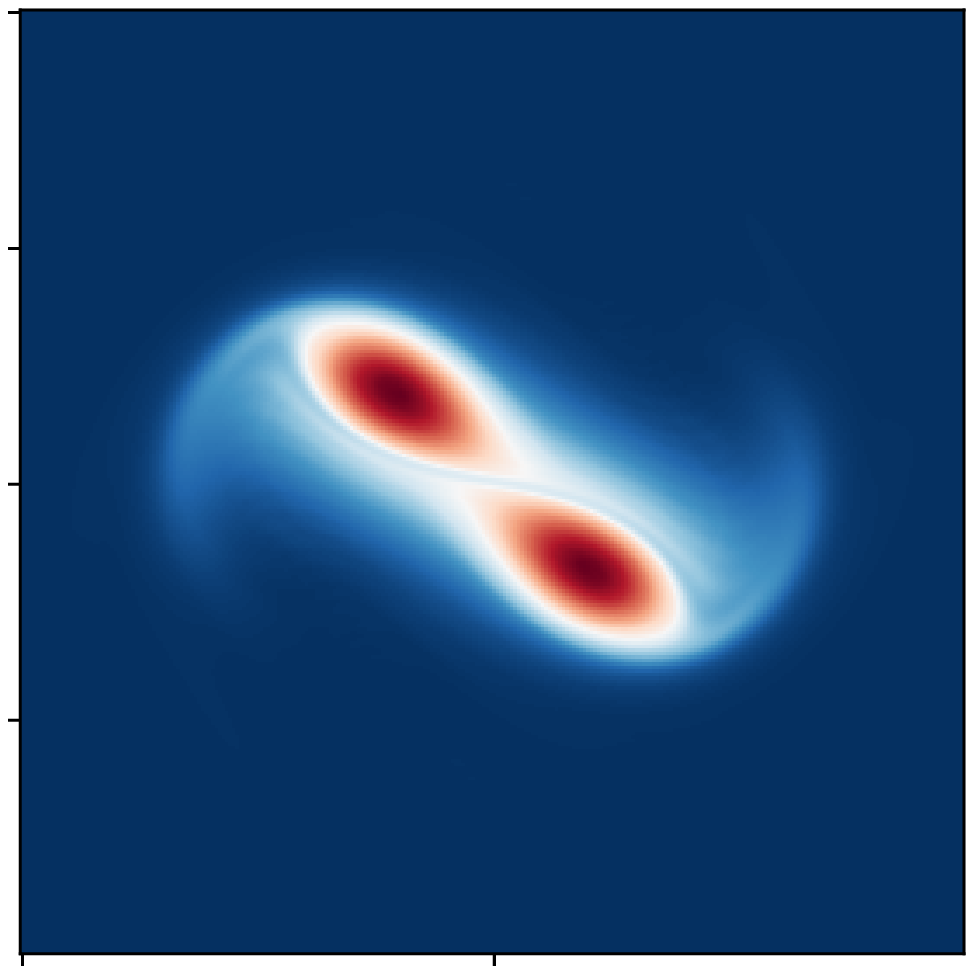} 
 \caption{ground truth, $\bfmu^{\text{test}}_2$}
\end{subfigure}
\begin{subfigure}[b]{0.24\textwidth}
\includegraphics[width=1.0\linewidth]{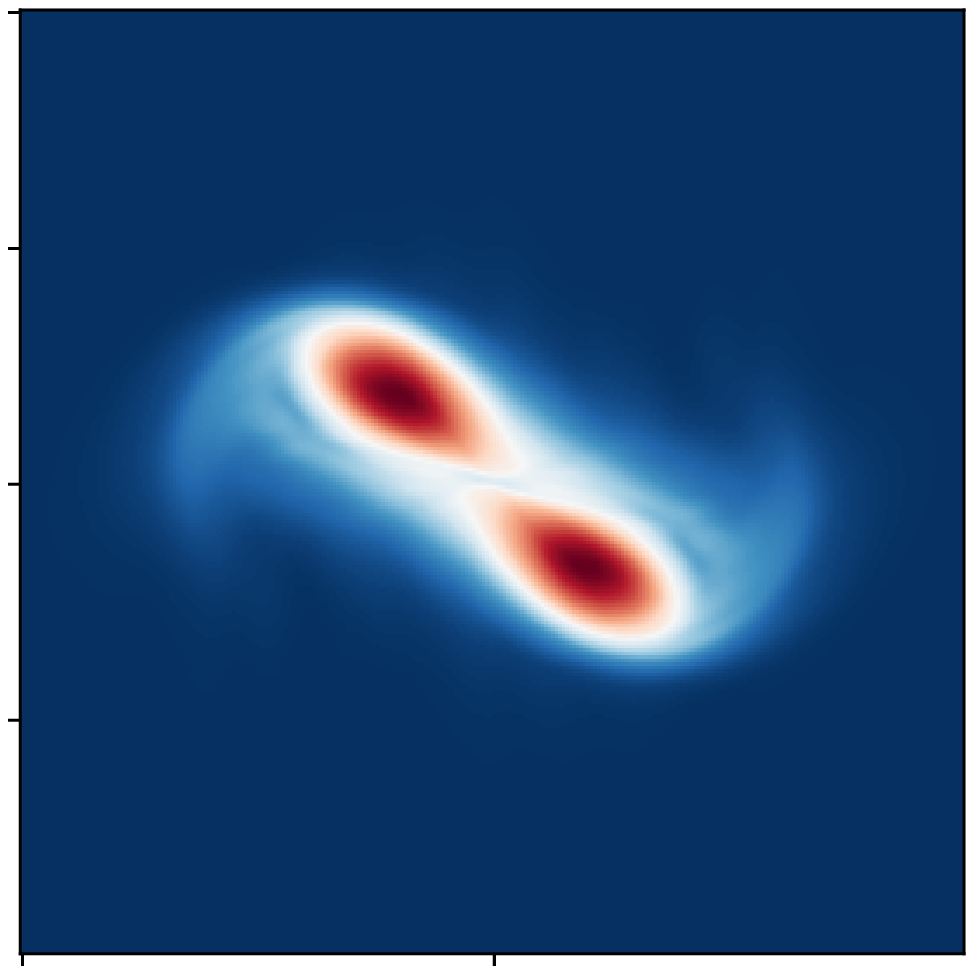} 
 \caption{projection, $\bfmu^{\text{test}}_2$}
\end{subfigure}
\begin{subfigure}[b]{0.24\textwidth}
\includegraphics[width=1.0\linewidth]{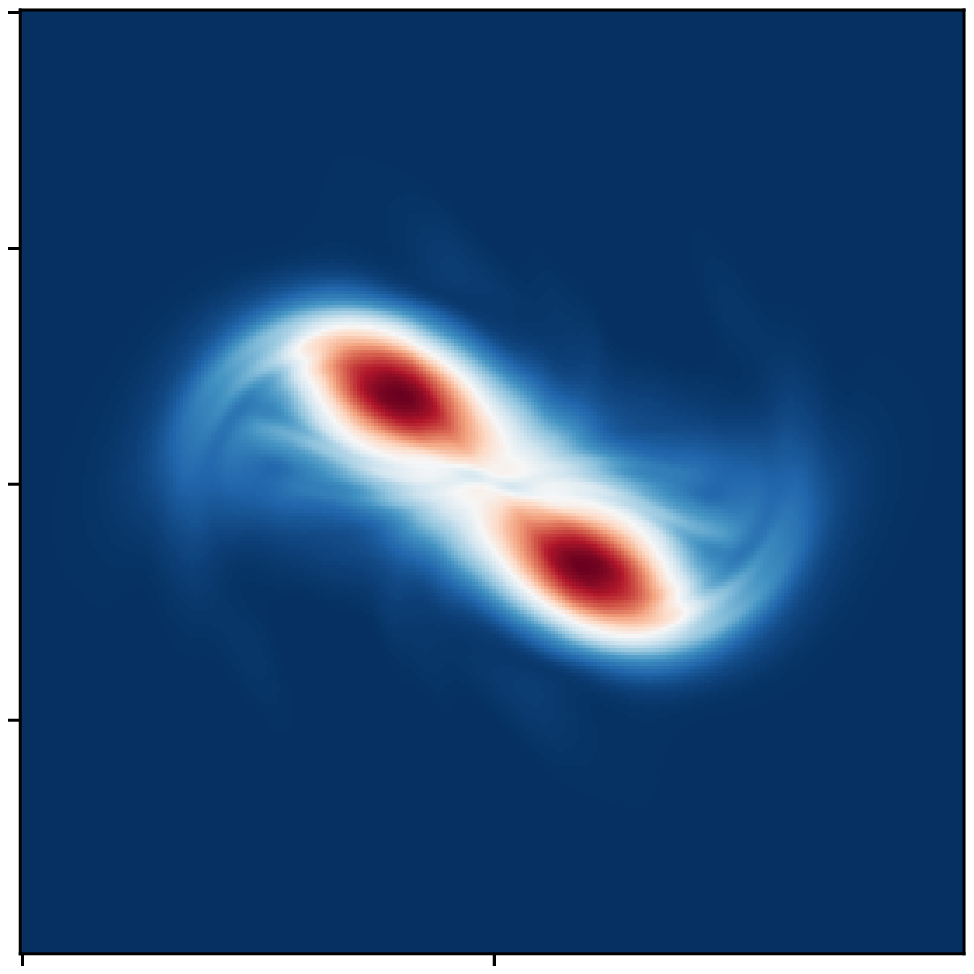} 
\caption{OpInf + roll outs, $\bfmu^{\text{test}}_2$}
\end{subfigure}
\begin{subfigure}[b]{0.24\textwidth}
\includegraphics[width=1.0\linewidth]{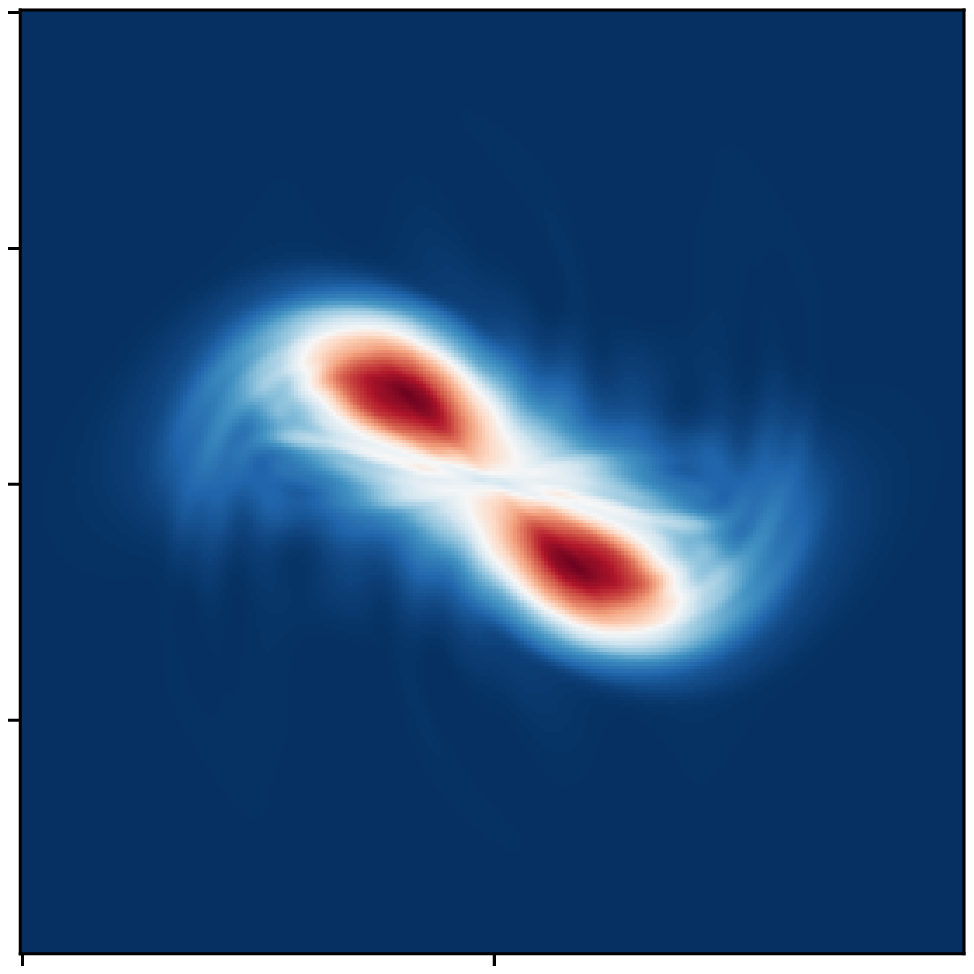} 
\caption{traditional OpInf, $\bfmu^{\text{test}}_2$}
\end{subfigure}

\begin{subfigure}[b]{0.24\textwidth}
\includegraphics[width=1.0\linewidth]{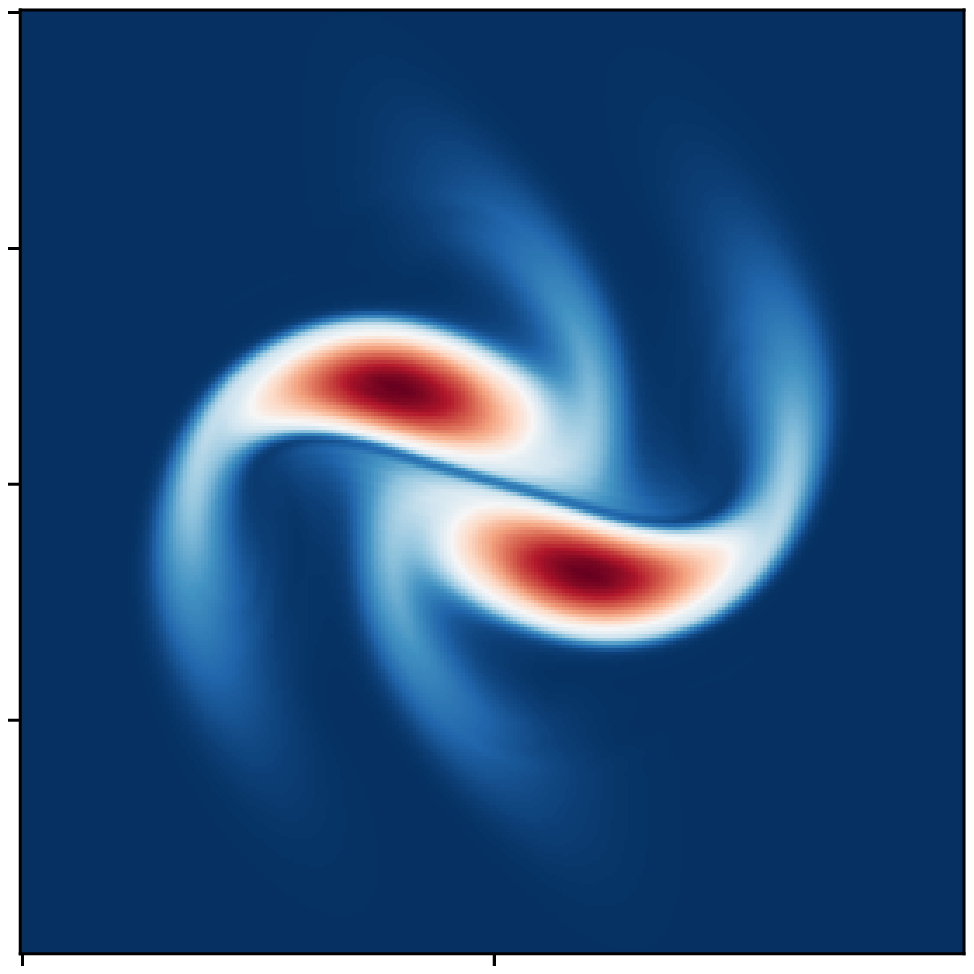} 
 \caption{ground truth, $\bfmu^{\text{test}}_3$}
\end{subfigure}
\begin{subfigure}[b]{0.24\textwidth}
\includegraphics[width=1.0\linewidth]{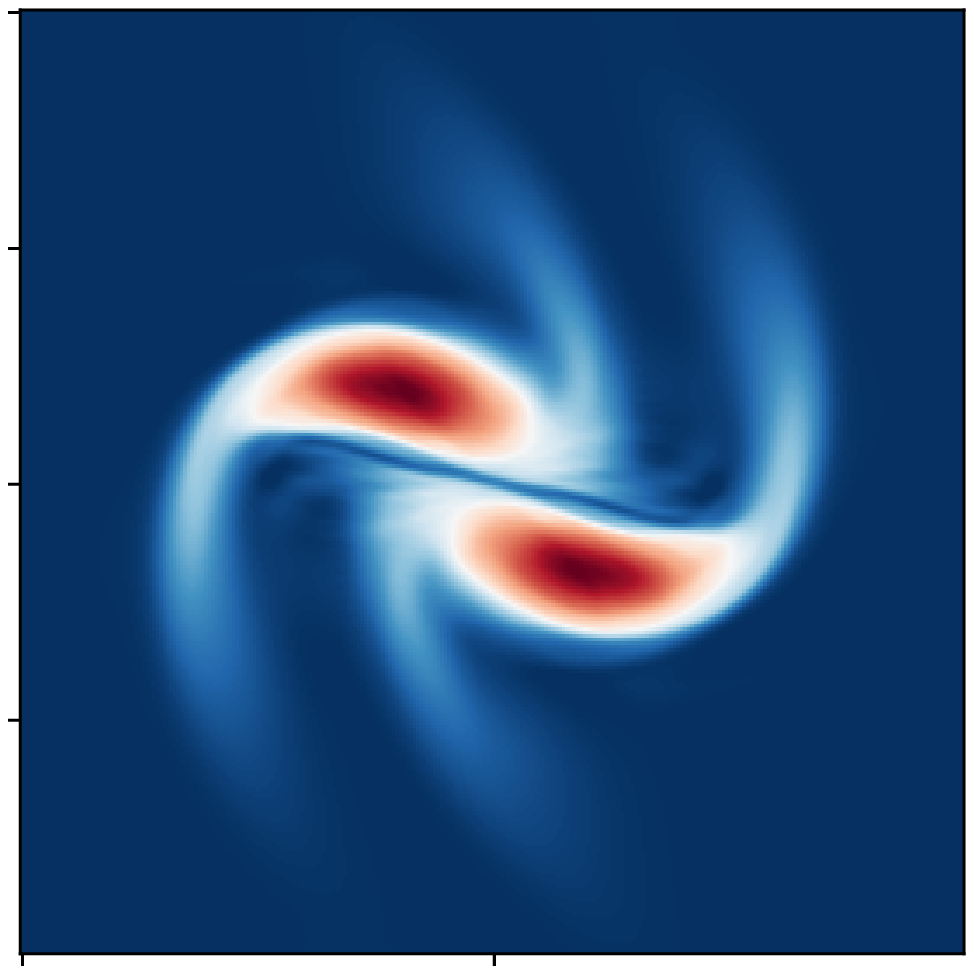} 
 \caption{projection, $\bfmu^{\text{test}}_3$}
\end{subfigure}
\begin{subfigure}[b]{0.24\textwidth}
\includegraphics[width=1.0\linewidth]{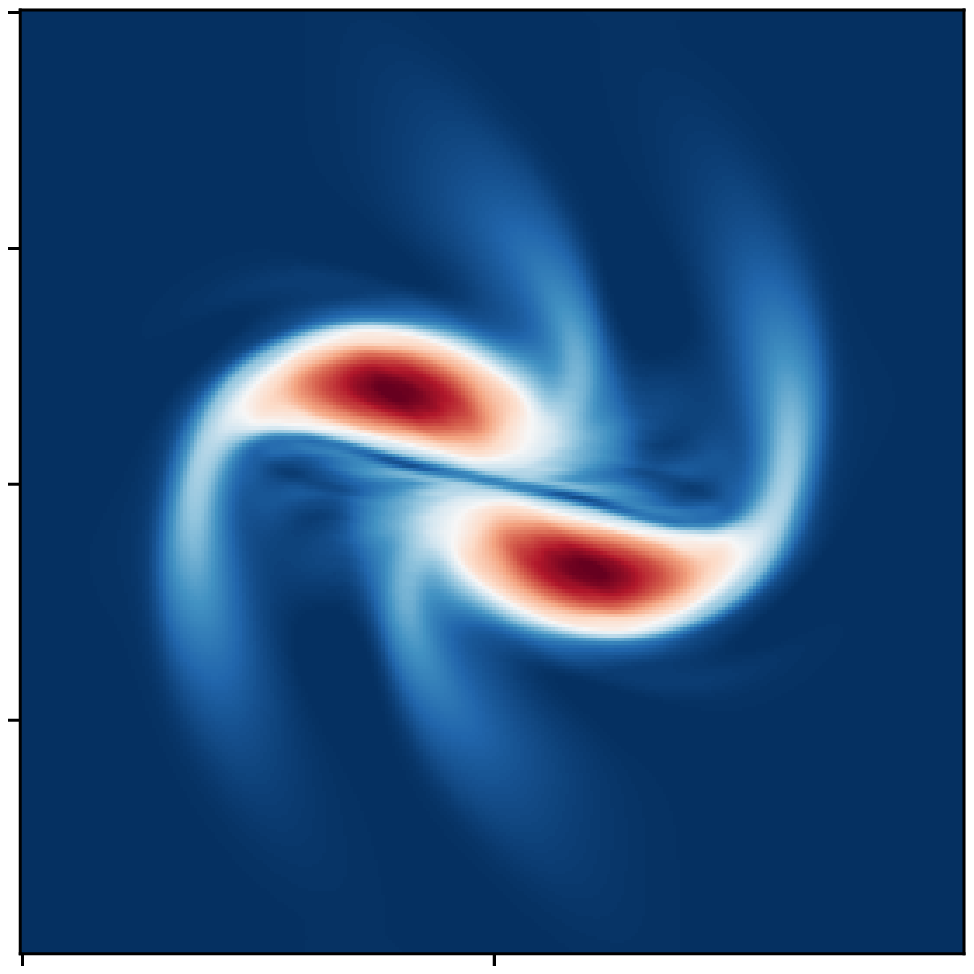} 
\caption{OpInf + roll outs, $\bfmu^{\text{test}}_3$}
\end{subfigure}
\begin{subfigure}[b]{0.24\textwidth}
\includegraphics[width=1.0\linewidth]{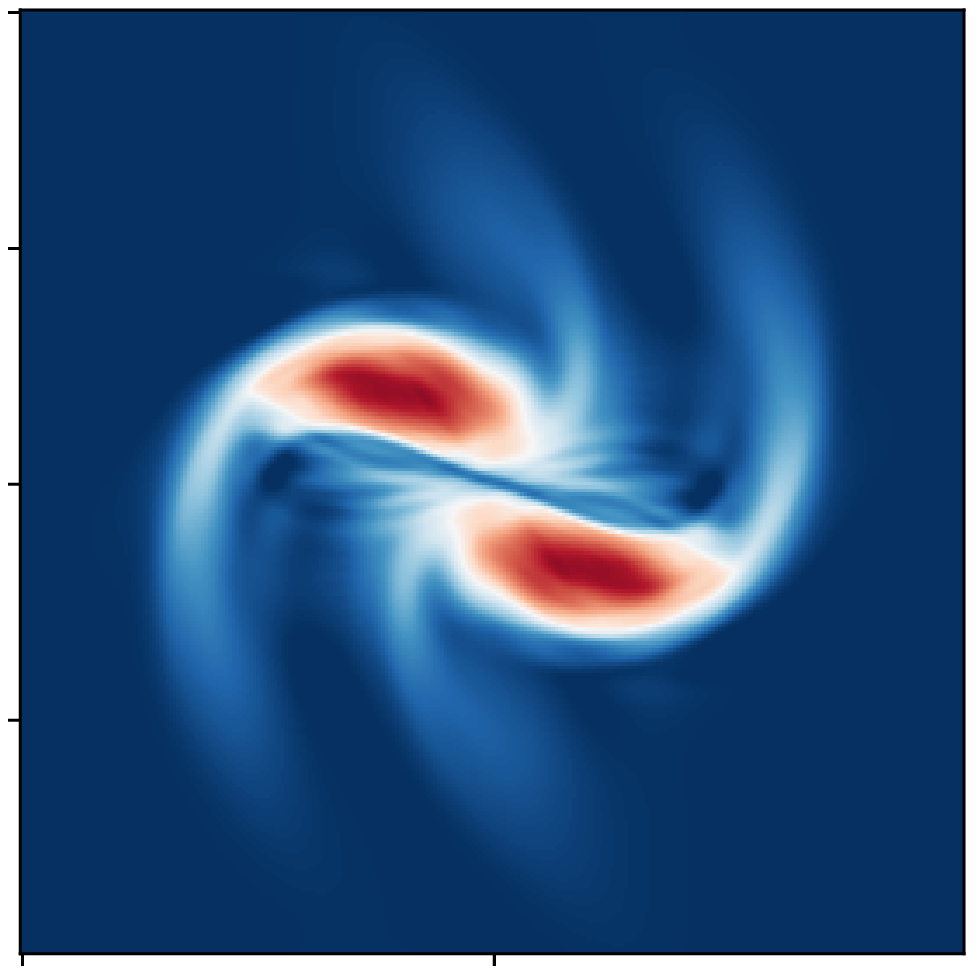} 
\caption{traditional OpInf, $\bfmu^{\text{test}}_3$}
\end{subfigure}

\begin{subfigure}[b]{0.24\textwidth}
\includegraphics[width=1.0\linewidth]{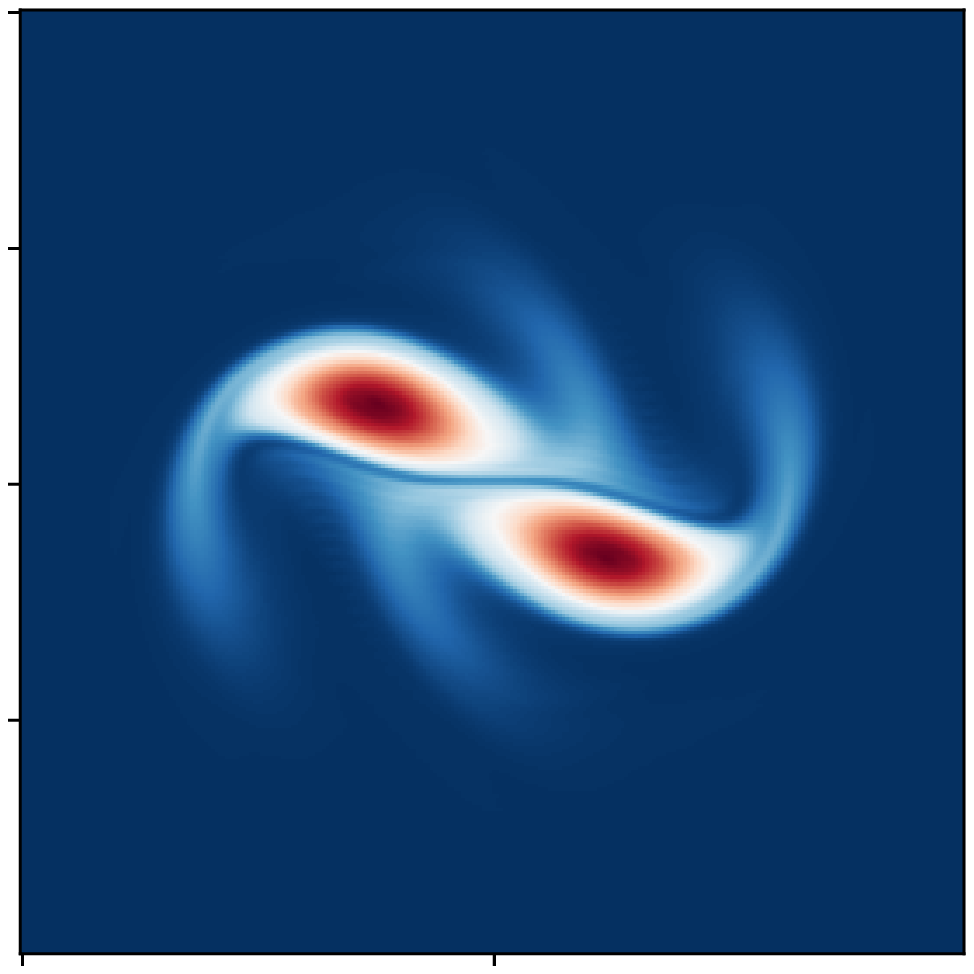} 
 \caption{ground truth, $\bfmu^{\text{test}}_4$}
\end{subfigure}
\begin{subfigure}[b]{0.24\textwidth}
\includegraphics[width=1.0\linewidth]{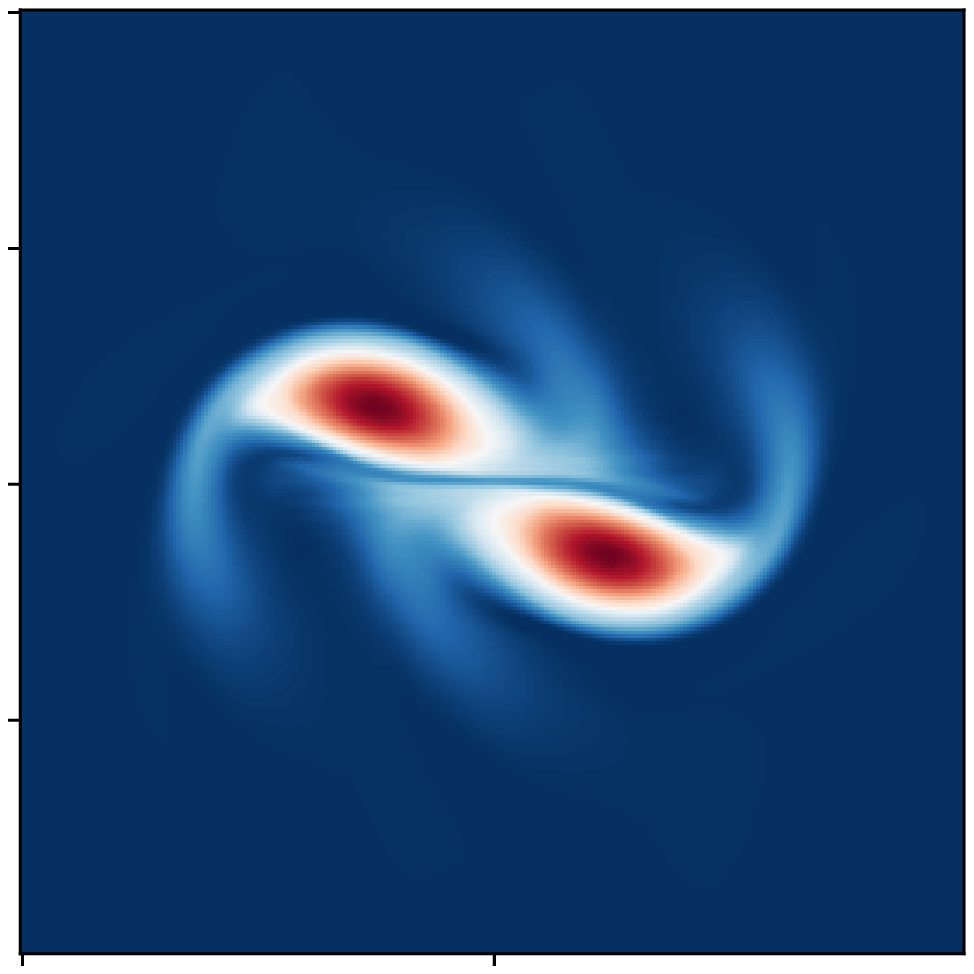} 
 \caption{projection, $\bfmu^{\text{test}}_4$}
\end{subfigure}
\begin{subfigure}[b]{0.24\textwidth}
\includegraphics[width=1.0\linewidth]{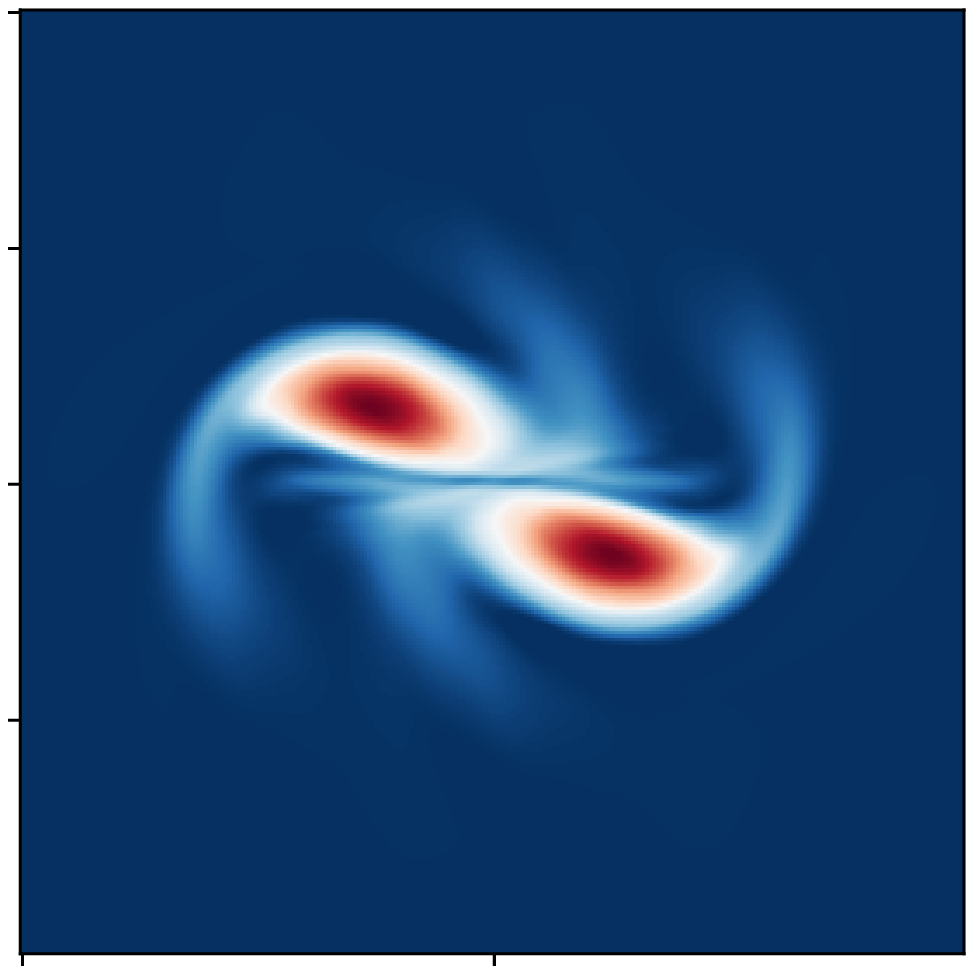} 
\caption{OpInf + roll outs, $\bfmu^{\text{test}}_4$}
\end{subfigure}
\begin{subfigure}[b]{0.24\textwidth}
\includegraphics[width=1.0\linewidth]{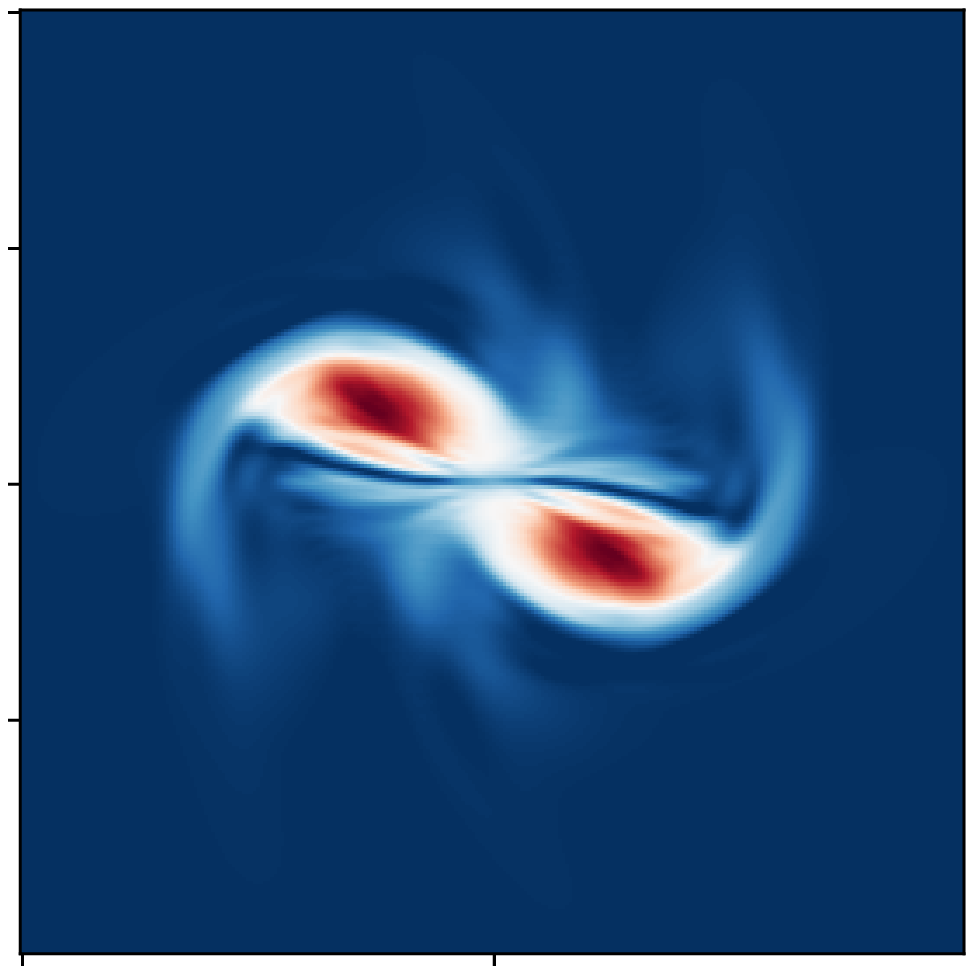} 
\caption{traditional OpInf, $\bfmu^{\text{test}}_4$}
\end{subfigure}

\caption{Surface quasi-geostrophic dynamics (Section~\ref{subsec:pyqg}): Operator inference with roll outs with sampling period $\xi=10$ leads to models that accurately predict the surface buoyancy at the test inputs. In contrast, traditional operator inference without roll outs results in almost unstable predictions for several test inputs and leads to visibly higher errors in the state predictions.}
\label{fig:sqg_IC}
\end{figure}

\begin{figure}[th]
\begin{subfigure}[b]{0.48\textwidth}
\begin{center}
{{\Large\resizebox{1\columnwidth}{!}{\input{SQG_errVsnTrainParam_sparse5}}}}
\end{center}
\caption{sampling period $\xi = 5$}
\label{fig:sqg_ErrVsRoll_sparse5}
\end{subfigure}
\begin{subfigure}[b]{0.48\textwidth}
\begin{center}
{{\Large\resizebox{1\columnwidth}{!}{\input{SQG_errVsnTrainParam_sparse10}}}}
\end{center}
\caption{sampling period $\xi = 10$}
\label{fig:sqg_ErrVsRoll_sparse10}
\end{subfigure}
\caption{Surface quasi-geostrophic dynamics (Section~\ref{subsec:pyqg}): For training data obtained with a large sampling period and from few trajectories only, roll outs compensate to some extent for the information loss by penalizing the misfit in the model predictions at future times. The error of the model obtained with roll outs approaches the projection error as the amount of information is increased. When training with traditional operator inference, the model predictions can be numerically unstable and inaccurate.}
\label{fig:sqg_ErrRollLen}
\end{figure}
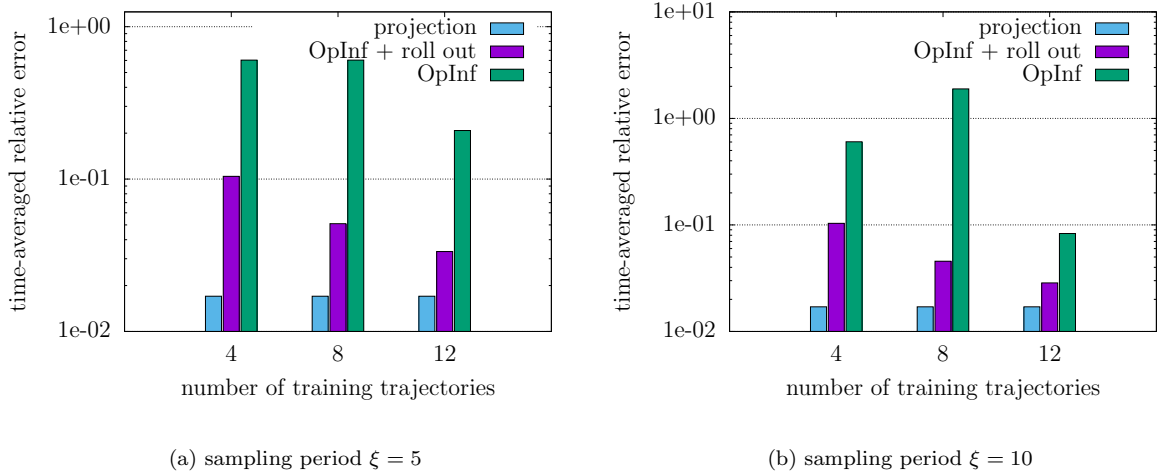

In the following, training data are scarce with respect to two properties: the number of training trajectories and the sampling period $\xi$, which corresponds the number of time steps that lie between two consecutive measurements.

Consider training data obtained with sampling period $\xi=10$ and $M_{\text{train}} = 12$ training trajectories. We set the roll-out length to $R = 200$. The state predictions are shown in Figure~\ref{fig:sqg_IC}.
Operator inference with roll outs trains models that accurately  predict the states corresponding to the test inputs. In contrast, the model obtained with traditional operator inference poorly generalizes to the test inputs. In particular, traditional operator inference leads to models that show instabilities, e.g., Figure~\ref{fig:sqgstates:TOpInfUnstable}. This is because when the training data is only available at sparsely sampled time points, recall that the sampling period is $\xi = 10$ in this experiment, then the approximation of time derivatives via finite differences becomes less reliable.

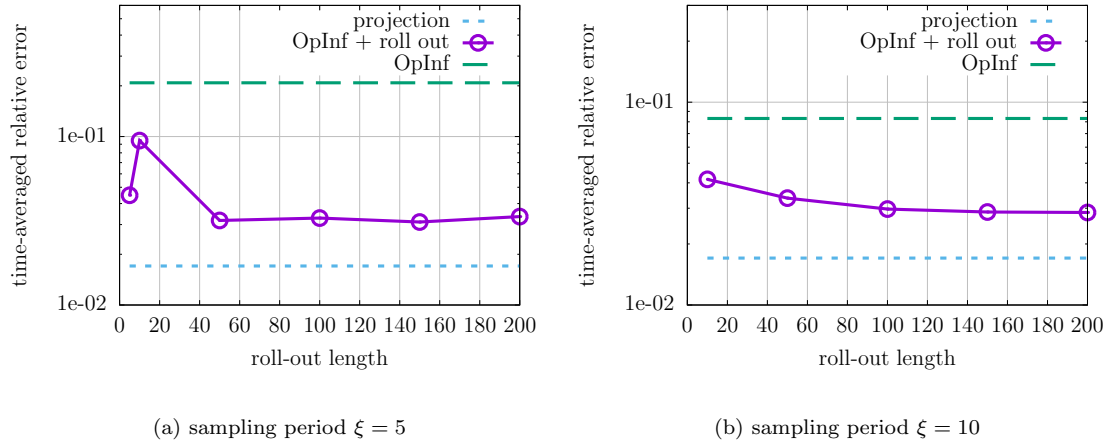
\begin{figure}[th]
\begin{subfigure}[b]{0.45\textwidth}
\begin{center}
{{\Large\resizebox{1\columnwidth}{!}{\input{SQG_errVsRoll_sparse5}}}}
\end{center}
\caption{sampling period $\xi=5$}
\label{fig:sqg_ErrVsnParam_sparse5}
\end{subfigure}
\begin{subfigure}[b]{0.45\textwidth}
\begin{center}
{{\Large\resizebox{1\columnwidth}{!}{\input{SQG_errVsRoll_sparse10}}}}
\end{center}
\caption{sampling period $\xi=10$}
\label{fig:sqg_ErrVsnParam_sparse10}
\end{subfigure}
\caption{Surface quasi-geostrophic dynamics (Section~\ref{subsec:pyqg}): In this example, increasing the roll-out length
improves the prediction accuracy of the learned model.}
\label{fig:sqg_nParamErr}
\end{figure}

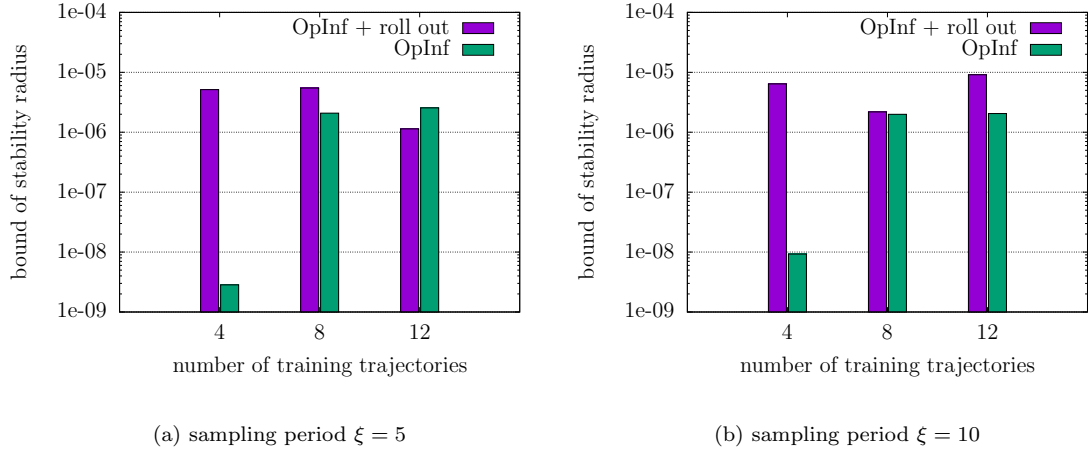
\begin{figure}
\begin{subfigure}[b]{0.45\textwidth}
\begin{center}
{{\Large\resizebox{1\columnwidth}{!}{\input{SQG_errVsnParam_Radius_sparse5}}}}
\end{center}
\caption{sampling period $\xi = 5$}
\label{fig:sqg_nParamRadius_sparse5}
\end{subfigure}
\begin{subfigure}[b]{0.45\textwidth}
\begin{center}
{{\Large\resizebox{1\columnwidth}{!}{\input{SQG_errVsnParam_Radius_sparse10}}}}
\end{center}
\caption{sampling period $\xi = 10$}
\label{fig:sqg_nParamRadius_sparse10}
\end{subfigure}
\caption{Surface quasi-geostrophic dynamics (Section~\ref{subsec:pyqg}): If data are scarce, e.g., the number of training trajectories is four, then roll outs help operator inference to achieve an orders of magnitude larger bound of stability radius in this example. As more training data are used, e.g., by increasing the number of training trajectories, operator inference without roll outs achieves comparable bounds of stability radii.}
\label{fig:sqg_nParam_radius}
\end{figure}

To quantify the accuracy of the learned models, Figure~\ref{fig:sqg_ErrRollLen} plots the test error \eqref{eq:TestRelErr} over the number of training trajectories for sampling periods $\xi=5$ and $\xi=10$.
The error of the model obtained with roll outs decreases as the amount of training data is increased. For traditional operator inference, several of the obtained models resulted in unstable predictions, and so we replaced their predictions by the initial conditions in computing the shown errors. The results indicate that the lack of training data is compensated by roll outs via solving a modified optimization problem that penalizes discrepancies between the data and model predictions at future times. In Figure~\ref{fig:sqg_nParamErr}, we fix the number of training trajectories to $M_{\text{train}}=12$ and plot the error over the roll length for sampling periods $\xi = 5$ and $\xi = 10$. In this example, increasing the roll-out length helps to decrease the prediction error of the learned model.

Figure~\ref{fig:sqg_nParam_radius} plots the bound of the stability radii, averaged over 1000 realizations of the random matrix $\bfL$ with standard normal entries.  We set $R = 200$. When the training data are very scarce, e.g., $M_{\text{train}}=4$, then the bound of the stability radius of the learned models from operator inference with roll outs is larger by up to three orders of magnitude compared to the models learned with traditional operator inference. However, as more training data are used, e.g., by increasing the sampling period and by increasing the number of trajectories, the bounds of the stability radii of the models learned with traditional operator inference improves. Indeed, the model learned via traditional operator inference in this example appears to be stable despite being inaccurate; see~Figure~\ref{fig:sqg_IC}.

\subsubsection{Surface quasi-geostrophic dynamics: Scarce data that are also noisy}

\begin{figure}
\begin{subfigure}[b]{0.24\textwidth}
\includegraphics[width=1.0\linewidth]{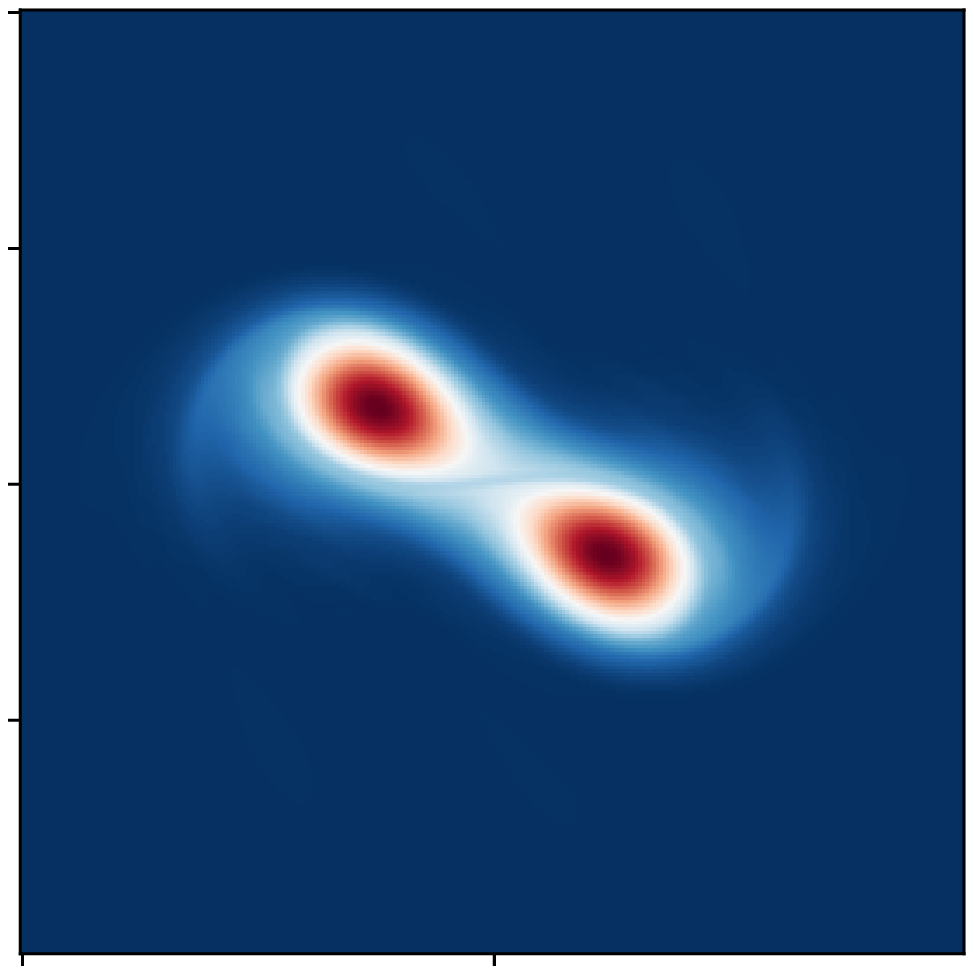} 
 \caption{ground truth, $\bfmu^{\text{test}}_1$}
\end{subfigure}
\begin{subfigure}[b]{0.24\textwidth}
\includegraphics[width=1.0\linewidth]{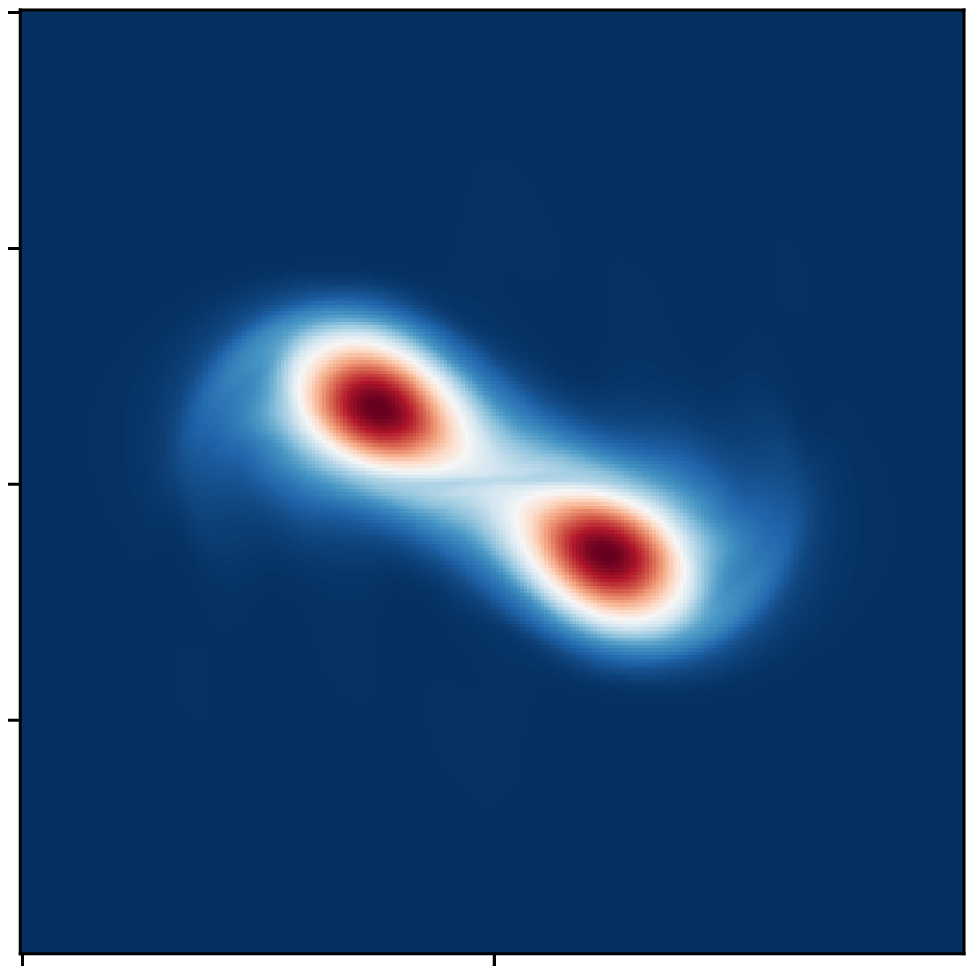} 
 \caption{projection, $\bfmu^{\text{test}}_1$}
\end{subfigure}
\begin{subfigure}[b]{0.24\textwidth}
\includegraphics[width=1.0\linewidth]{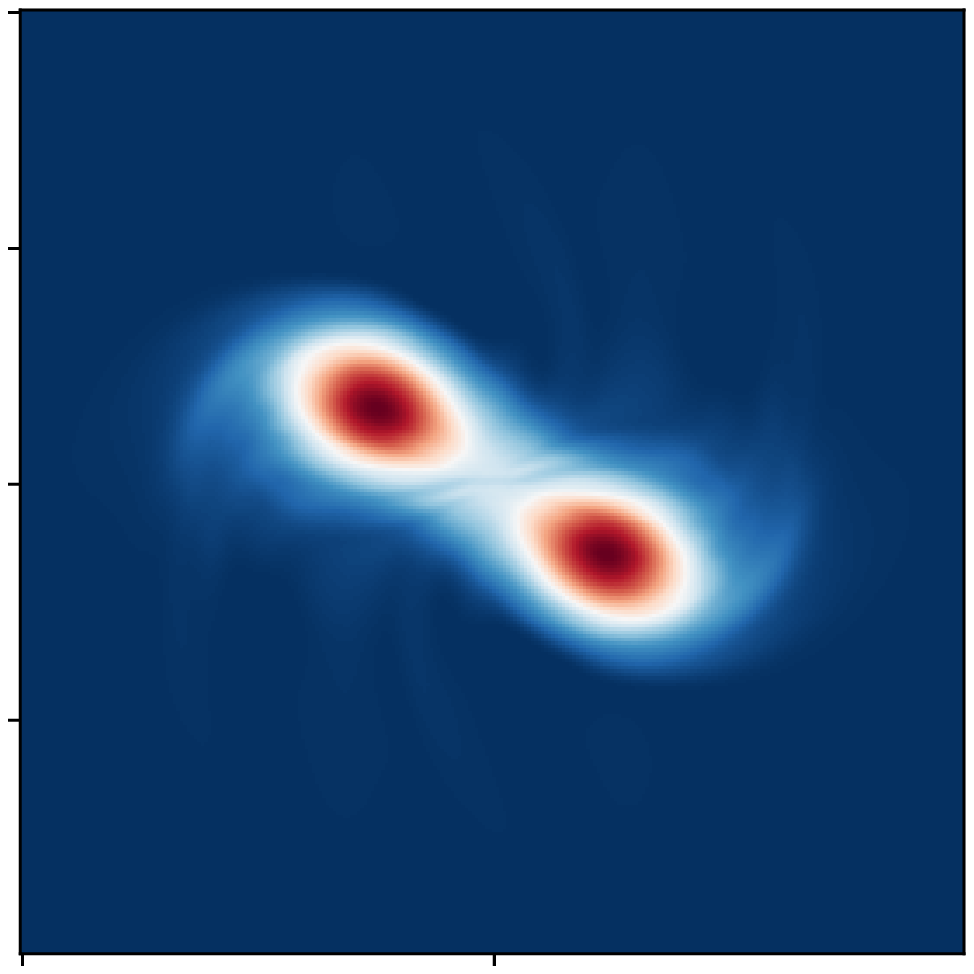} 
\caption{OpInf + roll outs, $\bfmu^{\text{test}}_1$}
\end{subfigure}
\begin{subfigure}[b]{0.24\textwidth}
\includegraphics[width=1.0\linewidth]{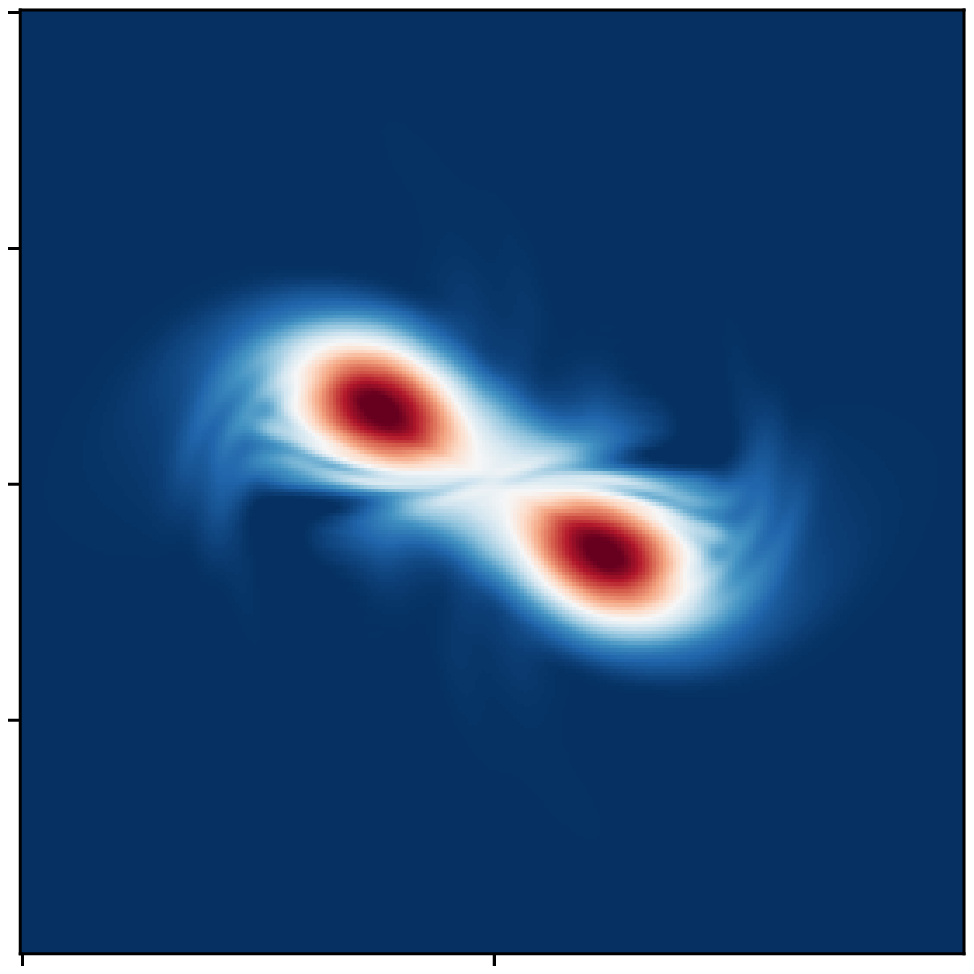} 
\caption{traditional OpInf, $\bfmu^{\text{test}}_1$}
\end{subfigure}

\begin{subfigure}[b]{0.24\textwidth}
\includegraphics[width=1.0\linewidth]{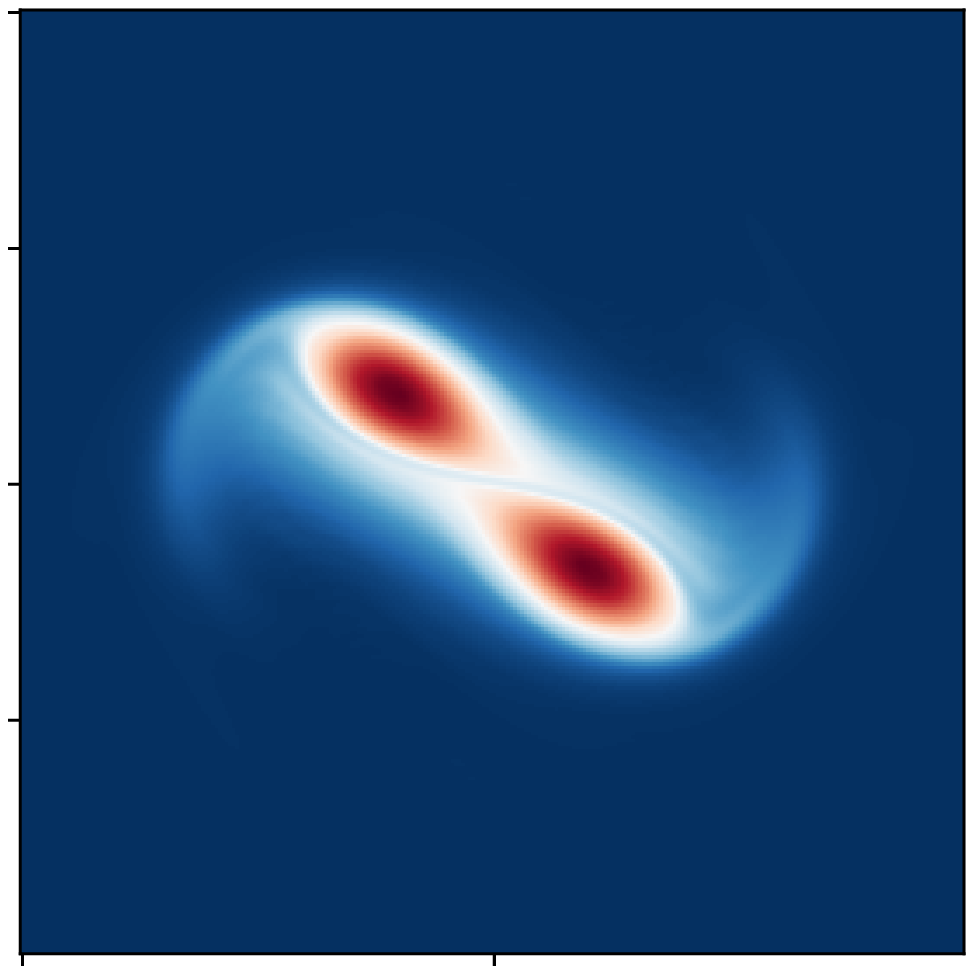} 
 \caption{ground truth, $\bfmu^{\text{test}}_2$}
\end{subfigure}
\begin{subfigure}[b]{0.24\textwidth}
\includegraphics[width=1.0\linewidth]{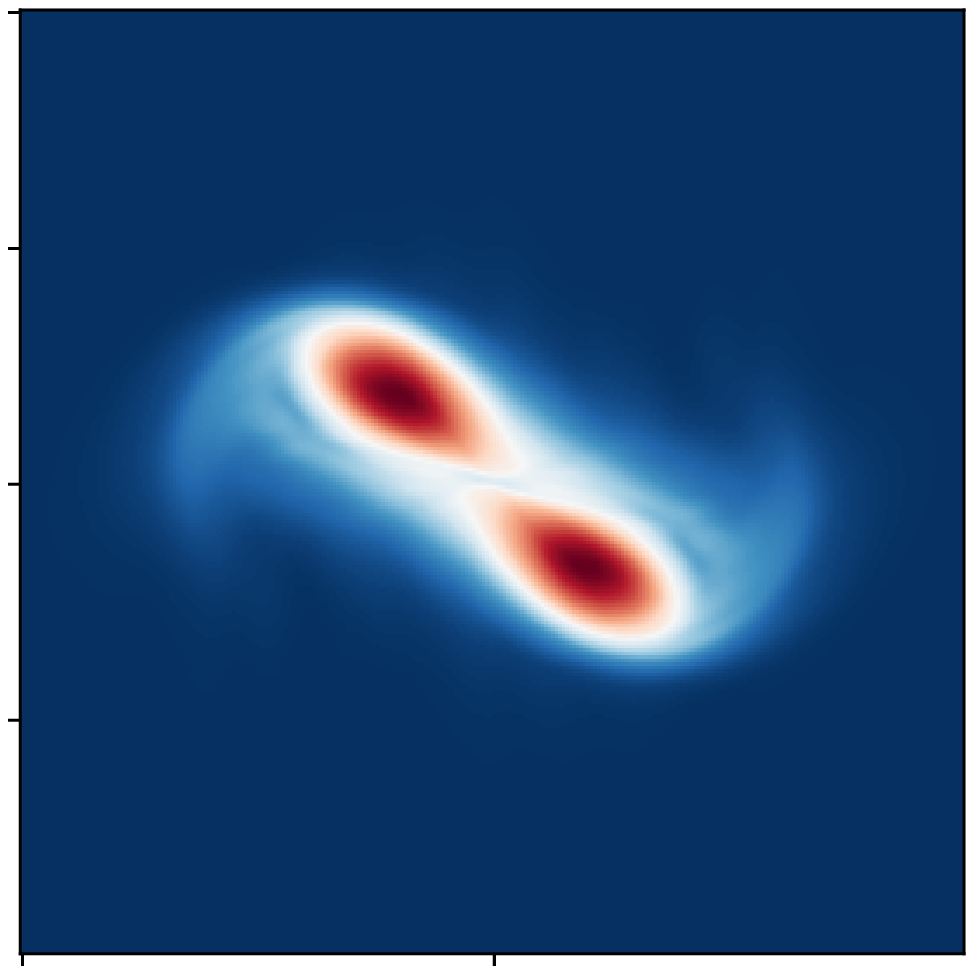} 
 \caption{projection, $\bfmu^{\text{test}}_2$}
\end{subfigure}
\begin{subfigure}[b]{0.24\textwidth}
\includegraphics[width=1.0\linewidth]{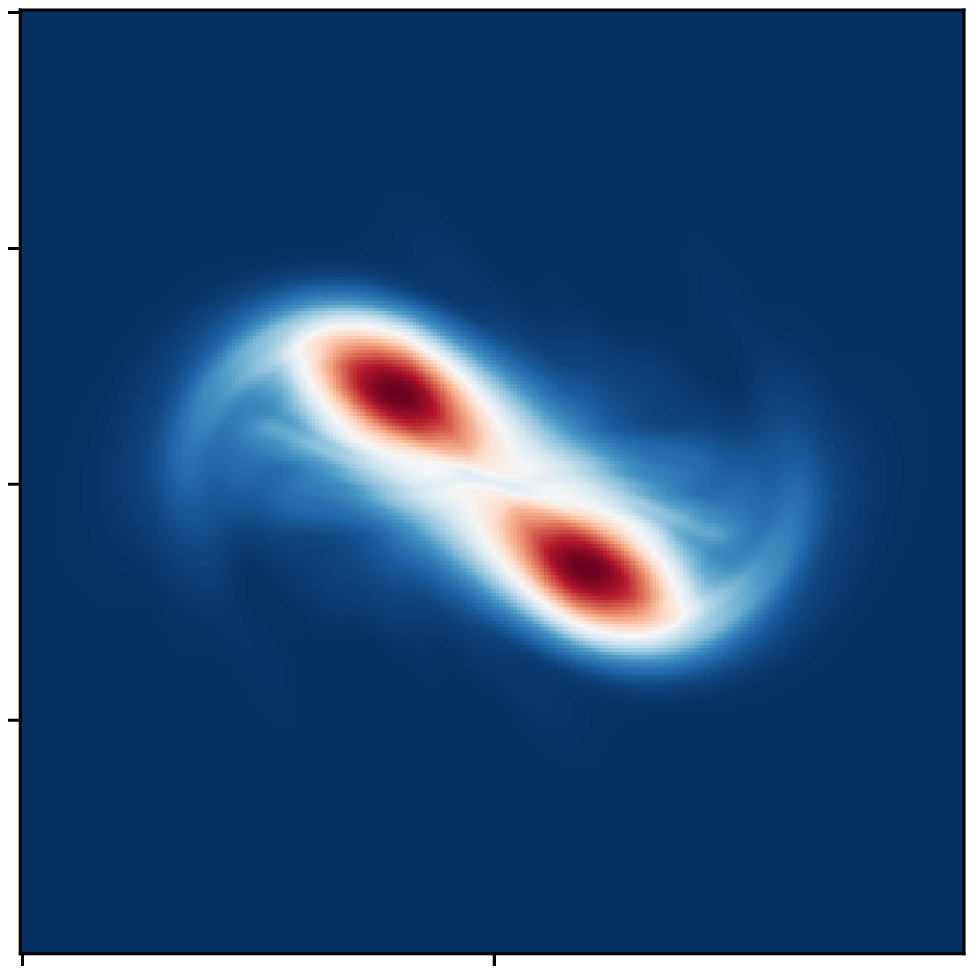} 
\caption{OpInf + roll outs, $\bfmu^{\text{test}}_2$}
\end{subfigure}
\begin{subfigure}[b]{0.24\textwidth}
\includegraphics[width=1.0\linewidth]{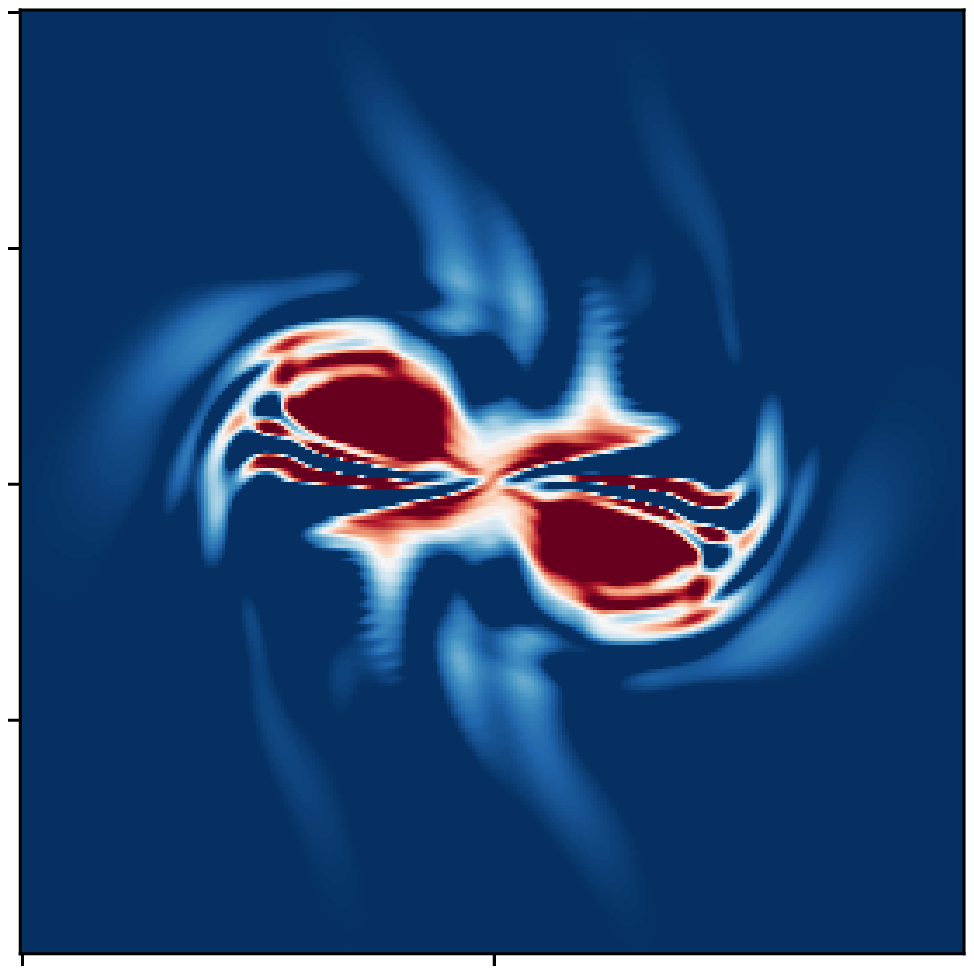} 
\caption{traditional OpInf, $\bfmu^{\text{test}}_2$}
\end{subfigure}

\begin{subfigure}[b]{0.24\textwidth}
\includegraphics[width=1.0\linewidth]{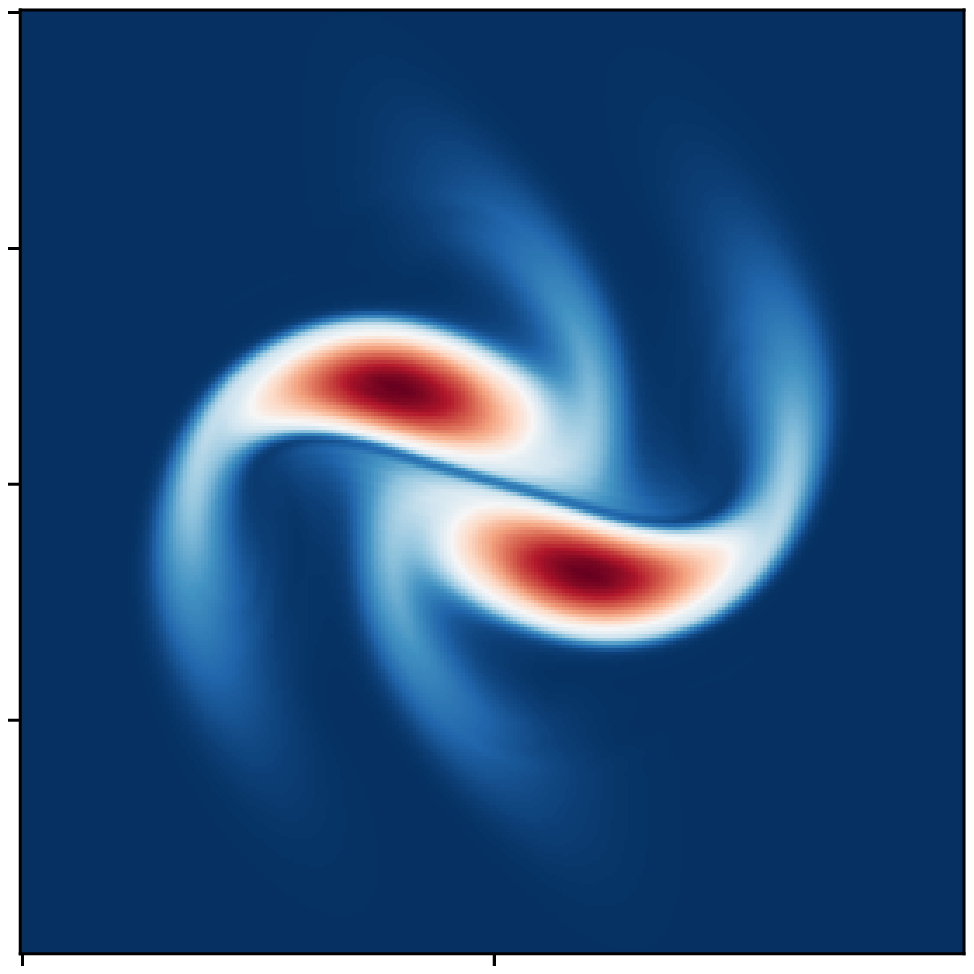} 
 \caption{ground truth, $\bfmu^{\text{test}}_3$}
\end{subfigure}
\begin{subfigure}[b]{0.24\textwidth}
\includegraphics[width=1.0\linewidth]{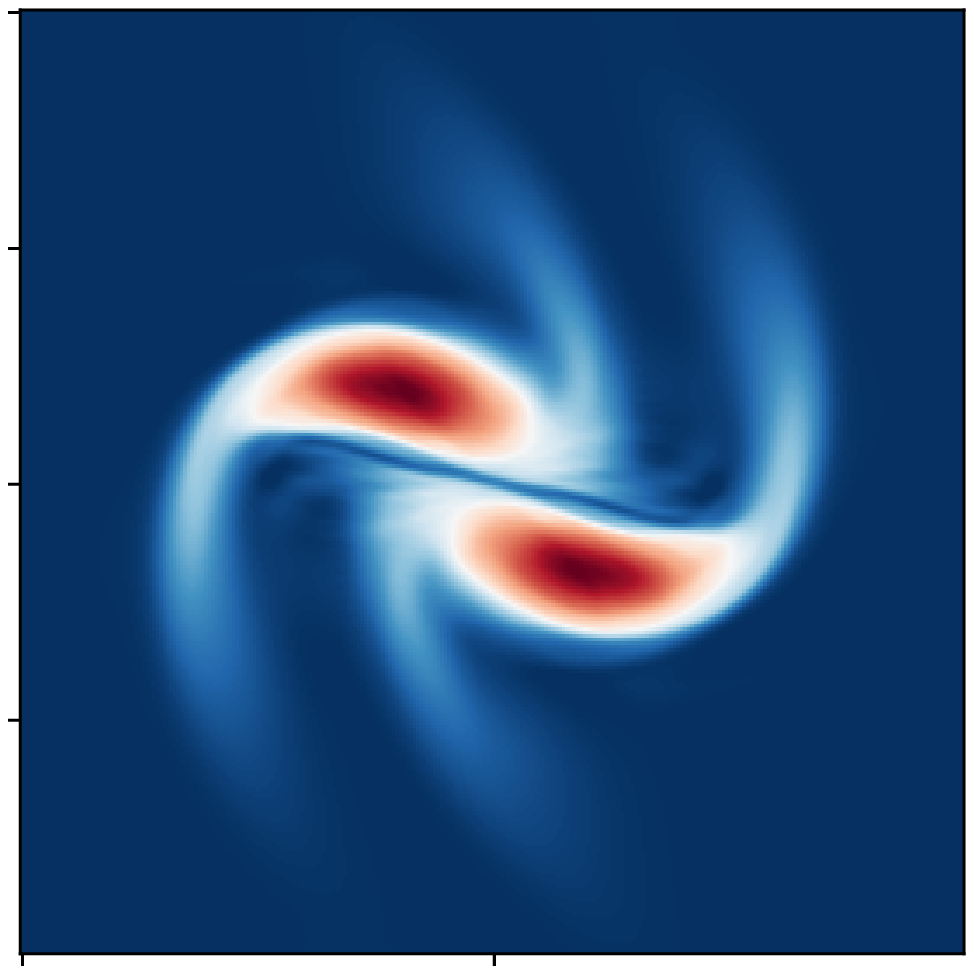} 
 \caption{projection, $\bfmu^{\text{test}}_3$}
\end{subfigure}
\begin{subfigure}[b]{0.24\textwidth}
\includegraphics[width=1.0\linewidth]{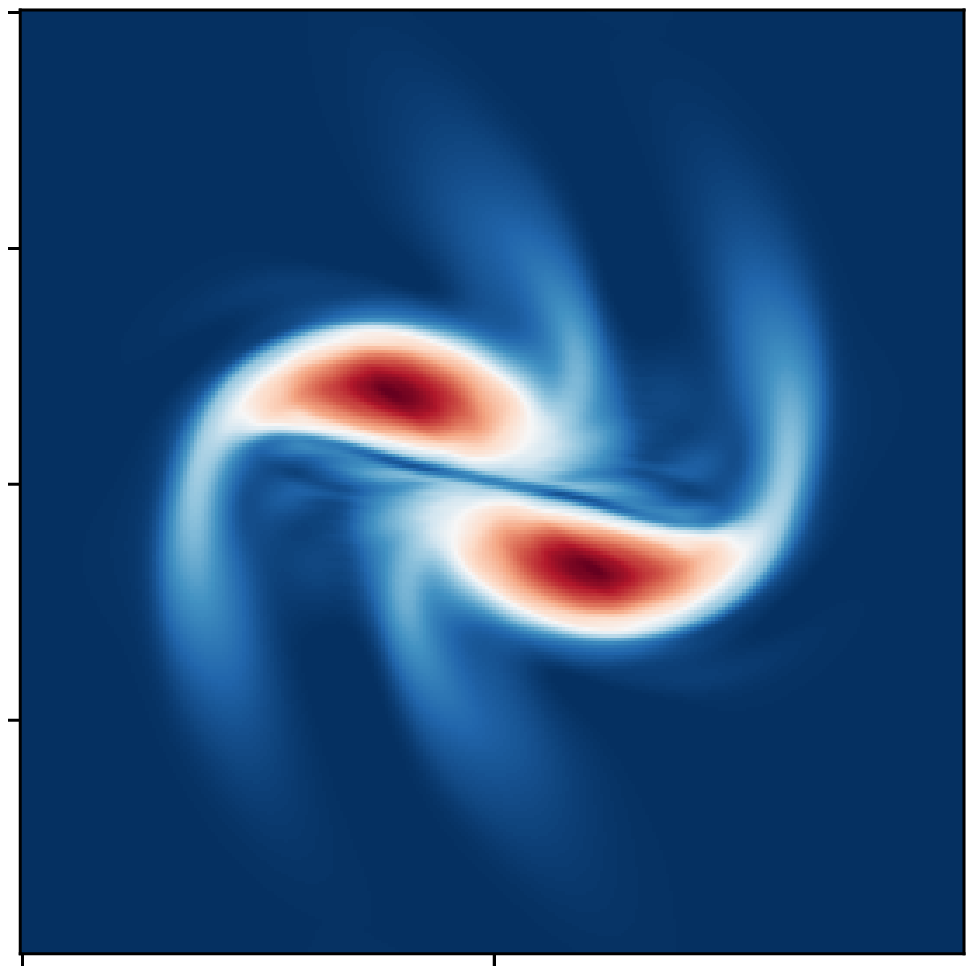} 
\caption{OpInf + roll outs, $\bfmu^{\text{test}}_3$}
\end{subfigure}
\begin{subfigure}[b]{0.24\textwidth}
\includegraphics[width=1.0\linewidth]{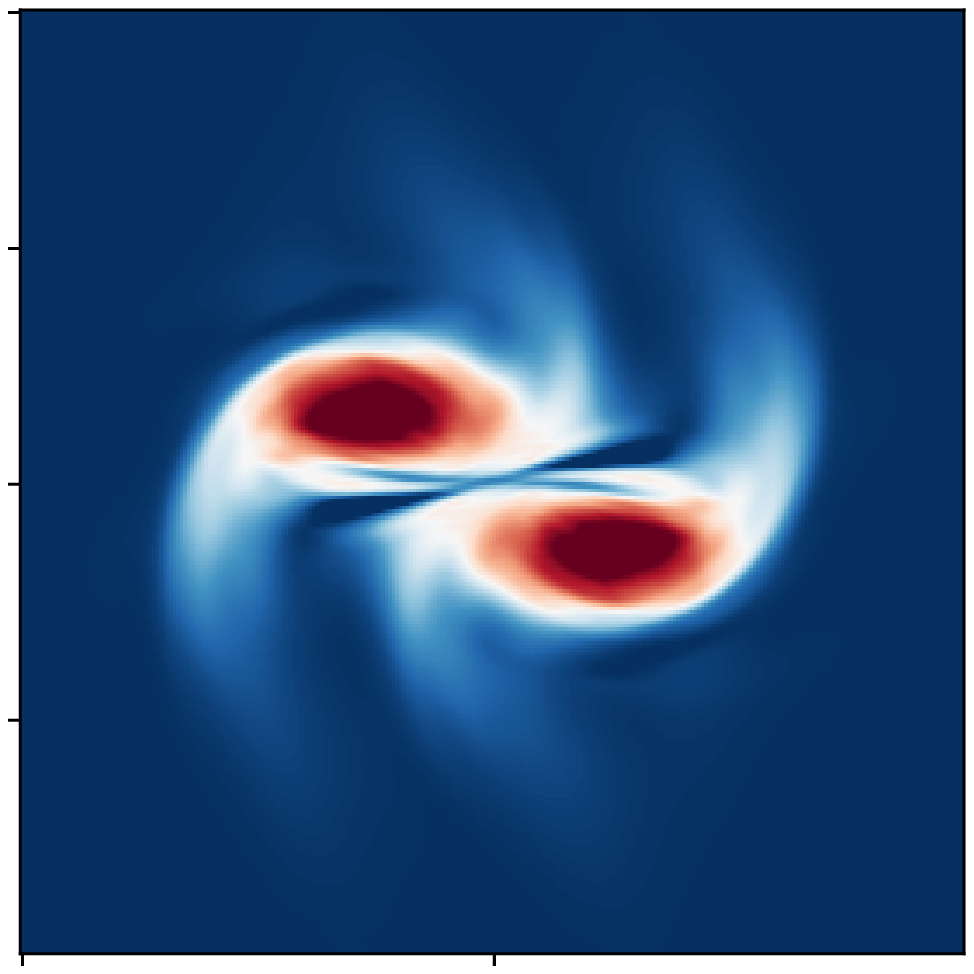} 
\caption{traditional OpInf, $\bfmu^{\text{test}}_3$}
\end{subfigure}

\begin{subfigure}[b]{0.24\textwidth}
\includegraphics[width=1.0\linewidth]{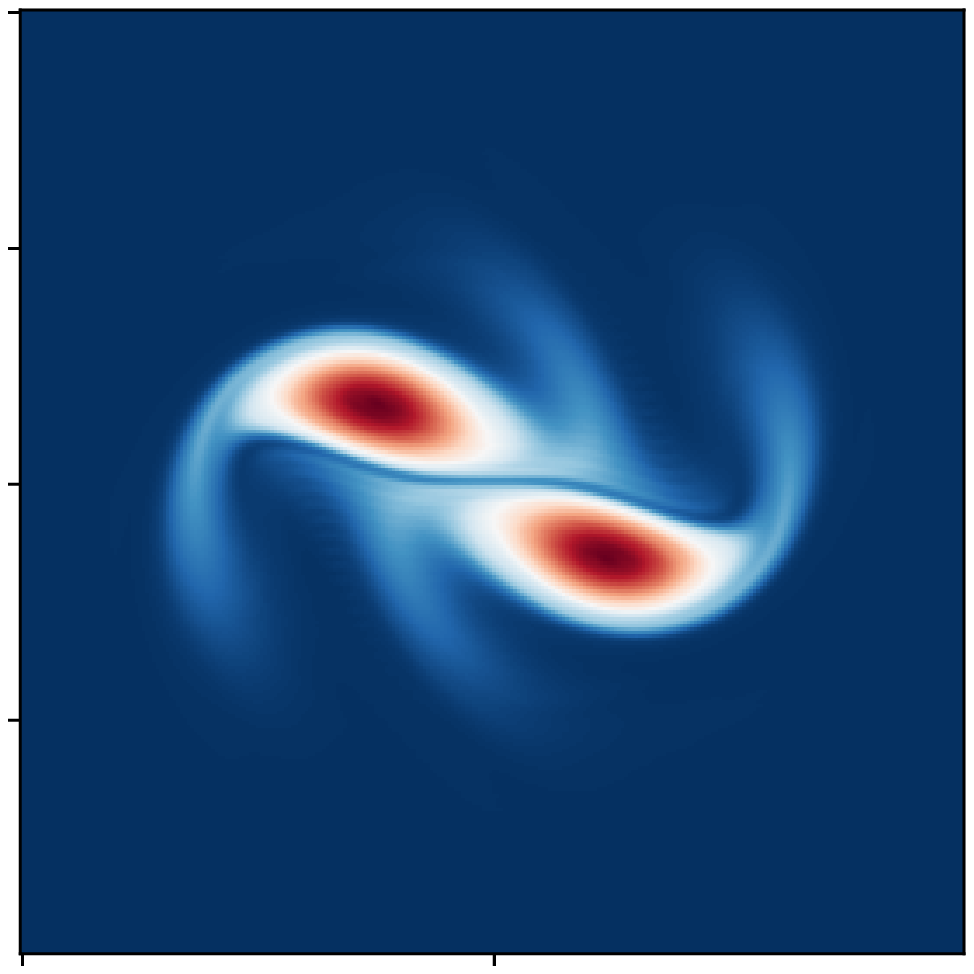} 
 \caption{ground truth, $\bfmu^{\text{test}}_4$}
\end{subfigure}
\begin{subfigure}[b]{0.24\textwidth}
\includegraphics[width=1.0\linewidth]{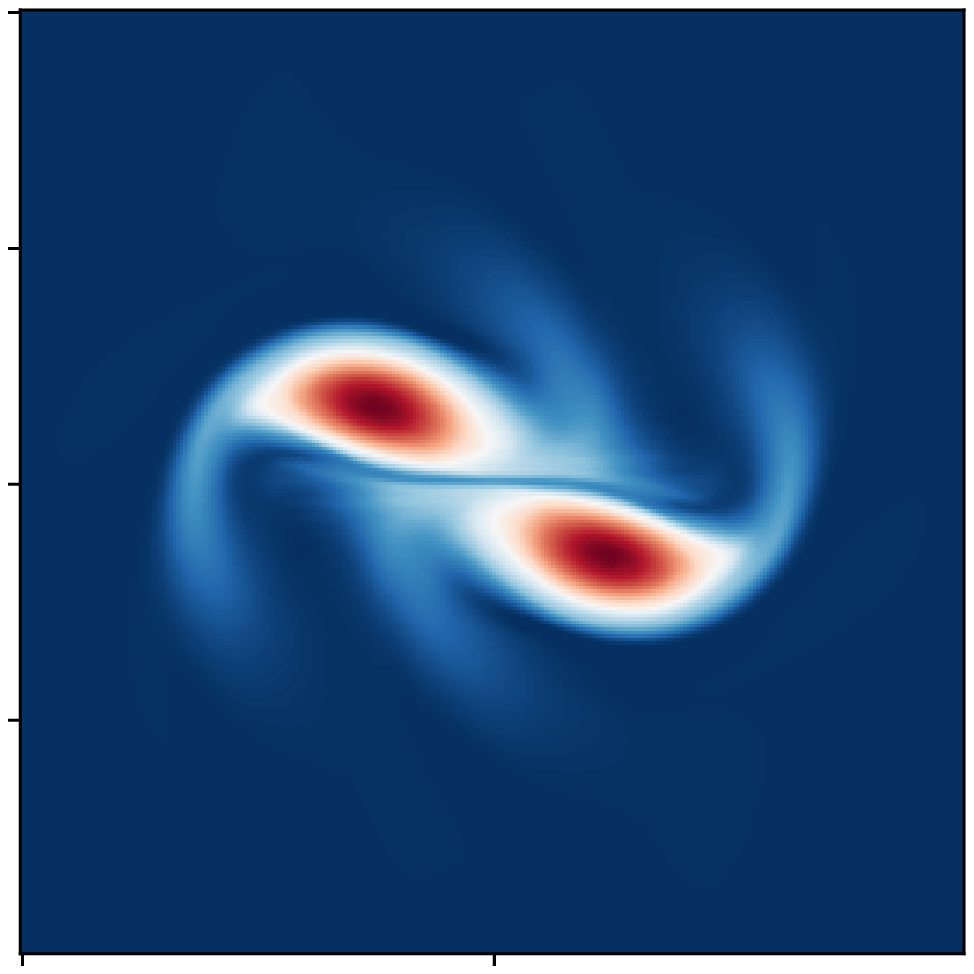} 
 \caption{projection, $\bfmu^{\text{test}}_4$}
\end{subfigure}
\begin{subfigure}[b]{0.24\textwidth}
\includegraphics[width=1.0\linewidth]{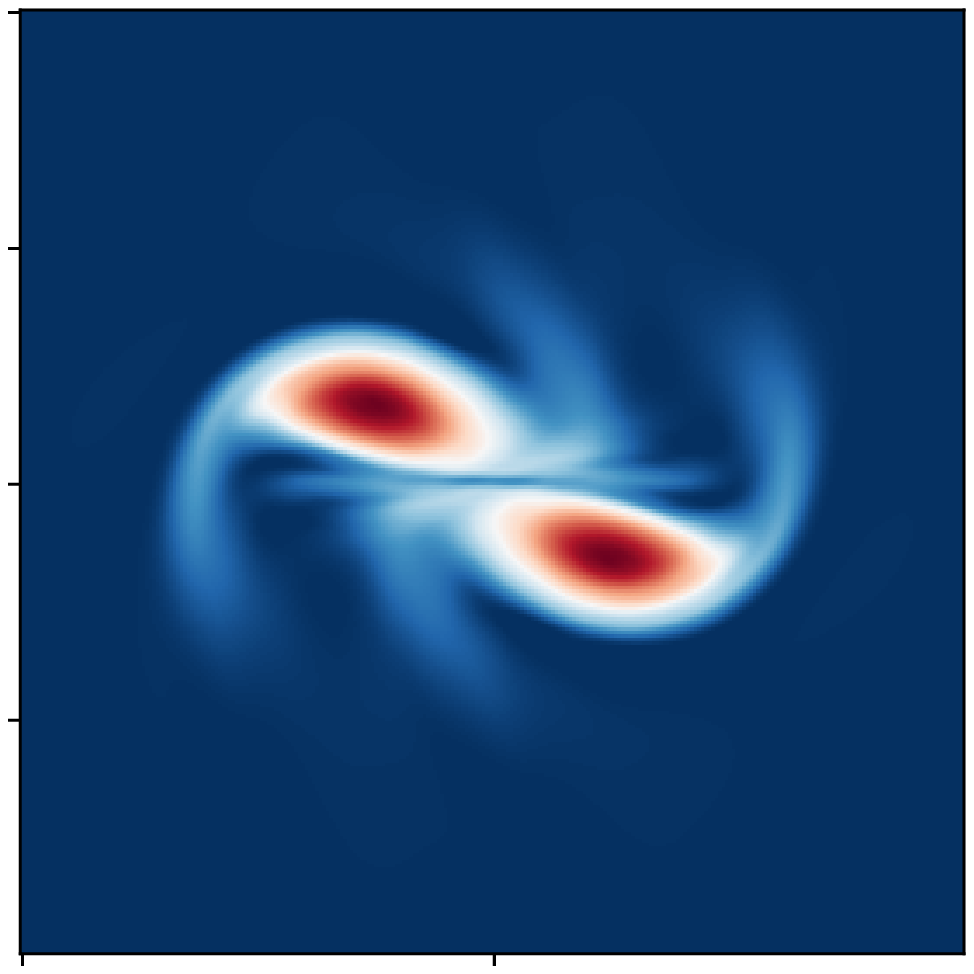} 
\caption{OpInf + roll outs, $\bfmu^{\text{test}}_4$}
\end{subfigure}
\begin{subfigure}[b]{0.24\textwidth}
\includegraphics[width=1.0\linewidth]{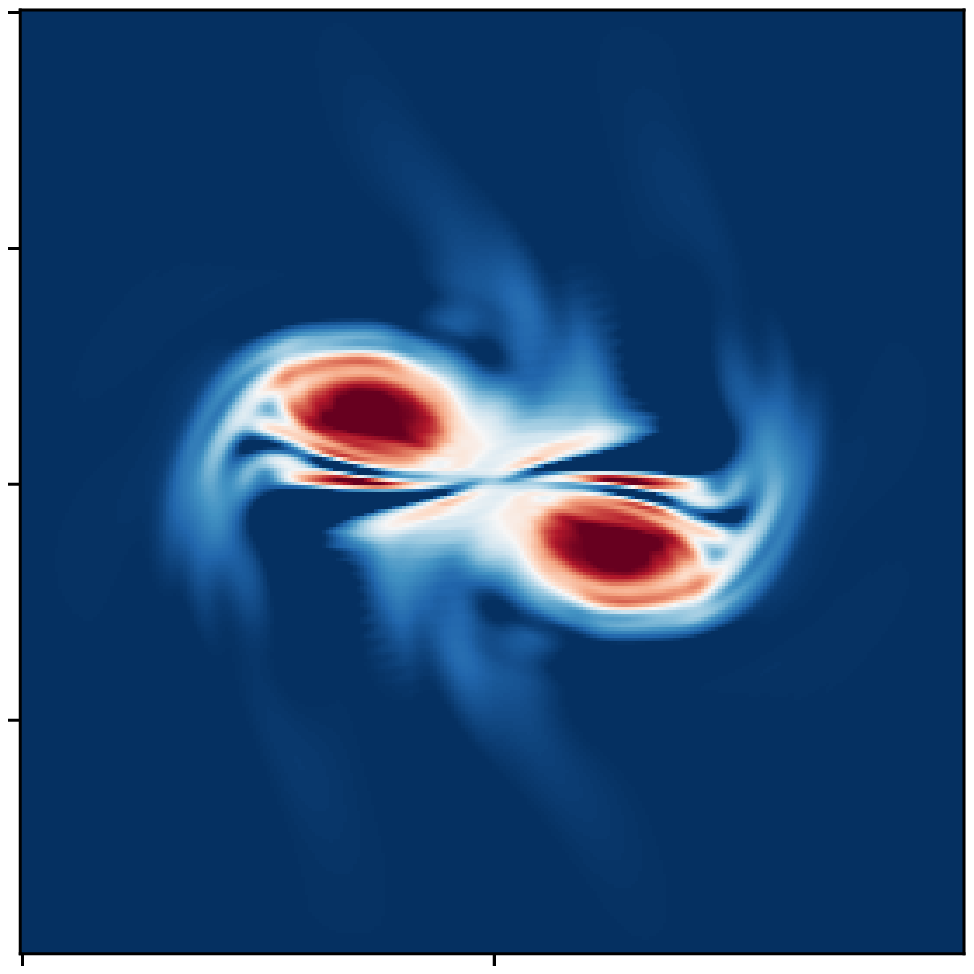} 
\caption{traditional OpInf, $\bfmu^{\text{test}}_4$}
\end{subfigure}

\caption{Surface quasi-geostrophic dynamics (Section~\ref{subsec:pyqg}): Operator inference with roll outs ($M_{\text{train}}=12, R=200, \xi=5)$ trained on data with $\rho=10\%$ noise, leads to models that accurately predict the surface buoyancy at the test parameters. In contrast, traditional operator inference without roll outs fails to make meaningful predictions across various test parameter values.}
\label{fig:sqg_noise_IC}
\end{figure}

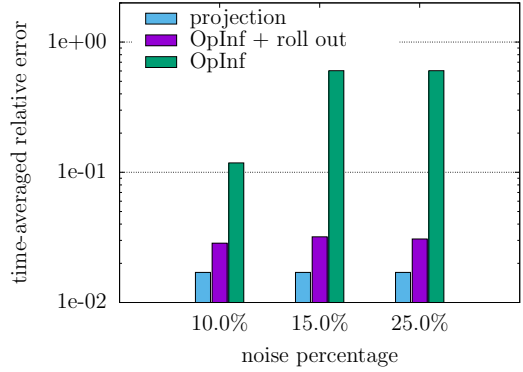
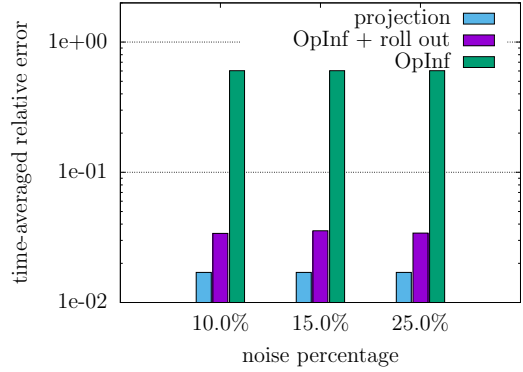
\begin{figure}
\begin{subfigure}[b]{0.45\textwidth}
\begin{center}
{{\Large\resizebox{1\columnwidth}{!}{\input{SQG_noise_errVsNoise_sparse5}}}}
\end{center}
\caption{sampling period $\xi=5$}
\label{fig:sqg_noiseVsErr_sparse5}
\end{subfigure}
\begin{subfigure}[b]{0.45\textwidth}
\begin{center}
{{\Large\resizebox{1\columnwidth}{!}{\input{SQG_noise_errVsNoise_sparse10}}}}
\end{center}
\caption{sampling period $\xi=10$}
\label{fig:sqg_noiseVsErr_sparse10}
\end{subfigure}
\caption{Surface quasi-geostrophic dynamics (Section~\ref{subsec:pyqg}): Operator inference with roll outs leads to models that are predictive even when the training data are polluted by 25\% noise. The accuracy achieved by operator inference with roll outs is close to the projection error of the ground-truth data on the reduced space. In contrast, traditional operator inference mostly learns unstable models that cannot make meaningful predictions at test inputs.}
\label{fig:sqg_noiseVsErr}
\end{figure}

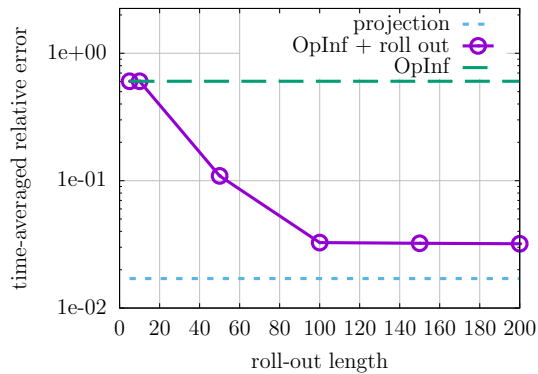
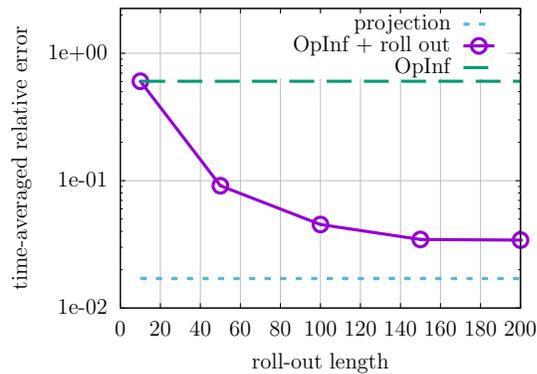
\begin{figure}
\begin{subfigure}[b]{0.45\textwidth}
\begin{center}
{{\Large\resizebox{1\columnwidth}{!}{\input{SQG_noise_errVsRoll_sparse5}}}}
\end{center}
\caption{sampling period $\xi = 5$}
\label{fig:sqg_noise_roll_sparse5}
\end{subfigure}
\begin{subfigure}[b]{0.45\textwidth}
\begin{center}
{{\Large\resizebox{1\columnwidth}{!}{\input{SQG_noise_errVsRoll_sparse10}}}}
\end{center}
\caption{sampling period $\xi = 10$}
\label{fig:sqg_noise_roll_sparse10}
\end{subfigure}
\caption{Surface quasi-geostrophic dynamics (Section~\ref{subsec:pyqg}):  When training data are both scarce and noisy, increasing the roll length leads to more accurate models because the objective function accounts for the model misfit at multiple times for larger roll-out lengths. Thus, operator inference with roll outs prevents overfitting to the noise and achieves a prediction error that is close to the projection error.
}
\label{fig:sqg_noise_rollVsErr}
\end{figure}

\begin{figure}
\begin{subfigure}[b]{0.45\textwidth}
\begin{center}
{{\Large\resizebox{1\columnwidth}{!}{\input{SQG_errVsNoise_radius_sparse5}}}}
\end{center}
\caption{sampling period $\xi=5$}
\label{fig:sqg_noise_radius_sparse5}
\end{subfigure}
\begin{subfigure}[b]{0.45\textwidth}
\begin{center}
{{\Large\resizebox{1\columnwidth}{!}{\input{SQG_errVsNoise_radius_sparse10}}}}
\end{center}
\caption{sampling period $\xi=10$}
\label{fig:sqg_noise_radius_sparse10}
\end{subfigure}
\caption{Surface quasi-geostrophic dynamics (Section~\ref{subsec:pyqg}): Models learned with roll outs achieve larger bounds of the stability radii  than the models obtained via traditional operator inference in this example.}
\label{fig:sqg_noise_radius}
\end{figure}
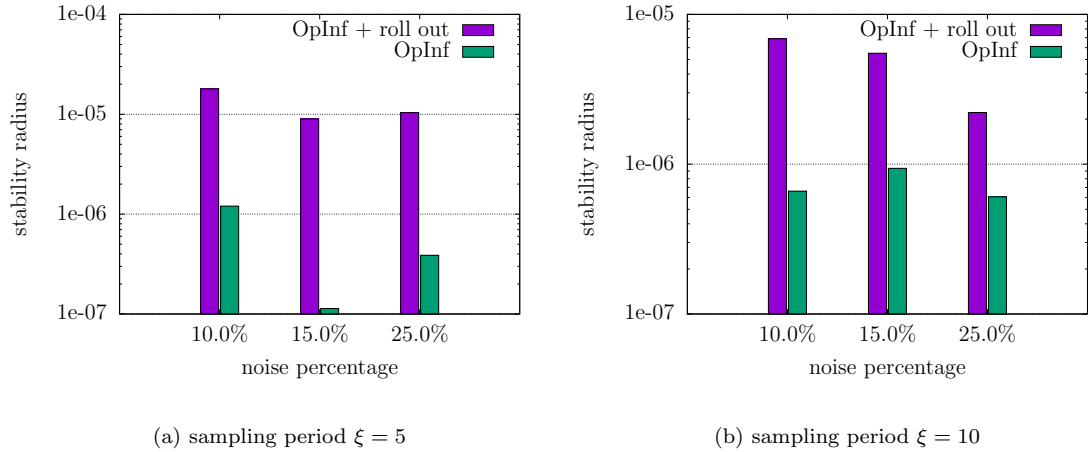

We now add noise to the training data in the same way as described in Section~\ref{subsubsec:SWE_scarceNoisy}, with $\rho$ controlling the standard deviation of the noise. We plot the buoyancy obtained with the learned models in Figure~\ref{fig:sqg_noise_IC},  where $R = 200, M_{\text{train}}= 12, \xi = 5, \rho = 10\%$. The  plots demonstrate that in the scarce and noisy data regime, operator inference with roll outs offers an approximation that closely resembles the projected states, which is the best approximation possible in the low-dimensional subspace. Meanwhile, traditional operator inference fails to provide meaningful predictions across various test inputs.

Figure~\ref{fig:sqg_noiseVsErr} shows the  test error \eqref{eq:TestRelErr} of the model predictions with respect to the standard deviation given by $\rho$ of the noise. The roll-out length is set to $R = 200$ and the number of training trajectories is $M_{\text{train}} = 12$.
As can be seen, despite the increase in the standard deviation of the noise, we only observe a modest increase in the prediction error of the model learned with roll outs.
In the scarce and noisy data regime, traditional operator inference leads to numerically unstable models quickly. In these cases, for computing the error, the initial condition serves as a surrogate for the model prediction at all times.

In Figure~\ref{fig:sqg_noise_rollVsErr}, we visualize the error over the roll length for $\rho=25\%$ and $M_{\text{train}}=12$. The results shown in the plots demonstrate that in this example, increasing the roll length improves the accuracy and stability of the learned model by up to two orders of magnitude. Increasing the roll length penalizes the misfit between data and predictions at more time points, thus, the learned model is less likely to overfit to the noise in the data.

Let us now investigate the stability radii of the learned models; we follow the same computation of the bounds of the stability radii as in the previous examples. The results are shown in Figure~\ref{fig:sqg_noise_radius} over the standard deviation given by $\rho$ for the sampling periods $\xi=5$ and $\xi=10$, with fixed $R=200,M_{\text{train}}=12$. The models learned with roll outs achieve bounds of the stability radii that are up to two orders of magnitude larger than those of the models learned with traditional operator inference. This is in agreement with our previous numerical experiments, where traditional operator inference leads to numerically unstable models when noise is added and data are scarce.

\section{Conclusions}\label{sec:Conc}
Operator inference with roll outs combines the interpretability, scalability, and structure preservation of traditional operator inference with the stability and robustness of dynamic training given by the roll outs of Neural ODEs. The roll outs are especially helpful to obtain stable and robust models when data are noisy and scarce.
Because we insist on the polynomial structure of the models, the learned models are amenable for analysis and interpretation. In particular, the stability bias imposed by the roll outs can be observed in increased bounds of the stability radii of the domain of attraction of the learned models. Furthermore, linear dynamics can be interpolated in a structure preserving way to make predictions at new, previously unseen physical inputs. Finally, the polynomial structure is also in agreement with wide engineering practice, where non-linear behavior is often approximated via polynomials corresponding to truncated Taylor expansions. While the training is more involved, operator inference with roll outs is straightforward to implement with automatic differentiation and tractable in large settings due to advances in stochastic gradient descent methods. The numerical experiments with shallow water wave and surface quasi-geostrophic dynamics demonstrate the increased robustness and stability of the models learned with roll outs.

\section*{Acknowledgements} The first and third author acknowledge partially supported by US Department of Energy, Office of Advanced
Scientific Computing Research, Applied Mathematics Program (Program Manager Dr. Steven Lee), DOE
Award DESC0019334. The third author also acknowledges funding from the National Science Foundation under Grant No.~1901091. This work was supported in part through the NYU IT High Performance Computing resources, services, and staff expertise.

\bibliography{main.bib}

\end{document}

%% file: shallow_errVsnTrainParam.tex
\begingroup
  \makeatletter
  \providecommand\color[2][]{%
    \GenericError{(gnuplot) \space\space\space\@spaces}{%
      Package color not loaded in conjunction with
      terminal option `colourtext'%
    }{See the gnuplot documentation for explanation.%
    }{Either use 'blacktext' in gnuplot or load the package
      color.sty in LaTeX.}%
    \renewcommand\color[2][]{}%
  }%
  \providecommand\includegraphics[2][]{%
    \GenericError{(gnuplot) \space\space\space\@spaces}{%
      Package graphicx or graphics not loaded%
    }{See the gnuplot documentation for explanation.%
    }{The gnuplot epslatex terminal needs graphicx.sty or graphics.sty.}%
    \renewcommand\includegraphics[2][]{}%
  }%
  \providecommand\rotatebox[2]{#2}%
  \@ifundefined{ifGPcolor}{%
    \newif\ifGPcolor
    \GPcolortrue
  }{}%
  \@ifundefined{ifGPblacktext}{%
    \newif\ifGPblacktext
    \GPblacktexttrue
  }{}%
  \let\gplgaddtomacro\g@addto@macro
  \gdef\gplbacktext{}%
  \gdef\gplfronttext{}%
  \makeatother
  \ifGPblacktext
    \def\colorrgb#1{}%
    \def\colorgray#1{}%
  \else
    \ifGPcolor
      \def\colorrgb#1{\color[rgb]{#1}}%
      \def\colorgray#1{\color[gray]{#1}}%
      \expandafter\def\csname LTw\endcsname{\color{white}}%
      \expandafter\def\csname LTb\endcsname{\color{black}}%
      \expandafter\def\csname LTa\endcsname{\color{black}}%
      \expandafter\def\csname LT0\endcsname{\color[rgb]{1,0,0}}%
      \expandafter\def\csname LT1\endcsname{\color[rgb]{0,1,0}}%
      \expandafter\def\csname LT2\endcsname{\color[rgb]{0,0,1}}%
      \expandafter\def\csname LT3\endcsname{\color[rgb]{1,0,1}}%
      \expandafter\def\csname LT4\endcsname{\color[rgb]{0,1,1}}%
      \expandafter\def\csname LT5\endcsname{\color[rgb]{1,1,0}}%
      \expandafter\def\csname LT6\endcsname{\color[rgb]{0,0,0}}%
      \expandafter\def\csname LT7\endcsname{\color[rgb]{1,0.3,0}}%
      \expandafter\def\csname LT8\endcsname{\color[rgb]{0.5,0.5,0.5}}%
    \else
      \def\colorrgb#1{\color{black}}%
      \def\colorgray#1{\color[gray]{#1}}%
      \expandafter\def\csname LTw\endcsname{\color{white}}%
      \expandafter\def\csname LTb\endcsname{\color{black}}%
      \expandafter\def\csname LTa\endcsname{\color{black}}%
      \expandafter\def\csname LT0\endcsname{\color{black}}%
      \expandafter\def\csname LT1\endcsname{\color{black}}%
      \expandafter\def\csname LT2\endcsname{\color{black}}%
      \expandafter\def\csname LT3\endcsname{\color{black}}%
      \expandafter\def\csname LT4\endcsname{\color{black}}%
      \expandafter\def\csname LT5\endcsname{\color{black}}%
      \expandafter\def\csname LT6\endcsname{\color{black}}%
      \expandafter\def\csname LT7\endcsname{\color{black}}%
      \expandafter\def\csname LT8\endcsname{\color{black}}%
    \fi
  \fi
    \setlength{\unitlength}{0.0500bp}%
    \ifx\gptboxheight\undefined%
      \newlength{\gptboxheight}%
      \newlength{\gptboxwidth}%
      \newsavebox{\gptboxtext}%
    \fi%
    \setlength{\fboxrule}{0.5pt}%
    \setlength{\fboxsep}{1pt}%
    \definecolor{tbcol}{rgb}{1,1,1}%
\begin{picture}(7200.00,5040.00)%
    \gplgaddtomacro\gplbacktext{%
      \csname LTb\endcsname
      \put(1372,896){\makebox(0,0)[r]{\strut{}1e-04}}%
      \csname LTb\endcsname
      \put(1372,2327){\makebox(0,0)[r]{\strut{}1e-03}}%
      \csname LTb\endcsname
      \put(1372,3759){\makebox(0,0)[r]{\strut{}1e-02}}%
      \put(2829,616){\makebox(0,0){\strut{}4}}%
      \put(4118,616){\makebox(0,0){\strut{}7}}%
      \put(5406,616){\makebox(0,0){\strut{}9}}%
    }%
    \gplgaddtomacro\gplfronttext{%
      \csname LTb\endcsname
      \put(266,2827){\rotatebox{-270}{\makebox(0,0){\strut{}time-averaged relative error}}}%
      \put(4117,196){\makebox(0,0){\strut{}number of trajectories in training data set}}%
      \csname LTb\endcsname
      \put(5792,4556){\makebox(0,0)[r]{\strut{}projection}}%
      \csname LTb\endcsname
      \put(5792,4276){\makebox(0,0)[r]{\strut{}OpInf + roll out}}%
      \csname LTb\endcsname
      \put(5792,3996){\makebox(0,0)[r]{\strut{}OpInf}}%
    }%
    \gplbacktext
    \put(0,0){\includegraphics[width={360.00bp},height={252.00bp}]{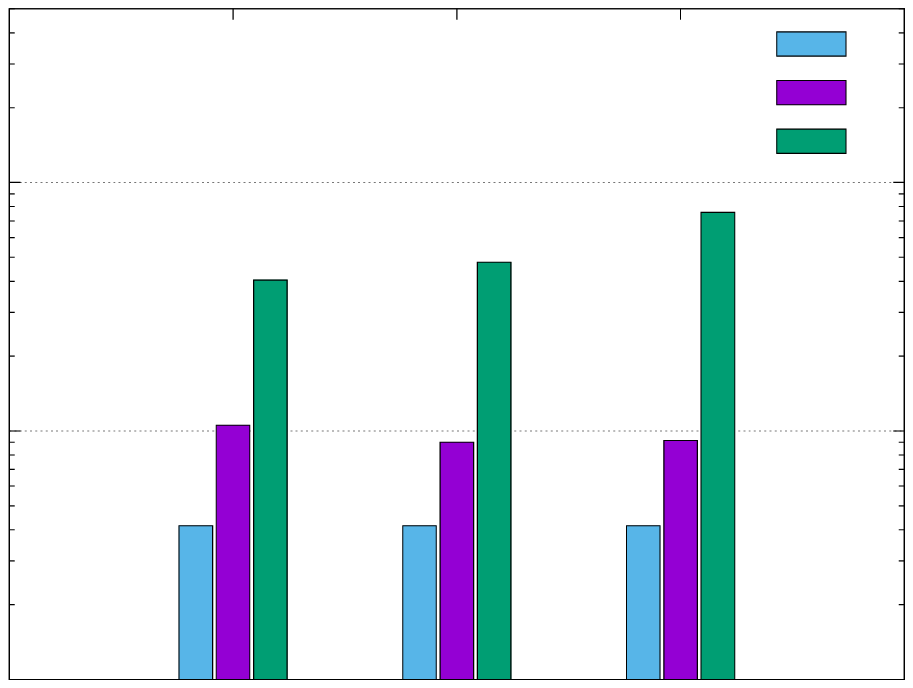}}%
    \gplfronttext
  \end{picture}%
\endgroup

%% file: shallow_errVsRollLen.tex
\begingroup
  \makeatletter
  \providecommand\color[2][]{%
    \GenericError{(gnuplot) \space\space\space\@spaces}{%
      Package color not loaded in conjunction with
      terminal option `colourtext'%
    }{See the gnuplot documentation for explanation.%
    }{Either use 'blacktext' in gnuplot or load the package
      color.sty in LaTeX.}%
    \renewcommand\color[2][]{}%
  }%
  \providecommand\includegraphics[2][]{%
    \GenericError{(gnuplot) \space\space\space\@spaces}{%
      Package graphicx or graphics not loaded%
    }{See the gnuplot documentation for explanation.%
    }{The gnuplot epslatex terminal needs graphicx.sty or graphics.sty.}%
    \renewcommand\includegraphics[2][]{}%
  }%
  \providecommand\rotatebox[2]{#2}%
  \@ifundefined{ifGPcolor}{%
    \newif\ifGPcolor
    \GPcolortrue
  }{}%
  \@ifundefined{ifGPblacktext}{%
    \newif\ifGPblacktext
    \GPblacktexttrue
  }{}%
  \let\gplgaddtomacro\g@addto@macro
  \gdef\gplbacktext{}%
  \gdef\gplfronttext{}%
  \makeatother
  \ifGPblacktext
    \def\colorrgb#1{}%
    \def\colorgray#1{}%
  \else
    \ifGPcolor
      \def\colorrgb#1{\color[rgb]{#1}}%
      \def\colorgray#1{\color[gray]{#1}}%
      \expandafter\def\csname LTw\endcsname{\color{white}}%
      \expandafter\def\csname LTb\endcsname{\color{black}}%
      \expandafter\def\csname LTa\endcsname{\color{black}}%
      \expandafter\def\csname LT0\endcsname{\color[rgb]{1,0,0}}%
      \expandafter\def\csname LT1\endcsname{\color[rgb]{0,1,0}}%
      \expandafter\def\csname LT2\endcsname{\color[rgb]{0,0,1}}%
      \expandafter\def\csname LT3\endcsname{\color[rgb]{1,0,1}}%
      \expandafter\def\csname LT4\endcsname{\color[rgb]{0,1,1}}%
      \expandafter\def\csname LT5\endcsname{\color[rgb]{1,1,0}}%
      \expandafter\def\csname LT6\endcsname{\color[rgb]{0,0,0}}%
      \expandafter\def\csname LT7\endcsname{\color[rgb]{1,0.3,0}}%
      \expandafter\def\csname LT8\endcsname{\color[rgb]{0.5,0.5,0.5}}%
    \else
      \def\colorrgb#1{\color{black}}%
      \def\colorgray#1{\color[gray]{#1}}%
      \expandafter\def\csname LTw\endcsname{\color{white}}%
      \expandafter\def\csname LTb\endcsname{\color{black}}%
      \expandafter\def\csname LTa\endcsname{\color{black}}%
      \expandafter\def\csname LT0\endcsname{\color{black}}%
      \expandafter\def\csname LT1\endcsname{\color{black}}%
      \expandafter\def\csname LT2\endcsname{\color{black}}%
      \expandafter\def\csname LT3\endcsname{\color{black}}%
      \expandafter\def\csname LT4\endcsname{\color{black}}%
      \expandafter\def\csname LT5\endcsname{\color{black}}%
      \expandafter\def\csname LT6\endcsname{\color{black}}%
      \expandafter\def\csname LT7\endcsname{\color{black}}%
      \expandafter\def\csname LT8\endcsname{\color{black}}%
    \fi
  \fi
    \setlength{\unitlength}{0.0500bp}%
    \ifx\gptboxheight\undefined%
      \newlength{\gptboxheight}%
      \newlength{\gptboxwidth}%
      \newsavebox{\gptboxtext}%
    \fi%
    \setlength{\fboxrule}{0.5pt}%
    \setlength{\fboxsep}{1pt}%
    \definecolor{tbcol}{rgb}{1,1,1}%
\begin{picture}(7200.00,5040.00)%
    \gplgaddtomacro\gplbacktext{%
      \csname LTb\endcsname
      \put(1372,2085){\makebox(0,0)[r]{\strut{}1e-03}}%
      \csname LTb\endcsname
      \put(1372,4359){\makebox(0,0)[r]{\strut{}1e-02}}%
      \csname LTb\endcsname
      \put(1540,616){\makebox(0,0){\strut{}$0$}}%
      \csname LTb\endcsname
      \put(2056,616){\makebox(0,0){\strut{}$20$}}%
      \csname LTb\endcsname
      \put(2571,616){\makebox(0,0){\strut{}$40$}}%
      \csname LTb\endcsname
      \put(3087,616){\makebox(0,0){\strut{}$60$}}%
      \csname LTb\endcsname
      \put(3602,616){\makebox(0,0){\strut{}$80$}}%
      \csname LTb\endcsname
      \put(4118,616){\makebox(0,0){\strut{}$100$}}%
      \csname LTb\endcsname
      \put(4633,616){\makebox(0,0){\strut{}$120$}}%
      \csname LTb\endcsname
      \put(5149,616){\makebox(0,0){\strut{}$140$}}%
      \csname LTb\endcsname
      \put(5664,616){\makebox(0,0){\strut{}$160$}}%
      \csname LTb\endcsname
      \put(6180,616){\makebox(0,0){\strut{}$180$}}%
      \csname LTb\endcsname
      \put(6695,616){\makebox(0,0){\strut{}$200$}}%
    }%
    \gplgaddtomacro\gplfronttext{%
      \csname LTb\endcsname
      \put(266,2827){\rotatebox{-270}{\makebox(0,0){\strut{}time-averaged relative error}}}%
      \put(4117,196){\makebox(0,0){\strut{}roll length}}%
      \csname LTb\endcsname
      \put(5792,4556){\makebox(0,0)[r]{\strut{}projection}}%
      \csname LTb\endcsname
      \put(5792,4276){\makebox(0,0)[r]{\strut{}OpInf + roll out}}%
      \csname LTb\endcsname
      \put(5792,3996){\makebox(0,0)[r]{\strut{}OpInf}}%
    }%
    \gplbacktext
    \put(0,0){\includegraphics[width={360.00bp},height={252.00bp}]{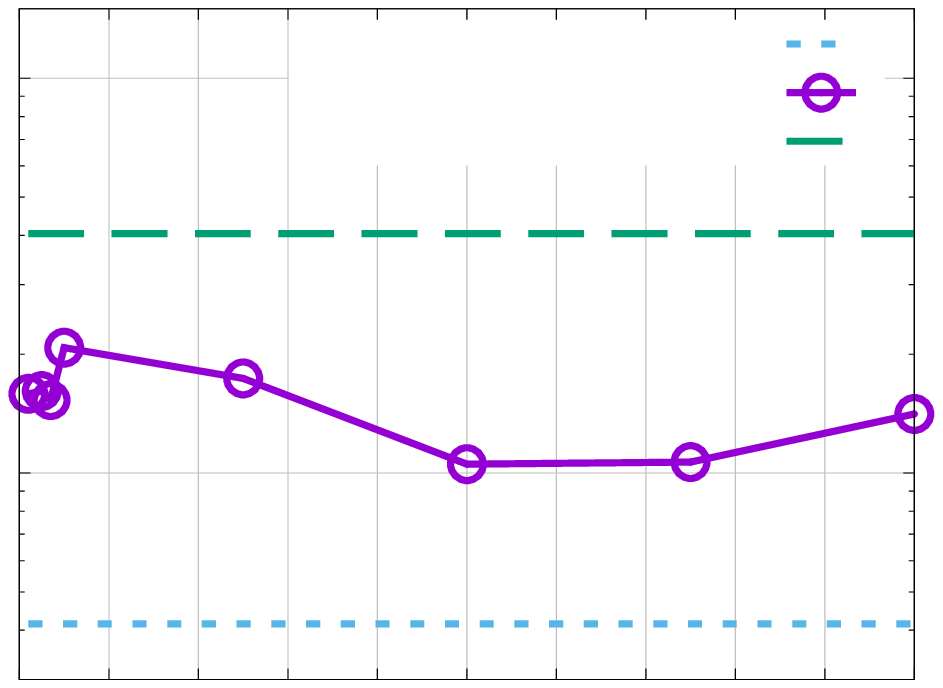}}%
    \gplfronttext
  \end{picture}%
\endgroup

%% file: shallow_radiusnParam.tex
\begingroup
  \makeatletter
  \providecommand\color[2][]{%
    \GenericError{(gnuplot) \space\space\space\@spaces}{%
      Package color not loaded in conjunction with
      terminal option `colourtext'%
    }{See the gnuplot documentation for explanation.%
    }{Either use 'blacktext' in gnuplot or load the package
      color.sty in LaTeX.}%
    \renewcommand\color[2][]{}%
  }%
  \providecommand\includegraphics[2][]{%
    \GenericError{(gnuplot) \space\space\space\@spaces}{%
      Package graphicx or graphics not loaded%
    }{See the gnuplot documentation for explanation.%
    }{The gnuplot epslatex terminal needs graphicx.sty or graphics.sty.}%
    \renewcommand\includegraphics[2][]{}%
  }%
  \providecommand\rotatebox[2]{#2}%
  \@ifundefined{ifGPcolor}{%
    \newif\ifGPcolor
    \GPcolortrue
  }{}%
  \@ifundefined{ifGPblacktext}{%
    \newif\ifGPblacktext
    \GPblacktexttrue
  }{}%
  \let\gplgaddtomacro\g@addto@macro
  \gdef\gplbacktext{}%
  \gdef\gplfronttext{}%
  \makeatother
  \ifGPblacktext
    \def\colorrgb#1{}%
    \def\colorgray#1{}%
  \else
    \ifGPcolor
      \def\colorrgb#1{\color[rgb]{#1}}%
      \def\colorgray#1{\color[gray]{#1}}%
      \expandafter\def\csname LTw\endcsname{\color{white}}%
      \expandafter\def\csname LTb\endcsname{\color{black}}%
      \expandafter\def\csname LTa\endcsname{\color{black}}%
      \expandafter\def\csname LT0\endcsname{\color[rgb]{1,0,0}}%
      \expandafter\def\csname LT1\endcsname{\color[rgb]{0,1,0}}%
      \expandafter\def\csname LT2\endcsname{\color[rgb]{0,0,1}}%
      \expandafter\def\csname LT3\endcsname{\color[rgb]{1,0,1}}%
      \expandafter\def\csname LT4\endcsname{\color[rgb]{0,1,1}}%
      \expandafter\def\csname LT5\endcsname{\color[rgb]{1,1,0}}%
      \expandafter\def\csname LT6\endcsname{\color[rgb]{0,0,0}}%
      \expandafter\def\csname LT7\endcsname{\color[rgb]{1,0.3,0}}%
      \expandafter\def\csname LT8\endcsname{\color[rgb]{0.5,0.5,0.5}}%
    \else
      \def\colorrgb#1{\color{black}}%
      \def\colorgray#1{\color[gray]{#1}}%
      \expandafter\def\csname LTw\endcsname{\color{white}}%
      \expandafter\def\csname LTb\endcsname{\color{black}}%
      \expandafter\def\csname LTa\endcsname{\color{black}}%
      \expandafter\def\csname LT0\endcsname{\color{black}}%
      \expandafter\def\csname LT1\endcsname{\color{black}}%
      \expandafter\def\csname LT2\endcsname{\color{black}}%
      \expandafter\def\csname LT3\endcsname{\color{black}}%
      \expandafter\def\csname LT4\endcsname{\color{black}}%
      \expandafter\def\csname LT5\endcsname{\color{black}}%
      \expandafter\def\csname LT6\endcsname{\color{black}}%
      \expandafter\def\csname LT7\endcsname{\color{black}}%
      \expandafter\def\csname LT8\endcsname{\color{black}}%
    \fi
  \fi
    \setlength{\unitlength}{0.0500bp}%
    \ifx\gptboxheight\undefined%
      \newlength{\gptboxheight}%
      \newlength{\gptboxwidth}%
      \newsavebox{\gptboxtext}%
    \fi%
    \setlength{\fboxrule}{0.5pt}%
    \setlength{\fboxsep}{1pt}%
    \definecolor{tbcol}{rgb}{1,1,1}%
\begin{picture}(7200.00,5040.00)%
    \gplgaddtomacro\gplbacktext{%
      \csname LTb\endcsname
      \put(1372,896){\makebox(0,0)[r]{\strut{}1e-11}}%
      \csname LTb\endcsname
      \put(1372,1448){\makebox(0,0)[r]{\strut{}1e-10}}%
      \csname LTb\endcsname
      \put(1372,2000){\makebox(0,0)[r]{\strut{}1e-09}}%
      \csname LTb\endcsname
      \put(1372,2552){\makebox(0,0)[r]{\strut{}1e-08}}%
      \csname LTb\endcsname
      \put(1372,3103){\makebox(0,0)[r]{\strut{}1e-07}}%
      \csname LTb\endcsname
      \put(1372,3655){\makebox(0,0)[r]{\strut{}1e-06}}%
      \csname LTb\endcsname
      \put(1372,4207){\makebox(0,0)[r]{\strut{}1e-05}}%
      \csname LTb\endcsname
      \put(1372,4759){\makebox(0,0)[r]{\strut{}1e-04}}%
      \put(2829,616){\makebox(0,0){\strut{}4}}%
      \put(4118,616){\makebox(0,0){\strut{}7}}%
      \put(5406,616){\makebox(0,0){\strut{}9}}%
    }%
    \gplgaddtomacro\gplfronttext{%
      \csname LTb\endcsname
      \put(266,2827){\rotatebox{-270}{\makebox(0,0){\strut{}bound of stability radius}}}%
      \put(4117,196){\makebox(0,0){\strut{}number of training trajectories}}%
      \csname LTb\endcsname
      \put(5792,4556){\makebox(0,0)[r]{\strut{}OpInf + roll out}}%
      \csname LTb\endcsname
      \put(5792,4276){\makebox(0,0)[r]{\strut{}OpInf}}%
    }%
    \gplbacktext
    \put(0,0){\includegraphics[width={360.00bp},height={252.00bp}]{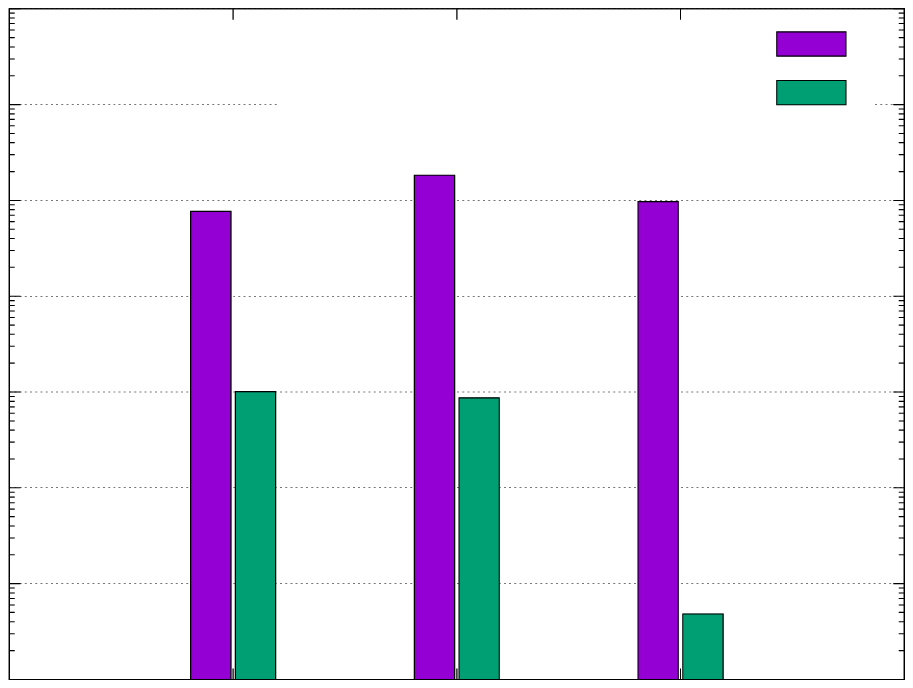}}%
    \gplfronttext
  \end{picture}%
\endgroup

%% file: shallow_radiusRollLen.tex
\begingroup
  \makeatletter
  \providecommand\color[2][]{%
    \GenericError{(gnuplot) \space\space\space\@spaces}{%
      Package color not loaded in conjunction with
      terminal option `colourtext'%
    }{See the gnuplot documentation for explanation.%
    }{Either use 'blacktext' in gnuplot or load the package
      color.sty in LaTeX.}%
    \renewcommand\color[2][]{}%
  }%
  \providecommand\includegraphics[2][]{%
    \GenericError{(gnuplot) \space\space\space\@spaces}{%
      Package graphicx or graphics not loaded%
    }{See the gnuplot documentation for explanation.%
    }{The gnuplot epslatex terminal needs graphicx.sty or graphics.sty.}%
    \renewcommand\includegraphics[2][]{}%
  }%
  \providecommand\rotatebox[2]{#2}%
  \@ifundefined{ifGPcolor}{%
    \newif\ifGPcolor
    \GPcolortrue
  }{}%
  \@ifundefined{ifGPblacktext}{%
    \newif\ifGPblacktext
    \GPblacktexttrue
  }{}%
  \let\gplgaddtomacro\g@addto@macro
  \gdef\gplbacktext{}%
  \gdef\gplfronttext{}%
  \makeatother
  \ifGPblacktext
    \def\colorrgb#1{}%
    \def\colorgray#1{}%
  \else
    \ifGPcolor
      \def\colorrgb#1{\color[rgb]{#1}}%
      \def\colorgray#1{\color[gray]{#1}}%
      \expandafter\def\csname LTw\endcsname{\color{white}}%
      \expandafter\def\csname LTb\endcsname{\color{black}}%
      \expandafter\def\csname LTa\endcsname{\color{black}}%
      \expandafter\def\csname LT0\endcsname{\color[rgb]{1,0,0}}%
      \expandafter\def\csname LT1\endcsname{\color[rgb]{0,1,0}}%
      \expandafter\def\csname LT2\endcsname{\color[rgb]{0,0,1}}%
      \expandafter\def\csname LT3\endcsname{\color[rgb]{1,0,1}}%
      \expandafter\def\csname LT4\endcsname{\color[rgb]{0,1,1}}%
      \expandafter\def\csname LT5\endcsname{\color[rgb]{1,1,0}}%
      \expandafter\def\csname LT6\endcsname{\color[rgb]{0,0,0}}%
      \expandafter\def\csname LT7\endcsname{\color[rgb]{1,0.3,0}}%
      \expandafter\def\csname LT8\endcsname{\color[rgb]{0.5,0.5,0.5}}%
    \else
      \def\colorrgb#1{\color{black}}%
      \def\colorgray#1{\color[gray]{#1}}%
      \expandafter\def\csname LTw\endcsname{\color{white}}%
      \expandafter\def\csname LTb\endcsname{\color{black}}%
      \expandafter\def\csname LTa\endcsname{\color{black}}%
      \expandafter\def\csname LT0\endcsname{\color{black}}%
      \expandafter\def\csname LT1\endcsname{\color{black}}%
      \expandafter\def\csname LT2\endcsname{\color{black}}%
      \expandafter\def\csname LT3\endcsname{\color{black}}%
      \expandafter\def\csname LT4\endcsname{\color{black}}%
      \expandafter\def\csname LT5\endcsname{\color{black}}%
      \expandafter\def\csname LT6\endcsname{\color{black}}%
      \expandafter\def\csname LT7\endcsname{\color{black}}%
      \expandafter\def\csname LT8\endcsname{\color{black}}%
    \fi
  \fi
    \setlength{\unitlength}{0.0500bp}%
    \ifx\gptboxheight\undefined%
      \newlength{\gptboxheight}%
      \newlength{\gptboxwidth}%
      \newsavebox{\gptboxtext}%
    \fi%
    \setlength{\fboxrule}{0.5pt}%
    \setlength{\fboxsep}{1pt}%
    \definecolor{tbcol}{rgb}{1,1,1}%
\begin{picture}(7200.00,5040.00)%
    \gplgaddtomacro\gplbacktext{%
      \csname LTb\endcsname
      \put(1372,1166){\makebox(0,0)[r]{\strut{}1e-08}}%
      \csname LTb\endcsname
      \put(1372,2065){\makebox(0,0)[r]{\strut{}1e-07}}%
      \csname LTb\endcsname
      \put(1372,2963){\makebox(0,0)[r]{\strut{}1e-06}}%
      \csname LTb\endcsname
      \put(1372,3861){\makebox(0,0)[r]{\strut{}1e-05}}%
      \csname LTb\endcsname
      \put(1372,4759){\makebox(0,0)[r]{\strut{}1e-04}}%
      \put(2113,616){\makebox(0,0){\strut{}2}}%
      \put(2686,616){\makebox(0,0){\strut{}5}}%
      \put(3258,616){\makebox(0,0){\strut{}7}}%
      \put(3831,616){\makebox(0,0){\strut{}10}}%
      \put(4404,616){\makebox(0,0){\strut{}50}}%
      \put(4977,616){\makebox(0,0){\strut{}100}}%
      \put(5549,616){\makebox(0,0){\strut{}150}}%
      \put(6122,616){\makebox(0,0){\strut{}200}}%
    }%
    \gplgaddtomacro\gplfronttext{%
      \csname LTb\endcsname
      \put(266,2827){\rotatebox{-270}{\makebox(0,0){\strut{}bound of stability radius}}}%
      \put(4117,196){\makebox(0,0){\strut{}roll-out length}}%
      \csname LTb\endcsname
      \put(5792,4556){\makebox(0,0)[r]{\strut{}OpInf + roll out}}%
      \csname LTb\endcsname
      \put(5792,4276){\makebox(0,0)[r]{\strut{}OpInf}}%
    }%
    \gplbacktext
    \put(0,0){\includegraphics[width={360.00bp},height={252.00bp}]{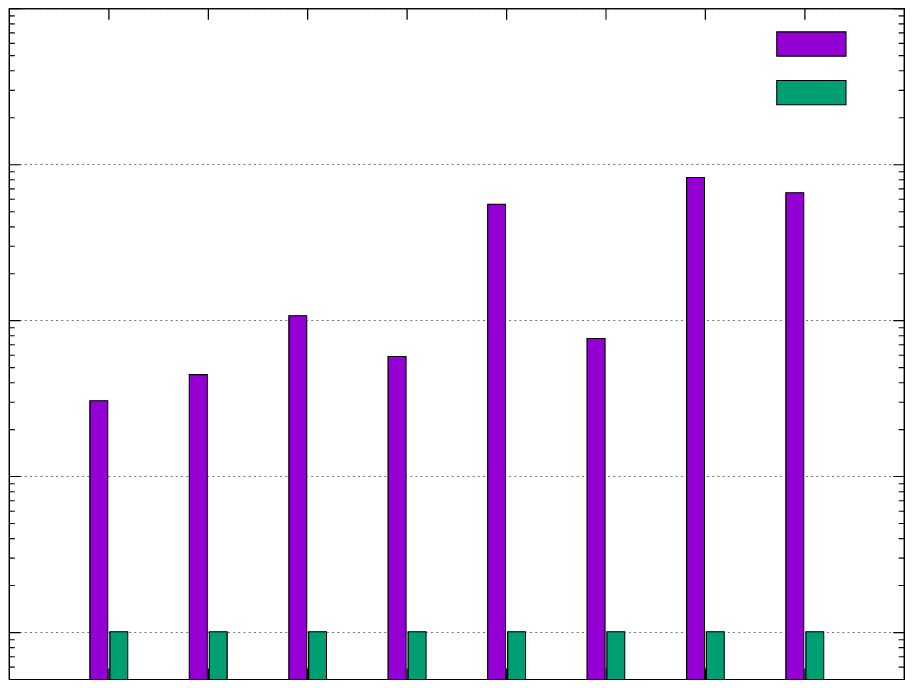}}%
    \gplfronttext
  \end{picture}%
\endgroup

%% file: shallow_errVsNoise_nTrain4.tex
\begingroup
  \makeatletter
  \providecommand\color[2][]{%
    \GenericError{(gnuplot) \space\space\space\@spaces}{%
      Package color not loaded in conjunction with
      terminal option `colourtext'%
    }{See the gnuplot documentation for explanation.%
    }{Either use 'blacktext' in gnuplot or load the package
      color.sty in LaTeX.}%
    \renewcommand\color[2][]{}%
  }%
  \providecommand\includegraphics[2][]{%
    \GenericError{(gnuplot) \space\space\space\@spaces}{%
      Package graphicx or graphics not loaded%
    }{See the gnuplot documentation for explanation.%
    }{The gnuplot epslatex terminal needs graphicx.sty or graphics.sty.}%
    \renewcommand\includegraphics[2][]{}%
  }%
  \providecommand\rotatebox[2]{#2}%
  \@ifundefined{ifGPcolor}{%
    \newif\ifGPcolor
    \GPcolortrue
  }{}%
  \@ifundefined{ifGPblacktext}{%
    \newif\ifGPblacktext
    \GPblacktexttrue
  }{}%
  \let\gplgaddtomacro\g@addto@macro
  \gdef\gplbacktext{}%
  \gdef\gplfronttext{}%
  \makeatother
  \ifGPblacktext
    \def\colorrgb#1{}%
    \def\colorgray#1{}%
  \else
    \ifGPcolor
      \def\colorrgb#1{\color[rgb]{#1}}%
      \def\colorgray#1{\color[gray]{#1}}%
      \expandafter\def\csname LTw\endcsname{\color{white}}%
      \expandafter\def\csname LTb\endcsname{\color{black}}%
      \expandafter\def\csname LTa\endcsname{\color{black}}%
      \expandafter\def\csname LT0\endcsname{\color[rgb]{1,0,0}}%
      \expandafter\def\csname LT1\endcsname{\color[rgb]{0,1,0}}%
      \expandafter\def\csname LT2\endcsname{\color[rgb]{0,0,1}}%
      \expandafter\def\csname LT3\endcsname{\color[rgb]{1,0,1}}%
      \expandafter\def\csname LT4\endcsname{\color[rgb]{0,1,1}}%
      \expandafter\def\csname LT5\endcsname{\color[rgb]{1,1,0}}%
      \expandafter\def\csname LT6\endcsname{\color[rgb]{0,0,0}}%
      \expandafter\def\csname LT7\endcsname{\color[rgb]{1,0.3,0}}%
      \expandafter\def\csname LT8\endcsname{\color[rgb]{0.5,0.5,0.5}}%
    \else
      \def\colorrgb#1{\color{black}}%
      \def\colorgray#1{\color[gray]{#1}}%
      \expandafter\def\csname LTw\endcsname{\color{white}}%
      \expandafter\def\csname LTb\endcsname{\color{black}}%
      \expandafter\def\csname LTa\endcsname{\color{black}}%
      \expandafter\def\csname LT0\endcsname{\color{black}}%
      \expandafter\def\csname LT1\endcsname{\color{black}}%
      \expandafter\def\csname LT2\endcsname{\color{black}}%
      \expandafter\def\csname LT3\endcsname{\color{black}}%
      \expandafter\def\csname LT4\endcsname{\color{black}}%
      \expandafter\def\csname LT5\endcsname{\color{black}}%
      \expandafter\def\csname LT6\endcsname{\color{black}}%
      \expandafter\def\csname LT7\endcsname{\color{black}}%
      \expandafter\def\csname LT8\endcsname{\color{black}}%
    \fi
  \fi
    \setlength{\unitlength}{0.0500bp}%
    \ifx\gptboxheight\undefined%
      \newlength{\gptboxheight}%
      \newlength{\gptboxwidth}%
      \newsavebox{\gptboxtext}%
    \fi%
    \setlength{\fboxrule}{0.5pt}%
    \setlength{\fboxsep}{1pt}%
    \definecolor{tbcol}{rgb}{1,1,1}%
\begin{picture}(7200.00,5040.00)%
    \gplgaddtomacro\gplbacktext{%
      \csname LTb\endcsname
      \put(1372,896){\makebox(0,0)[r]{\strut{}1e-04}}%
      \csname LTb\endcsname
      \put(1372,2184){\makebox(0,0)[r]{\strut{}1e-03}}%
      \csname LTb\endcsname
      \put(1372,3471){\makebox(0,0)[r]{\strut{}1e-02}}%
      \csname LTb\endcsname
      \put(1372,4759){\makebox(0,0)[r]{\strut{}1e-01}}%
      \put(2829,616){\makebox(0,0){\strut{}0.1\%}}%
      \put(4118,616){\makebox(0,0){\strut{}1.0\%}}%
      \put(5406,616){\makebox(0,0){\strut{}10.0\%}}%
    }%
    \gplgaddtomacro\gplfronttext{%
      \csname LTb\endcsname
      \put(182,2827){\rotatebox{-270}{\makebox(0,0){\strut{}time-averaged relative error}}}%
      \put(4117,196){\makebox(0,0){\strut{}noise percentage}}%
      \csname LTb\endcsname
      \put(2443,4486){\makebox(0,0)[l]{\strut{}projection}}%
      \csname LTb\endcsname
      \put(2443,4066){\makebox(0,0)[l]{\strut{}OpInf + roll out}}%
      \csname LTb\endcsname
      \put(2443,3646){\makebox(0,0)[l]{\strut{}OpInf}}%
    }%
    \gplbacktext
    \put(0,0){\includegraphics[width={360.00bp},height={252.00bp}]{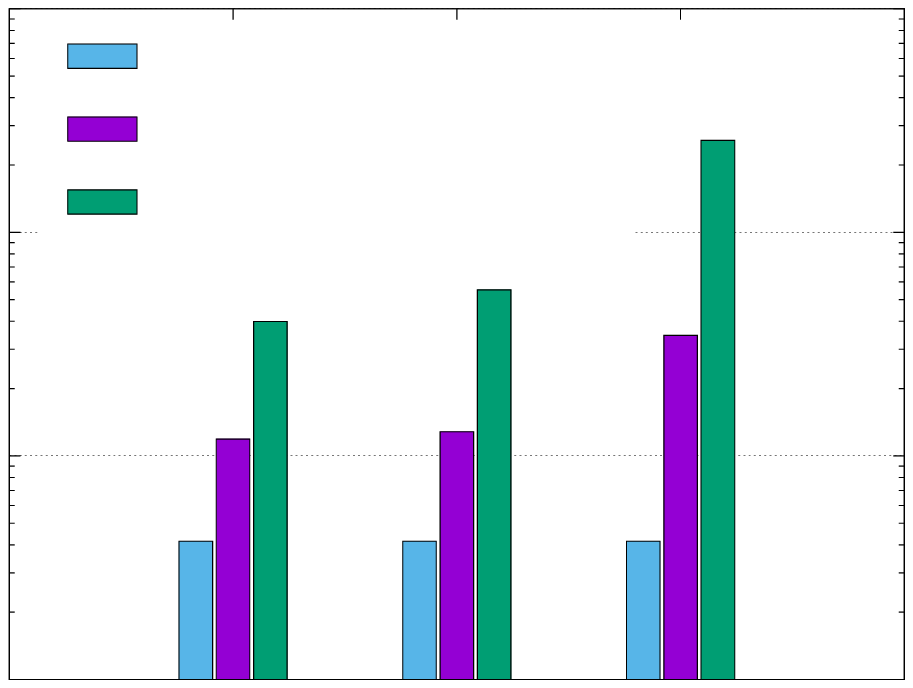}}%
    \gplfronttext
  \end{picture}%
\endgroup

%% file: shallow_errVsNoise_nTrain7.tex
\begingroup
  \makeatletter
  \providecommand\color[2][]{%
    \GenericError{(gnuplot) \space\space\space\@spaces}{%
      Package color not loaded in conjunction with
      terminal option `colourtext'%
    }{See the gnuplot documentation for explanation.%
    }{Either use 'blacktext' in gnuplot or load the package
      color.sty in LaTeX.}%
    \renewcommand\color[2][]{}%
  }%
  \providecommand\includegraphics[2][]{%
    \GenericError{(gnuplot) \space\space\space\@spaces}{%
      Package graphicx or graphics not loaded%
    }{See the gnuplot documentation for explanation.%
    }{The gnuplot epslatex terminal needs graphicx.sty or graphics.sty.}%
    \renewcommand\includegraphics[2][]{}%
  }%
  \providecommand\rotatebox[2]{#2}%
  \@ifundefined{ifGPcolor}{%
    \newif\ifGPcolor
    \GPcolortrue
  }{}%
  \@ifundefined{ifGPblacktext}{%
    \newif\ifGPblacktext
    \GPblacktexttrue
  }{}%
  \let\gplgaddtomacro\g@addto@macro
  \gdef\gplbacktext{}%
  \gdef\gplfronttext{}%
  \makeatother
  \ifGPblacktext
    \def\colorrgb#1{}%
    \def\colorgray#1{}%
  \else
    \ifGPcolor
      \def\colorrgb#1{\color[rgb]{#1}}%
      \def\colorgray#1{\color[gray]{#1}}%
      \expandafter\def\csname LTw\endcsname{\color{white}}%
      \expandafter\def\csname LTb\endcsname{\color{black}}%
      \expandafter\def\csname LTa\endcsname{\color{black}}%
      \expandafter\def\csname LT0\endcsname{\color[rgb]{1,0,0}}%
      \expandafter\def\csname LT1\endcsname{\color[rgb]{0,1,0}}%
      \expandafter\def\csname LT2\endcsname{\color[rgb]{0,0,1}}%
      \expandafter\def\csname LT3\endcsname{\color[rgb]{1,0,1}}%
      \expandafter\def\csname LT4\endcsname{\color[rgb]{0,1,1}}%
      \expandafter\def\csname LT5\endcsname{\color[rgb]{1,1,0}}%
      \expandafter\def\csname LT6\endcsname{\color[rgb]{0,0,0}}%
      \expandafter\def\csname LT7\endcsname{\color[rgb]{1,0.3,0}}%
      \expandafter\def\csname LT8\endcsname{\color[rgb]{0.5,0.5,0.5}}%
    \else
      \def\colorrgb#1{\color{black}}%
      \def\colorgray#1{\color[gray]{#1}}%
      \expandafter\def\csname LTw\endcsname{\color{white}}%
      \expandafter\def\csname LTb\endcsname{\color{black}}%
      \expandafter\def\csname LTa\endcsname{\color{black}}%
      \expandafter\def\csname LT0\endcsname{\color{black}}%
      \expandafter\def\csname LT1\endcsname{\color{black}}%
      \expandafter\def\csname LT2\endcsname{\color{black}}%
      \expandafter\def\csname LT3\endcsname{\color{black}}%
      \expandafter\def\csname LT4\endcsname{\color{black}}%
      \expandafter\def\csname LT5\endcsname{\color{black}}%
      \expandafter\def\csname LT6\endcsname{\color{black}}%
      \expandafter\def\csname LT7\endcsname{\color{black}}%
      \expandafter\def\csname LT8\endcsname{\color{black}}%
    \fi
  \fi
    \setlength{\unitlength}{0.0500bp}%
    \ifx\gptboxheight\undefined%
      \newlength{\gptboxheight}%
      \newlength{\gptboxwidth}%
      \newsavebox{\gptboxtext}%
    \fi%
    \setlength{\fboxrule}{0.5pt}%
    \setlength{\fboxsep}{1pt}%
    \definecolor{tbcol}{rgb}{1,1,1}%
\begin{picture}(7200.00,5040.00)%
    \gplgaddtomacro\gplbacktext{%
      \csname LTb\endcsname
      \put(1372,896){\makebox(0,0)[r]{\strut{}1e-04}}%
      \csname LTb\endcsname
      \put(1372,2184){\makebox(0,0)[r]{\strut{}1e-03}}%
      \csname LTb\endcsname
      \put(1372,3471){\makebox(0,0)[r]{\strut{}1e-02}}%
      \csname LTb\endcsname
      \put(1372,4759){\makebox(0,0)[r]{\strut{}1e-01}}%
      \put(2829,616){\makebox(0,0){\strut{}0.1\%}}%
      \put(4118,616){\makebox(0,0){\strut{}1.0\%}}%
      \put(5406,616){\makebox(0,0){\strut{}10.0\%}}%
    }%
    \gplgaddtomacro\gplfronttext{%
      \csname LTb\endcsname
      \put(182,2827){\rotatebox{-270}{\makebox(0,0){\strut{}time-averaged relative error}}}%
      \put(4117,196){\makebox(0,0){\strut{}noise percentage}}%
      \csname LTb\endcsname
      \put(2443,4486){\makebox(0,0)[l]{\strut{}projection}}%
      \csname LTb\endcsname
      \put(2443,4066){\makebox(0,0)[l]{\strut{}OpInf + roll out}}%
      \csname LTb\endcsname
      \put(2443,3646){\makebox(0,0)[l]{\strut{}OpInf}}%
    }%
    \gplbacktext
    \put(0,0){\includegraphics[width={360.00bp},height={252.00bp}]{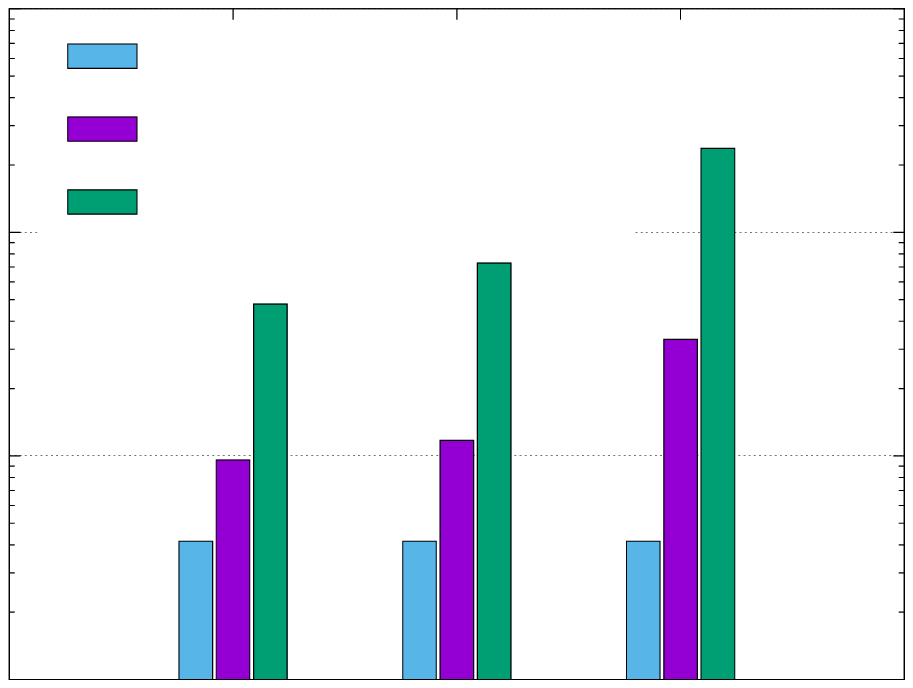}}%
    \gplfronttext
  \end{picture}%
\endgroup

%% file: shallow_errVsNoise_nTrain9.tex
\begingroup
  \makeatletter
  \providecommand\color[2][]{%
    \GenericError{(gnuplot) \space\space\space\@spaces}{%
      Package color not loaded in conjunction with
      terminal option `colourtext'%
    }{See the gnuplot documentation for explanation.%
    }{Either use 'blacktext' in gnuplot or load the package
      color.sty in LaTeX.}%
    \renewcommand\color[2][]{}%
  }%
  \providecommand\includegraphics[2][]{%
    \GenericError{(gnuplot) \space\space\space\@spaces}{%
      Package graphicx or graphics not loaded%
    }{See the gnuplot documentation for explanation.%
    }{The gnuplot epslatex terminal needs graphicx.sty or graphics.sty.}%
    \renewcommand\includegraphics[2][]{}%
  }%
  \providecommand\rotatebox[2]{#2}%
  \@ifundefined{ifGPcolor}{%
    \newif\ifGPcolor
    \GPcolortrue
  }{}%
  \@ifundefined{ifGPblacktext}{%
    \newif\ifGPblacktext
    \GPblacktexttrue
  }{}%
  \let\gplgaddtomacro\g@addto@macro
  \gdef\gplbacktext{}%
  \gdef\gplfronttext{}%
  \makeatother
  \ifGPblacktext
    \def\colorrgb#1{}%
    \def\colorgray#1{}%
  \else
    \ifGPcolor
      \def\colorrgb#1{\color[rgb]{#1}}%
      \def\colorgray#1{\color[gray]{#1}}%
      \expandafter\def\csname LTw\endcsname{\color{white}}%
      \expandafter\def\csname LTb\endcsname{\color{black}}%
      \expandafter\def\csname LTa\endcsname{\color{black}}%
      \expandafter\def\csname LT0\endcsname{\color[rgb]{1,0,0}}%
      \expandafter\def\csname LT1\endcsname{\color[rgb]{0,1,0}}%
      \expandafter\def\csname LT2\endcsname{\color[rgb]{0,0,1}}%
      \expandafter\def\csname LT3\endcsname{\color[rgb]{1,0,1}}%
      \expandafter\def\csname LT4\endcsname{\color[rgb]{0,1,1}}%
      \expandafter\def\csname LT5\endcsname{\color[rgb]{1,1,0}}%
      \expandafter\def\csname LT6\endcsname{\color[rgb]{0,0,0}}%
      \expandafter\def\csname LT7\endcsname{\color[rgb]{1,0.3,0}}%
      \expandafter\def\csname LT8\endcsname{\color[rgb]{0.5,0.5,0.5}}%
    \else
      \def\colorrgb#1{\color{black}}%
      \def\colorgray#1{\color[gray]{#1}}%
      \expandafter\def\csname LTw\endcsname{\color{white}}%
      \expandafter\def\csname LTb\endcsname{\color{black}}%
      \expandafter\def\csname LTa\endcsname{\color{black}}%
      \expandafter\def\csname LT0\endcsname{\color{black}}%
      \expandafter\def\csname LT1\endcsname{\color{black}}%
      \expandafter\def\csname LT2\endcsname{\color{black}}%
      \expandafter\def\csname LT3\endcsname{\color{black}}%
      \expandafter\def\csname LT4\endcsname{\color{black}}%
      \expandafter\def\csname LT5\endcsname{\color{black}}%
      \expandafter\def\csname LT6\endcsname{\color{black}}%
      \expandafter\def\csname LT7\endcsname{\color{black}}%
      \expandafter\def\csname LT8\endcsname{\color{black}}%
    \fi
  \fi
    \setlength{\unitlength}{0.0500bp}%
    \ifx\gptboxheight\undefined%
      \newlength{\gptboxheight}%
      \newlength{\gptboxwidth}%
      \newsavebox{\gptboxtext}%
    \fi%
    \setlength{\fboxrule}{0.5pt}%
    \setlength{\fboxsep}{1pt}%
    \definecolor{tbcol}{rgb}{1,1,1}%
\begin{picture}(7200.00,5040.00)%
    \gplgaddtomacro\gplbacktext{%
      \csname LTb\endcsname
      \put(1372,896){\makebox(0,0)[r]{\strut{}1e-04}}%
      \csname LTb\endcsname
      \put(1372,2184){\makebox(0,0)[r]{\strut{}1e-03}}%
      \csname LTb\endcsname
      \put(1372,3471){\makebox(0,0)[r]{\strut{}1e-02}}%
      \csname LTb\endcsname
      \put(1372,4759){\makebox(0,0)[r]{\strut{}1e-01}}%
      \put(2829,616){\makebox(0,0){\strut{}0.1\%}}%
      \put(4118,616){\makebox(0,0){\strut{}1.0\%}}%
      \put(5406,616){\makebox(0,0){\strut{}10.0\%}}%
    }%
    \gplgaddtomacro\gplfronttext{%
      \csname LTb\endcsname
      \put(182,2827){\rotatebox{-270}{\makebox(0,0){\strut{}time-averaged relative error}}}%
      \put(4117,196){\makebox(0,0){\strut{}noise percentage}}%
      \csname LTb\endcsname
      \put(2443,4486){\makebox(0,0)[l]{\strut{}projection}}%
      \csname LTb\endcsname
      \put(2443,4066){\makebox(0,0)[l]{\strut{}OpInf + roll out}}%
      \csname LTb\endcsname
      \put(2443,3646){\makebox(0,0)[l]{\strut{}OpInf}}%
    }%
    \gplbacktext
    \put(0,0){\includegraphics[width={360.00bp},height={252.00bp}]{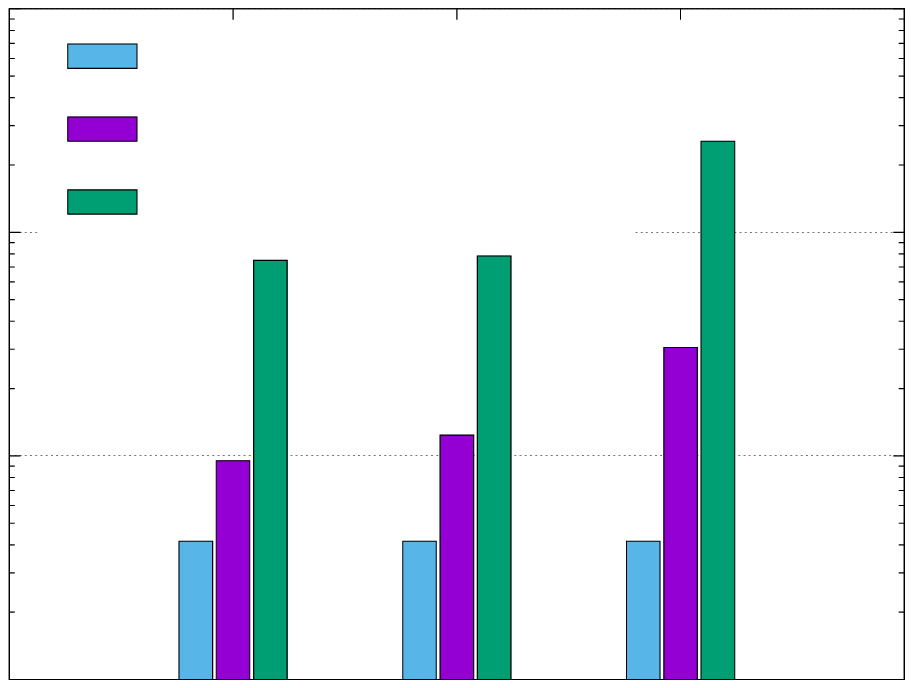}}%
    \gplfronttext
  \end{picture}%
\endgroup

%% file: shallow_errVsRollLen_noise1.tex
\begingroup
  \makeatletter
  \providecommand\color[2][]{%
    \GenericError{(gnuplot) \space\space\space\@spaces}{%
      Package color not loaded in conjunction with
      terminal option `colourtext'%
    }{See the gnuplot documentation for explanation.%
    }{Either use 'blacktext' in gnuplot or load the package
      color.sty in LaTeX.}%
    \renewcommand\color[2][]{}%
  }%
  \providecommand\includegraphics[2][]{%
    \GenericError{(gnuplot) \space\space\space\@spaces}{%
      Package graphicx or graphics not loaded%
    }{See the gnuplot documentation for explanation.%
    }{The gnuplot epslatex terminal needs graphicx.sty or graphics.sty.}%
    \renewcommand\includegraphics[2][]{}%
  }%
  \providecommand\rotatebox[2]{#2}%
  \@ifundefined{ifGPcolor}{%
    \newif\ifGPcolor
    \GPcolortrue
  }{}%
  \@ifundefined{ifGPblacktext}{%
    \newif\ifGPblacktext
    \GPblacktexttrue
  }{}%
  \let\gplgaddtomacro\g@addto@macro
  \gdef\gplbacktext{}%
  \gdef\gplfronttext{}%
  \makeatother
  \ifGPblacktext
    \def\colorrgb#1{}%
    \def\colorgray#1{}%
  \else
    \ifGPcolor
      \def\colorrgb#1{\color[rgb]{#1}}%
      \def\colorgray#1{\color[gray]{#1}}%
      \expandafter\def\csname LTw\endcsname{\color{white}}%
      \expandafter\def\csname LTb\endcsname{\color{black}}%
      \expandafter\def\csname LTa\endcsname{\color{black}}%
      \expandafter\def\csname LT0\endcsname{\color[rgb]{1,0,0}}%
      \expandafter\def\csname LT1\endcsname{\color[rgb]{0,1,0}}%
      \expandafter\def\csname LT2\endcsname{\color[rgb]{0,0,1}}%
      \expandafter\def\csname LT3\endcsname{\color[rgb]{1,0,1}}%
      \expandafter\def\csname LT4\endcsname{\color[rgb]{0,1,1}}%
      \expandafter\def\csname LT5\endcsname{\color[rgb]{1,1,0}}%
      \expandafter\def\csname LT6\endcsname{\color[rgb]{0,0,0}}%
      \expandafter\def\csname LT7\endcsname{\color[rgb]{1,0.3,0}}%
      \expandafter\def\csname LT8\endcsname{\color[rgb]{0.5,0.5,0.5}}%
    \else
      \def\colorrgb#1{\color{black}}%
      \def\colorgray#1{\color[gray]{#1}}%
      \expandafter\def\csname LTw\endcsname{\color{white}}%
      \expandafter\def\csname LTb\endcsname{\color{black}}%
      \expandafter\def\csname LTa\endcsname{\color{black}}%
      \expandafter\def\csname LT0\endcsname{\color{black}}%
      \expandafter\def\csname LT1\endcsname{\color{black}}%
      \expandafter\def\csname LT2\endcsname{\color{black}}%
      \expandafter\def\csname LT3\endcsname{\color{black}}%
      \expandafter\def\csname LT4\endcsname{\color{black}}%
      \expandafter\def\csname LT5\endcsname{\color{black}}%
      \expandafter\def\csname LT6\endcsname{\color{black}}%
      \expandafter\def\csname LT7\endcsname{\color{black}}%
      \expandafter\def\csname LT8\endcsname{\color{black}}%
    \fi
  \fi
    \setlength{\unitlength}{0.0500bp}%
    \ifx\gptboxheight\undefined%
      \newlength{\gptboxheight}%
      \newlength{\gptboxwidth}%
      \newsavebox{\gptboxtext}%
    \fi%
    \setlength{\fboxrule}{0.5pt}%
    \setlength{\fboxsep}{1pt}%
    \definecolor{tbcol}{rgb}{1,1,1}%
\begin{picture}(7200.00,5040.00)%
    \gplgaddtomacro\gplbacktext{%
      \csname LTb\endcsname
      \put(1372,2003){\makebox(0,0)[r]{\strut{}1e-03}}%
      \csname LTb\endcsname
      \put(1372,4121){\makebox(0,0)[r]{\strut{}1e-02}}%
      \csname LTb\endcsname
      \put(1540,616){\makebox(0,0){\strut{}$0$}}%
      \csname LTb\endcsname
      \put(2056,616){\makebox(0,0){\strut{}$20$}}%
      \csname LTb\endcsname
      \put(2571,616){\makebox(0,0){\strut{}$40$}}%
      \csname LTb\endcsname
      \put(3087,616){\makebox(0,0){\strut{}$60$}}%
      \csname LTb\endcsname
      \put(3602,616){\makebox(0,0){\strut{}$80$}}%
      \csname LTb\endcsname
      \put(4118,616){\makebox(0,0){\strut{}$100$}}%
      \csname LTb\endcsname
      \put(4633,616){\makebox(0,0){\strut{}$120$}}%
      \csname LTb\endcsname
      \put(5149,616){\makebox(0,0){\strut{}$140$}}%
      \csname LTb\endcsname
      \put(5664,616){\makebox(0,0){\strut{}$160$}}%
      \csname LTb\endcsname
      \put(6180,616){\makebox(0,0){\strut{}$180$}}%
      \csname LTb\endcsname
      \put(6695,616){\makebox(0,0){\strut{}$200$}}%
    }%
    \gplgaddtomacro\gplfronttext{%
      \csname LTb\endcsname
      \put(266,2827){\rotatebox{-270}{\makebox(0,0){\strut{}time-averaged relative error}}}%
      \put(4117,196){\makebox(0,0){\strut{}roll-out length}}%
      \csname LTb\endcsname
      \put(5792,4556){\makebox(0,0)[r]{\strut{}projection}}%
      \csname LTb\endcsname
      \put(5792,4276){\makebox(0,0)[r]{\strut{}OpInf + roll out}}%
      \csname LTb\endcsname
      \put(5792,3996){\makebox(0,0)[r]{\strut{}OpInf}}%
    }%
    \gplbacktext
    \put(0,0){\includegraphics[width={360.00bp},height={252.00bp}]{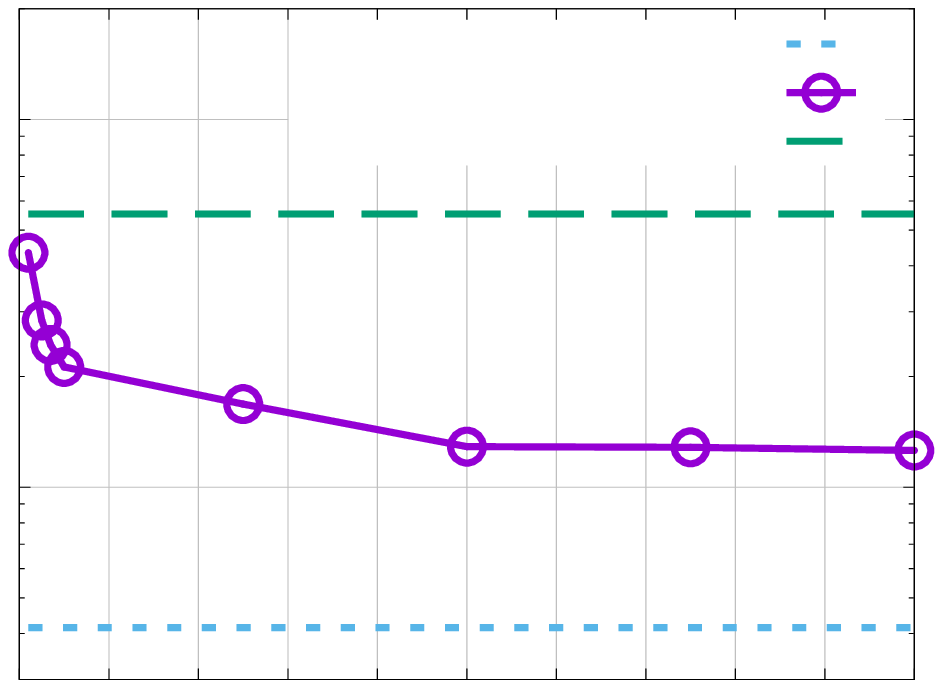}}%
    \gplfronttext
  \end{picture}%
\endgroup

%% file: shallow_errVsRollLen_noise10.tex
\begingroup
  \makeatletter
  \providecommand\color[2][]{%
    \GenericError{(gnuplot) \space\space\space\@spaces}{%
      Package color not loaded in conjunction with
      terminal option `colourtext'%
    }{See the gnuplot documentation for explanation.%
    }{Either use 'blacktext' in gnuplot or load the package
      color.sty in LaTeX.}%
    \renewcommand\color[2][]{}%
  }%
  \providecommand\includegraphics[2][]{%
    \GenericError{(gnuplot) \space\space\space\@spaces}{%
      Package graphicx or graphics not loaded%
    }{See the gnuplot documentation for explanation.%
    }{The gnuplot epslatex terminal needs graphicx.sty or graphics.sty.}%
    \renewcommand\includegraphics[2][]{}%
  }%
  \providecommand\rotatebox[2]{#2}%
  \@ifundefined{ifGPcolor}{%
    \newif\ifGPcolor
    \GPcolortrue
  }{}%
  \@ifundefined{ifGPblacktext}{%
    \newif\ifGPblacktext
    \GPblacktexttrue
  }{}%
  \let\gplgaddtomacro\g@addto@macro
  \gdef\gplbacktext{}%
  \gdef\gplfronttext{}%
  \makeatother
  \ifGPblacktext
    \def\colorrgb#1{}%
    \def\colorgray#1{}%
  \else
    \ifGPcolor
      \def\colorrgb#1{\color[rgb]{#1}}%
      \def\colorgray#1{\color[gray]{#1}}%
      \expandafter\def\csname LTw\endcsname{\color{white}}%
      \expandafter\def\csname LTb\endcsname{\color{black}}%
      \expandafter\def\csname LTa\endcsname{\color{black}}%
      \expandafter\def\csname LT0\endcsname{\color[rgb]{1,0,0}}%
      \expandafter\def\csname LT1\endcsname{\color[rgb]{0,1,0}}%
      \expandafter\def\csname LT2\endcsname{\color[rgb]{0,0,1}}%
      \expandafter\def\csname LT3\endcsname{\color[rgb]{1,0,1}}%
      \expandafter\def\csname LT4\endcsname{\color[rgb]{0,1,1}}%
      \expandafter\def\csname LT5\endcsname{\color[rgb]{1,1,0}}%
      \expandafter\def\csname LT6\endcsname{\color[rgb]{0,0,0}}%
      \expandafter\def\csname LT7\endcsname{\color[rgb]{1,0.3,0}}%
      \expandafter\def\csname LT8\endcsname{\color[rgb]{0.5,0.5,0.5}}%
    \else
      \def\colorrgb#1{\color{black}}%
      \def\colorgray#1{\color[gray]{#1}}%
      \expandafter\def\csname LTw\endcsname{\color{white}}%
      \expandafter\def\csname LTb\endcsname{\color{black}}%
      \expandafter\def\csname LTa\endcsname{\color{black}}%
      \expandafter\def\csname LT0\endcsname{\color{black}}%
      \expandafter\def\csname LT1\endcsname{\color{black}}%
      \expandafter\def\csname LT2\endcsname{\color{black}}%
      \expandafter\def\csname LT3\endcsname{\color{black}}%
      \expandafter\def\csname LT4\endcsname{\color{black}}%
      \expandafter\def\csname LT5\endcsname{\color{black}}%
      \expandafter\def\csname LT6\endcsname{\color{black}}%
      \expandafter\def\csname LT7\endcsname{\color{black}}%
      \expandafter\def\csname LT8\endcsname{\color{black}}%
    \fi
  \fi
    \setlength{\unitlength}{0.0500bp}%
    \ifx\gptboxheight\undefined%
      \newlength{\gptboxheight}%
      \newlength{\gptboxwidth}%
      \newsavebox{\gptboxtext}%
    \fi%
    \setlength{\fboxrule}{0.5pt}%
    \setlength{\fboxsep}{1pt}%
    \definecolor{tbcol}{rgb}{1,1,1}%
\begin{picture}(7200.00,5040.00)%
    \gplgaddtomacro\gplbacktext{%
      \csname LTb\endcsname
      \put(1372,1644){\makebox(0,0)[r]{\strut{}1e-03}}%
      \csname LTb\endcsname
      \put(1372,3076){\makebox(0,0)[r]{\strut{}1e-02}}%
      \csname LTb\endcsname
      \put(1372,4507){\makebox(0,0)[r]{\strut{}1e-01}}%
      \csname LTb\endcsname
      \put(1540,616){\makebox(0,0){\strut{}$0$}}%
      \csname LTb\endcsname
      \put(2056,616){\makebox(0,0){\strut{}$20$}}%
      \csname LTb\endcsname
      \put(2571,616){\makebox(0,0){\strut{}$40$}}%
      \csname LTb\endcsname
      \put(3087,616){\makebox(0,0){\strut{}$60$}}%
      \csname LTb\endcsname
      \put(3602,616){\makebox(0,0){\strut{}$80$}}%
      \csname LTb\endcsname
      \put(4118,616){\makebox(0,0){\strut{}$100$}}%
      \csname LTb\endcsname
      \put(4633,616){\makebox(0,0){\strut{}$120$}}%
      \csname LTb\endcsname
      \put(5149,616){\makebox(0,0){\strut{}$140$}}%
      \csname LTb\endcsname
      \put(5664,616){\makebox(0,0){\strut{}$160$}}%
      \csname LTb\endcsname
      \put(6180,616){\makebox(0,0){\strut{}$180$}}%
      \csname LTb\endcsname
      \put(6695,616){\makebox(0,0){\strut{}$200$}}%
    }%
    \gplgaddtomacro\gplfronttext{%
      \csname LTb\endcsname
      \put(266,2827){\rotatebox{-270}{\makebox(0,0){\strut{}time-averaged relative error}}}%
      \put(4117,196){\makebox(0,0){\strut{}roll-out length}}%
      \csname LTb\endcsname
      \put(5792,4556){\makebox(0,0)[r]{\strut{}projection}}%
      \csname LTb\endcsname
      \put(5792,4276){\makebox(0,0)[r]{\strut{}OpInf + roll out}}%
      \csname LTb\endcsname
      \put(5792,3996){\makebox(0,0)[r]{\strut{}OpInf}}%
    }%
    \gplbacktext
    \put(0,0){\includegraphics[width={360.00bp},height={252.00bp}]{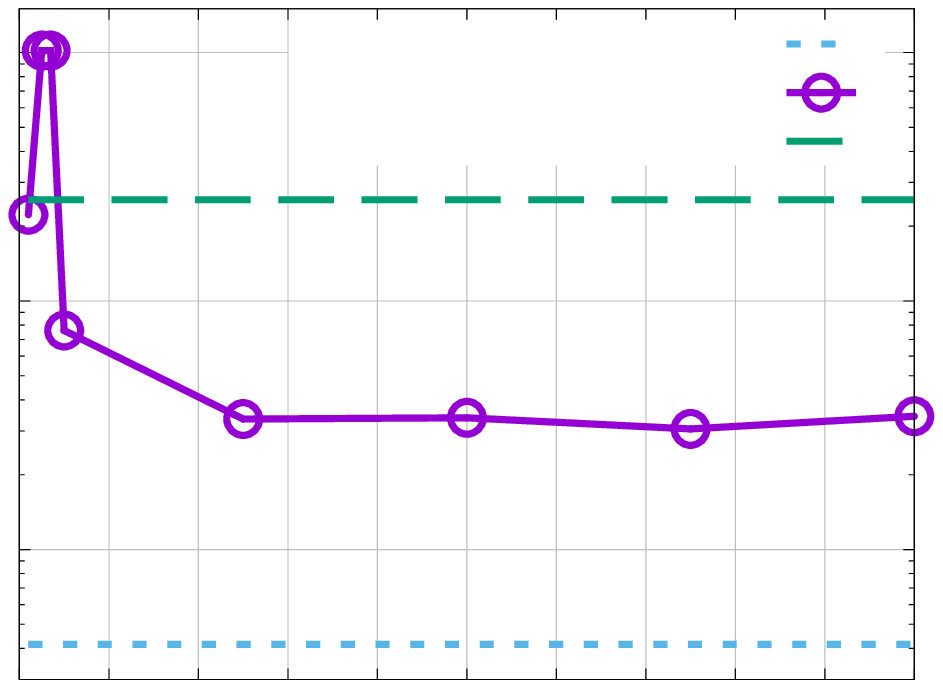}}%
    \gplfronttext
  \end{picture}%
\endgroup

%% file: shallow_errVsNoise_radius_nTrain4.tex
\begingroup
  \makeatletter
  \providecommand\color[2][]{%
    \GenericError{(gnuplot) \space\space\space\@spaces}{%
      Package color not loaded in conjunction with
      terminal option `colourtext'%
    }{See the gnuplot documentation for explanation.%
    }{Either use 'blacktext' in gnuplot or load the package
      color.sty in LaTeX.}%
    \renewcommand\color[2][]{}%
  }%
  \providecommand\includegraphics[2][]{%
    \GenericError{(gnuplot) \space\space\space\@spaces}{%
      Package graphicx or graphics not loaded%
    }{See the gnuplot documentation for explanation.%
    }{The gnuplot epslatex terminal needs graphicx.sty or graphics.sty.}%
    \renewcommand\includegraphics[2][]{}%
  }%
  \providecommand\rotatebox[2]{#2}%
  \@ifundefined{ifGPcolor}{%
    \newif\ifGPcolor
    \GPcolortrue
  }{}%
  \@ifundefined{ifGPblacktext}{%
    \newif\ifGPblacktext
    \GPblacktexttrue
  }{}%
  \let\gplgaddtomacro\g@addto@macro
  \gdef\gplbacktext{}%
  \gdef\gplfronttext{}%
  \makeatother
  \ifGPblacktext
    \def\colorrgb#1{}%
    \def\colorgray#1{}%
  \else
    \ifGPcolor
      \def\colorrgb#1{\color[rgb]{#1}}%
      \def\colorgray#1{\color[gray]{#1}}%
      \expandafter\def\csname LTw\endcsname{\color{white}}%
      \expandafter\def\csname LTb\endcsname{\color{black}}%
      \expandafter\def\csname LTa\endcsname{\color{black}}%
      \expandafter\def\csname LT0\endcsname{\color[rgb]{1,0,0}}%
      \expandafter\def\csname LT1\endcsname{\color[rgb]{0,1,0}}%
      \expandafter\def\csname LT2\endcsname{\color[rgb]{0,0,1}}%
      \expandafter\def\csname LT3\endcsname{\color[rgb]{1,0,1}}%
      \expandafter\def\csname LT4\endcsname{\color[rgb]{0,1,1}}%
      \expandafter\def\csname LT5\endcsname{\color[rgb]{1,1,0}}%
      \expandafter\def\csname LT6\endcsname{\color[rgb]{0,0,0}}%
      \expandafter\def\csname LT7\endcsname{\color[rgb]{1,0.3,0}}%
      \expandafter\def\csname LT8\endcsname{\color[rgb]{0.5,0.5,0.5}}%
    \else
      \def\colorrgb#1{\color{black}}%
      \def\colorgray#1{\color[gray]{#1}}%
      \expandafter\def\csname LTw\endcsname{\color{white}}%
      \expandafter\def\csname LTb\endcsname{\color{black}}%
      \expandafter\def\csname LTa\endcsname{\color{black}}%
      \expandafter\def\csname LT0\endcsname{\color{black}}%
      \expandafter\def\csname LT1\endcsname{\color{black}}%
      \expandafter\def\csname LT2\endcsname{\color{black}}%
      \expandafter\def\csname LT3\endcsname{\color{black}}%
      \expandafter\def\csname LT4\endcsname{\color{black}}%
      \expandafter\def\csname LT5\endcsname{\color{black}}%
      \expandafter\def\csname LT6\endcsname{\color{black}}%
      \expandafter\def\csname LT7\endcsname{\color{black}}%
      \expandafter\def\csname LT8\endcsname{\color{black}}%
    \fi
  \fi
    \setlength{\unitlength}{0.0500bp}%
    \ifx\gptboxheight\undefined%
      \newlength{\gptboxheight}%
      \newlength{\gptboxwidth}%
      \newsavebox{\gptboxtext}%
    \fi%
    \setlength{\fboxrule}{0.5pt}%
    \setlength{\fboxsep}{1pt}%
    \definecolor{tbcol}{rgb}{1,1,1}%
\begin{picture}(7200.00,5040.00)%
    \gplgaddtomacro\gplbacktext{%
      \csname LTb\endcsname
      \put(1372,896){\makebox(0,0)[r]{\strut{}1e-07}}%
      \csname LTb\endcsname
      \put(1372,2184){\makebox(0,0)[r]{\strut{}1e-06}}%
      \csname LTb\endcsname
      \put(1372,3471){\makebox(0,0)[r]{\strut{}1e-05}}%
      \csname LTb\endcsname
      \put(1372,4759){\makebox(0,0)[r]{\strut{}1e-04}}%
      \put(2829,616){\makebox(0,0){\strut{}0.1\%}}%
      \put(4118,616){\makebox(0,0){\strut{}1.0\%}}%
      \put(5406,616){\makebox(0,0){\strut{}10.0\%}}%
    }%
    \gplgaddtomacro\gplfronttext{%
      \csname LTb\endcsname
      \put(266,2827){\rotatebox{-270}{\makebox(0,0){\strut{}stability radius}}}%
      \put(4117,196){\makebox(0,0){\strut{}noise percentage}}%
      \csname LTb\endcsname
      \put(5792,4486){\makebox(0,0)[r]{\strut{}OpInf + roll out}}%
      \csname LTb\endcsname
      \put(5792,4066){\makebox(0,0)[r]{\strut{}OpInf}}%
    }%
    \gplbacktext
    \put(0,0){\includegraphics[width={360.00bp},height={252.00bp}]{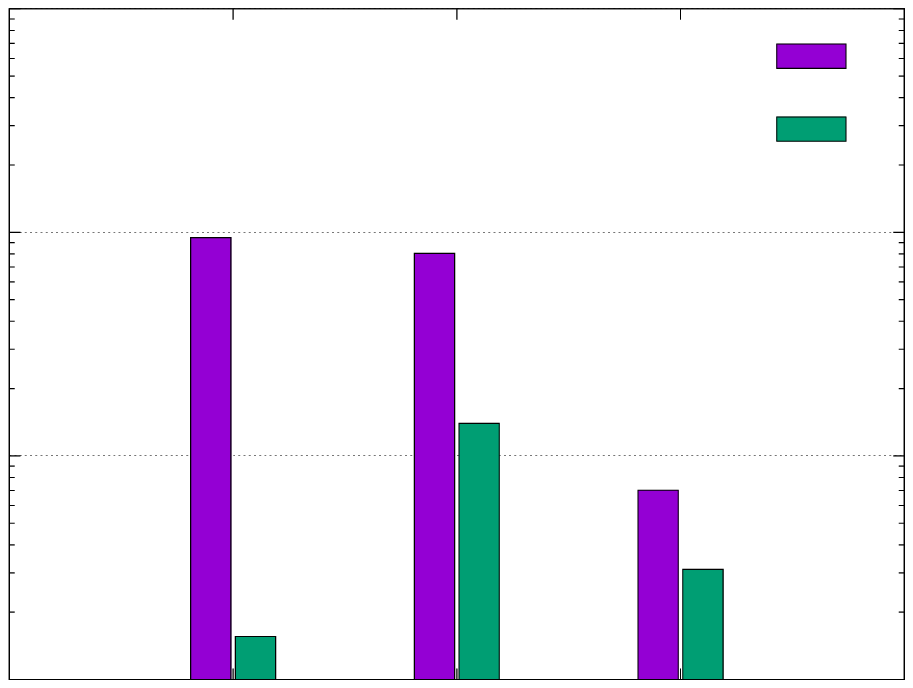}}%
    \gplfronttext
  \end{picture}%
\endgroup

%% file: shallow_errVsNoise_radius_nTrain7.tex
\begingroup
  \makeatletter
  \providecommand\color[2][]{%
    \GenericError{(gnuplot) \space\space\space\@spaces}{%
      Package color not loaded in conjunction with
      terminal option `colourtext'%
    }{See the gnuplot documentation for explanation.%
    }{Either use 'blacktext' in gnuplot or load the package
      color.sty in LaTeX.}%
    \renewcommand\color[2][]{}%
  }%
  \providecommand\includegraphics[2][]{%
    \GenericError{(gnuplot) \space\space\space\@spaces}{%
      Package graphicx or graphics not loaded%
    }{See the gnuplot documentation for explanation.%
    }{The gnuplot epslatex terminal needs graphicx.sty or graphics.sty.}%
    \renewcommand\includegraphics[2][]{}%
  }%
  \providecommand\rotatebox[2]{#2}%
  \@ifundefined{ifGPcolor}{%
    \newif\ifGPcolor
    \GPcolortrue
  }{}%
  \@ifundefined{ifGPblacktext}{%
    \newif\ifGPblacktext
    \GPblacktexttrue
  }{}%
  \let\gplgaddtomacro\g@addto@macro
  \gdef\gplbacktext{}%
  \gdef\gplfronttext{}%
  \makeatother
  \ifGPblacktext
    \def\colorrgb#1{}%
    \def\colorgray#1{}%
  \else
    \ifGPcolor
      \def\colorrgb#1{\color[rgb]{#1}}%
      \def\colorgray#1{\color[gray]{#1}}%
      \expandafter\def\csname LTw\endcsname{\color{white}}%
      \expandafter\def\csname LTb\endcsname{\color{black}}%
      \expandafter\def\csname LTa\endcsname{\color{black}}%
      \expandafter\def\csname LT0\endcsname{\color[rgb]{1,0,0}}%
      \expandafter\def\csname LT1\endcsname{\color[rgb]{0,1,0}}%
      \expandafter\def\csname LT2\endcsname{\color[rgb]{0,0,1}}%
      \expandafter\def\csname LT3\endcsname{\color[rgb]{1,0,1}}%
      \expandafter\def\csname LT4\endcsname{\color[rgb]{0,1,1}}%
      \expandafter\def\csname LT5\endcsname{\color[rgb]{1,1,0}}%
      \expandafter\def\csname LT6\endcsname{\color[rgb]{0,0,0}}%
      \expandafter\def\csname LT7\endcsname{\color[rgb]{1,0.3,0}}%
      \expandafter\def\csname LT8\endcsname{\color[rgb]{0.5,0.5,0.5}}%
    \else
      \def\colorrgb#1{\color{black}}%
      \def\colorgray#1{\color[gray]{#1}}%
      \expandafter\def\csname LTw\endcsname{\color{white}}%
      \expandafter\def\csname LTb\endcsname{\color{black}}%
      \expandafter\def\csname LTa\endcsname{\color{black}}%
      \expandafter\def\csname LT0\endcsname{\color{black}}%
      \expandafter\def\csname LT1\endcsname{\color{black}}%
      \expandafter\def\csname LT2\endcsname{\color{black}}%
      \expandafter\def\csname LT3\endcsname{\color{black}}%
      \expandafter\def\csname LT4\endcsname{\color{black}}%
      \expandafter\def\csname LT5\endcsname{\color{black}}%
      \expandafter\def\csname LT6\endcsname{\color{black}}%
      \expandafter\def\csname LT7\endcsname{\color{black}}%
      \expandafter\def\csname LT8\endcsname{\color{black}}%
    \fi
  \fi
    \setlength{\unitlength}{0.0500bp}%
    \ifx\gptboxheight\undefined%
      \newlength{\gptboxheight}%
      \newlength{\gptboxwidth}%
      \newsavebox{\gptboxtext}%
    \fi%
    \setlength{\fboxrule}{0.5pt}%
    \setlength{\fboxsep}{1pt}%
    \definecolor{tbcol}{rgb}{1,1,1}%
\begin{picture}(7200.00,5040.00)%
    \gplgaddtomacro\gplbacktext{%
      \csname LTb\endcsname
      \put(1372,896){\makebox(0,0)[r]{\strut{}1e-07}}%
      \csname LTb\endcsname
      \put(1372,2828){\makebox(0,0)[r]{\strut{}1e-06}}%
      \csname LTb\endcsname
      \put(1372,4759){\makebox(0,0)[r]{\strut{}1e-05}}%
      \put(2829,616){\makebox(0,0){\strut{}0.1\%}}%
      \put(4118,616){\makebox(0,0){\strut{}1.0\%}}%
      \put(5406,616){\makebox(0,0){\strut{}10.0\%}}%
    }%
    \gplgaddtomacro\gplfronttext{%
      \csname LTb\endcsname
      \put(266,2827){\rotatebox{-270}{\makebox(0,0){\strut{}stability radius}}}%
      \put(4117,196){\makebox(0,0){\strut{}noise percentage}}%
      \csname LTb\endcsname
      \put(5792,4486){\makebox(0,0)[r]{\strut{}OpInf + roll out}}%
      \csname LTb\endcsname
      \put(5792,4066){\makebox(0,0)[r]{\strut{}OpInf}}%
    }%
    \gplbacktext
    \put(0,0){\includegraphics[width={360.00bp},height={252.00bp}]{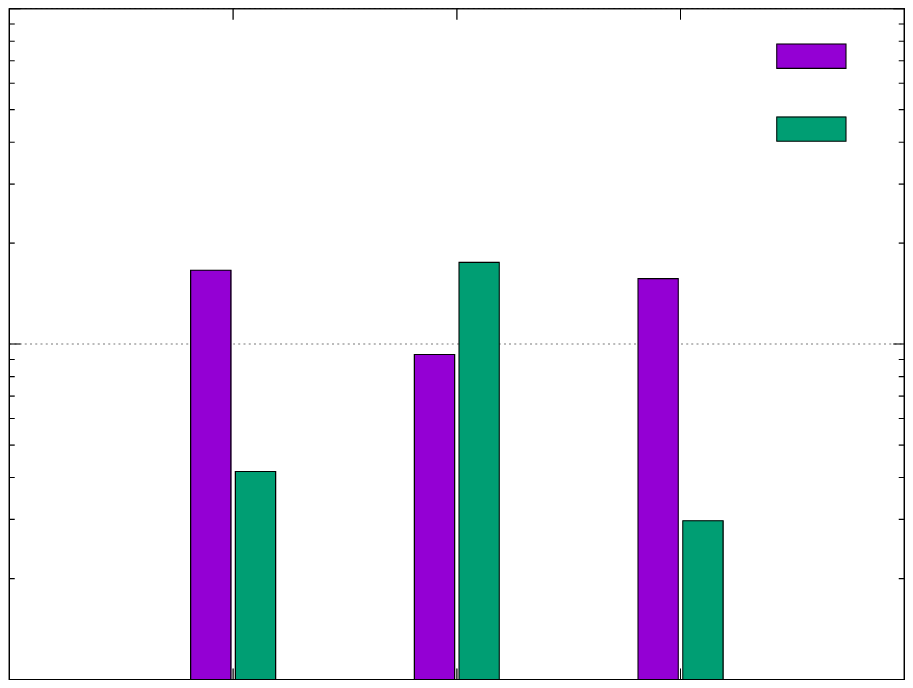}}%
    \gplfronttext
  \end{picture}%
\endgroup

%% file: shallow_errVsNoise_radius_nTrain9.tex
\begingroup
  \makeatletter
  \providecommand\color[2][]{%
    \GenericError{(gnuplot) \space\space\space\@spaces}{%
      Package color not loaded in conjunction with
      terminal option `colourtext'%
    }{See the gnuplot documentation for explanation.%
    }{Either use 'blacktext' in gnuplot or load the package
      color.sty in LaTeX.}%
    \renewcommand\color[2][]{}%
  }%
  \providecommand\includegraphics[2][]{%
    \GenericError{(gnuplot) \space\space\space\@spaces}{%
      Package graphicx or graphics not loaded%
    }{See the gnuplot documentation for explanation.%
    }{The gnuplot epslatex terminal needs graphicx.sty or graphics.sty.}%
    \renewcommand\includegraphics[2][]{}%
  }%
  \providecommand\rotatebox[2]{#2}%
  \@ifundefined{ifGPcolor}{%
    \newif\ifGPcolor
    \GPcolortrue
  }{}%
  \@ifundefined{ifGPblacktext}{%
    \newif\ifGPblacktext
    \GPblacktexttrue
  }{}%
  \let\gplgaddtomacro\g@addto@macro
  \gdef\gplbacktext{}%
  \gdef\gplfronttext{}%
  \makeatother
  \ifGPblacktext
    \def\colorrgb#1{}%
    \def\colorgray#1{}%
  \else
    \ifGPcolor
      \def\colorrgb#1{\color[rgb]{#1}}%
      \def\colorgray#1{\color[gray]{#1}}%
      \expandafter\def\csname LTw\endcsname{\color{white}}%
      \expandafter\def\csname LTb\endcsname{\color{black}}%
      \expandafter\def\csname LTa\endcsname{\color{black}}%
      \expandafter\def\csname LT0\endcsname{\color[rgb]{1,0,0}}%
      \expandafter\def\csname LT1\endcsname{\color[rgb]{0,1,0}}%
      \expandafter\def\csname LT2\endcsname{\color[rgb]{0,0,1}}%
      \expandafter\def\csname LT3\endcsname{\color[rgb]{1,0,1}}%
      \expandafter\def\csname LT4\endcsname{\color[rgb]{0,1,1}}%
      \expandafter\def\csname LT5\endcsname{\color[rgb]{1,1,0}}%
      \expandafter\def\csname LT6\endcsname{\color[rgb]{0,0,0}}%
      \expandafter\def\csname LT7\endcsname{\color[rgb]{1,0.3,0}}%
      \expandafter\def\csname LT8\endcsname{\color[rgb]{0.5,0.5,0.5}}%
    \else
      \def\colorrgb#1{\color{black}}%
      \def\colorgray#1{\color[gray]{#1}}%
      \expandafter\def\csname LTw\endcsname{\color{white}}%
      \expandafter\def\csname LTb\endcsname{\color{black}}%
      \expandafter\def\csname LTa\endcsname{\color{black}}%
      \expandafter\def\csname LT0\endcsname{\color{black}}%
      \expandafter\def\csname LT1\endcsname{\color{black}}%
      \expandafter\def\csname LT2\endcsname{\color{black}}%
      \expandafter\def\csname LT3\endcsname{\color{black}}%
      \expandafter\def\csname LT4\endcsname{\color{black}}%
      \expandafter\def\csname LT5\endcsname{\color{black}}%
      \expandafter\def\csname LT6\endcsname{\color{black}}%
      \expandafter\def\csname LT7\endcsname{\color{black}}%
      \expandafter\def\csname LT8\endcsname{\color{black}}%
    \fi
  \fi
    \setlength{\unitlength}{0.0500bp}%
    \ifx\gptboxheight\undefined%
      \newlength{\gptboxheight}%
      \newlength{\gptboxwidth}%
      \newsavebox{\gptboxtext}%
    \fi%
    \setlength{\fboxrule}{0.5pt}%
    \setlength{\fboxsep}{1pt}%
    \definecolor{tbcol}{rgb}{1,1,1}%
\begin{picture}(7200.00,5040.00)%
    \gplgaddtomacro\gplbacktext{%
      \csname LTb\endcsname
      \put(1372,896){\makebox(0,0)[r]{\strut{}1e-09}}%
      \csname LTb\endcsname
      \put(1372,1862){\makebox(0,0)[r]{\strut{}1e-08}}%
      \csname LTb\endcsname
      \put(1372,2828){\makebox(0,0)[r]{\strut{}1e-07}}%
      \csname LTb\endcsname
      \put(1372,3793){\makebox(0,0)[r]{\strut{}1e-06}}%
      \csname LTb\endcsname
      \put(1372,4759){\makebox(0,0)[r]{\strut{}1e-05}}%
      \put(2829,616){\makebox(0,0){\strut{}0.1\%}}%
      \put(4118,616){\makebox(0,0){\strut{}1.0\%}}%
      \put(5406,616){\makebox(0,0){\strut{}10.0\%}}%
    }%
    \gplgaddtomacro\gplfronttext{%
      \csname LTb\endcsname
      \put(266,2827){\rotatebox{-270}{\makebox(0,0){\strut{}stability radius}}}%
      \put(4117,196){\makebox(0,0){\strut{}noise percentage}}%
      \csname LTb\endcsname
      \put(5792,4486){\makebox(0,0)[r]{\strut{}OpInf + roll out}}%
      \csname LTb\endcsname
      \put(5792,4066){\makebox(0,0)[r]{\strut{}OpInf}}%
    }%
    \gplbacktext
    \put(0,0){\includegraphics[width={360.00bp},height={252.00bp}]{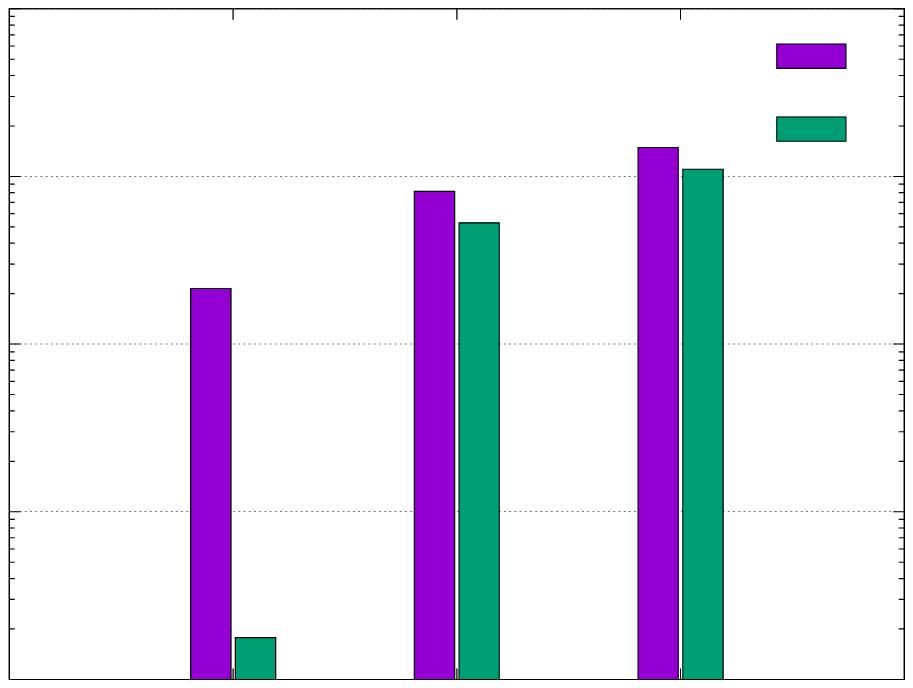}}%
    \gplfronttext
  \end{picture}%
\endgroup

%% file: SQG_errVsnTrainParam_sparse5.tex
\begingroup
  \makeatletter
  \providecommand\color[2][]{%
    \GenericError{(gnuplot) \space\space\space\@spaces}{%
      Package color not loaded in conjunction with
      terminal option `colourtext'%
    }{See the gnuplot documentation for explanation.%
    }{Either use 'blacktext' in gnuplot or load the package
      color.sty in LaTeX.}%
    \renewcommand\color[2][]{}%
  }%
  \providecommand\includegraphics[2][]{%
    \GenericError{(gnuplot) \space\space\space\@spaces}{%
      Package graphicx or graphics not loaded%
    }{See the gnuplot documentation for explanation.%
    }{The gnuplot epslatex terminal needs graphicx.sty or graphics.sty.}%
    \renewcommand\includegraphics[2][]{}%
  }%
  \providecommand\rotatebox[2]{#2}%
  \@ifundefined{ifGPcolor}{%
    \newif\ifGPcolor
    \GPcolortrue
  }{}%
  \@ifundefined{ifGPblacktext}{%
    \newif\ifGPblacktext
    \GPblacktexttrue
  }{}%
  \let\gplgaddtomacro\g@addto@macro
  \gdef\gplbacktext{}%
  \gdef\gplfronttext{}%
  \makeatother
  \ifGPblacktext
    \def\colorrgb#1{}%
    \def\colorgray#1{}%
  \else
    \ifGPcolor
      \def\colorrgb#1{\color[rgb]{#1}}%
      \def\colorgray#1{\color[gray]{#1}}%
      \expandafter\def\csname LTw\endcsname{\color{white}}%
      \expandafter\def\csname LTb\endcsname{\color{black}}%
      \expandafter\def\csname LTa\endcsname{\color{black}}%
      \expandafter\def\csname LT0\endcsname{\color[rgb]{1,0,0}}%
      \expandafter\def\csname LT1\endcsname{\color[rgb]{0,1,0}}%
      \expandafter\def\csname LT2\endcsname{\color[rgb]{0,0,1}}%
      \expandafter\def\csname LT3\endcsname{\color[rgb]{1,0,1}}%
      \expandafter\def\csname LT4\endcsname{\color[rgb]{0,1,1}}%
      \expandafter\def\csname LT5\endcsname{\color[rgb]{1,1,0}}%
      \expandafter\def\csname LT6\endcsname{\color[rgb]{0,0,0}}%
      \expandafter\def\csname LT7\endcsname{\color[rgb]{1,0.3,0}}%
      \expandafter\def\csname LT8\endcsname{\color[rgb]{0.5,0.5,0.5}}%
    \else
      \def\colorrgb#1{\color{black}}%
      \def\colorgray#1{\color[gray]{#1}}%
      \expandafter\def\csname LTw\endcsname{\color{white}}%
      \expandafter\def\csname LTb\endcsname{\color{black}}%
      \expandafter\def\csname LTa\endcsname{\color{black}}%
      \expandafter\def\csname LT0\endcsname{\color{black}}%
      \expandafter\def\csname LT1\endcsname{\color{black}}%
      \expandafter\def\csname LT2\endcsname{\color{black}}%
      \expandafter\def\csname LT3\endcsname{\color{black}}%
      \expandafter\def\csname LT4\endcsname{\color{black}}%
      \expandafter\def\csname LT5\endcsname{\color{black}}%
      \expandafter\def\csname LT6\endcsname{\color{black}}%
      \expandafter\def\csname LT7\endcsname{\color{black}}%
      \expandafter\def\csname LT8\endcsname{\color{black}}%
    \fi
  \fi
    \setlength{\unitlength}{0.0500bp}%
    \ifx\gptboxheight\undefined%
      \newlength{\gptboxheight}%
      \newlength{\gptboxwidth}%
      \newsavebox{\gptboxtext}%
    \fi%
    \setlength{\fboxrule}{0.5pt}%
    \setlength{\fboxsep}{1pt}%
    \definecolor{tbcol}{rgb}{1,1,1}%
\begin{picture}(7200.00,5040.00)%
    \gplgaddtomacro\gplbacktext{%
      \csname LTb\endcsname
      \put(1372,896){\makebox(0,0)[r]{\strut{}1e-02}}%
      \csname LTb\endcsname
      \put(1372,2738){\makebox(0,0)[r]{\strut{}1e-01}}%
      \csname LTb\endcsname
      \put(1372,4580){\makebox(0,0)[r]{\strut{}1e+00}}%
      \put(2829,616){\makebox(0,0){\strut{}4}}%
      \put(4118,616){\makebox(0,0){\strut{}8}}%
      \put(5406,616){\makebox(0,0){\strut{}12}}%
    }%
    \gplgaddtomacro\gplfronttext{%
      \csname LTb\endcsname
      \put(266,2827){\rotatebox{-270}{\makebox(0,0){\strut{}time-averaged relative error}}}%
      \put(4117,196){\makebox(0,0){\strut{}number of training trajectories}}%
      \csname LTb\endcsname
      \put(5792,4556){\makebox(0,0)[r]{\strut{}projection}}%
      \csname LTb\endcsname
      \put(5792,4276){\makebox(0,0)[r]{\strut{}OpInf + roll out}}%
      \csname LTb\endcsname
      \put(5792,3996){\makebox(0,0)[r]{\strut{}OpInf}}%
    }%
    \gplbacktext
    \put(0,0){\includegraphics[width={360.00bp},height={252.00bp}]{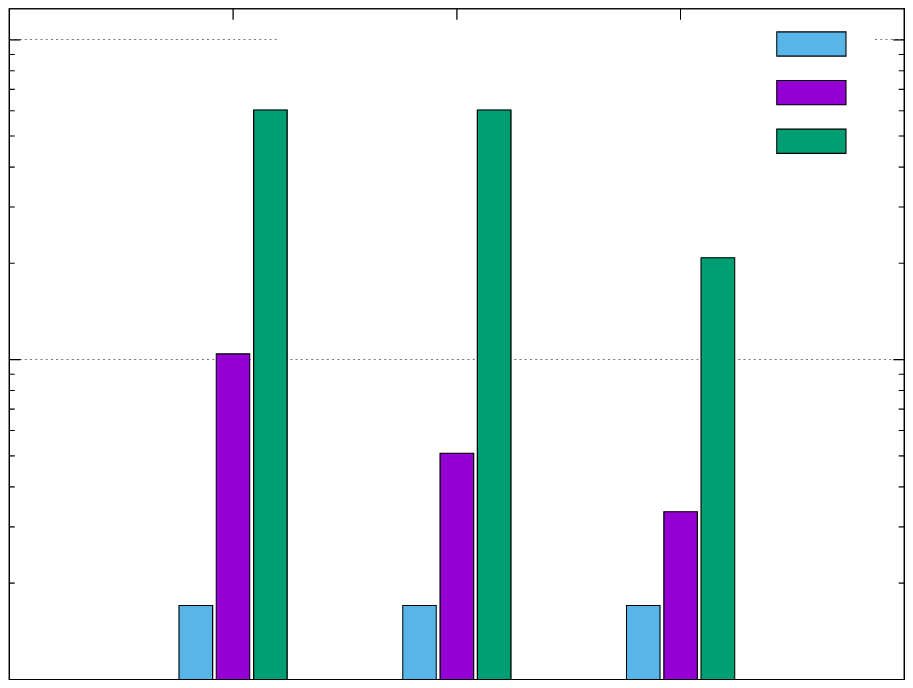}}%
    \gplfronttext
  \end{picture}%
\endgroup

%% file: SQG_errVsnTrainParam_sparse10.tex
\begingroup
  \makeatletter
  \providecommand\color[2][]{%
    \GenericError{(gnuplot) \space\space\space\@spaces}{%
      Package color not loaded in conjunction with
      terminal option `colourtext'%
    }{See the gnuplot documentation for explanation.%
    }{Either use 'blacktext' in gnuplot or load the package
      color.sty in LaTeX.}%
    \renewcommand\color[2][]{}%
  }%
  \providecommand\includegraphics[2][]{%
    \GenericError{(gnuplot) \space\space\space\@spaces}{%
      Package graphicx or graphics not loaded%
    }{See the gnuplot documentation for explanation.%
    }{The gnuplot epslatex terminal needs graphicx.sty or graphics.sty.}%
    \renewcommand\includegraphics[2][]{}%
  }%
  \providecommand\rotatebox[2]{#2}%
  \@ifundefined{ifGPcolor}{%
    \newif\ifGPcolor
    \GPcolortrue
  }{}%
  \@ifundefined{ifGPblacktext}{%
    \newif\ifGPblacktext
    \GPblacktexttrue
  }{}%
  \let\gplgaddtomacro\g@addto@macro
  \gdef\gplbacktext{}%
  \gdef\gplfronttext{}%
  \makeatother
  \ifGPblacktext
    \def\colorrgb#1{}%
    \def\colorgray#1{}%
  \else
    \ifGPcolor
      \def\colorrgb#1{\color[rgb]{#1}}%
      \def\colorgray#1{\color[gray]{#1}}%
      \expandafter\def\csname LTw\endcsname{\color{white}}%
      \expandafter\def\csname LTb\endcsname{\color{black}}%
      \expandafter\def\csname LTa\endcsname{\color{black}}%
      \expandafter\def\csname LT0\endcsname{\color[rgb]{1,0,0}}%
      \expandafter\def\csname LT1\endcsname{\color[rgb]{0,1,0}}%
      \expandafter\def\csname LT2\endcsname{\color[rgb]{0,0,1}}%
      \expandafter\def\csname LT3\endcsname{\color[rgb]{1,0,1}}%
      \expandafter\def\csname LT4\endcsname{\color[rgb]{0,1,1}}%
      \expandafter\def\csname LT5\endcsname{\color[rgb]{1,1,0}}%
      \expandafter\def\csname LT6\endcsname{\color[rgb]{0,0,0}}%
      \expandafter\def\csname LT7\endcsname{\color[rgb]{1,0.3,0}}%
      \expandafter\def\csname LT8\endcsname{\color[rgb]{0.5,0.5,0.5}}%
    \else
      \def\colorrgb#1{\color{black}}%
      \def\colorgray#1{\color[gray]{#1}}%
      \expandafter\def\csname LTw\endcsname{\color{white}}%
      \expandafter\def\csname LTb\endcsname{\color{black}}%
      \expandafter\def\csname LTa\endcsname{\color{black}}%
      \expandafter\def\csname LT0\endcsname{\color{black}}%
      \expandafter\def\csname LT1\endcsname{\color{black}}%
      \expandafter\def\csname LT2\endcsname{\color{black}}%
      \expandafter\def\csname LT3\endcsname{\color{black}}%
      \expandafter\def\csname LT4\endcsname{\color{black}}%
      \expandafter\def\csname LT5\endcsname{\color{black}}%
      \expandafter\def\csname LT6\endcsname{\color{black}}%
      \expandafter\def\csname LT7\endcsname{\color{black}}%
      \expandafter\def\csname LT8\endcsname{\color{black}}%
    \fi
  \fi
    \setlength{\unitlength}{0.0500bp}%
    \ifx\gptboxheight\undefined%
      \newlength{\gptboxheight}%
      \newlength{\gptboxwidth}%
      \newsavebox{\gptboxtext}%
    \fi%
    \setlength{\fboxrule}{0.5pt}%
    \setlength{\fboxsep}{1pt}%
    \definecolor{tbcol}{rgb}{1,1,1}%
\begin{picture}(7200.00,5040.00)%
    \gplgaddtomacro\gplbacktext{%
      \csname LTb\endcsname
      \put(1372,896){\makebox(0,0)[r]{\strut{}1e-02}}%
      \csname LTb\endcsname
      \put(1372,2184){\makebox(0,0)[r]{\strut{}1e-01}}%
      \csname LTb\endcsname
      \put(1372,3471){\makebox(0,0)[r]{\strut{}1e+00}}%
      \csname LTb\endcsname
      \put(1372,4759){\makebox(0,0)[r]{\strut{}1e+01}}%
      \put(2829,616){\makebox(0,0){\strut{}4}}%
      \put(4118,616){\makebox(0,0){\strut{}8}}%
      \put(5406,616){\makebox(0,0){\strut{}12}}%
    }%
    \gplgaddtomacro\gplfronttext{%
      \csname LTb\endcsname
      \put(266,2827){\rotatebox{-270}{\makebox(0,0){\strut{}time-averaged relative error}}}%
      \put(4117,196){\makebox(0,0){\strut{}number of training trajectories}}%
      \csname LTb\endcsname
      \put(5792,4556){\makebox(0,0)[r]{\strut{}projection}}%
      \csname LTb\endcsname
      \put(5792,4276){\makebox(0,0)[r]{\strut{}OpInf + roll out}}%
      \csname LTb\endcsname
      \put(5792,3996){\makebox(0,0)[r]{\strut{}OpInf}}%
    }%
    \gplbacktext
    \put(0,0){\includegraphics[width={360.00bp},height={252.00bp}]{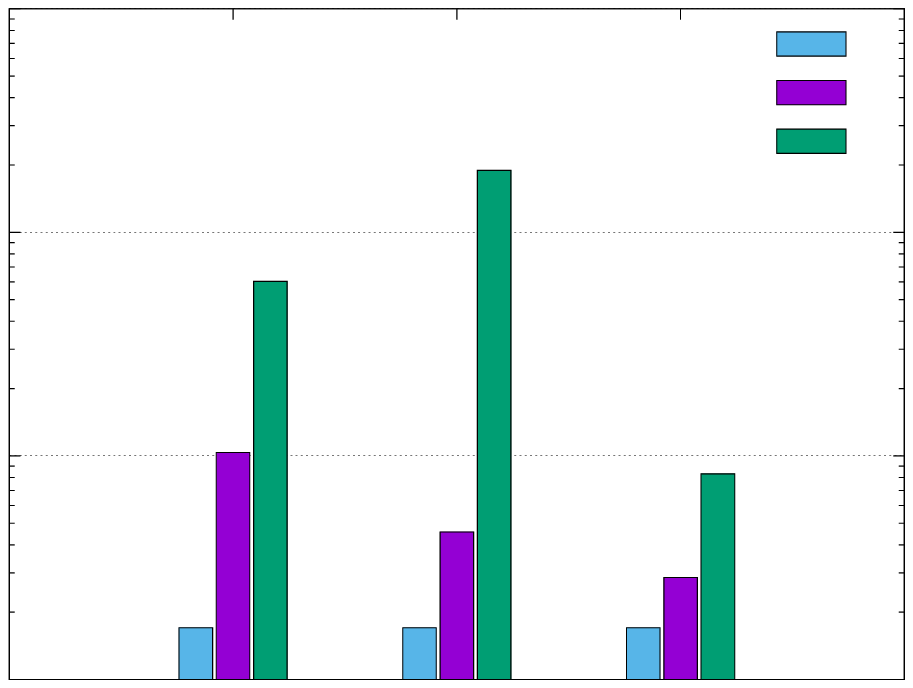}}%
    \gplfronttext
  \end{picture}%
\endgroup

%% file: SQG_errVsRoll_sparse5.tex
\begingroup
  \makeatletter
  \providecommand\color[2][]{%
    \GenericError{(gnuplot) \space\space\space\@spaces}{%
      Package color not loaded in conjunction with
      terminal option `colourtext'%
    }{See the gnuplot documentation for explanation.%
    }{Either use 'blacktext' in gnuplot or load the package
      color.sty in LaTeX.}%
    \renewcommand\color[2][]{}%
  }%
  \providecommand\includegraphics[2][]{%
    \GenericError{(gnuplot) \space\space\space\@spaces}{%
      Package graphicx or graphics not loaded%
    }{See the gnuplot documentation for explanation.%
    }{The gnuplot epslatex terminal needs graphicx.sty or graphics.sty.}%
    \renewcommand\includegraphics[2][]{}%
  }%
  \providecommand\rotatebox[2]{#2}%
  \@ifundefined{ifGPcolor}{%
    \newif\ifGPcolor
    \GPcolortrue
  }{}%
  \@ifundefined{ifGPblacktext}{%
    \newif\ifGPblacktext
    \GPblacktexttrue
  }{}%
  \let\gplgaddtomacro\g@addto@macro
  \gdef\gplbacktext{}%
  \gdef\gplfronttext{}%
  \makeatother
  \ifGPblacktext
    \def\colorrgb#1{}%
    \def\colorgray#1{}%
  \else
    \ifGPcolor
      \def\colorrgb#1{\color[rgb]{#1}}%
      \def\colorgray#1{\color[gray]{#1}}%
      \expandafter\def\csname LTw\endcsname{\color{white}}%
      \expandafter\def\csname LTb\endcsname{\color{black}}%
      \expandafter\def\csname LTa\endcsname{\color{black}}%
      \expandafter\def\csname LT0\endcsname{\color[rgb]{1,0,0}}%
      \expandafter\def\csname LT1\endcsname{\color[rgb]{0,1,0}}%
      \expandafter\def\csname LT2\endcsname{\color[rgb]{0,0,1}}%
      \expandafter\def\csname LT3\endcsname{\color[rgb]{1,0,1}}%
      \expandafter\def\csname LT4\endcsname{\color[rgb]{0,1,1}}%
      \expandafter\def\csname LT5\endcsname{\color[rgb]{1,1,0}}%
      \expandafter\def\csname LT6\endcsname{\color[rgb]{0,0,0}}%
      \expandafter\def\csname LT7\endcsname{\color[rgb]{1,0.3,0}}%
      \expandafter\def\csname LT8\endcsname{\color[rgb]{0.5,0.5,0.5}}%
    \else
      \def\colorrgb#1{\color{black}}%
      \def\colorgray#1{\color[gray]{#1}}%
      \expandafter\def\csname LTw\endcsname{\color{white}}%
      \expandafter\def\csname LTb\endcsname{\color{black}}%
      \expandafter\def\csname LTa\endcsname{\color{black}}%
      \expandafter\def\csname LT0\endcsname{\color{black}}%
      \expandafter\def\csname LT1\endcsname{\color{black}}%
      \expandafter\def\csname LT2\endcsname{\color{black}}%
      \expandafter\def\csname LT3\endcsname{\color{black}}%
      \expandafter\def\csname LT4\endcsname{\color{black}}%
      \expandafter\def\csname LT5\endcsname{\color{black}}%
      \expandafter\def\csname LT6\endcsname{\color{black}}%
      \expandafter\def\csname LT7\endcsname{\color{black}}%
      \expandafter\def\csname LT8\endcsname{\color{black}}%
    \fi
  \fi
    \setlength{\unitlength}{0.0500bp}%
    \ifx\gptboxheight\undefined%
      \newlength{\gptboxheight}%
      \newlength{\gptboxwidth}%
      \newsavebox{\gptboxtext}%
    \fi%
    \setlength{\fboxrule}{0.5pt}%
    \setlength{\fboxsep}{1pt}%
    \definecolor{tbcol}{rgb}{1,1,1}%
\begin{picture}(7200.00,5040.00)%
    \gplgaddtomacro\gplbacktext{%
      \csname LTb\endcsname
      \put(1372,896){\makebox(0,0)[r]{\strut{}1e-02}}%
      \csname LTb\endcsname
      \put(1372,3068){\makebox(0,0)[r]{\strut{}1e-01}}%
      \csname LTb\endcsname
      \put(1540,616){\makebox(0,0){\strut{}$0$}}%
      \csname LTb\endcsname
      \put(2056,616){\makebox(0,0){\strut{}$20$}}%
      \csname LTb\endcsname
      \put(2571,616){\makebox(0,0){\strut{}$40$}}%
      \csname LTb\endcsname
      \put(3087,616){\makebox(0,0){\strut{}$60$}}%
      \csname LTb\endcsname
      \put(3602,616){\makebox(0,0){\strut{}$80$}}%
      \csname LTb\endcsname
      \put(4118,616){\makebox(0,0){\strut{}$100$}}%
      \csname LTb\endcsname
      \put(4633,616){\makebox(0,0){\strut{}$120$}}%
      \csname LTb\endcsname
      \put(5149,616){\makebox(0,0){\strut{}$140$}}%
      \csname LTb\endcsname
      \put(5664,616){\makebox(0,0){\strut{}$160$}}%
      \csname LTb\endcsname
      \put(6180,616){\makebox(0,0){\strut{}$180$}}%
      \csname LTb\endcsname
      \put(6695,616){\makebox(0,0){\strut{}$200$}}%
    }%
    \gplgaddtomacro\gplfronttext{%
      \csname LTb\endcsname
      \put(266,2827){\rotatebox{-270}{\makebox(0,0){\strut{}time-averaged relative error}}}%
      \put(4117,196){\makebox(0,0){\strut{}roll-out length}}%
      \csname LTb\endcsname
      \put(5792,4556){\makebox(0,0)[r]{\strut{}projection}}%
      \csname LTb\endcsname
      \put(5792,4276){\makebox(0,0)[r]{\strut{}OpInf + roll out}}%
      \csname LTb\endcsname
      \put(5792,3996){\makebox(0,0)[r]{\strut{}OpInf}}%
    }%
    \gplbacktext
    \put(0,0){\includegraphics[width={360.00bp},height={252.00bp}]{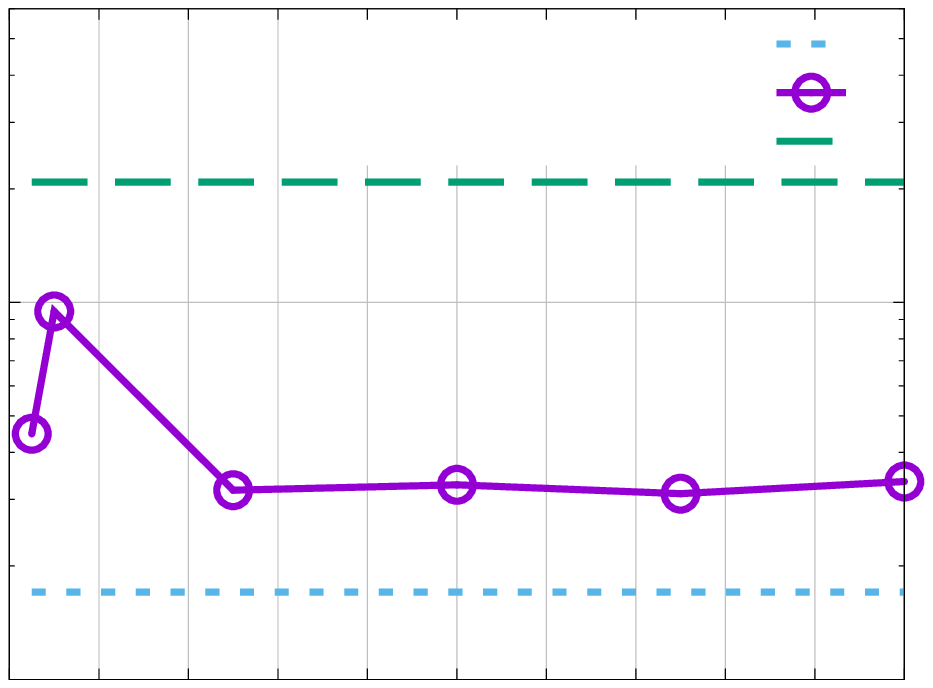}}%
    \gplfronttext
  \end{picture}%
\endgroup

%% file: SQG_errVsRoll_sparse10.tex
\begingroup
  \makeatletter
  \providecommand\color[2][]{%
    \GenericError{(gnuplot) \space\space\space\@spaces}{%
      Package color not loaded in conjunction with
      terminal option `colourtext'%
    }{See the gnuplot documentation for explanation.%
    }{Either use 'blacktext' in gnuplot or load the package
      color.sty in LaTeX.}%
    \renewcommand\color[2][]{}%
  }%
  \providecommand\includegraphics[2][]{%
    \GenericError{(gnuplot) \space\space\space\@spaces}{%
      Package graphicx or graphics not loaded%
    }{See the gnuplot documentation for explanation.%
    }{The gnuplot epslatex terminal needs graphicx.sty or graphics.sty.}%
    \renewcommand\includegraphics[2][]{}%
  }%
  \providecommand\rotatebox[2]{#2}%
  \@ifundefined{ifGPcolor}{%
    \newif\ifGPcolor
    \GPcolortrue
  }{}%
  \@ifundefined{ifGPblacktext}{%
    \newif\ifGPblacktext
    \GPblacktexttrue
  }{}%
  \let\gplgaddtomacro\g@addto@macro
  \gdef\gplbacktext{}%
  \gdef\gplfronttext{}%
  \makeatother
  \ifGPblacktext
    \def\colorrgb#1{}%
    \def\colorgray#1{}%
  \else
    \ifGPcolor
      \def\colorrgb#1{\color[rgb]{#1}}%
      \def\colorgray#1{\color[gray]{#1}}%
      \expandafter\def\csname LTw\endcsname{\color{white}}%
      \expandafter\def\csname LTb\endcsname{\color{black}}%
      \expandafter\def\csname LTa\endcsname{\color{black}}%
      \expandafter\def\csname LT0\endcsname{\color[rgb]{1,0,0}}%
      \expandafter\def\csname LT1\endcsname{\color[rgb]{0,1,0}}%
      \expandafter\def\csname LT2\endcsname{\color[rgb]{0,0,1}}%
      \expandafter\def\csname LT3\endcsname{\color[rgb]{1,0,1}}%
      \expandafter\def\csname LT4\endcsname{\color[rgb]{0,1,1}}%
      \expandafter\def\csname LT5\endcsname{\color[rgb]{1,1,0}}%
      \expandafter\def\csname LT6\endcsname{\color[rgb]{0,0,0}}%
      \expandafter\def\csname LT7\endcsname{\color[rgb]{1,0.3,0}}%
      \expandafter\def\csname LT8\endcsname{\color[rgb]{0.5,0.5,0.5}}%
    \else
      \def\colorrgb#1{\color{black}}%
      \def\colorgray#1{\color[gray]{#1}}%
      \expandafter\def\csname LTw\endcsname{\color{white}}%
      \expandafter\def\csname LTb\endcsname{\color{black}}%
      \expandafter\def\csname LTa\endcsname{\color{black}}%
      \expandafter\def\csname LT0\endcsname{\color{black}}%
      \expandafter\def\csname LT1\endcsname{\color{black}}%
      \expandafter\def\csname LT2\endcsname{\color{black}}%
      \expandafter\def\csname LT3\endcsname{\color{black}}%
      \expandafter\def\csname LT4\endcsname{\color{black}}%
      \expandafter\def\csname LT5\endcsname{\color{black}}%
      \expandafter\def\csname LT6\endcsname{\color{black}}%
      \expandafter\def\csname LT7\endcsname{\color{black}}%
      \expandafter\def\csname LT8\endcsname{\color{black}}%
    \fi
  \fi
    \setlength{\unitlength}{0.0500bp}%
    \ifx\gptboxheight\undefined%
      \newlength{\gptboxheight}%
      \newlength{\gptboxwidth}%
      \newsavebox{\gptboxtext}%
    \fi%
    \setlength{\fboxrule}{0.5pt}%
    \setlength{\fboxsep}{1pt}%
    \definecolor{tbcol}{rgb}{1,1,1}%
\begin{picture}(7200.00,5040.00)%
    \gplgaddtomacro\gplbacktext{%
      \csname LTb\endcsname
      \put(1372,896){\makebox(0,0)[r]{\strut{}1e-02}}%
      \csname LTb\endcsname
      \put(1372,3511){\makebox(0,0)[r]{\strut{}1e-01}}%
      \csname LTb\endcsname
      \put(1540,616){\makebox(0,0){\strut{}$0$}}%
      \csname LTb\endcsname
      \put(2056,616){\makebox(0,0){\strut{}$20$}}%
      \csname LTb\endcsname
      \put(2571,616){\makebox(0,0){\strut{}$40$}}%
      \csname LTb\endcsname
      \put(3087,616){\makebox(0,0){\strut{}$60$}}%
      \csname LTb\endcsname
      \put(3602,616){\makebox(0,0){\strut{}$80$}}%
      \csname LTb\endcsname
      \put(4118,616){\makebox(0,0){\strut{}$100$}}%
      \csname LTb\endcsname
      \put(4633,616){\makebox(0,0){\strut{}$120$}}%
      \csname LTb\endcsname
      \put(5149,616){\makebox(0,0){\strut{}$140$}}%
      \csname LTb\endcsname
      \put(5664,616){\makebox(0,0){\strut{}$160$}}%
      \csname LTb\endcsname
      \put(6180,616){\makebox(0,0){\strut{}$180$}}%
      \csname LTb\endcsname
      \put(6695,616){\makebox(0,0){\strut{}$200$}}%
    }%
    \gplgaddtomacro\gplfronttext{%
      \csname LTb\endcsname
      \put(266,2827){\rotatebox{-270}{\makebox(0,0){\strut{}time-averaged relative error}}}%
      \put(4117,196){\makebox(0,0){\strut{}roll-out length}}%
      \csname LTb\endcsname
      \put(5792,4556){\makebox(0,0)[r]{\strut{}projection}}%
      \csname LTb\endcsname
      \put(5792,4276){\makebox(0,0)[r]{\strut{}OpInf + roll out}}%
      \csname LTb\endcsname
      \put(5792,3996){\makebox(0,0)[r]{\strut{}OpInf}}%
    }%
    \gplbacktext
    \put(0,0){\includegraphics[width={360.00bp},height={252.00bp}]{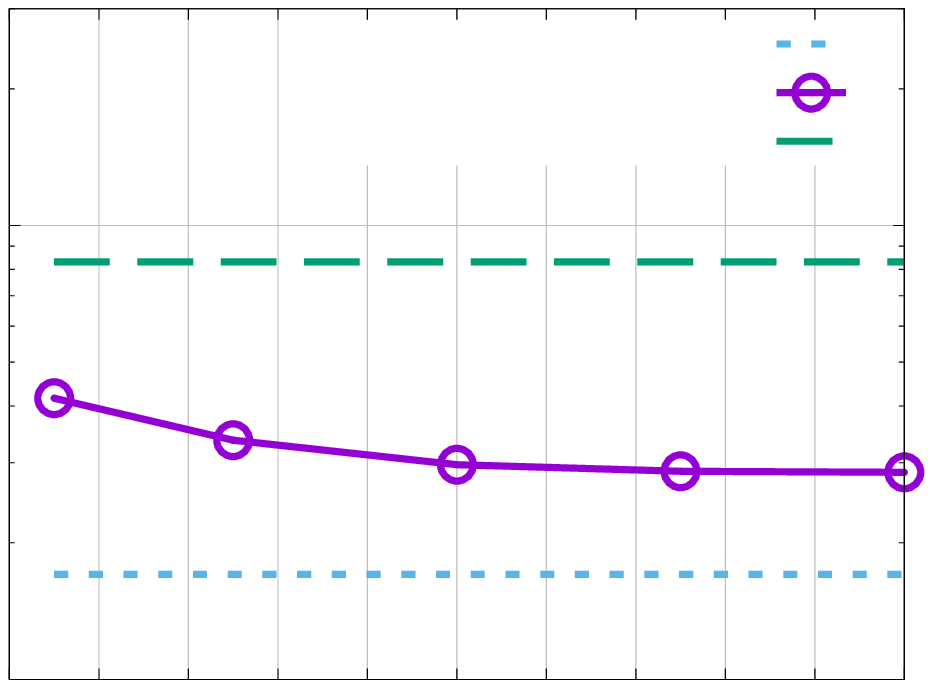}}%
    \gplfronttext
  \end{picture}%
\endgroup

%% file: SQG_errVsnParam_Radius_sparse5.tex
\begingroup
  \makeatletter
  \providecommand\color[2][]{%
    \GenericError{(gnuplot) \space\space\space\@spaces}{%
      Package color not loaded in conjunction with
      terminal option `colourtext'%
    }{See the gnuplot documentation for explanation.%
    }{Either use 'blacktext' in gnuplot or load the package
      color.sty in LaTeX.}%
    \renewcommand\color[2][]{}%
  }%
  \providecommand\includegraphics[2][]{%
    \GenericError{(gnuplot) \space\space\space\@spaces}{%
      Package graphicx or graphics not loaded%
    }{See the gnuplot documentation for explanation.%
    }{The gnuplot epslatex terminal needs graphicx.sty or graphics.sty.}%
    \renewcommand\includegraphics[2][]{}%
  }%
  \providecommand\rotatebox[2]{#2}%
  \@ifundefined{ifGPcolor}{%
    \newif\ifGPcolor
    \GPcolortrue
  }{}%
  \@ifundefined{ifGPblacktext}{%
    \newif\ifGPblacktext
    \GPblacktexttrue
  }{}%
  \let\gplgaddtomacro\g@addto@macro
  \gdef\gplbacktext{}%
  \gdef\gplfronttext{}%
  \makeatother
  \ifGPblacktext
    \def\colorrgb#1{}%
    \def\colorgray#1{}%
  \else
    \ifGPcolor
      \def\colorrgb#1{\color[rgb]{#1}}%
      \def\colorgray#1{\color[gray]{#1}}%
      \expandafter\def\csname LTw\endcsname{\color{white}}%
      \expandafter\def\csname LTb\endcsname{\color{black}}%
      \expandafter\def\csname LTa\endcsname{\color{black}}%
      \expandafter\def\csname LT0\endcsname{\color[rgb]{1,0,0}}%
      \expandafter\def\csname LT1\endcsname{\color[rgb]{0,1,0}}%
      \expandafter\def\csname LT2\endcsname{\color[rgb]{0,0,1}}%
      \expandafter\def\csname LT3\endcsname{\color[rgb]{1,0,1}}%
      \expandafter\def\csname LT4\endcsname{\color[rgb]{0,1,1}}%
      \expandafter\def\csname LT5\endcsname{\color[rgb]{1,1,0}}%
      \expandafter\def\csname LT6\endcsname{\color[rgb]{0,0,0}}%
      \expandafter\def\csname LT7\endcsname{\color[rgb]{1,0.3,0}}%
      \expandafter\def\csname LT8\endcsname{\color[rgb]{0.5,0.5,0.5}}%
    \else
      \def\colorrgb#1{\color{black}}%
      \def\colorgray#1{\color[gray]{#1}}%
      \expandafter\def\csname LTw\endcsname{\color{white}}%
      \expandafter\def\csname LTb\endcsname{\color{black}}%
      \expandafter\def\csname LTa\endcsname{\color{black}}%
      \expandafter\def\csname LT0\endcsname{\color{black}}%
      \expandafter\def\csname LT1\endcsname{\color{black}}%
      \expandafter\def\csname LT2\endcsname{\color{black}}%
      \expandafter\def\csname LT3\endcsname{\color{black}}%
      \expandafter\def\csname LT4\endcsname{\color{black}}%
      \expandafter\def\csname LT5\endcsname{\color{black}}%
      \expandafter\def\csname LT6\endcsname{\color{black}}%
      \expandafter\def\csname LT7\endcsname{\color{black}}%
      \expandafter\def\csname LT8\endcsname{\color{black}}%
    \fi
  \fi
    \setlength{\unitlength}{0.0500bp}%
    \ifx\gptboxheight\undefined%
      \newlength{\gptboxheight}%
      \newlength{\gptboxwidth}%
      \newsavebox{\gptboxtext}%
    \fi%
    \setlength{\fboxrule}{0.5pt}%
    \setlength{\fboxsep}{1pt}%
    \definecolor{tbcol}{rgb}{1,1,1}%
\begin{picture}(7200.00,5040.00)%
    \gplgaddtomacro\gplbacktext{%
      \csname LTb\endcsname
      \put(1372,896){\makebox(0,0)[r]{\strut{}1e-09}}%
      \csname LTb\endcsname
      \put(1372,1669){\makebox(0,0)[r]{\strut{}1e-08}}%
      \csname LTb\endcsname
      \put(1372,2441){\makebox(0,0)[r]{\strut{}1e-07}}%
      \csname LTb\endcsname
      \put(1372,3214){\makebox(0,0)[r]{\strut{}1e-06}}%
      \csname LTb\endcsname
      \put(1372,3986){\makebox(0,0)[r]{\strut{}1e-05}}%
      \csname LTb\endcsname
      \put(1372,4759){\makebox(0,0)[r]{\strut{}1e-04}}%
      \put(2829,616){\makebox(0,0){\strut{}4}}%
      \put(4118,616){\makebox(0,0){\strut{}8}}%
      \put(5406,616){\makebox(0,0){\strut{}12}}%
    }%
    \gplgaddtomacro\gplfronttext{%
      \csname LTb\endcsname
      \put(266,2827){\rotatebox{-270}{\makebox(0,0){\strut{}bound of stability radius}}}%
      \put(4117,196){\makebox(0,0){\strut{}number of training trajectories}}%
      \csname LTb\endcsname
      \put(5792,4556){\makebox(0,0)[r]{\strut{}OpInf + roll out}}%
      \csname LTb\endcsname
      \put(5792,4276){\makebox(0,0)[r]{\strut{}OpInf}}%
    }%
    \gplbacktext
    \put(0,0){\includegraphics[width={360.00bp},height={252.00bp}]{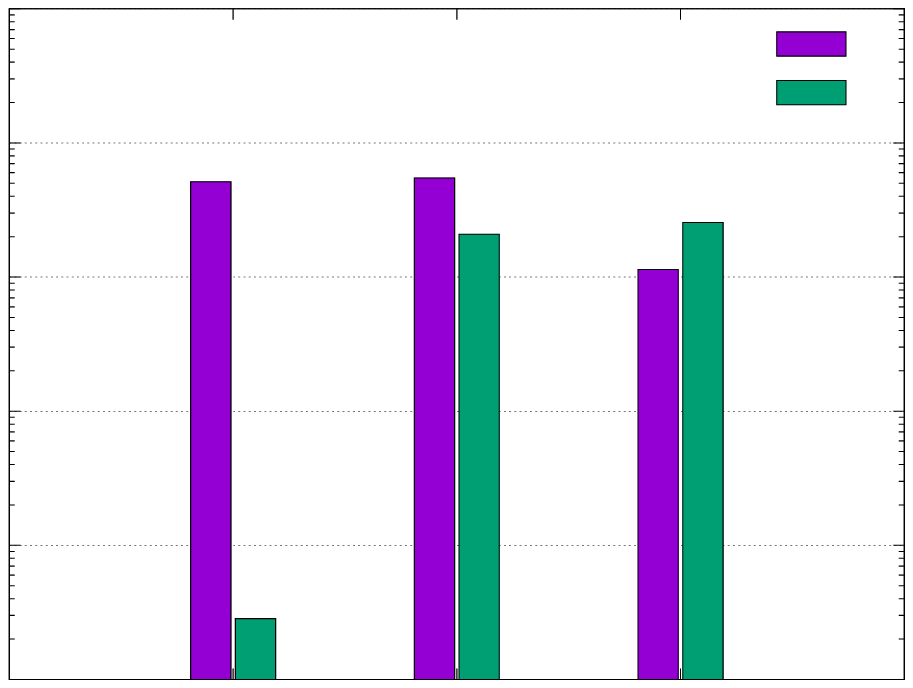}}%
    \gplfronttext
  \end{picture}%
\endgroup

%% file: SQG_errVsnParam_Radius_sparse10.tex
\begingroup
  \makeatletter
  \providecommand\color[2][]{%
    \GenericError{(gnuplot) \space\space\space\@spaces}{%
      Package color not loaded in conjunction with
      terminal option `colourtext'%
    }{See the gnuplot documentation for explanation.%
    }{Either use 'blacktext' in gnuplot or load the package
      color.sty in LaTeX.}%
    \renewcommand\color[2][]{}%
  }%
  \providecommand\includegraphics[2][]{%
    \GenericError{(gnuplot) \space\space\space\@spaces}{%
      Package graphicx or graphics not loaded%
    }{See the gnuplot documentation for explanation.%
    }{The gnuplot epslatex terminal needs graphicx.sty or graphics.sty.}%
    \renewcommand\includegraphics[2][]{}%
  }%
  \providecommand\rotatebox[2]{#2}%
  \@ifundefined{ifGPcolor}{%
    \newif\ifGPcolor
    \GPcolortrue
  }{}%
  \@ifundefined{ifGPblacktext}{%
    \newif\ifGPblacktext
    \GPblacktexttrue
  }{}%
  \let\gplgaddtomacro\g@addto@macro
  \gdef\gplbacktext{}%
  \gdef\gplfronttext{}%
  \makeatother
  \ifGPblacktext
    \def\colorrgb#1{}%
    \def\colorgray#1{}%
  \else
    \ifGPcolor
      \def\colorrgb#1{\color[rgb]{#1}}%
      \def\colorgray#1{\color[gray]{#1}}%
      \expandafter\def\csname LTw\endcsname{\color{white}}%
      \expandafter\def\csname LTb\endcsname{\color{black}}%
      \expandafter\def\csname LTa\endcsname{\color{black}}%
      \expandafter\def\csname LT0\endcsname{\color[rgb]{1,0,0}}%
      \expandafter\def\csname LT1\endcsname{\color[rgb]{0,1,0}}%
      \expandafter\def\csname LT2\endcsname{\color[rgb]{0,0,1}}%
      \expandafter\def\csname LT3\endcsname{\color[rgb]{1,0,1}}%
      \expandafter\def\csname LT4\endcsname{\color[rgb]{0,1,1}}%
      \expandafter\def\csname LT5\endcsname{\color[rgb]{1,1,0}}%
      \expandafter\def\csname LT6\endcsname{\color[rgb]{0,0,0}}%
      \expandafter\def\csname LT7\endcsname{\color[rgb]{1,0.3,0}}%
      \expandafter\def\csname LT8\endcsname{\color[rgb]{0.5,0.5,0.5}}%
    \else
      \def\colorrgb#1{\color{black}}%
      \def\colorgray#1{\color[gray]{#1}}%
      \expandafter\def\csname LTw\endcsname{\color{white}}%
      \expandafter\def\csname LTb\endcsname{\color{black}}%
      \expandafter\def\csname LTa\endcsname{\color{black}}%
      \expandafter\def\csname LT0\endcsname{\color{black}}%
      \expandafter\def\csname LT1\endcsname{\color{black}}%
      \expandafter\def\csname LT2\endcsname{\color{black}}%
      \expandafter\def\csname LT3\endcsname{\color{black}}%
      \expandafter\def\csname LT4\endcsname{\color{black}}%
      \expandafter\def\csname LT5\endcsname{\color{black}}%
      \expandafter\def\csname LT6\endcsname{\color{black}}%
      \expandafter\def\csname LT7\endcsname{\color{black}}%
      \expandafter\def\csname LT8\endcsname{\color{black}}%
    \fi
  \fi
    \setlength{\unitlength}{0.0500bp}%
    \ifx\gptboxheight\undefined%
      \newlength{\gptboxheight}%
      \newlength{\gptboxwidth}%
      \newsavebox{\gptboxtext}%
    \fi%
    \setlength{\fboxrule}{0.5pt}%
    \setlength{\fboxsep}{1pt}%
    \definecolor{tbcol}{rgb}{1,1,1}%
\begin{picture}(7200.00,5040.00)%
    \gplgaddtomacro\gplbacktext{%
      \csname LTb\endcsname
      \put(1372,896){\makebox(0,0)[r]{\strut{}1e-09}}%
      \csname LTb\endcsname
      \put(1372,1669){\makebox(0,0)[r]{\strut{}1e-08}}%
      \csname LTb\endcsname
      \put(1372,2441){\makebox(0,0)[r]{\strut{}1e-07}}%
      \csname LTb\endcsname
      \put(1372,3214){\makebox(0,0)[r]{\strut{}1e-06}}%
      \csname LTb\endcsname
      \put(1372,3986){\makebox(0,0)[r]{\strut{}1e-05}}%
      \csname LTb\endcsname
      \put(1372,4759){\makebox(0,0)[r]{\strut{}1e-04}}%
      \put(2829,616){\makebox(0,0){\strut{}4}}%
      \put(4118,616){\makebox(0,0){\strut{}8}}%
      \put(5406,616){\makebox(0,0){\strut{}12}}%
    }%
    \gplgaddtomacro\gplfronttext{%
      \csname LTb\endcsname
      \put(266,2827){\rotatebox{-270}{\makebox(0,0){\strut{}bound of stability radius}}}%
      \put(4117,196){\makebox(0,0){\strut{}number of training trajectories}}%
      \csname LTb\endcsname
      \put(5792,4556){\makebox(0,0)[r]{\strut{}OpInf + roll out}}%
      \csname LTb\endcsname
      \put(5792,4276){\makebox(0,0)[r]{\strut{}OpInf}}%
    }%
    \gplbacktext
    \put(0,0){\includegraphics[width={360.00bp},height={252.00bp}]{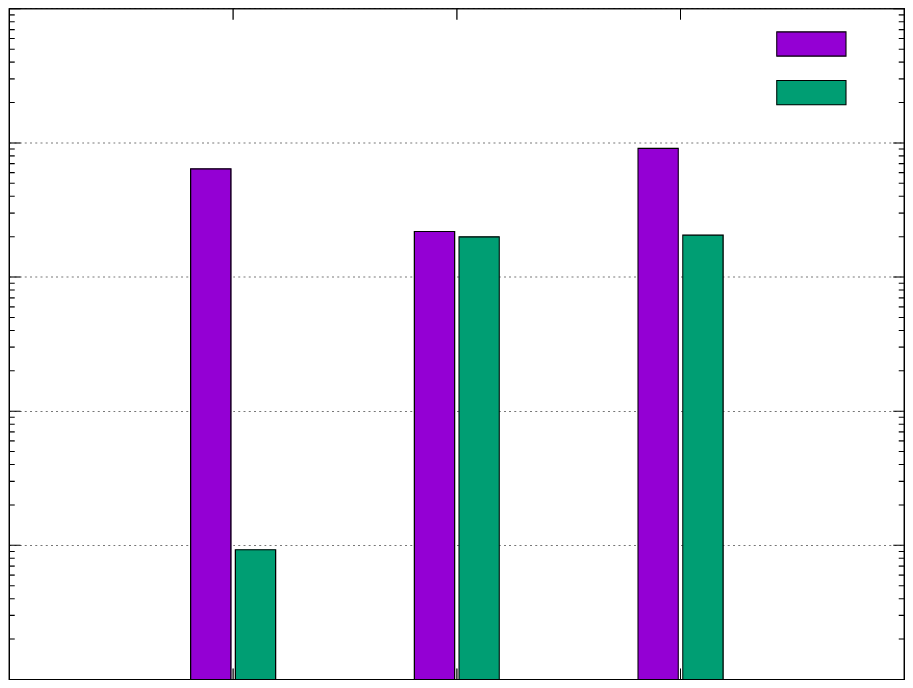}}%
    \gplfronttext
  \end{picture}%
\endgroup

%% file: SQG_noise_errVsNoise_sparse5.tex
\begingroup
  \makeatletter
  \providecommand\color[2][]{%
    \GenericError{(gnuplot) \space\space\space\@spaces}{%
      Package color not loaded in conjunction with
      terminal option `colourtext'%
    }{See the gnuplot documentation for explanation.%
    }{Either use 'blacktext' in gnuplot or load the package
      color.sty in LaTeX.}%
    \renewcommand\color[2][]{}%
  }%
  \providecommand\includegraphics[2][]{%
    \GenericError{(gnuplot) \space\space\space\@spaces}{%
      Package graphicx or graphics not loaded%
    }{See the gnuplot documentation for explanation.%
    }{The gnuplot epslatex terminal needs graphicx.sty or graphics.sty.}%
    \renewcommand\includegraphics[2][]{}%
  }%
  \providecommand\rotatebox[2]{#2}%
  \@ifundefined{ifGPcolor}{%
    \newif\ifGPcolor
    \GPcolortrue
  }{}%
  \@ifundefined{ifGPblacktext}{%
    \newif\ifGPblacktext
    \GPblacktexttrue
  }{}%
  \let\gplgaddtomacro\g@addto@macro
  \gdef\gplbacktext{}%
  \gdef\gplfronttext{}%
  \makeatother
  \ifGPblacktext
    \def\colorrgb#1{}%
    \def\colorgray#1{}%
  \else
    \ifGPcolor
      \def\colorrgb#1{\color[rgb]{#1}}%
      \def\colorgray#1{\color[gray]{#1}}%
      \expandafter\def\csname LTw\endcsname{\color{white}}%
      \expandafter\def\csname LTb\endcsname{\color{black}}%
      \expandafter\def\csname LTa\endcsname{\color{black}}%
      \expandafter\def\csname LT0\endcsname{\color[rgb]{1,0,0}}%
      \expandafter\def\csname LT1\endcsname{\color[rgb]{0,1,0}}%
      \expandafter\def\csname LT2\endcsname{\color[rgb]{0,0,1}}%
      \expandafter\def\csname LT3\endcsname{\color[rgb]{1,0,1}}%
      \expandafter\def\csname LT4\endcsname{\color[rgb]{0,1,1}}%
      \expandafter\def\csname LT5\endcsname{\color[rgb]{1,1,0}}%
      \expandafter\def\csname LT6\endcsname{\color[rgb]{0,0,0}}%
      \expandafter\def\csname LT7\endcsname{\color[rgb]{1,0.3,0}}%
      \expandafter\def\csname LT8\endcsname{\color[rgb]{0.5,0.5,0.5}}%
    \else
      \def\colorrgb#1{\color{black}}%
      \def\colorgray#1{\color[gray]{#1}}%
      \expandafter\def\csname LTw\endcsname{\color{white}}%
      \expandafter\def\csname LTb\endcsname{\color{black}}%
      \expandafter\def\csname LTa\endcsname{\color{black}}%
      \expandafter\def\csname LT0\endcsname{\color{black}}%
      \expandafter\def\csname LT1\endcsname{\color{black}}%
      \expandafter\def\csname LT2\endcsname{\color{black}}%
      \expandafter\def\csname LT3\endcsname{\color{black}}%
      \expandafter\def\csname LT4\endcsname{\color{black}}%
      \expandafter\def\csname LT5\endcsname{\color{black}}%
      \expandafter\def\csname LT6\endcsname{\color{black}}%
      \expandafter\def\csname LT7\endcsname{\color{black}}%
      \expandafter\def\csname LT8\endcsname{\color{black}}%
    \fi
  \fi
    \setlength{\unitlength}{0.0500bp}%
    \ifx\gptboxheight\undefined%
      \newlength{\gptboxheight}%
      \newlength{\gptboxwidth}%
      \newsavebox{\gptboxtext}%
    \fi%
    \setlength{\fboxrule}{0.5pt}%
    \setlength{\fboxsep}{1pt}%
    \definecolor{tbcol}{rgb}{1,1,1}%
\begin{picture}(7200.00,5040.00)%
    \gplgaddtomacro\gplbacktext{%
      \csname LTb\endcsname
      \put(1372,896){\makebox(0,0)[r]{\strut{}1e-02}}%
      \csname LTb\endcsname
      \put(1372,2575){\makebox(0,0)[r]{\strut{}1e-01}}%
      \csname LTb\endcsname
      \put(1372,4254){\makebox(0,0)[r]{\strut{}1e+00}}%
      \put(2829,616){\makebox(0,0){\strut{}10.0\%}}%
      \put(4118,616){\makebox(0,0){\strut{}15.0\%}}%
      \put(5406,616){\makebox(0,0){\strut{}25.0\%}}%
    }%
    \gplgaddtomacro\gplfronttext{%
      \csname LTb\endcsname
      \put(266,2827){\rotatebox{-270}{\makebox(0,0){\strut{}time-averaged relative error}}}%
      \put(4117,196){\makebox(0,0){\strut{}noise percentage}}%
      \csname LTb\endcsname
      \put(2443,4556){\makebox(0,0)[l]{\strut{}projection}}%
      \csname LTb\endcsname
      \put(2443,4276){\makebox(0,0)[l]{\strut{}OpInf + roll out}}%
      \csname LTb\endcsname
      \put(2443,3996){\makebox(0,0)[l]{\strut{}OpInf}}%
    }%
    \gplbacktext
    \put(0,0){\includegraphics[width={360.00bp},height={252.00bp}]{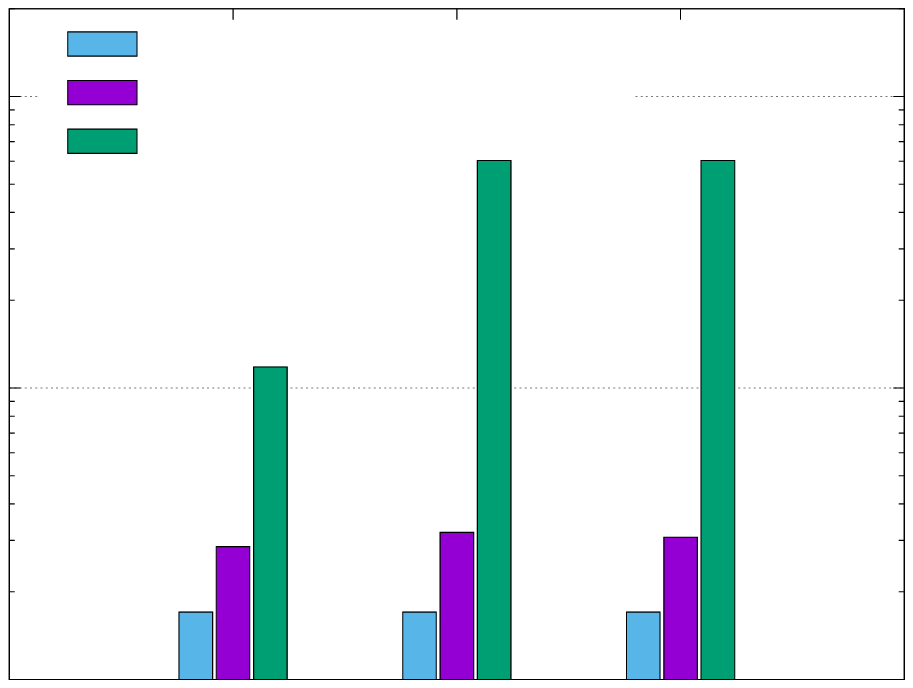}}%
    \gplfronttext
  \end{picture}%
\endgroup

%% file: SQG_noise_errVsNoise_sparse10.tex
\begingroup
  \makeatletter
  \providecommand\color[2][]{%
    \GenericError{(gnuplot) \space\space\space\@spaces}{%
      Package color not loaded in conjunction with
      terminal option `colourtext'%
    }{See the gnuplot documentation for explanation.%
    }{Either use 'blacktext' in gnuplot or load the package
      color.sty in LaTeX.}%
    \renewcommand\color[2][]{}%
  }%
  \providecommand\includegraphics[2][]{%
    \GenericError{(gnuplot) \space\space\space\@spaces}{%
      Package graphicx or graphics not loaded%
    }{See the gnuplot documentation for explanation.%
    }{The gnuplot epslatex terminal needs graphicx.sty or graphics.sty.}%
    \renewcommand\includegraphics[2][]{}%
  }%
  \providecommand\rotatebox[2]{#2}%
  \@ifundefined{ifGPcolor}{%
    \newif\ifGPcolor
    \GPcolortrue
  }{}%
  \@ifundefined{ifGPblacktext}{%
    \newif\ifGPblacktext
    \GPblacktexttrue
  }{}%
  \let\gplgaddtomacro\g@addto@macro
  \gdef\gplbacktext{}%
  \gdef\gplfronttext{}%
  \makeatother
  \ifGPblacktext
    \def\colorrgb#1{}%
    \def\colorgray#1{}%
  \else
    \ifGPcolor
      \def\colorrgb#1{\color[rgb]{#1}}%
      \def\colorgray#1{\color[gray]{#1}}%
      \expandafter\def\csname LTw\endcsname{\color{white}}%
      \expandafter\def\csname LTb\endcsname{\color{black}}%
      \expandafter\def\csname LTa\endcsname{\color{black}}%
      \expandafter\def\csname LT0\endcsname{\color[rgb]{1,0,0}}%
      \expandafter\def\csname LT1\endcsname{\color[rgb]{0,1,0}}%
      \expandafter\def\csname LT2\endcsname{\color[rgb]{0,0,1}}%
      \expandafter\def\csname LT3\endcsname{\color[rgb]{1,0,1}}%
      \expandafter\def\csname LT4\endcsname{\color[rgb]{0,1,1}}%
      \expandafter\def\csname LT5\endcsname{\color[rgb]{1,1,0}}%
      \expandafter\def\csname LT6\endcsname{\color[rgb]{0,0,0}}%
      \expandafter\def\csname LT7\endcsname{\color[rgb]{1,0.3,0}}%
      \expandafter\def\csname LT8\endcsname{\color[rgb]{0.5,0.5,0.5}}%
    \else
      \def\colorrgb#1{\color{black}}%
      \def\colorgray#1{\color[gray]{#1}}%
      \expandafter\def\csname LTw\endcsname{\color{white}}%
      \expandafter\def\csname LTb\endcsname{\color{black}}%
      \expandafter\def\csname LTa\endcsname{\color{black}}%
      \expandafter\def\csname LT0\endcsname{\color{black}}%
      \expandafter\def\csname LT1\endcsname{\color{black}}%
      \expandafter\def\csname LT2\endcsname{\color{black}}%
      \expandafter\def\csname LT3\endcsname{\color{black}}%
      \expandafter\def\csname LT4\endcsname{\color{black}}%
      \expandafter\def\csname LT5\endcsname{\color{black}}%
      \expandafter\def\csname LT6\endcsname{\color{black}}%
      \expandafter\def\csname LT7\endcsname{\color{black}}%
      \expandafter\def\csname LT8\endcsname{\color{black}}%
    \fi
  \fi
    \setlength{\unitlength}{0.0500bp}%
    \ifx\gptboxheight\undefined%
      \newlength{\gptboxheight}%
      \newlength{\gptboxwidth}%
      \newsavebox{\gptboxtext}%
    \fi%
    \setlength{\fboxrule}{0.5pt}%
    \setlength{\fboxsep}{1pt}%
    \definecolor{tbcol}{rgb}{1,1,1}%
\begin{picture}(7200.00,5040.00)%
    \gplgaddtomacro\gplbacktext{%
      \csname LTb\endcsname
      \put(1372,896){\makebox(0,0)[r]{\strut{}1e-02}}%
      \csname LTb\endcsname
      \put(1372,2575){\makebox(0,0)[r]{\strut{}1e-01}}%
      \csname LTb\endcsname
      \put(1372,4254){\makebox(0,0)[r]{\strut{}1e+00}}%
      \put(2829,616){\makebox(0,0){\strut{}10.0\%}}%
      \put(4118,616){\makebox(0,0){\strut{}15.0\%}}%
      \put(5406,616){\makebox(0,0){\strut{}25.0\%}}%
    }%
    \gplgaddtomacro\gplfronttext{%
      \csname LTb\endcsname
      \put(266,2827){\rotatebox{-270}{\makebox(0,0){\strut{}time-averaged relative error}}}%
      \put(4117,196){\makebox(0,0){\strut{}noise percentage}}%
      \csname LTb\endcsname
      \put(5792,4556){\makebox(0,0)[r]{\strut{}projection}}%
      \csname LTb\endcsname
      \put(5792,4276){\makebox(0,0)[r]{\strut{}OpInf + roll out}}%
      \csname LTb\endcsname
      \put(5792,3996){\makebox(0,0)[r]{\strut{}OpInf}}%
    }%
    \gplbacktext
    \put(0,0){\includegraphics[width={360.00bp},height={252.00bp}]{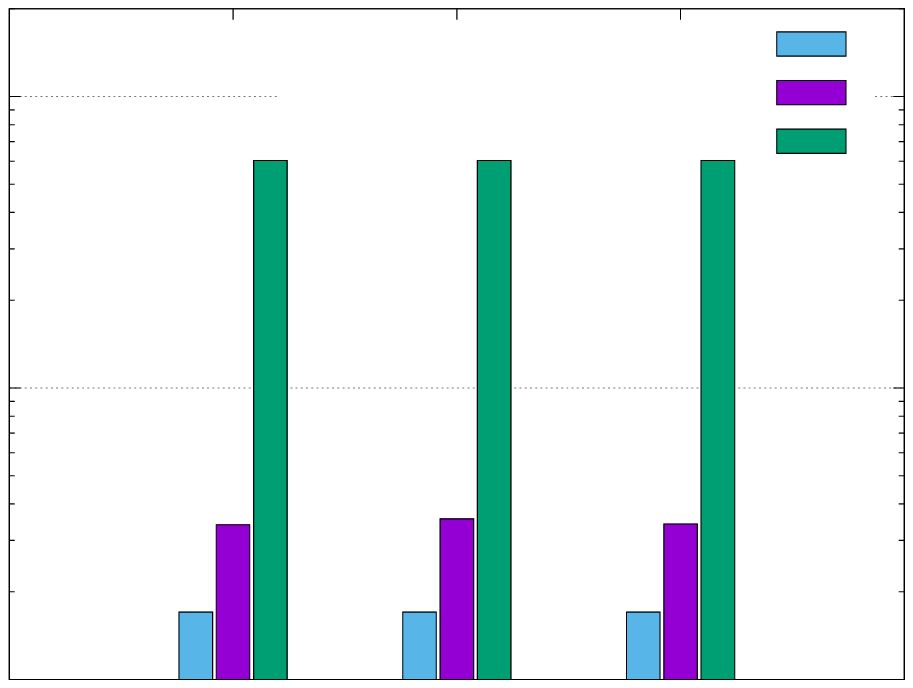}}%
    \gplfronttext
  \end{picture}%
\endgroup

%% file: SQG_noise_errVsRoll_sparse5.tex
\begingroup
  \makeatletter
  \providecommand\color[2][]{%
    \GenericError{(gnuplot) \space\space\space\@spaces}{%
      Package color not loaded in conjunction with
      terminal option `colourtext'%
    }{See the gnuplot documentation for explanation.%
    }{Either use 'blacktext' in gnuplot or load the package
      color.sty in LaTeX.}%
    \renewcommand\color[2][]{}%
  }%
  \providecommand\includegraphics[2][]{%
    \GenericError{(gnuplot) \space\space\space\@spaces}{%
      Package graphicx or graphics not loaded%
    }{See the gnuplot documentation for explanation.%
    }{The gnuplot epslatex terminal needs graphicx.sty or graphics.sty.}%
    \renewcommand\includegraphics[2][]{}%
  }%
  \providecommand\rotatebox[2]{#2}%
  \@ifundefined{ifGPcolor}{%
    \newif\ifGPcolor
    \GPcolortrue
  }{}%
  \@ifundefined{ifGPblacktext}{%
    \newif\ifGPblacktext
    \GPblacktexttrue
  }{}%
  \let\gplgaddtomacro\g@addto@macro
  \gdef\gplbacktext{}%
  \gdef\gplfronttext{}%
  \makeatother
  \ifGPblacktext
    \def\colorrgb#1{}%
    \def\colorgray#1{}%
  \else
    \ifGPcolor
      \def\colorrgb#1{\color[rgb]{#1}}%
      \def\colorgray#1{\color[gray]{#1}}%
      \expandafter\def\csname LTw\endcsname{\color{white}}%
      \expandafter\def\csname LTb\endcsname{\color{black}}%
      \expandafter\def\csname LTa\endcsname{\color{black}}%
      \expandafter\def\csname LT0\endcsname{\color[rgb]{1,0,0}}%
      \expandafter\def\csname LT1\endcsname{\color[rgb]{0,1,0}}%
      \expandafter\def\csname LT2\endcsname{\color[rgb]{0,0,1}}%
      \expandafter\def\csname LT3\endcsname{\color[rgb]{1,0,1}}%
      \expandafter\def\csname LT4\endcsname{\color[rgb]{0,1,1}}%
      \expandafter\def\csname LT5\endcsname{\color[rgb]{1,1,0}}%
      \expandafter\def\csname LT6\endcsname{\color[rgb]{0,0,0}}%
      \expandafter\def\csname LT7\endcsname{\color[rgb]{1,0.3,0}}%
      \expandafter\def\csname LT8\endcsname{\color[rgb]{0.5,0.5,0.5}}%
    \else
      \def\colorrgb#1{\color{black}}%
      \def\colorgray#1{\color[gray]{#1}}%
      \expandafter\def\csname LTw\endcsname{\color{white}}%
      \expandafter\def\csname LTb\endcsname{\color{black}}%
      \expandafter\def\csname LTa\endcsname{\color{black}}%
      \expandafter\def\csname LT0\endcsname{\color{black}}%
      \expandafter\def\csname LT1\endcsname{\color{black}}%
      \expandafter\def\csname LT2\endcsname{\color{black}}%
      \expandafter\def\csname LT3\endcsname{\color{black}}%
      \expandafter\def\csname LT4\endcsname{\color{black}}%
      \expandafter\def\csname LT5\endcsname{\color{black}}%
      \expandafter\def\csname LT6\endcsname{\color{black}}%
      \expandafter\def\csname LT7\endcsname{\color{black}}%
      \expandafter\def\csname LT8\endcsname{\color{black}}%
    \fi
  \fi
    \setlength{\unitlength}{0.0500bp}%
    \ifx\gptboxheight\undefined%
      \newlength{\gptboxheight}%
      \newlength{\gptboxwidth}%
      \newsavebox{\gptboxtext}%
    \fi%
    \setlength{\fboxrule}{0.5pt}%
    \setlength{\fboxsep}{1pt}%
    \definecolor{tbcol}{rgb}{1,1,1}%
\begin{picture}(7200.00,5040.00)%
    \gplgaddtomacro\gplbacktext{%
      \csname LTb\endcsname
      \put(1372,896){\makebox(0,0)[r]{\strut{}1e-02}}%
      \csname LTb\endcsname
      \put(1372,2538){\makebox(0,0)[r]{\strut{}1e-01}}%
      \csname LTb\endcsname
      \put(1372,4181){\makebox(0,0)[r]{\strut{}1e+00}}%
      \csname LTb\endcsname
      \put(1540,616){\makebox(0,0){\strut{}$0$}}%
      \csname LTb\endcsname
      \put(2056,616){\makebox(0,0){\strut{}$20$}}%
      \csname LTb\endcsname
      \put(2571,616){\makebox(0,0){\strut{}$40$}}%
      \csname LTb\endcsname
      \put(3087,616){\makebox(0,0){\strut{}$60$}}%
      \csname LTb\endcsname
      \put(3602,616){\makebox(0,0){\strut{}$80$}}%
      \csname LTb\endcsname
      \put(4118,616){\makebox(0,0){\strut{}$100$}}%
      \csname LTb\endcsname
      \put(4633,616){\makebox(0,0){\strut{}$120$}}%
      \csname LTb\endcsname
      \put(5149,616){\makebox(0,0){\strut{}$140$}}%
      \csname LTb\endcsname
      \put(5664,616){\makebox(0,0){\strut{}$160$}}%
      \csname LTb\endcsname
      \put(6180,616){\makebox(0,0){\strut{}$180$}}%
      \csname LTb\endcsname
      \put(6695,616){\makebox(0,0){\strut{}$200$}}%
    }%
    \gplgaddtomacro\gplfronttext{%
      \csname LTb\endcsname
      \put(266,2827){\rotatebox{-270}{\makebox(0,0){\strut{}time-averaged relative error}}}%
      \put(4117,196){\makebox(0,0){\strut{}roll-out length}}%
      \csname LTb\endcsname
      \put(5792,4556){\makebox(0,0)[r]{\strut{}projection}}%
      \csname LTb\endcsname
      \put(5792,4276){\makebox(0,0)[r]{\strut{}OpInf + roll out}}%
      \csname LTb\endcsname
      \put(5792,3996){\makebox(0,0)[r]{\strut{}OpInf}}%
    }%
    \gplbacktext
    \put(0,0){\includegraphics[width={360.00bp},height={252.00bp}]{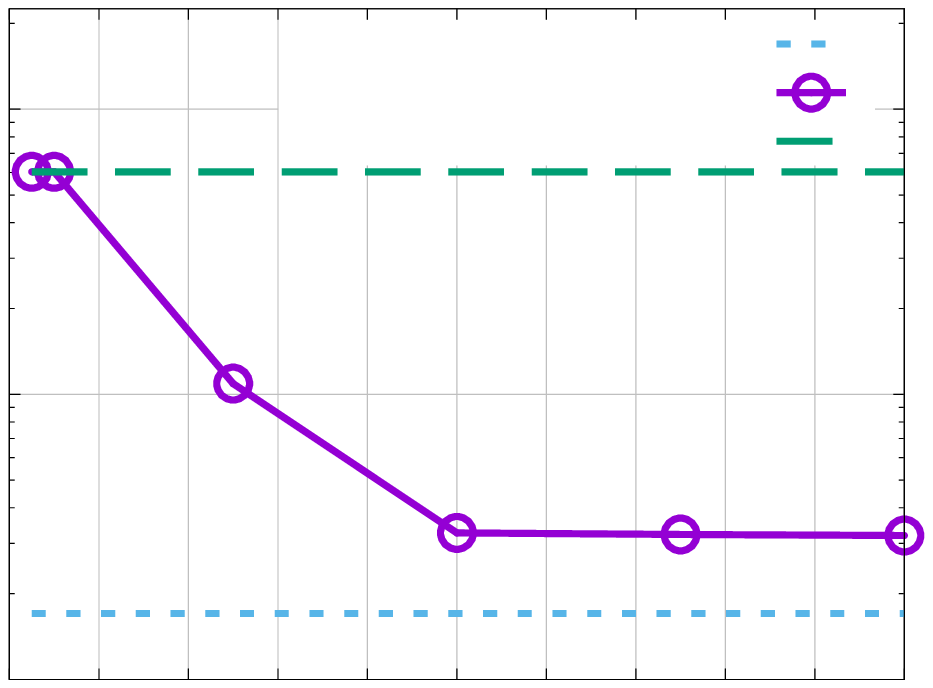}}%
    \gplfronttext
  \end{picture}%
\endgroup

%% file: SQG_noise_errVsRoll_sparse10.tex
\begingroup
  \makeatletter
  \providecommand\color[2][]{%
    \GenericError{(gnuplot) \space\space\space\@spaces}{%
      Package color not loaded in conjunction with
      terminal option `colourtext'%
    }{See the gnuplot documentation for explanation.%
    }{Either use 'blacktext' in gnuplot or load the package
      color.sty in LaTeX.}%
    \renewcommand\color[2][]{}%
  }%
  \providecommand\includegraphics[2][]{%
    \GenericError{(gnuplot) \space\space\space\@spaces}{%
      Package graphicx or graphics not loaded%
    }{See the gnuplot documentation for explanation.%
    }{The gnuplot epslatex terminal needs graphicx.sty or graphics.sty.}%
    \renewcommand\includegraphics[2][]{}%
  }%
  \providecommand\rotatebox[2]{#2}%
  \@ifundefined{ifGPcolor}{%
    \newif\ifGPcolor
    \GPcolortrue
  }{}%
  \@ifundefined{ifGPblacktext}{%
    \newif\ifGPblacktext
    \GPblacktexttrue
  }{}%
  \let\gplgaddtomacro\g@addto@macro
  \gdef\gplbacktext{}%
  \gdef\gplfronttext{}%
  \makeatother
  \ifGPblacktext
    \def\colorrgb#1{}%
    \def\colorgray#1{}%
  \else
    \ifGPcolor
      \def\colorrgb#1{\color[rgb]{#1}}%
      \def\colorgray#1{\color[gray]{#1}}%
      \expandafter\def\csname LTw\endcsname{\color{white}}%
      \expandafter\def\csname LTb\endcsname{\color{black}}%
      \expandafter\def\csname LTa\endcsname{\color{black}}%
      \expandafter\def\csname LT0\endcsname{\color[rgb]{1,0,0}}%
      \expandafter\def\csname LT1\endcsname{\color[rgb]{0,1,0}}%
      \expandafter\def\csname LT2\endcsname{\color[rgb]{0,0,1}}%
      \expandafter\def\csname LT3\endcsname{\color[rgb]{1,0,1}}%
      \expandafter\def\csname LT4\endcsname{\color[rgb]{0,1,1}}%
      \expandafter\def\csname LT5\endcsname{\color[rgb]{1,1,0}}%
      \expandafter\def\csname LT6\endcsname{\color[rgb]{0,0,0}}%
      \expandafter\def\csname LT7\endcsname{\color[rgb]{1,0.3,0}}%
      \expandafter\def\csname LT8\endcsname{\color[rgb]{0.5,0.5,0.5}}%
    \else
      \def\colorrgb#1{\color{black}}%
      \def\colorgray#1{\color[gray]{#1}}%
      \expandafter\def\csname LTw\endcsname{\color{white}}%
      \expandafter\def\csname LTb\endcsname{\color{black}}%
      \expandafter\def\csname LTa\endcsname{\color{black}}%
      \expandafter\def\csname LT0\endcsname{\color{black}}%
      \expandafter\def\csname LT1\endcsname{\color{black}}%
      \expandafter\def\csname LT2\endcsname{\color{black}}%
      \expandafter\def\csname LT3\endcsname{\color{black}}%
      \expandafter\def\csname LT4\endcsname{\color{black}}%
      \expandafter\def\csname LT5\endcsname{\color{black}}%
      \expandafter\def\csname LT6\endcsname{\color{black}}%
      \expandafter\def\csname LT7\endcsname{\color{black}}%
      \expandafter\def\csname LT8\endcsname{\color{black}}%
    \fi
  \fi
    \setlength{\unitlength}{0.0500bp}%
    \ifx\gptboxheight\undefined%
      \newlength{\gptboxheight}%
      \newlength{\gptboxwidth}%
      \newsavebox{\gptboxtext}%
    \fi%
    \setlength{\fboxrule}{0.5pt}%
    \setlength{\fboxsep}{1pt}%
    \definecolor{tbcol}{rgb}{1,1,1}%
\begin{picture}(7200.00,5040.00)%
    \gplgaddtomacro\gplbacktext{%
      \csname LTb\endcsname
      \put(1372,896){\makebox(0,0)[r]{\strut{}1e-02}}%
      \csname LTb\endcsname
      \put(1372,2538){\makebox(0,0)[r]{\strut{}1e-01}}%
      \csname LTb\endcsname
      \put(1372,4181){\makebox(0,0)[r]{\strut{}1e+00}}%
      \csname LTb\endcsname
      \put(1540,616){\makebox(0,0){\strut{}$0$}}%
      \csname LTb\endcsname
      \put(2056,616){\makebox(0,0){\strut{}$20$}}%
      \csname LTb\endcsname
      \put(2571,616){\makebox(0,0){\strut{}$40$}}%
      \csname LTb\endcsname
      \put(3087,616){\makebox(0,0){\strut{}$60$}}%
      \csname LTb\endcsname
      \put(3602,616){\makebox(0,0){\strut{}$80$}}%
      \csname LTb\endcsname
      \put(4118,616){\makebox(0,0){\strut{}$100$}}%
      \csname LTb\endcsname
      \put(4633,616){\makebox(0,0){\strut{}$120$}}%
      \csname LTb\endcsname
      \put(5149,616){\makebox(0,0){\strut{}$140$}}%
      \csname LTb\endcsname
      \put(5664,616){\makebox(0,0){\strut{}$160$}}%
      \csname LTb\endcsname
      \put(6180,616){\makebox(0,0){\strut{}$180$}}%
      \csname LTb\endcsname
      \put(6695,616){\makebox(0,0){\strut{}$200$}}%
    }%
    \gplgaddtomacro\gplfronttext{%
      \csname LTb\endcsname
      \put(266,2827){\rotatebox{-270}{\makebox(0,0){\strut{}time-averaged relative error}}}%
      \put(4117,196){\makebox(0,0){\strut{}roll-out length}}%
      \csname LTb\endcsname
      \put(5792,4556){\makebox(0,0)[r]{\strut{}projection}}%
      \csname LTb\endcsname
      \put(5792,4276){\makebox(0,0)[r]{\strut{}OpInf + roll out}}%
      \csname LTb\endcsname
      \put(5792,3996){\makebox(0,0)[r]{\strut{}OpInf}}%
    }%
    \gplbacktext
    \put(0,0){\includegraphics[width={360.00bp},height={252.00bp}]{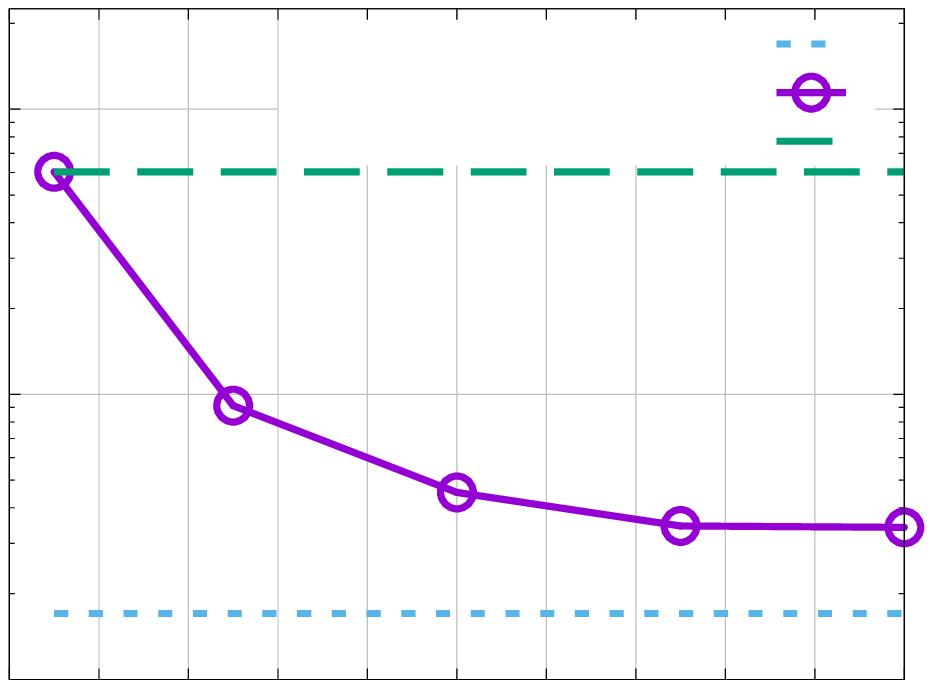}}%
    \gplfronttext
  \end{picture}%
\endgroup

%% file: SQG_errVsNoise_radius_sparse5.tex
\begingroup
  \makeatletter
  \providecommand\color[2][]{%
    \GenericError{(gnuplot) \space\space\space\@spaces}{%
      Package color not loaded in conjunction with
      terminal option `colourtext'%
    }{See the gnuplot documentation for explanation.%
    }{Either use 'blacktext' in gnuplot or load the package
      color.sty in LaTeX.}%
    \renewcommand\color[2][]{}%
  }%
  \providecommand\includegraphics[2][]{%
    \GenericError{(gnuplot) \space\space\space\@spaces}{%
      Package graphicx or graphics not loaded%
    }{See the gnuplot documentation for explanation.%
    }{The gnuplot epslatex terminal needs graphicx.sty or graphics.sty.}%
    \renewcommand\includegraphics[2][]{}%
  }%
  \providecommand\rotatebox[2]{#2}%
  \@ifundefined{ifGPcolor}{%
    \newif\ifGPcolor
    \GPcolortrue
  }{}%
  \@ifundefined{ifGPblacktext}{%
    \newif\ifGPblacktext
    \GPblacktexttrue
  }{}%
  \let\gplgaddtomacro\g@addto@macro
  \gdef\gplbacktext{}%
  \gdef\gplfronttext{}%
  \makeatother
  \ifGPblacktext
    \def\colorrgb#1{}%
    \def\colorgray#1{}%
  \else
    \ifGPcolor
      \def\colorrgb#1{\color[rgb]{#1}}%
      \def\colorgray#1{\color[gray]{#1}}%
      \expandafter\def\csname LTw\endcsname{\color{white}}%
      \expandafter\def\csname LTb\endcsname{\color{black}}%
      \expandafter\def\csname LTa\endcsname{\color{black}}%
      \expandafter\def\csname LT0\endcsname{\color[rgb]{1,0,0}}%
      \expandafter\def\csname LT1\endcsname{\color[rgb]{0,1,0}}%
      \expandafter\def\csname LT2\endcsname{\color[rgb]{0,0,1}}%
      \expandafter\def\csname LT3\endcsname{\color[rgb]{1,0,1}}%
      \expandafter\def\csname LT4\endcsname{\color[rgb]{0,1,1}}%
      \expandafter\def\csname LT5\endcsname{\color[rgb]{1,1,0}}%
      \expandafter\def\csname LT6\endcsname{\color[rgb]{0,0,0}}%
      \expandafter\def\csname LT7\endcsname{\color[rgb]{1,0.3,0}}%
      \expandafter\def\csname LT8\endcsname{\color[rgb]{0.5,0.5,0.5}}%
    \else
      \def\colorrgb#1{\color{black}}%
      \def\colorgray#1{\color[gray]{#1}}%
      \expandafter\def\csname LTw\endcsname{\color{white}}%
      \expandafter\def\csname LTb\endcsname{\color{black}}%
      \expandafter\def\csname LTa\endcsname{\color{black}}%
      \expandafter\def\csname LT0\endcsname{\color{black}}%
      \expandafter\def\csname LT1\endcsname{\color{black}}%
      \expandafter\def\csname LT2\endcsname{\color{black}}%
      \expandafter\def\csname LT3\endcsname{\color{black}}%
      \expandafter\def\csname LT4\endcsname{\color{black}}%
      \expandafter\def\csname LT5\endcsname{\color{black}}%
      \expandafter\def\csname LT6\endcsname{\color{black}}%
      \expandafter\def\csname LT7\endcsname{\color{black}}%
      \expandafter\def\csname LT8\endcsname{\color{black}}%
    \fi
  \fi
    \setlength{\unitlength}{0.0500bp}%
    \ifx\gptboxheight\undefined%
      \newlength{\gptboxheight}%
      \newlength{\gptboxwidth}%
      \newsavebox{\gptboxtext}%
    \fi%
    \setlength{\fboxrule}{0.5pt}%
    \setlength{\fboxsep}{1pt}%
    \definecolor{tbcol}{rgb}{1,1,1}%
\begin{picture}(7200.00,5040.00)%
    \gplgaddtomacro\gplbacktext{%
      \csname LTb\endcsname
      \put(1372,896){\makebox(0,0)[r]{\strut{}1e-07}}%
      \csname LTb\endcsname
      \put(1372,2184){\makebox(0,0)[r]{\strut{}1e-06}}%
      \csname LTb\endcsname
      \put(1372,3471){\makebox(0,0)[r]{\strut{}1e-05}}%
      \csname LTb\endcsname
      \put(1372,4759){\makebox(0,0)[r]{\strut{}1e-04}}%
      \put(2829,616){\makebox(0,0){\strut{}10.0\%}}%
      \put(4118,616){\makebox(0,0){\strut{}15.0\%}}%
      \put(5406,616){\makebox(0,0){\strut{}25.0\%}}%
    }%
    \gplgaddtomacro\gplfronttext{%
      \csname LTb\endcsname
      \put(266,2827){\rotatebox{-270}{\makebox(0,0){\strut{}stability radius}}}%
      \put(4117,196){\makebox(0,0){\strut{}noise percentage}}%
      \csname LTb\endcsname
      \put(5792,4556){\makebox(0,0)[r]{\strut{}OpInf + roll out}}%
      \csname LTb\endcsname
      \put(5792,4276){\makebox(0,0)[r]{\strut{}OpInf}}%
    }%
    \gplbacktext
    \put(0,0){\includegraphics[width={360.00bp},height={252.00bp}]{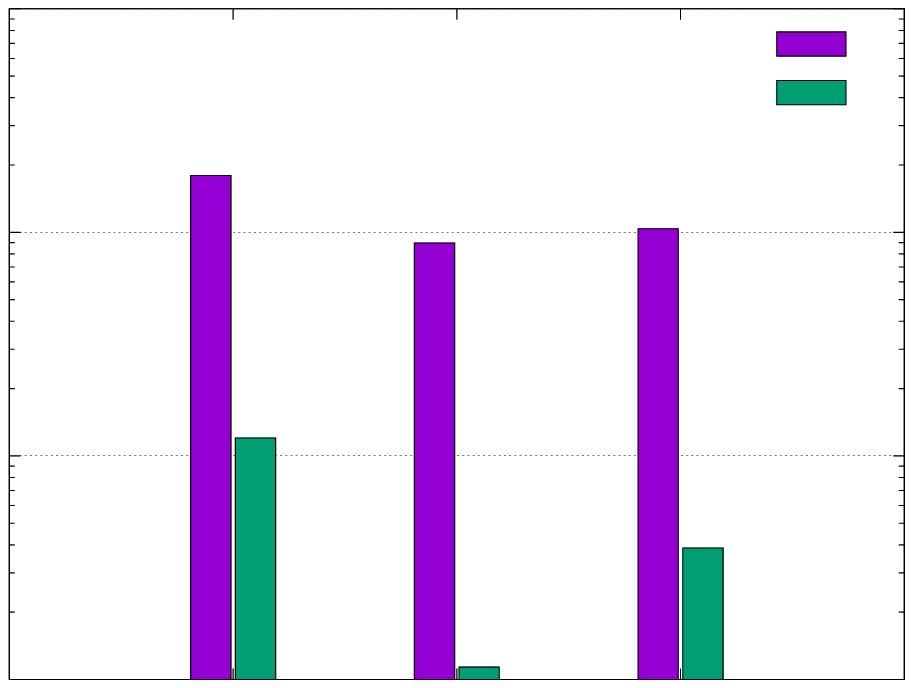}}%
    \gplfronttext
  \end{picture}%
\endgroup

%% file: SQG_errVsNoise_radius_sparse10.tex
\begingroup
  \makeatletter
  \providecommand\color[2][]{%
    \GenericError{(gnuplot) \space\space\space\@spaces}{%
      Package color not loaded in conjunction with
      terminal option `colourtext'%
    }{See the gnuplot documentation for explanation.%
    }{Either use 'blacktext' in gnuplot or load the package
      color.sty in LaTeX.}%
    \renewcommand\color[2][]{}%
  }%
  \providecommand\includegraphics[2][]{%
    \GenericError{(gnuplot) \space\space\space\@spaces}{%
      Package graphicx or graphics not loaded%
    }{See the gnuplot documentation for explanation.%
    }{The gnuplot epslatex terminal needs graphicx.sty or graphics.sty.}%
    \renewcommand\includegraphics[2][]{}%
  }%
  \providecommand\rotatebox[2]{#2}%
  \@ifundefined{ifGPcolor}{%
    \newif\ifGPcolor
    \GPcolortrue
  }{}%
  \@ifundefined{ifGPblacktext}{%
    \newif\ifGPblacktext
    \GPblacktexttrue
  }{}%
  \let\gplgaddtomacro\g@addto@macro
  \gdef\gplbacktext{}%
  \gdef\gplfronttext{}%
  \makeatother
  \ifGPblacktext
    \def\colorrgb#1{}%
    \def\colorgray#1{}%
  \else
    \ifGPcolor
      \def\colorrgb#1{\color[rgb]{#1}}%
      \def\colorgray#1{\color[gray]{#1}}%
      \expandafter\def\csname LTw\endcsname{\color{white}}%
      \expandafter\def\csname LTb\endcsname{\color{black}}%
      \expandafter\def\csname LTa\endcsname{\color{black}}%
      \expandafter\def\csname LT0\endcsname{\color[rgb]{1,0,0}}%
      \expandafter\def\csname LT1\endcsname{\color[rgb]{0,1,0}}%
      \expandafter\def\csname LT2\endcsname{\color[rgb]{0,0,1}}%
      \expandafter\def\csname LT3\endcsname{\color[rgb]{1,0,1}}%
      \expandafter\def\csname LT4\endcsname{\color[rgb]{0,1,1}}%
      \expandafter\def\csname LT5\endcsname{\color[rgb]{1,1,0}}%
      \expandafter\def\csname LT6\endcsname{\color[rgb]{0,0,0}}%
      \expandafter\def\csname LT7\endcsname{\color[rgb]{1,0.3,0}}%
      \expandafter\def\csname LT8\endcsname{\color[rgb]{0.5,0.5,0.5}}%
    \else
      \def\colorrgb#1{\color{black}}%
      \def\colorgray#1{\color[gray]{#1}}%
      \expandafter\def\csname LTw\endcsname{\color{white}}%
      \expandafter\def\csname LTb\endcsname{\color{black}}%
      \expandafter\def\csname LTa\endcsname{\color{black}}%
      \expandafter\def\csname LT0\endcsname{\color{black}}%
      \expandafter\def\csname LT1\endcsname{\color{black}}%
      \expandafter\def\csname LT2\endcsname{\color{black}}%
      \expandafter\def\csname LT3\endcsname{\color{black}}%
      \expandafter\def\csname LT4\endcsname{\color{black}}%
      \expandafter\def\csname LT5\endcsname{\color{black}}%
      \expandafter\def\csname LT6\endcsname{\color{black}}%
      \expandafter\def\csname LT7\endcsname{\color{black}}%
      \expandafter\def\csname LT8\endcsname{\color{black}}%
    \fi
  \fi
    \setlength{\unitlength}{0.0500bp}%
    \ifx\gptboxheight\undefined%
      \newlength{\gptboxheight}%
      \newlength{\gptboxwidth}%
      \newsavebox{\gptboxtext}%
    \fi%
    \setlength{\fboxrule}{0.5pt}%
    \setlength{\fboxsep}{1pt}%
    \definecolor{tbcol}{rgb}{1,1,1}%
\begin{picture}(7200.00,5040.00)%
    \gplgaddtomacro\gplbacktext{%
      \csname LTb\endcsname
      \put(1372,896){\makebox(0,0)[r]{\strut{}1e-07}}%
      \csname LTb\endcsname
      \put(1372,2828){\makebox(0,0)[r]{\strut{}1e-06}}%
      \csname LTb\endcsname
      \put(1372,4759){\makebox(0,0)[r]{\strut{}1e-05}}%
      \put(2829,616){\makebox(0,0){\strut{}10.0\%}}%
      \put(4118,616){\makebox(0,0){\strut{}15.0\%}}%
      \put(5406,616){\makebox(0,0){\strut{}25.0\%}}%
    }%
    \gplgaddtomacro\gplfronttext{%
      \csname LTb\endcsname
      \put(266,2827){\rotatebox{-270}{\makebox(0,0){\strut{}stability radius}}}%
      \put(4117,196){\makebox(0,0){\strut{}noise percentage}}%
      \csname LTb\endcsname
      \put(5792,4556){\makebox(0,0)[r]{\strut{}OpInf + roll out}}%
      \csname LTb\endcsname
      \put(5792,4276){\makebox(0,0)[r]{\strut{}OpInf}}%
    }%
    \gplbacktext
    \put(0,0){\includegraphics[width={360.00bp},height={252.00bp}]{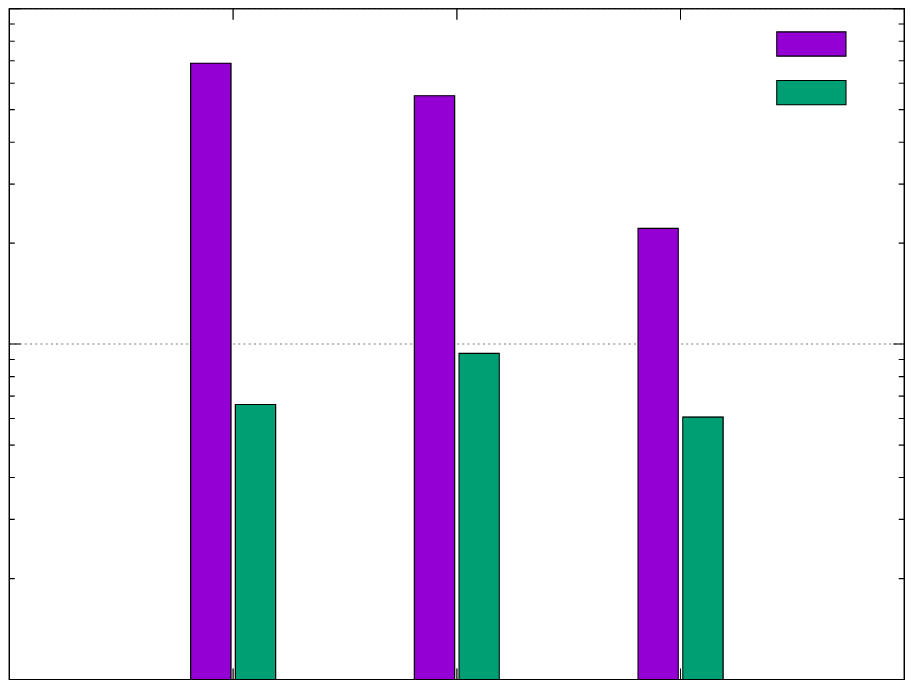}}%
    \gplfronttext
  \end{picture}%
\endgroup

%% file: main.bbl
\begin{thebibliography}{10}
\expandafter\ifx\csname url\endcsname\relax
  \def\url#1{\texttt{#1}}\fi
\expandafter\ifx\csname urlprefix\endcsname\relax\def\urlprefix{URL }\fi
\expandafter\ifx\csname href\endcsname\relax
  \def\href#1#2{#2} \def\path#1{#1}\fi

\bibitem{paper:PeherstorferW2016}
B.~Peherstorfer, K.~Willcox, Data-driven operator inference for nonintrusive
  projection-based model reduction, Computer Methods in Applied Mechanics and
  Engineering 306 (2016) 196--215.

\bibitem{NEURIPS2018_69386f6b}
R.~T.~Q. Chen, Y.~Rubanova, J.~Bettencourt, D.~K. Duvenaud, Neural ordinary
  differential equations, in: S.~Bengio, H.~Wallach, H.~Larochelle, K.~Grauman,
  N.~Cesa-Bianchi, R.~Garnett (Eds.), Advances in Neural Information Processing
  Systems, Vol.~31, Curran Associates, Inc., 2018, pp. 1--13.

\bibitem{paper:Schmid2010}
P.~J. Schmid, Dynamic mode decomposition of numerical and experimental data,
  Journal of Fluid Mechanics 656 (2010) 5--28.

\bibitem{FLM:6837872}
C.~Rowley, I.~Mezi\'{c}, S.~Bagheri, P.~Schlatter, D.~Henningson, Spectral
  analysis of nonlinear flows, Journal of Fluid Mechanics 641 (2009) 115--127.

\bibitem{Tu2014391}
J.~H. Tu, C.~W. Rowley, D.~M. Luchtenburg, S.~L. Brunton, J.~N. Kutz, On
  dynamic mode decomposition: Theory and applications, Journal of Computational
  Dynamics 1~(2) (2014) 391--421.

\bibitem{NathanBook}
J.~N. Kutz, S.~L. Brunton, B.~W. Brunton, J.~L. Proctor, Dynamic mode
  decomposition: Data-driven modeling of complex systems, SIAM, 2016.

\bibitem{Williams2015}
M.~O. Williams, I.~G. Kevrekidis, C.~W. Rowley, A data{\textendash}driven
  approximation of the {K}oopman operator: Extending dynamic mode
  decomposition, Journal of Nonlinear Science 25~(6) (2015) 1307--1346.

\bibitem{doi:10.1137/20M1338289}
D.~Burov, D.~Giannakis, K.~Manohar, A.~Stuart, Kernel analog forecasting:
  Multiscale test problems, Multiscale Modeling \& Simulation 19~(2) (2021)
  1011--1040.

\bibitem{paper:BruntonPK2016}
S.~L. Brunton, J.~L. Proctor, J.~N. Kutz, Discovering governing equations from
  data by sparse identification of nonlinear dynamical systems, Proceedings of
  the National Academy of Sciences 113~(15) (2016) 3932--3937.

\bibitem{Schaeffer6634}
H.~Schaeffer, R.~Caflisch, C.~D. Hauck, S.~Osher, Sparse dynamics for partial
  differential equations, Proceedings of the National Academy of Sciences
  110~(17) (2013) 6634--6639.

\bibitem{Schaeffer2018}
H.~Schaeffer, G.~Tran, R.~Ward, Extracting sparse high-dimensional dynamics
  from limited data, {SIAM} Journal on Applied Mathematics 78~(6) (2018)
  3279--3295.

\bibitem{Rudye1602614}
S.~H. Rudy, S.~L. Brunton, J.~L. Proctor, J.~N. Kutz, Data-driven discovery of
  partial differential equations, Science Advances 3~(4) (2017).

\bibitem{doi:10.1137/17M1145136}
X.~Xie, M.~Mohebujjaman, L.~G. Rebholz, T.~Iliescu, Data-driven filtered
  reduced order modeling of fluid flows, SIAM Journal on Scientific Computing
  40~(3) (2018) B834--B857.

\bibitem{MOU2021113470}
C.~Mou, B.~Koc, O.~San, L.~G. Rebholz, T.~Iliescu, Data-driven variational
  multiscale reduced order models, Computer Methods in Applied Mechanics and
  Engineering 373 (2021) 113470.

\bibitem{Hesthaven2018}
J.~S. Hesthaven, S.~Ubbiali, Non-intrusive reduced order modeling of nonlinear
  problems using neural networks, Journal of Computational Physics 363 (2018)
  55--78.

\bibitem{DEMO2019873}
N.~Demo, M.~Tezzele, G.~Rozza, A non-intrusive approach for the reconstruction
  of {POD} modal coefficients through active subspaces, Comptes Rendus
  Mécanique 347~(11) (2019) 873--881.

\bibitem{Guo2018}
M.~Guo, J.~S. Hesthaven, Reduced order modeling for nonlinear structural
  analysis using {Gaussian} process regression, Computer Methods in Applied
  Mechanics and Engineering 341 (2018) 807--826.

\bibitem{doi:10.1080/10618562.2014.932352}
D.~Forti, G.~Rozza, Efficient geometrical parametrisation techniques of
  interfaces for reduced-order modelling: application to fluid–structure
  interaction coupling problems, International Journal of Computational Fluid
  Dynamics 28~(3-4) (2014) 158--169.

\bibitem{Guo2019}
M.~Guo, J.~S. Hesthaven, Data-driven reduced order modeling for time-dependent
  problems, Computer Methods in Applied Mechanics and Engineering 345 (2019)
  75--99.

\bibitem{LARIO2022111475}
A.~Lario, R.~Maulik, O.~T. Schmidt, G.~Rozza, G.~Mengaldo, Neural-network
  learning of {SPOD} latent dynamics, Journal of Computational Physics 468
  (2022) 111475.

\bibitem{PhysRevE.100.053306}
S.~M. Rahman, S.~Pawar, O.~San, A.~Rasheed, T.~Iliescu, Nonintrusive reduced
  order modeling framework for quasigeostrophic turbulence, Phys. Rev. E 100
  (2019) 053306.

\bibitem{Demo2018}
N.~Demo, M.~Tezzele, G.~Rozza, {EZyRB}: Easy reduced basis method, Journal of
  Open Source Software 3~(24) (2018) 661.

\bibitem{ANTOULAS01011986}
A.~C. Antoulas, B.~D.~O. Anderson, On the scalar rational interpolation
  problem, IMA Journal of Mathematical Control \& Information 3~(2-3) (1986)
  61--88.

\bibitem{5356286}
S.~Lefteriu, A.~C. Antoulas, A new approach to modeling multiport systems from
  frequency-domain data, Computer-Aided Design of Integrated Circuits and
  Systems, IEEE Transactions on 29~(1) (2010) 14--27.

\bibitem{Mayo2007634}
A.~J. Mayo, A.~C. Antoulas, A framework for the solution of the generalized
  realization problem, Linear Algebra and its Applications 425~(2–3) (2007)
  634--662.

\bibitem{BeaG12}
C.~Beattie, S.~Gugercin, {Realization-independent
  $\mathcal{H}_2$-approximation}, in: Proc. IEEE Conf. Decis. Control, Maui,
  HI, USA, 2012, pp. 4953--4958.

\bibitem{paper:AntoulasGI2016}
A.~C. Antoulas, I.~V. Gosea, A.~C. Ionita, Model reduction of bilinear systems
  in the {Loewner} framework, {SIAM} Journal on Scientific Computing 38~(5)
  (2016) B889--B916.

\bibitem{paper:GoseaA2018}
I.~V. Gosea, A.~C. Antoulas, Data-driven model order reduction of
  quadratic-bilinear systems, Numerical Linear Algebra with Applications 25~(6)
  (2018) e2200.

\bibitem{paper:IonitaA2014}
A.~C. Ionita, A.~C. Antoulas, Data-driven parametrized model reduction in the
  {Loewner} framework, {SIAM} Journal on Scientific Computing 36~(3) (2014)
  A984--A1007.

\bibitem{Drmac2022}
Z.~Drma{\v{c}}, B.~Peherstorfer, Learning low-dimensional dynamical-system
  models from noisy frequency-response data with {Loewner} rational
  interpolation, in: C.~Beattie, P.~Benner, M.~Embree, S.~Gugercin, S.~Lefteriu
  (Eds.), Realization and Model Reduction of Dynamical Systems: A Festschrift
  in Honor of the 70th Birthday of Thanos Antoulas, Springer International
  Publishing, Cham, 2022, pp. 39--57.

\bibitem{PSW16TLoewner}
B.~Peherstorfer, S.~Gugercin, K.~Willcox, Data-driven reduced model
  construction with time-domain {Loewner} models, SIAM Journal on Scientific
  Computing 39~(5) (2017) A2152--A2178.

\bibitem{772353}
B.~{Gustavsen}, A.~{Semlyen}, Rational approximation of frequency domain
  responses by vector fitting, IEEE Transactions on Power Delivery 14~(3)
  (1999) 1052--1061.

\bibitem{paper:DrmacGB2015}
Z.~Drma{\v{c}}, S.~Gugercin, C.~Beattie, Vector fitting for matrix-valued
  rational approximation, {SIAM} Journal on Scientific Computing 37~(5) (2015)
  A2346--A2379.

\bibitem{doi:10.2514/3.20031}
J.-N. Juang, R.~S. Pappa, An eigensystem realization algorithm for modal
  parameter identification and model reduction, Journal of Guidance, Control,
  and Dynamics 8~(5) (1985) 620--627.

\bibitem{KramerERZ}
B.~Kramer, S.~Gugercin, Tangential interpolation-based eigensystem realization
  algorithm for {MIMO} systems, Mathematical and Computer Modelling of
  Dynamical Systems 22~(4) (2016) 282--306.

\bibitem{QIAN2020132401}
E.~Qian, B.~Kramer, B.~Peherstorfer, K.~Willcox, Lift \& learn:
  Physics-informed machine learning for large-scale nonlinear dynamical
  systems, Physica D: Nonlinear Phenomena 406 (2020) 132401.

\bibitem{https://doi.org/10.48550/arxiv.2110.07653}
S.~A. McQuarrie, P.~Khodabakhshi, K.~E. Willcox, Non-intrusive reduced-order
  models for parametric partial differential equations via data-driven operator
  inference, arXiv 2110.07653 (2021).

\bibitem{khodabakhshi2022non}
P.~Khodabakhshi, K.~E. Willcox, Non-intrusive data-driven model reduction for
  differential--algebraic equations derived from lifting transformations,
  Computer Methods in Applied Mechanics and Engineering 389 (2022) 114296.

\bibitem{Uy2021}
W.~I.~T. Uy, B.~Peherstorfer, Operator inference of non-{Markovian} terms for
  learning reduced models from partially observed state trajectories, Journal
  of Scientific Computing 88~(3) (2021) 91.

\bibitem{UP20OpInfError}
W.~I.~T. Uy, B.~Peherstorfer, Probabilistic error estimation for non-intrusive
  reduced models learned from data of systems governed by linear parabolic
  partial differential equations, ESAIM: Mathematical Modelling and Numerical
  Analysis (M2AN) 55~(3) (2021) 735--761.

\bibitem{BENNER2020113433}
P.~Benner, P.~Goyal, B.~Kramer, B.~Peherstorfer, K.~Willcox, Operator inference
  for non-intrusive model reduction of systems with non-polynomial nonlinear
  terms, Computer Methods in Applied Mechanics and Engineering 372 (2020)
  113433.

\bibitem{SWK21-HOPINF}
H.~Sharma, Z.~Wang, B.~Kramer, {H}amiltonian operator inference:
  Physics-preserving learning of reduced-order models for {H}amiltonian
  systems, Physica {D}: {N}onlinear {P}henomena 431 (2022) 133122.

\bibitem{doi:10.1080/03036758.2020.1863237}
S.~A. McQuarrie, C.~Huang, K.~E. Willcox, Data-driven reduced-order models via
  regularised operator inference for a single-injector combustion process,
  Journal of the Royal Society of New Zealand 51~(2) (2021) 194--211.

\bibitem{Guo_2022}
M.~Guo, S.~A. McQuarrie, K.~E. Willcox, Bayesian operator inference for
  data-driven reduced-order modeling, Computer Methods in Applied Mechanics and
  Engineering (2022) 115336.

\bibitem{doi:10.2514/1.J058943}
R.~Swischuk, B.~Kramer, C.~Huang, K.~Willcox, Learning physics-based
  reduced-order models for a single-injector combustion process, AIAA Journal
  58~(6) (2020) 2658--2672.

\bibitem{QianThesis}
E.~Qian, A scientific machine learning approach to learning reduced models for
  nonlinear partial differential equations, Ph.D. thesis, Massachusetts
  Institute of Technology (2021).

\bibitem{paper:Peherstorfer2019}
B.~Peherstorfer, Sampling low-dimensional markovian dynamics for
  pre-asymptotically recovering reduced models from data with operator
  inference, SIAM Journal on Scientific Computing 42 (2020) A3489--A3515.

\bibitem{https://doi.org/10.48550/arxiv.2107.02597}
N.~Sawant, B.~Kramer, B.~Peherstorfer, Physics-informed regularization and
  structure preservation for learning stable reduced models from data with
  operator inference, arXiv 2107.02597 (2021).

\bibitem{AOpInf}
W.~I.~T. Uy, Y.~Wang, Y.~Wen, B.~Peherstorfer, Active operator inference for
  learning low-dimensional dynamical-system models from noisy data, arXiv
  2107.09256 (2021).

\bibitem{HartmannFailerOpInf}
D.~Hartmann, L.~Failer, A differentiable solver approach to operator inference,
  arXiv 2107.02093 (2021).

\bibitem{NEURIPS2019_42a6845a}
Y.~Rubanova, R.~T.~Q. Chen, D.~K. Duvenaud, Latent ordinary differential
  equations for irregularly-sampled time series, in: H.~Wallach, H.~Larochelle,
  A.~Beygelzimer, F.~d'Alch\'{e} Buc, E.~Fox, R.~Garnett (Eds.), Advances in
  Neural Information Processing Systems, Vol.~32, Curran Associates, Inc.,
  2019, pp. 1--11.

\bibitem{doi:10.1098/rspa.2021.0162}
K.~Lee, E.~J. Parish, Parameterized neural ordinary differential equations:
  applications to computational physics problems, Proceedings of the Royal
  Society A: Mathematical, Physical and Engineering Sciences 477~(2253) (2021)
  20210162.

\bibitem{https://doi.org/10.48550/arxiv.2202.12373}
J.~Baker, E.~Cherkaev, A.~Narayan, B.~Wang, Learning {POD} of complex dynamics
  using heavy-ball neural {ODEs}, arXiv 2202.12373 (2022).

\bibitem{SAN2019271}
O.~San, R.~Maulik, M.~Ahmed, An artificial neural network framework for reduced
  order modeling of transient flows, Communications in Nonlinear Science and
  Numerical Simulation 77 (2019) 271--287.

\bibitem{Fresca2021}
S.~Fresca, L.~Dede', A.~Manzoni, A comprehensive deep learning-based approach
  to reduced order modeling of nonlinear time-dependent parametrized {PDEs},
  Journal of Scientific Computing 87~(2) (2021) 61.

\bibitem{doi:10.1098/rspa.2020.0290}
Y.~Yang, M.~Aziz~Bhouri, P.~Perdikaris, Bayesian differential programming for
  robust systems identification under uncertainty, Proceedings of the Royal
  Society A: Mathematical, Physical and Engineering Sciences 476~(2243) (2020)
  20200290.

\bibitem{RUDY2019483}
S.~H. Rudy, J.~{Nathan Kutz}, S.~L. Brunton, Deep learning of dynamics and
  signal-noise decomposition with time-stepping constraints, Journal of
  Computational Physics 396 (2019) 483--506.

\bibitem{https://doi.org/10.48550/arxiv.2205.09479}
P.~Goyal, P.~Benner, Neural {ODEs} with irregular and noisy data, arXiv
  2205.09479 (2022).

\bibitem{NME:NME2681}
D.~Amsallem, J.~Cortial, K.~Carlberg, C.~Farhat, A method for interpolating on
  manifolds structural dynamics reduced-order models, International Journal for
  Numerical Methods in Engineering 80~(9) (2009) 1241--1258.

\bibitem{panzer_parametric_2010}
H.~Panzer, J.~Mohring, R.~Eid, B.~Lohmann, Parametric model order reduction by
  matrix interpolation, at -- Automatisierungstechnik 58~(8) (2010) 475--484.

\bibitem{degroote_interpolation_2010}
J.~Degroote, J.~Vierendeels, K.~Willcox, Interpolation among reduced-order
  matrices to obtain parameterized models for design, optimization and
  probabilistic analysis, International Journal for Numerical Methods in Fluids
  63~(2) (2010) 207--230.

\bibitem{amsallem_online_2011}
D.~Amsallem, C.~Farhat, An online method for interpolating linear parametric
  reduced-order models, {SIAM} Journal on Scientific Computing 33~(5) (2011)
  2169--2198.

\bibitem{doi:10.1098/rspa.2021.0883}
P.~Goyal, P.~Benner, Discovery of nonlinear dynamical systems using a
  {Runge-Kutta} inspired dictionary-based sparse regression approach,
  Proceedings of the Royal Society A: Mathematical, Physical and Engineering
  Sciences 478~(2262) (2022) 20210883.

\bibitem{book:Antoulas2005}
A.~C. Antoulas, Approximation of Large-Scale Dynamical Systems, Society for
  Industrial and Applied Mathematics, 2005.

\bibitem{RozzaPateraSurvey}
G.~Rozza, D.~Huynh, A.~T. Patera, Reduced basis approximation and a posteriori
  error estimation for affinely parametrized elliptic coercive partial
  differential equations, Archives of Computational Methods in Engineering
  15~(3) (2008) 1--47.

\bibitem{paper:BennerGW2015}
P.~Benner, S.~Gugercin, K.~Willcox, A survey of projection-based model
  reduction methods for parametric dynamical systems, {SIAM} Review 57~(4)
  (2015) 483--531.

\bibitem{P22AMS}
B.~Peherstorfer, Breaking the {Kolmogorov} barrier with nonlinear model
  reduction, Notices of the American Mathematical Society 69 (2022) 725--733.

\bibitem{rawlings2017model}
J.~B. Rawlings, D.~Q. Mayne, M.~Diehl, Model predictive control: theory,
  computation, and design, Vol.~2, Nob Hill Publishing Madison, 2017.

\bibitem{um2020solver}
K.~Um, R.~Brand, Y.~R. Fei, P.~Holl, N.~Thuerey, Solver-in-the-loop: Learning
  from differentiable physics to interact with iterative {PDE}-solvers,
  Advances in Neural Information Processing Systems 33 (2020) 6111--6122.

\bibitem{kochkov2021machine}
D.~Kochkov, J.~A. Smith, A.~Alieva, Q.~Wang, M.~P. Brenner, S.~Hoyer, Machine
  learning--accelerated computational fluid dynamics, Proceedings of the
  National Academy of Sciences 118~(21) (2021) e2101784118.

\bibitem{schoenholz2020jax}
S.~Schoenholz, E.~D. Cubuk, Jax {MD}: a framework for differentiable physics,
  Advances in Neural Information Processing Systems 33 (2020) 11428--11441.

\bibitem{genesio1989stability}
R.~Genesio, A.~Tesi, Stability analysis of quadratic systems, IFAC Proceedings
  Volumes 22~(3) (1989) 195--199.

\bibitem{411100}
A.~Tesi, F.~Villoresi, R.~Genesio, On stability domain estimation via a
  quadratic {Lyapunov} function: convexity and optimality properties for
  polynomial systems, in: Proceedings of 1994 33rd IEEE Conference on Decision
  and Control, Vol.~2, 1994, pp. 1907--1912.

\bibitem{K2020_stability_domains_QBROMs}
B.~Kramer, Stability domains for quadratic-bilinear reduced-order models, SIAM
  Journal on Applied Dynamical Systems 20~(2) (2021) 981--996.

\bibitem{kaptanoglu2021promoting}
A.~A. Kaptanoglu, J.~L. Callaham, C.~J. Hansen, A.~Aravkin, S.~L. Brunton,
  Promoting global stability in data-driven models of quadratic nonlinear
  dynamics, arXiv 2105~(01843) (2021) 1--29.

\bibitem{5991229}
C.~Gu, {QLMOR}: A projection-based nonlinear model order reduction approach
  using quadratic-linear representation of nonlinear systems, IEEE Transactions
  on Computer-Aided Design of Integrated Circuits and Systems 30~(9) (2011)
  1307--1320.

\bibitem{47008}
R.~Frostig, M.~Johnson, C.~Leary,
  \href{https://mlsys.org/Conferences/doc/2018/146.pdf}{Compiling machine
  learning programs via high-level tracing}, in: MLSys, 2018, pp. 1--3.
\newline\urlprefix\url{https://mlsys.org/Conferences/doc/2018/146.pdf}

\bibitem{jax2018github}
J.~Bradbury, R.~Frostig, P.~Hawkins, M.~J. Johnson, C.~Leary, D.~Maclaurin,
  G.~Necula, A.~Paszke, J.~Vander{P}las, S.~Wanderman-{M}ilne, Q.~Zhang,
  \href{http://github.com/google/jax}{{JAX}: composable transformations of
  {P}ython+{N}um{P}y programs} (2018).
\newline\urlprefix\url{http://github.com/google/jax}

\bibitem{rahman2022neural}
A.~Rahman, J.~Drgo{\v{n}}a, A.~Tuor, J.~Strube, Neural ordinary differential
  equations for nonlinear system identification, arXiv preprint
  arXiv:2203.00120 (2022).

\bibitem{ADAMOptimizer}
D.~P. Kingma, J.~Ba, Adam: A method for stochastic optimization., in:
  Y.~Bengio, Y.~LeCun (Eds.), ICLR (Poster), 2015, pp. 1--1.

\bibitem{held_pierrehumbert_garner_swanson_1995}
I.~M. Held, R.~T. Pierrehumbert, S.~T. Garner, K.~L. Swanson, Surface
  quasi-geostrophic dynamics, Journal of Fluid Mechanics 282 (1995) 1–20.

\end{thebibliography}
